\documentclass{article}

\PassOptionsToPackage{numbers, compress}{natbib}

\usepackage[preprint]{neurips_2025}

\usepackage[utf8]{inputenc} 
\usepackage[T1]{fontenc}    
\usepackage{hyperref}       
\usepackage{url}            
\usepackage{booktabs}       
\usepackage{amsfonts}       
\usepackage{nicefrac}       
\usepackage{microtype}      
\usepackage{xcolor}         
\usepackage{amsmath}
\usepackage{graphicx}
\usepackage{subfigure}
\usepackage{placeins}
\usepackage[ruled,vlined]{algorithm2e} 
\usepackage{multirow}
\usepackage{soul}           
\usepackage{listings}
\usepackage{enumitem}
\usepackage{color, colortbl}
\usepackage{fancyvrb}
\usepackage{wrapfig}
\usepackage{floatflt}
\usepackage{makecell}
\usepackage{tabularray}
\usepackage{array}

\title{ScreenExplorer: Training a Vision–Language Model for Diverse Exploration in Open GUI World}

\author{%
  Runliang Niu\footnotemark[1], Jinglong Ji, Yi Chang, Qi Wang\footnotemark[2] \\
  School of Artificial Intelligence, Jilin University\\
  ChangChun, China. 
}

\begin{document}

\footnotetext[1]{Email: \texttt{niurl19@mails.jlu.edu.cn} }
\footnotetext[2]{Corresponding author, email: \texttt{qiwang@jlu.edu.cn }}

\maketitle

\begin{abstract}\label{sec:abstract}

The rapid progress of large language models (LLMs) has sparked growing interest in building Artificial General Intelligence (AGI) within Graphical User Interface (GUI) environments. However, existing GUI agents based on LLMs or vision-language models (VLMs) often fail to generalize to novel environments and rely heavily on manually curated, diverse datasets. To overcome these limitations, we introduce ScreenExplorer, a VLM trained via Group Relative Policy Optimization(GRPO) in real, dynamic, and open-ended GUI environments. Innovatively, we introduced a world-model-based curiosity reward function to help the agent overcome the cold-start phase of exploration. Additionally, distilling experience streams further enhances the model's exploration capabilities. Our training framework enhances model exploration in open GUI environments, with trained models showing better environmental adaptation and sustained exploration compared to static deployment models. Our findings offer a scalable pathway toward AGI systems with self-improving capabilities in complex interactive settings. 

\end{abstract}

\section{Introduction}\label{sec:introduction}
A central goal of Artificial General Intelligence (AGI) is to build agents that can autonomously explore open-world environments and continuously interact with them. With the recent advances in large language models (LLMs) and visual language models (VLMs), agents have shown impressive task generalization capabilities across domains \cite{bubeck2023sparksartificialgeneralintelligence}. Among open-world environments, Graphical User Interfaces (GUIs) are one of the most ubiquitous areas of human-computer interaction, making GUI agents a compelling research direction \cite{zhang2025largelanguagemodelbrainedgui,tan2024cradle}.

Open-world GUI environments pose two major challenges: First, the content of the interface is dynamic. For example, pages in news websites or file browsers often change over time, requiring agents to adapt to evolving states. Second, the state space is virtually unbounded, making it infeasible to manually collect sufficient diverse data to support robust generalization. However, most existing GUI agents powered by frozen models, which fundamentally implies that they cannot adapt to new environments through parameter updates. Thus, they fail to learn from the trial and error process \cite{zhang2024ufo,zhang2025ufo2,wu2024oscopilot,Agent-S,Agent-S2,hong2024cogagentvisuallanguagemodel}. Moreover, these models rely on costly human-collected datasets \cite{Rico2017Deka,zhang2021screenrecognitioncreatingaccessibility,deng2023mind2web,wu2024mobilevlmvisionlanguagemodelbetter,ScreenAgent2024niu,GUICourse2024Wentong}, where annotators must thoroughly explore the environment to collect diverse samples. 
Therefore, the ability to effectively interact with new environments and autonomously perform diverse/ active exploration is crucial for the evolution of GUI agents.

In this work, we build a real, dynamic, and open-ended GUI environment and propose ScreenExplorer, a VLM agent trained via reinforcement learning (RL) to both interact effectively and explore actively. \begin{wrapfigure}{r}{0.4\linewidth}
\centering
\includegraphics[width=\linewidth]{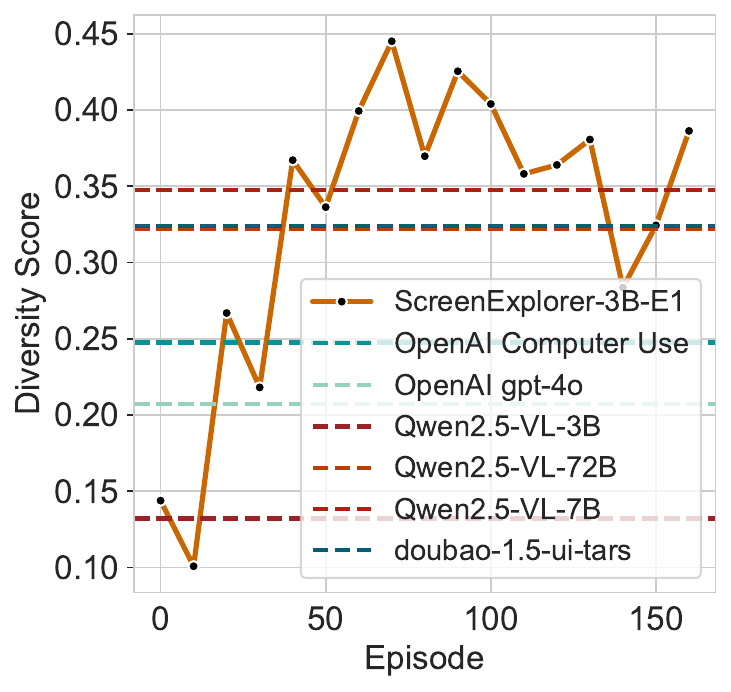}
\caption{ScreenExplorer-3B-E1's RL training leads to better GUI exploration diversity versus static models.}\label{trianse}
\vspace{-0.4cm}
\end{wrapfigure}  During training, we design rewards to encourage actions that lead to successful interaction in novel environments. To further incentivize exploration, we introduce a World Model(WM) that learns the environment's transition dynamics. 
We quantify the novelty of visited states by measuring the discrepancy between the predicted and actual state transitions, 
combined with other environment change-related rewards to form a curiosity-driven reward signal for guiding exploration.
As shown in Figure \ref{trianse}, our ScreenExplorer demonstrates significant improvement from the worst to the best in exploring diversity through RL training.

Additionally, inspired by DeepSeek-R1\cite{deepseekai2025deepseekr1incentivizingreasoningcapability}, we collect the experience streams generated during exploration and distill the original model. This learning strategy is designed to mitigate exploration bottlenecks and enable sustained capability improvement, aiming to lead to better initialization.  We find that fine-tuning agents on experience from the previous phase leads to better initialization for exploration, promoting both efficiency and diversity.

Our results suggest that training agents via RL in open GUI environments fosters both effective interaction and diverse/autonomous exploration. By combining these abilities with experience stream distillation, agents can gradually reduce reliance on manually collected data and continuously evolve. This offers a promising path toward building AGI systems capable of continual self-improvement. Our contributions are as follows:

\begin{itemize}[leftmargin=*]
\item We propose ScreenExplorer, a vision-language model (VLM) agent trained via world-model-based RL in a real, dynamic, and open-ended GUI environment. The agent is rewarded for both successful interaction and exploration novelty, enabling generalization to previously unseen interface states.
\item We introduce a curiosity mechanism that leverages a world model to estimate state novelty in transition dynamics. This encourages the agent to actively explore diverse states, addressing the challenge of sparse supervision in open-ended GUI environments.
\item We develop an experience stream distillation pipeline inspired by DeepSeek-R1, where each generation’s exploration experience is reused to fine-tune future agents. This strategy improves exploration efficiency, reduces reliance on manually curated datasets, and enables continuous capability evolution.
\end{itemize}

\footnotetext{Our source code and visualisation are available at \url{https://github.com/niuzaisheng/ScreenExplorer}.}

\section{Related Work}

\subsection{Open-World Exploration}
Open-world or open-ended environments are characterized by vast state spaces, variable objectives, and sparse rewards, where agents must engage in active exploration to acquire rewards and accomplish tasks. Researchers have focused on enhancing RL agents' exploration capabilities through various approaches, including intrinsic motivation (such as curiosity-driven rewards) 
\cite{burda2018explorationrandomnetworkdistillation,NovelD2021Zhang}, unsupervised skill discovery \cite{eysenbach2018diversityneedlearningskills,sharma2020dynamicsawareunsuperviseddiscoveryskills,wang2023voyageropenendedembodiedagent}, goal-oriented policy learning \cite{campero2021learningamigoadversariallymotivated}, and exploration rewards \cite{li2025openworldreinforcementlearninglong}.
Random Network Distillation (RND) \cite{burda2018explorationrandomnetworkdistillation} efficiently drives agent exploration to novel states by utilizing the prediction error between a fixed random target network and a trainable predictor network as an intrinsic reward signal.
These methods aim to enable agents to autonomously discover meaningful behavioral patterns in the absence of external rewards, thereby achieving broader skill sets and higher sample efficiency in complex environments.
In \cite{li2025openworldreinforcementlearninglong}, the authors tackle the open-world 3D environment of Minecraft by first generating simulated trajectories that progressively zoom in on a target object, then use a task reward model to score these image sequences and train a world model to predict environmental dynamics. This world model produces an affordance map indicating feasible regions for the target object, which is then used as an intrinsic reward to drive the agent to approach the object and complete the task. In our work, inspired by \cite{burda2018explorationrandomnetworkdistillation} and \cite{li2025openworldreinforcementlearninglong}, we introduce a world model into the exploration process to approximate environmental state transitions. We use the discrepancy between the world model's predictions and actual post-action states as an intrinsic exploration reward, driving the model to interact effectively with the environment and reach unexplored states.

\subsection{Diverse Exploration by LM}
For statically deployed models, various decoding strategies can be employed to control output diversity, such as adjusting temperature sampling or top-p decoding parameters. Additionally, researchers have explored using Monte Carlo Tree Search (MCTS) to generate diverse solution paths, followed by validation through external tools or Process-supervised Reward Models (PRM) to verify path correctness, thereby enhancing the model's capability to solve practical problems \cite{xin2024deepseekproverv15harnessingproofassistant,lightman2023letsverifystepstep,qi2024mutualreasoningmakessmaller}.
In \cite{pan2025largelanguagemodelsthink}, researchers evaluate LLMs' exploration capabilities through the Little Alchemy 2 game, revealing that most LLMs—with the exception of the OpenAI o1 model—performed worse than humans, highlighting the need to improve LLMs' ability to explore open environments. In \cite{dou2025improvingrlexplorationllm}, researchers find that LLMs' early promising solutions are often forgotten during RL fine-tuning due to policy gradient updates. Their proposed RRL algorithm stores and replays valuable early trajectories, enabling models to revisit previous approaches as their capabilities grow. This highlights the value of sustained exploration in RL-based LLM optimization. Another research \cite{ragen2025} shows that training LLMs by RL can reduce exploration as models tend to repeat early high-reward responses, leading to repetitive outputs rather than exploring diverse solutions.

\subsection{LLM- or VLM-Brained GUI Agent}

The GUI environment presents an ideal platform for training LLM/VLM agent as it combines both textual and visual elements, leveraging the inherent knowledge capabilities of large models. Recent research has focused on end-to-end training of VLMs to generate GUI control instructions for directly manipulating desktops \cite{zhang2024ufo,lin2024showuivisionlanguageactionmodelgui}, 
mobile phones \cite{wu2024foundations,wang2024mobileagentautonomousmultimodalmobile}, browsers \cite{deng2023mind2web}, or games \cite{tan2024towards,tan2024cradle}. Notable commercial products include OpenAI Operator\footnote{\url{https://openai.com/index/introducing-operator/}} and Manus \footnote{\url{https://manus.im/}}. Voyager \cite{wang2023voyageropenendedembodiedagent}, powered by GPT-4, operates in the open-world environment of Minecraft. These products employ an unsupervised skill discovery mechanism to automatically generate and iteratively refine an executable code skill library, thereby achieving continuous autonomous exploration and diverse skill acquisition. Beyond static model deployment, training agents using RL in real environments can enhance their adaptability and improve task performance. DigiRL \cite{bai2024digirltraininginthewilddevicecontrol} establishes an offline-to-online RL training pipeline by using offline RL to train VLM GUI control on existing trajectories. Then, the process transfers to online training, which incorporates autonomous evaluation from \cite{pan2024autonomousevaluationrefinementdigital}.
DistRL \cite{wang2025distrlasynchronousdistributedreinforcement} employs large-scale RL to train VLM agents for Android task completion, utilizing Gemini-1.5-pro as an evaluator to determine task success.
Most existing approaches are costly as they need to collect and label data manually. Our method employs a heuristic-based reward, offering a more cost-effective solution.

\section{Framework}

In this section, we present the overall framework of our ScreenExplorer model, which is designed to train Vision-Language Model (VLM) agents to perform autonomous and exploratory interactions within graphical user interface (GUI) environments. To support this, we construct a GUI-based operating system as a RL environment, where the VLM agent interacts with GUI in a human-like manner by outputting function calls for mouse and keyboard operations. Technical details of this environment are provided in Appendix \ref{appendix:environment}. We formulate GUI exploration as a Markov Decision Process (MDP), where the agent perceives the environment through visual and textual signals, and outputs structured actions and intent descriptions using a VLM-based policy. The framework integrates several core components: a reward system tailored for encouraging exploration and meaningful interaction, a world model that predicts state transitions to support curiosity-driven behavior, and a training pipeline that combines RL with supervised distillation. The experience stream distillation allows the model to efficiently inherit and build upon past exploration knowledge, improving both exploration breadth and behavioral robustness. The subsequent subsections detail the MDP formulation, reward design, world model learning, and policy optimization with GRPO.

\begin{figure}
    \centering
    \subfigure[One episode rollout collection and training progress.]{
            \includegraphics[width=0.56\textwidth]{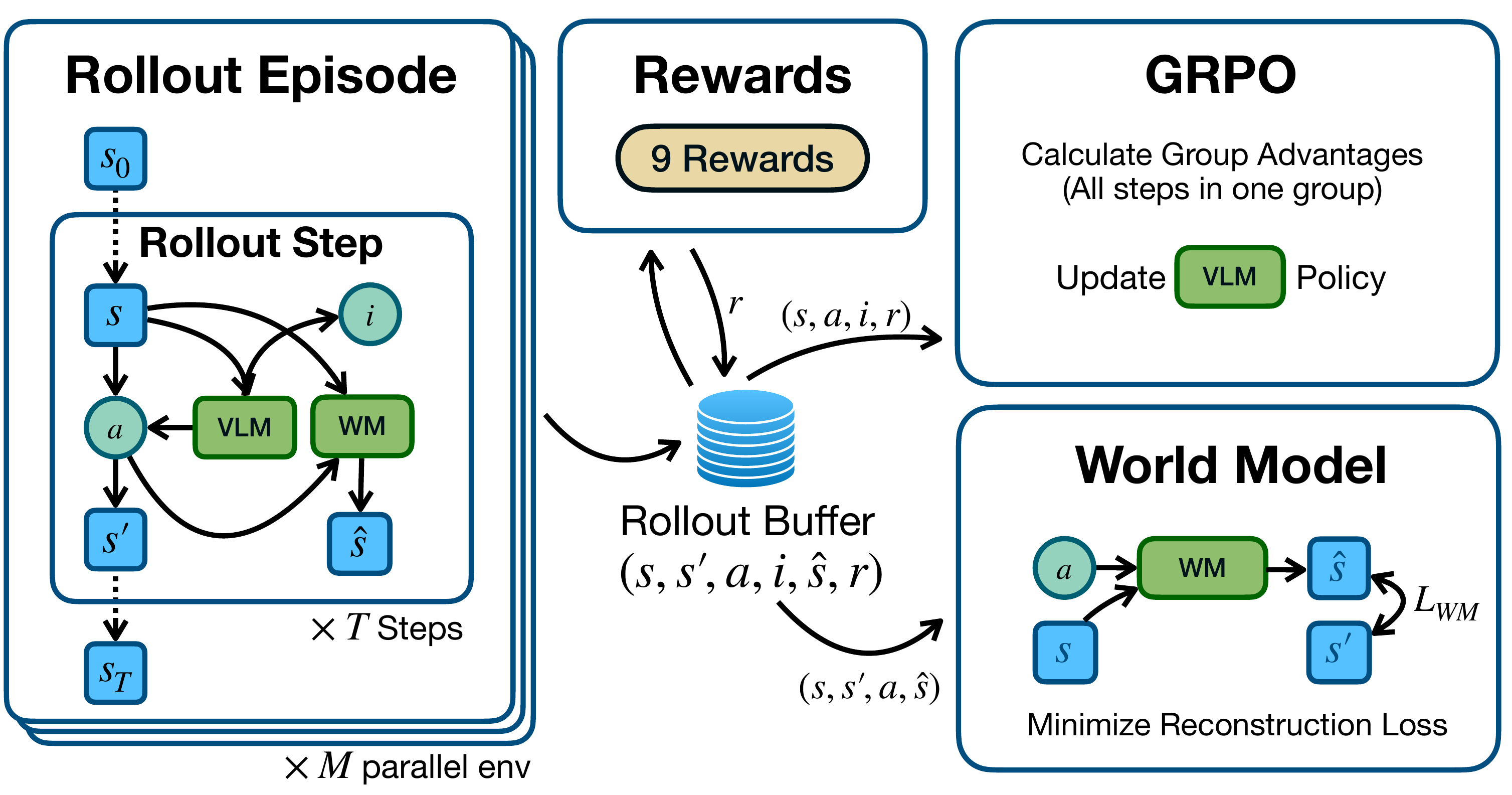}
            \label{fig:framework}
    }
    \subfigure[Reward function.]{
            \includegraphics[width=0.39\textwidth]{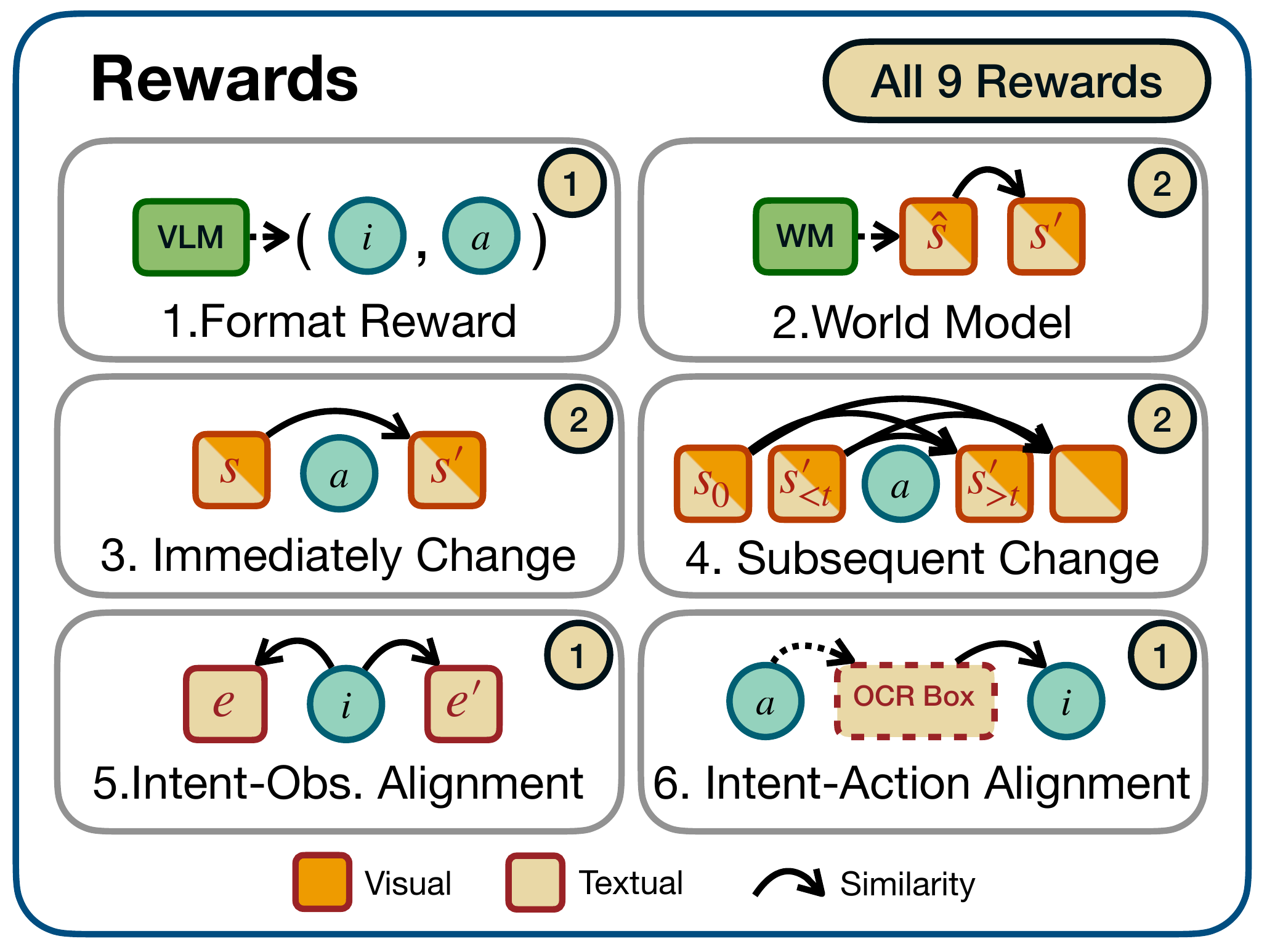}
            \label{fig:pipeline}
    }
    \caption{Framework overview: (a)We run $M$ parallel environments for $T$ steps per episode. At each step, the VLM takes state $s$ and outputs an intent $i$ and action $a$, the environment returns the post-action state $s'$, and the world model predicts the next state $\hat{s}$. All transitions are stored in a rollout buffer, where a reward function computes an exploration reward for each action. The VLM is then updated via GRPO, while the world model learns transitions by minimizing reconstruction error; (b)The reward function consists of nine terms that enforce correct action formatting, encourage large state changes, and align intents with observed states.}
\end{figure}

\subsection{Modeling Exploration in GUI as a Markov Decision Process (MDP)}

The current environment state $s := (o, e)$ is a tuple, where $o$ represents the screenshot image and $e$ is the text embedding extracted from the image using OCR. We use a Vision-Language Model (VLM) as the policy function $\pi_\theta$. At each decision step, the VLM receives a fixed prompt along with the current environment state $s$, and outputs a structured textual response containing an action $a$ and an intent description $i$, denoted as: $(a, i) = \pi_\theta(s)$.
The action $a$ is represented in the form of a function call as text. After executing $a$, the environment transitions from the current state $s$ to a new state $s'$. Meanwhile, we introduce a world model to simulate the state transition: given the current state $s$ and action $a$, it predicts the next state $\hat{s}$.

Based on policy design, we use a 5-tuple $(s, a, i, s', \hat{s})$ to compute reward $r$ corresponding to the action. Each decision step therefore generates a 6-tuple: $(s, a, i, s', \hat{s}, r)$, which is stored in the rollout buffer $\mathcal{B}$.
Each episode starts from a fixed initial state $s_0$ and proceeds for $T$ decision steps to reach a terminal state $s_T$. In a GUI environment, $s_0$ corresponds to the initial desktop state and $s_T$ represents the final screen state.
If $N$ environments run in parallel, a total of $N \times T$ 6-tuples are collected into the buffer. The data in the buffer is used both to update the policy function $\pi_\theta$ (e.g., via the GRPO algorithm), and to train the world model for better state prediction.
The complete MDP, along with the policy network update and world model training procedures, is provided in Appendix \ref{appendix:optim-step}.

\subsection{Reward Function}\label{section:reward}

To encourage the agent's effective interaction with the environment while promoting exploration of unseen states, we design a set of reward functions. 
Effective interaction requires the model to output correct actions that induce state changes in the environment. In GUI environments, this means the Vision-Language Model (VLM) must output actions in the proper format, and the execution of the action should result in substantial changes to the screen's visual or textual content. Specifically, our designed rewards include the following categories:

\begin{itemize}[leftmargin=*]
    \item \textbf{Format Reward ($r_{format}$):\ \ }
    At each action step, the environment verifies whether the model's output is in the correct format.
    \begin{itemize}[leftmargin=*]
        \item[i)] If the format is correct, the reward $r_{format} \leftarrow 1$.
        \item[ii)] If the format is incorrect, the reward $r_{format} \leftarrow 0$, and the action $a$ is set to a null action.
    \end{itemize}

    \item \textbf{Exploration Reward:\ \ }
    To encourage the agent to explore novel environment states, we design the exploration reward. We use cosine similarity, denoted as $sim(\cdot, \cdot)$, to measure the similarity between states (e.g., visual and textual representations). Lower similarity between states indicates higher exploration diversity. For action $a_t$, the exploration reward is composed of the following terms:
    \begin{itemize}[leftmargin=*]
        \item[i)] \textit{Instantaneous Change Reward ($r^{vis}_{inst}$ \& $r^{text}_{inst}$):\ \ } Measures the degree of state change (e.g., visual and textual representations) immediately before and after the action:
        $r^{vis}_{inst}(t) := 1 - sim(o_t, o'_t)$,
        $r^{text}_{inst}(t) := 1 - sim(e_t, e'_t)$. 
        Where $o_t, e_t$ are the visual and textual representations before the action, and $o'_t, e'_t$ are those after the action.

        \item[ii)] \textit{Subsequent Change Reward ($r^{vis}_{seq}$ \& $r^{text}_{seq}$  ):\ \ } To measure the diversity gain brought by action $a_t$ to the overall state sequence, we evaluate the average similarity between all state pairs before and after the action occurs. The difference from this average similarity is attributed to the action's subsequent change reward, encouraging the agent to take actions that lead to novel long-term state sequences:
        $r^{vis}_{seq}(t) := E_{i<t, j>t}[1 - sim(o'_i, o'_j)]$,
        $r^{text}_{seq}(t) := E_{i<t, j>t}[1 - sim(e'_i, e'_j)]$. 
        Where $E_{i<t, j>t}[\cdot]$ denotes the expectation of similarity calculated over all state pairs with $i < t$ and $j > t$.

        \item[iii)] \textit{World Model Curiosity Reward ($r^{vis}_{world}$ \& $r^{text}_{world}$):\ \ } Captures the agent's "surprise" by comparing the actual post-action states with the states predicted by a world model, thereby encouraging the agent to explore uncertain or difficult-to-predict areas of the environment:
        $r^{vis}_{world}(t) := 1 - sim(o'_t, \hat{o}_t)$,
        $r^{text}_{world}(t) := 1 - sim(e'_t, \hat{e}_t)$. 
        Where $\hat{o}_t, \hat{e}_t$ are the visual and textual states predicted by the world model after the action.
    \end{itemize}
    \item \textbf{Intent-State Alignment Rewards:\ \ }
    To encourage the model to observe the environment more carefully and generate intents (or explanations) related to its actions or the environment state, we design two intent-state alignment rewards. These rewards are computed by comparing the text embeddings of the agent's output intent string $i_t$ with text embeddings obtained from screen OCR:
    \begin{itemize}[leftmargin=*]
        \item[i)] \textit{Environment Description Reward ($r_{des}$):\ \ } Encourages the agent's intent string $i_t$ to be related to the overall environment content by calculating similarity with pre- and post-action text $e_t, e'_t$:
        $r_{des}(t) := sim(i_t, e_t) + sim(i_t, e'_t)$.
        \item[ii)] \textit{Intent Interpretation Reward ($r_{inter}$):\ \ } Encourages the agent's intent string $i_t$ to be related to the specific UI element located at the action coordinates. We extract the text $e_{box}$ from the OCR box where the action coordinates are located and compute its similarity with the intent text:
        $r_{inter}(t) := sim(i_t, e_{box})$. 
        Where $e_{box}$ denotes the text embedding generated from the text within the OCR box containing the action coordinates.
    \end{itemize}
\end{itemize}

\paragraph{Overall Reward:}
We combine the above reward terms to construct the final reward function as:
\begin{equation}
r := r_{format} \times (r^{vis}_{inst} + r^{text}_{inst} + r^{vis}_{seq} + r^{text}_{seq} + r^{vis}_{world} + r^{text}_{world} + r_{des} + r_{inter}). \label{eq:overall_reward}
\end{equation}\label{equ:overall-reward}
Here, $r_{format}, r^{vis}_{inst}, \dots$ are used as shorthand for $r_{format}(t), r^{vis}_{inst}(t), \dots$ to simplify notation. This formula indicates that the sum of other reward terms is included in the total reward only when the action format is correct ($r_{format}=1$); otherwise ($r_{format}=0$), the total reward is zero.


\subsection{Learning State Transitions with a World Model}\label{section:worldmodel}

The world model $\mathcal{M}_\phi: \mathcal{S} \times \mathcal{A} \rightarrow \mathcal{S}$ predicts the next state given the current state and action: $ \hat{s} \leftarrow \mathcal{M}_{\phi}(s, a)$.
The action's word tokens, image's visual tokens and text embedding $e$ are concatenated to form the input sequence for the world model, illustrate as, $\hat{s} := (\hat{o}, \hat{e}) = \mathcal{M}_{\phi}(o, e, a)$.
The world model parameters $\phi$ are optimized to minimize the reconstruction loss between predicted and actual next states as follows. More details of the world model can be found in Appendix \ref{appendix:worldmodel}.
\begin{align}
    \mathcal{L}_{\text{WM}}(\phi) &= \mathbb{E}_{(s, a, s') \sim \mathcal{B}} \left[ \| \mathcal{M}_\phi(s, a) - s' \|^2 \right] \\ \nonumber
     &= \mathbb{E}_{(o, e, a, o', e') \sim \mathcal{B}}  \left[ \| \hat{o} - o' \|^2 +  \| \hat{e} - e' \|^2\right].
\end{align}

\subsection{Training VLM Agent with GRPO}\label{section:grpo}
We optimize VLM Agent policy by Group Relative Policy Optimization (GRPO)\cite{shao2024deepseekmathpushinglimitsmathematical}. All samples in the buffer are treated as a GRPO group and the advantage of each action $A_i$ is estimated as: $A_i=(r_i-\mathrm{mean}(\{r_j\}_{j=1}^{|\mathcal B|}))/\mathrm{std}(\{r_j\}_{j=1}^{|\mathcal B|})$.

GRPO optimizes the policy model by maximizing the following objective function:
\begin{equation} \begin{aligned}
& \mathcal{J}_{GRPO}(\theta) =\mathbb{E}\left[(o,a,r) \sim \mathcal{B}, \left\{a_i\right\}_{i=1}^{|\mathcal{B}|} \sim \pi_{\theta_{\text {old}}}(a\mid o)\right] \\
& \frac{1}{|\mathcal{B}|} \sum_{i=1}^\mathcal{B} \left\{ \min \left[\frac{\pi_\theta\left(a_i \mid o_i\right)}{\pi_{\theta_{\text {old }}}\left(a_i \mid o_i\right)} A_i, \operatorname{clip}\left(\frac{\pi_\theta\left(a_i \mid o_i\right)}{\pi_{\theta_{\text {old}}}\left(a_i \mid o_i\right)}, 1-\varepsilon_{low}, 1+\varepsilon_{high} \right) A_i\right] - \beta \mathbb{D}_{KL}\left(\pi_\theta \| \pi_{r e f}\right)\right\},
\end{aligned} \end{equation}

where $\epsilon_{low}$ and $\epsilon_{high}$ controls the clip range; $\mathbb{D}_{KL}\left(\pi_\theta \| \pi_{ref}\right)$ measures the KL divergence between the current policy $\pi_\theta$ and a reference policy $\pi_{ref}$. $\beta$ weights the KL divergence regularization term, which prevents the optimization process from deviating too far from the original policy:
\begin{equation}
    \mathbb{D}_{KL}\left(\pi_\theta \| \pi_{ref}\right)=\frac{\pi_{ref}\left(a_i \mid o_i\right)}{\pi_\theta\left(a_i \mid o_i\right)}-\log \frac{\pi_{ref}\left(a_i \mid o_i\right)}{\pi_\theta\left(a_i \mid o_i\right)}-1.
\end{equation}

\subsection{Experience Stream Distillation}\label{sec:distillation}
In the RL phase, starting from the base model, the agent learns to interact with the environment, explores it diversely, and collects trajectories to form a data set of experience stream. Next, these streams are passed through a quality and diversity filter, which selects the most informative and diverse trajectories. 
In the distillation phase, we reinitialize from the base model and perform supervised fine-tuning (SFT) using the filtered data to distill a new model. Through supervised fine-tuning, the experience gained from RL training can be efficiently transferred to the distilled model, avoiding the need to learn from scratch.
Our training approach, which combines RL and distillation, gradually enhances the capacity of VLM agents to perform meaningful, diverse actions in complex GUI environments. More details of the distillation process are provided in Appendix \ref{appendix:distillation}.

\section{Experiment}

\subsection{Evaluation Metrics on Exploration Diversity}
We evaluate the performance of agents' diverse exploration in GUI environments by measuring both the diversity of environmental states within individual trajectories and the diversity of states across all trajectories within a group. We use the cosine similarity $sim(\cdot,\cdot)$ to measure the similarity between environmental states, as defined in subsection \ref{section:reward}. 

\paragraph{Trajectory-level Diversity:} 
For a trajectory $\tau$ consists of $T$ states $\{s_1, \dots, s_T\}$, the corresponding visual representations are $\{o_1, \dots, o_T\}$, and textual representations are $\{e_1, \dots, e_T\}$, define the visual and textual sequence diversity of trajectory $\tau$ as:
\begin{equation}
  d_{\mathrm{seq}}^{\mathrm{vis}}(\tau) 
  = \frac{1}{T(T-1)}
    \sum_{1\le k<l\le T} \bigl[1 - sim(o'_k, o'_l)\bigr];
  d_{\mathrm{seq}}^{\mathrm{text}}(\tau)
  = \frac{1}{T(T-1)}
    \sum_{1\le k<l\le T} \bigl[1 - sim(e'_k, e'_l)\bigr].
\end{equation}

\paragraph{Group-level Diversity:}
For a group of trajectories $\mathcal{G} = \{\tau^1,\dots,\tau^M\}$, each trajectory $\tau$ consists of $T$ states. 
We flatten all $M$ trajectories into a set of $N=M\times T$ different states and denote their visual and textual embeddings by $\{(o_k,e_k)\}_{k=1}^N$, define the visual and textual diversity of group $\mathcal{G}$ as:
\begin{equation}
  D_{\mathrm{grp}}^{\mathrm{vis}}(\mathcal{G})
  = \frac{1}{N(N-1)}
    \sum_{1\le k<l\le N}
      \bigl[\,1 - sim(o'_k, o'_l)\bigr];
  D_{\mathrm{grp}}^{\mathrm{text}}(\mathcal{G})
  = \frac{1}{N(N-1)}
    \sum_{1\le k<l\le N}
      \bigl[\,1 - sim(e'_k, e'_l)\bigr].
\end{equation}

\subsection{Exploration Performance} 

We adopt \textit{Qwen2.5-VL-3B} as the backbone of \textit{ScreenExplorer-3B} and apply a distillation procedure as illustrated in Section \ref{sec:distillation}. All hyperparameters and additional training specifics are outlined in Appendix \ref{appendix:optim-step}. The baselines are categorized into two groups: \textit{General Used Models} and \textit{GUI-specific models}, which are accessible via API or static deployment.
Our model includes two variants: \textit{ScreenExplorer-3B-E1} denotes the best-performing checkpoint from the RL phase, \textit{ScreenExplorer-3B-Distill} refers to the model after experience stream distillation. Our variant model is statically deployed under identical experimental conditions as baselines, with trajectory sampling replicated for comparative analysis. Each experimental condition is held constant: we sampled 20 episodes per model, each episode comprising 10 steps, yielding 200 frames for diversity computation. Except for OpenAI Computer Use, all other models are tested under two sampling temperatures, $t=1.0$ and $t=0.5$. Table 1 summarizes the overall exploration performance and the findings are:

\newcommand{\paleredbox}[1]{\colorbox{red!10}{#1}}
\begin{table}[t]
  \centering
  \small
  \caption{Overall Performance in Correct Formatting and Exploration. \paleredbox{box} indicates the lowest scores, while \textbf{bold} text denotes the highest scores.}
  \label{tab:overall_performance}
  \resizebox{\textwidth}{!}{
  \begin{tabular}{@{} l l c c c c c c @{}}
    \toprule
      \multirow{2}{*}{\textbf{Model}}
      & \multirow{2}{*}{\textbf{Setting}}
      & \multirow{2}{*}{\textbf{\makecell{Correct \\ Format}}}
      & \multicolumn{2}{c}{\textbf{Trajectory-level}}
      & 
      \multicolumn{2}{c}{\textbf{Group-level}}
      & 
      \multirow{2}{*}{\textbf{\makecell{Avg. \\ Diversity}}} \\
      \cmidrule(lr){4-5} \cmidrule(lr){6-7}
      & & &
      $d_{\mathrm{seq}}^{\mathrm{vis}}(\tau)$ 
      & $d_{\mathrm{seq}}^{\mathrm{text}}(\tau)$ 
      & $D_{\mathrm{grp}}^{\mathrm{vis}}(\mathcal{G})$ 
      & $D_{\mathrm{grp}}^{\mathrm{text}}(\mathcal{G})$ 
      & \\
    \midrule
    \rowcolor{gray!15}
    \multicolumn{8}{l}{\textit{General Used Models}} \\
    \multirow{2}{*}{OpenAI gpt-4o}
       & $t=1.0$ & 0.95 & 0.25 & 0.16 & 0.35 & 0.25 & 0.25 \\
       & $t=0.5$ & \textbf{1.00} & \paleredbox{0.13} & 0.17 & \paleredbox{0.18} & 0.26 & 0.18 \\
    \midrule
    \multirow{2}{*}{Qwen2.5-VL-72B}
      & $t=1.0$ & 0.96 & 0.39  & 0.26 & 0.69 & 0.39 & 0.43 \\
      & $t=0.5$ & \textbf{1.00} & 0.23  & 0.16 & 0.39 & 0.22 & 0.25 \\
    \midrule
    \multirow{2}{*}{Qwen2.5-VL-7B}
      & $t=1.0$ & 0.68 & 0.44 & 0.26  & 0.57 & 0.37 & 0.41  \\
      & $t=0.5$ & 0.75 & 0.37 & 0.21 & 0.34 & 0.29 & 0.32 \\
    \midrule
    \multirow{2}{*}{Qwen2.5-VL-3B}
      & $t=1.0$ & \paleredbox{0.62} & 0.16 & 0.10 & 0.40 & 0.19 & 0.21 \\
      & $t=0.5$ & 0.84 & 0.15 & \paleredbox{0.08} & 0.31 & \paleredbox{0.14} & \paleredbox{0.17} \\
    \midrule
    \rowcolor{gray!15}
    \multicolumn{8}{l}{\textit{GUI-specific Models}} \\
     OpenAI Computer Use
        & default & 0.95 & 0.28  & 0.21 & 0.60 & 0.32 & 0.35 \\
    \midrule
      \multirow{2}{*}{doubao-1.5-ui-tars}
        &  $t=1.0$ & 0.82 & 0.41 & 0.24 & \textbf{0.76} & 0.38 & 0.45 \\
        &  $t=0.5$ & 0.75 & 0.30 & 0.17  & 0.64 & 0.32 & 0.36 \\
    \midrule
    \rowcolor{gray!15}
    \multicolumn{8}{l}{\textit{Ours}} \\
     \multirow{2}{*}{ScreenExplorer-3B-E1}
        & $t=1.0$ & 0.99 & 0.57  & 0.33 & 0.68 & 0.45 & 0.51 \\
        & $t=0.5$ & \textbf{1.00} & 0.57  & 0.33 & 0.72 & \textbf{0.46} & 0.52\\
     \midrule
     \multirow{2}{*}{ScreenExplorer-3B-Distill}
        & $t=1.0$ & 0.93 & 0.64 & 0.37 & 0.68 & 0.43 & 0.53 \\
        & $t=0.5$ & 0.99 & \textbf{0.66} & \textbf{0.41} & 0.67 & 0.44 & \textbf{0.55}  \\
    \bottomrule
  \end{tabular}
  }
\end{table}

\begin{itemize}[leftmargin=*]
    \item Our model excelled in all exploration diversity metrics, showing superior adaptation to GUI environment and ability to find diverse paths. Among General Used Models, OpenAI gpt-4o excels at scene recognition but struggles with coordinate localization, leading to ineffective actions. Static deployment prevents strategy adaptation, causing repeated failed attempts. In the Qwen2.5-VL series, 72B version achieves effective interaction and task completion, while 3B performs poorly. Within the Qwen2.5-VL series, Qwen2.5-VL-72B performs best, achieving competitive performance in sequential tasks, while Qwen2.5-VL-3B shows the poorest performance.
    \item In GUI-specific Models, OpenAI Computer Use and doubao-1.5-ui-tars achieve precise UI element localization and responsive operations, resulting in superior sequence- and group-level diversity. Through RL and experience stream distillation of Qwen2.5-VL-3B, we improve average diversity scores from 0.21/0.17 to 0.53/0.54 (at t=1.0/0.5). This showcases both the potential of smaller models and the effectiveness of real-world RL for environmental adaptation.
    \item We observed that sampling temperature has a significant impact on instruction-following and exploration diversity: for the majority of models, higher temperatures ($t=1.0$) encourage the generation of more diverse actions, leading to richer trajectories. However, higher temperatures tend to compromise instruction-following capabilities, as evidenced by the decline in correct format.
    \item Our model is capable of generating function calls in the correct format. Among general used models, gpt-4o and \textit{Qwen2.5-VL-72B} generated correct function calls effectively, while \textit{Qwen2.5-VL-3B} underperformed due to its size limitations. After our training method, \textit{ScreenExplorer-3B}'s correct format metric improved from 0.62 to 0.99 at $t = 1.0$, and from 0.84 to 1.00 at $t = 0.5$.
\end{itemize}

\begin{figure}
    \centering
    \includegraphics[width=1\linewidth]{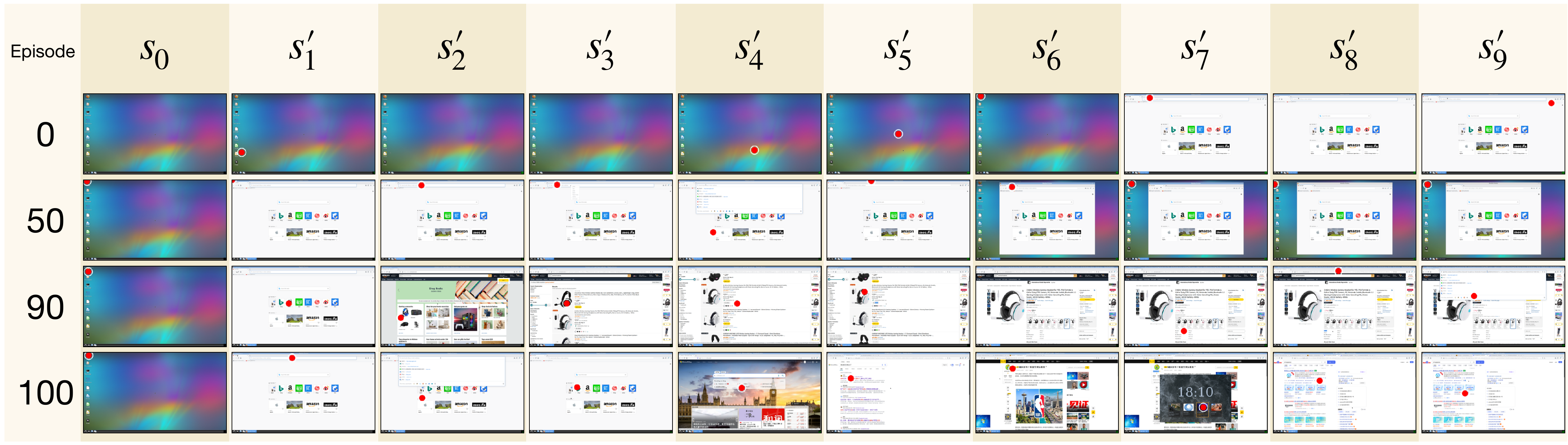}
    \caption{Examples of Trajectories from \textit{ScreenExplorer-3B-E1}. Through RL training, the model developed increasingly effective interactions with the environment, enabling exploration of deeper pages.}
    \label{fig:case_e1}
\end{figure}

We trained the \textit{ScreenExplorer-3B-Distill} version through distillation with a small amount of experience stream data and achieved optimal performance in both trajectory-level visual and textual diversity metrics, as well as the highest average diversity scores. This demonstrates that fine-tuning with a limited amount of experience stream data enables efficient transfer of exploration knowledge from the previous generation to subsequent iterations. Figure \ref{fig:case_e1} shows training trajectories that demonstrate how the model developed effective environmental interactions and explored deeper pages through RL. More details regarding prompts and other  settings are provided in Appendix \ref{appendix:baseline-settings}.

\subsection{RL training Process}
\begin{figure}
    \centering
    \includegraphics[width=1\linewidth]{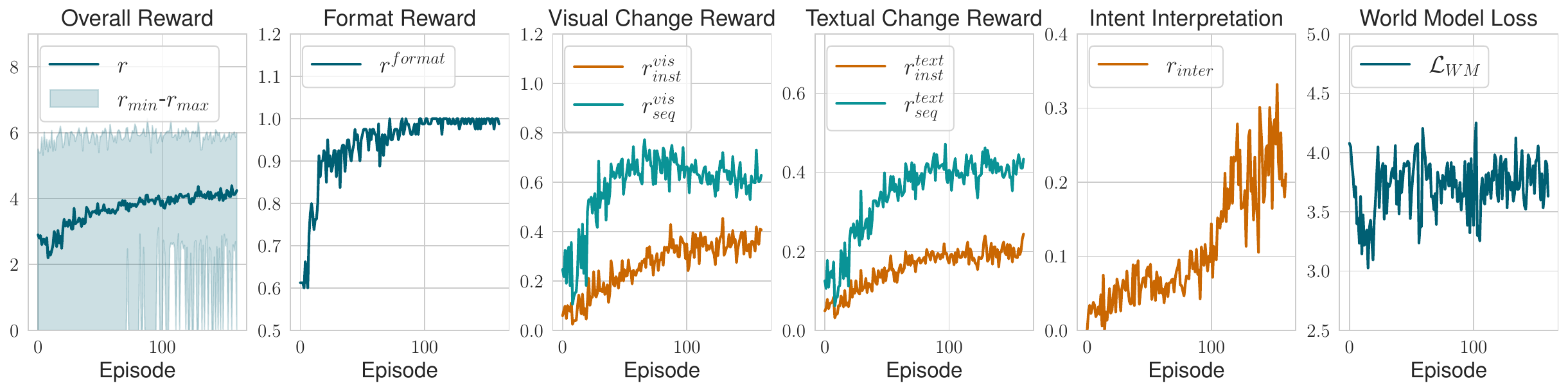}
    \caption{Indicators of \textit{ScreenExplorer-3B-E1} in training. During RL training, the VLM actor's increasing rewards show improved environment interaction and state space exploration. The world model loss demonstrates sustained curiosity that drives further exploration.}
    \label{fig:all_rewards}
\end{figure}

To better understand the progression of model's different capabilities, we analyze the key metrics in RL training process of \textit{ScreenExplorer-3B-E1}. As illustrated in Figure \ref{fig:all_rewards}, all reward values increase as training progresses, leading to a gradual improvement in Overall Reward. 
In the initial training phase, the Format Reward shows rapid improvement first. According to Equation \ref{equ:overall-reward}, format errors incur the heaviest penalties, prompting the agent's model to prioritize optimizing its output format. Subsequently, four exploration diversity rewards begin to rise, indicating that the model gradually learns to interact effectively with the environment and explore deeper into pages. After about 100 steps, visual and textual exploration diversity rewards gradually saturate, leading the model to enhance its intent outputs to achieve higher intent interpretation rewards.
Throughout the training process, the world model loss $\mathcal{L}_{WM}$ first slightly decreases, then oscillates at a high value, indicating that the world model maintains a consistently high level of curiosity.

\subsection{Ablation Study on Reward Components}

\begin{figure}[t]
    \centering
    \includegraphics[width=1\linewidth]{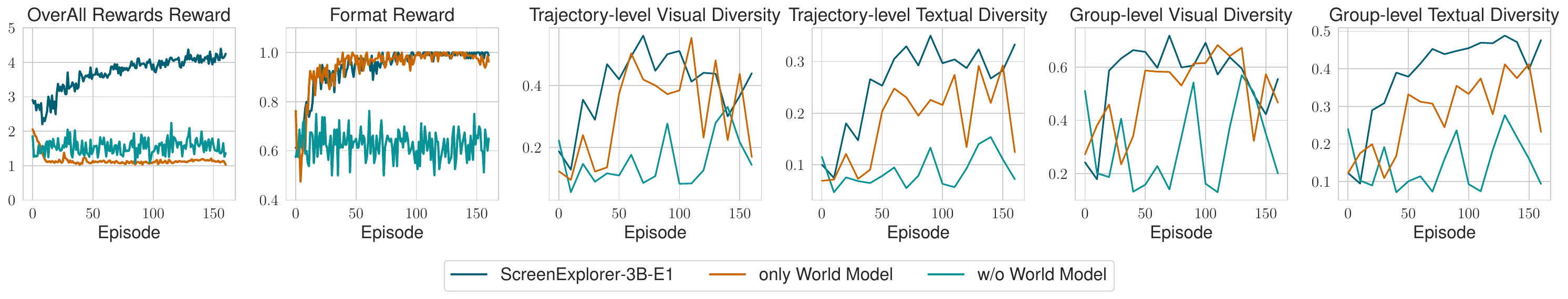}
    \caption{Rewards and metrics of exploration diversity in world model ablation study.}
    \label{fig:world_model_ablation}
\end{figure}

In this section, we conduct ablation studies on various components of the reward function. To evaluate the effectiveness of the world model, we designed two comparative experiments—"w/o world model" and "only world model"—and compared them with our complete experiment \textit{ScreenExplorer-3B-E1}. Figure \ref{fig:world_model_ablation} illustrates the progression of overall reward and format reward, along with four diversity metrics. In the "w/o world model" experiment where the world model is removed, the policy network struggles to improve format rewards and establish effective interaction with the environment. In contrast, both experiments incorporating the world model successfully overcome the cold start phase, as evidenced by continuous improvements in both reward and diversity metrics. For additional ablation studies on reward components, please refer to Appendix \ref{appendix:reward-ablation}.

\begin{figure}[t]
    \centering
    \includegraphics[width=0.9\linewidth]{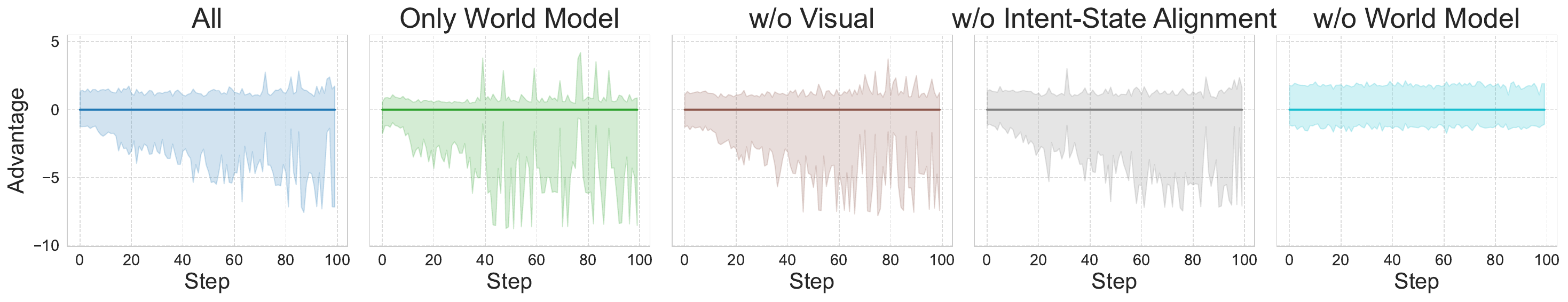}
    \caption{Advantage Range. By introducing the world model reward increases the variance of GRPO advantage, enabling smoother cold-start optimization and more effective exploration through better gradient differentiation across sample groups. }\label{fig:advantage_range}
    \label{fig:advantage}
\end{figure}

\paragraph{Why World Model?}
To study the role of world model rewards, we analyze advantage values from ablation experiments. The results are illustrated in Figure \ref{fig:advantage}. Adding curiosity rewards from the world model increased advantage variance, which smooths the score gradients within a group of samples, enabling the RL process to more rapidly identify optimization directions — this is important in the cold start phase of exploration. Similar observations were also noted in \cite{razin2025makesrewardmodelgood}, which argues that an effective reward function not only needs to be accurate but also requires sufficient variance. 

\section{Discussion}\label{sec:discussion}

\textbf{Findings.} \textit{Capability Compositionality.} Small-scale VLMs possess basic capabilities like OCR, computer knowledge, and visual detection/localization, but struggle to combine these effectively.  In Appendix \ref{appendix:llms-capability}, our cases show these models often fail to coordinate their skills, with conflicts arising between language generation and visual processing abilities. Through training in GUI environment, models can combine these inherent capabilities into higher-level exploration skills; \textit{Noisy TV Problem and Exploration Dilemma.}

Curiosity-driven exploration in RL faces the "Noisy TV problem."\cite{burda2018explorationrandomnetworkdistillation,mavorparker2024staycuriousavoidingnoisy} We find that agents get stuck on irrelevant, perpetually novel stimuli (e.g., clicking the first news item or opening video sites). We found long period of training LLMs with RL can reduce exploration, as they tend to repeat early high-reward responses, leading to exploration traps. To improve exploration diversity and prevent stagnation, experience stream distillation was implemented.

\textbf{Limitations.}
We mainly use screenshots and text content to measure environmental state similarity. However, structured state information from operating system, such as process running status and variables, are also suitable choices. But, achieving this requires a considerable amount of systems engineering work. Although we introduced vLLM \cite{kwon2023efficient} to accelerate VLM decoding, the time cost of sampling and training the VLM agent in a real GUI environment is prohibitively high, which limits our ability to conduct larger-scale experiments. We plan to implement distributed sampling and training methods in the future to speed up the learning process.

\section{Conclusion}

In this paper, we trained a Qwen2.5-VL-3B-based agent for exploration in a open-world GUI environment. Using exploration rewards, a world model, and GRPO reinforcement learning, we improved the agent's GUI interaction capabilities. 
And distilling experience streams can further enhance the model's exploration capabilities.
Our approach unlocks the potential of a 3B-parameter model, enabling it to adapt to our GUI environment more effectively than even larger models, highlighting the importance of real-time learning in ever-evolving open-world settings.
By leveraging robust exploration to harvest experience directly from its environment, the agent can steadily wean itself off human-annotated data—charting a clear course toward AGI.

\bibliographystyle{unsrt}
\bibliography{ref}

\begin{thebibliography}{10}

\bibitem{bubeck2023sparksartificialgeneralintelligence}
Sébastien Bubeck, Varun Chandrasekaran, Ronen Eldan, Johannes Gehrke, Eric Horvitz, Ece Kamar, Peter Lee, Yin~Tat Lee, Yuanzhi Li, Scott Lundberg, Harsha Nori, Hamid Palangi, Marco~Tulio Ribeiro, and Yi~Zhang.
\newblock Sparks of artificial general intelligence: Early experiments with gpt-4, 2023.

\bibitem{zhang2025largelanguagemodelbrainedgui}
Chaoyun Zhang, Shilin He, Jiaxu Qian, Bowen Li, Liqun Li, Si~Qin, Yu~Kang, Minghua Ma, Guyue Liu, Qingwei Lin, Saravan Rajmohan, Dongmei Zhang, and Qi~Zhang.
\newblock Large language model-brained gui agents: A survey, 2025.

\bibitem{tan2024cradle}
Weihao Tan, Wentao Zhang, Xinrun Xu, Haochong Xia, Ziluo Ding, Boyu Li, Bohan Zhou, Junpeng Yue, Jiechuan Jiang, Yewen Li, Ruyi An, Molei Qin, Chuqiao Zong, Longtao Zheng, Yujie Wu, Xiaoqiang Chai, Yifei Bi, Tianbao Xie, Pengjie Gu, Xiyun Li, Ceyao Zhang, Long Tian, Chaojie Wang, Xinrun Wang, Börje~F. Karlsson, Bo~An, Shuicheng Yan, and Zongqing Lu.
\newblock Cradle: Empowering foundation agents towards general computer control.
\newblock {\em arXiv preprint arXiv:2403.03186}, 2024.

\bibitem{zhang2024ufo}
Chaoyun Zhang, Liqun Li, Shilin He, Xu~Zhang, Bo~Qiao, Si~Qin, Minghua Ma, Yu~Kang, Qingwei Lin, Saravan Rajmohan, Dongmei Zhang, and Qi~Zhang.
\newblock {UFO: A UI-Focused Agent for Windows OS Interaction}.
\newblock {\em arXiv preprint arXiv:2402.07939}, 2024.

\bibitem{zhang2025ufo2}
Chaoyun Zhang, He~Huang, Chiming Ni, Jian Mu, Si~Qin, Shilin He, Lu~Wang, Fangkai Yang, Pu~Zhao, Chao Du, Liqun Li, Yu~Kang, Zhao Jiang, Suzhen Zheng, Rujia Wang, Jiaxu Qian, Minghua Ma, Jian-Guang Lou, Qingwei Lin, Saravan Rajmohan, and Dongmei Zhang.
\newblock {UFO2: The Desktop AgentOS}.
\newblock {\em arXiv preprint arXiv:2504.14603}, 2025.

\bibitem{wu2024oscopilot}
Zhiyong Wu, Chengcheng Han, Zichen Ding, Zhenmin Weng, Zhoumianze Liu, Shunyu Yao, Tao Yu, and Lingpeng Kong.
\newblock Os-copilot: Towards generalist computer agents with self-improvement, 2024.

\bibitem{Agent-S}
Saaket Agashe, Jiuzhou Han, Shuyu Gan, Jiachen Yang, Ang Li, and Xin~Eric Wang.
\newblock {Agent S: An Open Agentic Framework that Uses Computers Like a Human}.
\newblock In {\em International Conference on Learning Representations (ICLR)}, 2025.

\bibitem{Agent-S2}
Saaket Agashe, Kyle Wong, Vincent Tu, Jiachen Yang, Ang Li, and Xin~Eric Wang.
\newblock Agent s2: A compositional generalist-specialist framework for computer use agents, 2025.

\bibitem{hong2024cogagentvisuallanguagemodel}
Wenyi Hong, Weihan Wang, Qingsong Lv, Jiazheng Xu, Wenmeng Yu, Junhui Ji, Yan Wang, Zihan Wang, Yuxuan Zhang, Juanzi Li, Bin Xu, Yuxiao Dong, Ming Ding, and Jie Tang.
\newblock Cogagent: A visual language model for gui agents, 2024.

\bibitem{Rico2017Deka}
Biplab Deka, Zifeng Huang, Chad Franzen, Joshua Hibschman, Daniel Afergan, Yang Li, Jeffrey Nichols, and Ranjitha Kumar.
\newblock Rico: A mobile app dataset for building data-driven design applications.
\newblock In {\em Proceedings of the 30th Annual ACM Symposium on User Interface Software and Technology}, UIST '17, page 845–854, New York, NY, USA, 2017. Association for Computing Machinery.

\bibitem{zhang2021screenrecognitioncreatingaccessibility}
Xiaoyi Zhang, Lilian de~Greef, Amanda Swearngin, Samuel White, Kyle Murray, Lisa Yu, Qi~Shan, Jeffrey Nichols, Jason Wu, Chris Fleizach, Aaron Everitt, and Jeffrey~P. Bigham.
\newblock Screen recognition: Creating accessibility metadata for mobile applications from pixels, 2021.

\bibitem{deng2023mind2web}
Xiang Deng, Yu~Gu, Boyuan Zheng, Shijie Chen, Samuel Stevens, Boshi Wang, Huan Sun, and Yu~Su.
\newblock Mind2web: Towards a generalist agent for the web, 2023.

\bibitem{wu2024mobilevlmvisionlanguagemodelbetter}
Qinzhuo Wu, Weikai Xu, Wei Liu, Tao Tan, Jianfeng Liu, Ang Li, Jian Luan, Bin Wang, and Shuo Shang.
\newblock Mobilevlm: A vision-language model for better intra- and inter-ui understanding, 2024.

\bibitem{ScreenAgent2024niu}
Runliang Niu, Jindong Li, Shiqi Wang, Yali Fu, Xiyu Hu, Xueyuan Leng, He~Kong, Yi~Chang, and Qi~Wang.
\newblock Screenagent: A vision language model-driven computer control agent.
\newblock In Kate Larson, editor, {\em Proceedings of the Thirty-Third International Joint Conference on Artificial Intelligence, {IJCAI-24}}, pages 6433--6441. International Joint Conferences on Artificial Intelligence Organization, 8 2024.
\newblock Main Track.

\bibitem{GUICourse2024Wentong}
Wentong Chen, Junbo Cui, Jinyi Hu, Yujia Qin, Junjie Fang, Yue Zhao, Chongyi Wang, Jun Liu, Guirong Chen, Yupeng Huo, Yuan Yao, Yankai Lin, Zhiyuan Liu, and Maosong Sun.
\newblock Guicourse: From general vision language models to versatile gui agents, 2024.

\bibitem{deepseekai2025deepseekr1incentivizingreasoningcapability}
DeepSeek-AI, Daya Guo, Dejian Yang, Haowei Zhang, Junxiao Song, Ruoyu Zhang, Runxin Xu, Qihao Zhu, Shirong Ma, Peiyi Wang, Xiao Bi, Xiaokang Zhang, Xingkai Yu, Yu~Wu, Z.~F. Wu, Zhibin Gou, Zhihong Shao, Zhuoshu Li, Ziyi Gao, Aixin Liu, Bing Xue, Bingxuan Wang, Bochao Wu, Bei Feng, Chengda Lu, Chenggang Zhao, Chengqi Deng, Chenyu Zhang, Chong Ruan, Damai Dai, Deli Chen, Dongjie Ji, Erhang Li, Fangyun Lin, Fucong Dai, Fuli Luo, Guangbo Hao, Guanting Chen, Guowei Li, H.~Zhang, Han Bao, Hanwei Xu, Haocheng Wang, Honghui Ding, Huajian Xin, Huazuo Gao, Hui Qu, Hui Li, Jianzhong Guo, Jiashi Li, Jiawei Wang, Jingchang Chen, Jingyang Yuan, Junjie Qiu, Junlong Li, J.~L. Cai, Jiaqi Ni, Jian Liang, Jin Chen, Kai Dong, Kai Hu, Kaige Gao, Kang Guan, Kexin Huang, Kuai Yu, Lean Wang, Lecong Zhang, Liang Zhao, Litong Wang, Liyue Zhang, Lei Xu, Leyi Xia, Mingchuan Zhang, Minghua Zhang, Minghui Tang, Meng Li, Miaojun Wang, Mingming Li, Ning Tian, Panpan Huang, Peng Zhang, Qiancheng Wang, Qinyu Chen, Qiushi Du, Ruiqi Ge, Ruisong
  Zhang, Ruizhe Pan, Runji Wang, R.~J. Chen, R.~L. Jin, Ruyi Chen, Shanghao Lu, Shangyan Zhou, Shanhuang Chen, Shengfeng Ye, Shiyu Wang, Shuiping Yu, Shunfeng Zhou, Shuting Pan, S.~S. Li, Shuang Zhou, Shaoqing Wu, Shengfeng Ye, Tao Yun, Tian Pei, Tianyu Sun, T.~Wang, Wangding Zeng, Wanjia Zhao, Wen Liu, Wenfeng Liang, Wenjun Gao, Wenqin Yu, Wentao Zhang, W.~L. Xiao, Wei An, Xiaodong Liu, Xiaohan Wang, Xiaokang Chen, Xiaotao Nie, Xin Cheng, Xin Liu, Xin Xie, Xingchao Liu, Xinyu Yang, Xinyuan Li, Xuecheng Su, Xuheng Lin, X.~Q. Li, Xiangyue Jin, Xiaojin Shen, Xiaosha Chen, Xiaowen Sun, Xiaoxiang Wang, Xinnan Song, Xinyi Zhou, Xianzu Wang, Xinxia Shan, Y.~K. Li, Y.~Q. Wang, Y.~X. Wei, Yang Zhang, Yanhong Xu, Yao Li, Yao Zhao, Yaofeng Sun, Yaohui Wang, Yi~Yu, Yichao Zhang, Yifan Shi, Yiliang Xiong, Ying He, Yishi Piao, Yisong Wang, Yixuan Tan, Yiyang Ma, Yiyuan Liu, Yongqiang Guo, Yuan Ou, Yuduan Wang, Yue Gong, Yuheng Zou, Yujia He, Yunfan Xiong, Yuxiang Luo, Yuxiang You, Yuxuan Liu, Yuyang Zhou, Y.~X. Zhu,
  Yanhong Xu, Yanping Huang, Yaohui Li, Yi~Zheng, Yuchen Zhu, Yunxian Ma, Ying Tang, Yukun Zha, Yuting Yan, Z.~Z. Ren, Zehui Ren, Zhangli Sha, Zhe Fu, Zhean Xu, Zhenda Xie, Zhengyan Zhang, Zhewen Hao, Zhicheng Ma, Zhigang Yan, Zhiyu Wu, Zihui Gu, Zijia Zhu, Zijun Liu, Zilin Li, Ziwei Xie, Ziyang Song, Zizheng Pan, Zhen Huang, Zhipeng Xu, Zhongyu Zhang, and Zhen Zhang.
\newblock Deepseek-r1: Incentivizing reasoning capability in llms via reinforcement learning, 2025.

\bibitem{burda2018explorationrandomnetworkdistillation}
Yuri Burda, Harrison Edwards, Amos Storkey, and Oleg Klimov.
\newblock Exploration by random network distillation, 2018.

\bibitem{NovelD2021Zhang}
Tianjun Zhang, Huazhe Xu, Xiaolong Wang, Yi~Wu, Kurt Keutzer, Joseph~E Gonzalez, and Yuandong Tian.
\newblock Noveld: A simple yet effective exploration criterion.
\newblock In M.~Ranzato, A.~Beygelzimer, Y.~Dauphin, P.S. Liang, and J.~Wortman Vaughan, editors, {\em Advances in Neural Information Processing Systems}, volume~34, pages 25217--25230. Curran Associates, Inc., 2021.

\bibitem{eysenbach2018diversityneedlearningskills}
Benjamin Eysenbach, Abhishek Gupta, Julian Ibarz, and Sergey Levine.
\newblock Diversity is all you need: Learning skills without a reward function, 2018.

\bibitem{sharma2020dynamicsawareunsuperviseddiscoveryskills}
Archit Sharma, Shixiang Gu, Sergey Levine, Vikash Kumar, and Karol Hausman.
\newblock Dynamics-aware unsupervised discovery of skills, 2020.

\bibitem{wang2023voyageropenendedembodiedagent}
Guanzhi Wang, Yuqi Xie, Yunfan Jiang, Ajay Mandlekar, Chaowei Xiao, Yuke Zhu, Linxi Fan, and Anima Anandkumar.
\newblock Voyager: An open-ended embodied agent with large language models, 2023.

\bibitem{campero2021learningamigoadversariallymotivated}
Andres Campero, Roberta Raileanu, Heinrich Küttler, Joshua~B. Tenenbaum, Tim Rocktäschel, and Edward Grefenstette.
\newblock Learning with amigo: Adversarially motivated intrinsic goals, 2021.

\bibitem{li2025openworldreinforcementlearninglong}
Jiajian Li, Qi~Wang, Yunbo Wang, Xin Jin, Yang Li, Wenjun Zeng, and Xiaokang Yang.
\newblock Open-world reinforcement learning over long short-term imagination, 2025.

\bibitem{xin2024deepseekproverv15harnessingproofassistant}
Huajian Xin, Z.~Z. Ren, Junxiao Song, Zhihong Shao, Wanjia Zhao, Haocheng Wang, Bo~Liu, Liyue Zhang, Xuan Lu, Qiushi Du, Wenjun Gao, Qihao Zhu, Dejian Yang, Zhibin Gou, Z.~F. Wu, Fuli Luo, and Chong Ruan.
\newblock Deepseek-prover-v1.5: Harnessing proof assistant feedback for reinforcement learning and monte-carlo tree search.
\newblock 2024.

\bibitem{lightman2023letsverifystepstep}
Hunter Lightman, Vineet Kosaraju, Yura Burda, Harri Edwards, Bowen Baker, Teddy Lee, Jan Leike, John Schulman, Ilya Sutskever, and Karl Cobbe.
\newblock Let's verify step by step, 2023.

\bibitem{qi2024mutualreasoningmakessmaller}
Zhenting Qi, Mingyuan Ma, Jiahang Xu, Li~Lyna Zhang, Fan Yang, and Mao Yang.
\newblock Mutual reasoning makes smaller llms stronger problem-solvers, 2024.

\bibitem{pan2025largelanguagemodelsthink}
Lan Pan, Hanbo Xie, and Robert~C. Wilson.
\newblock Large language models think too fast to explore effectively, 2025.

\bibitem{dou2025improvingrlexplorationllm}
Shihan Dou, Muling Wu, Jingwen Xu, Rui Zheng, Tao Gui, Qi~Zhang, and Xuanjing Huang.
\newblock Improving rl exploration for llm reasoning through retrospective replay, 2025.

\bibitem{ragen2025}
Zihan Wang, Kangrui Wang, Qineng Wang, Pingyue Zhang, Linjie Li, Zhengyuan Yang, Xing Jin, Kefan Yu, Minh~Nhat Nguyen, Licheng Liu, Eli Gottlieb, Yiping Lu, Kyunghyun Cho, Jiajun Wu, Li~Fei-Fei, Lijuan Wang, Yejin Choi, and Manling Li.
\newblock Ragen: Understanding self-evolution in llm agents via multi-turn reinforcement learning, 2025.

\bibitem{lin2024showuivisionlanguageactionmodelgui}
Kevin~Qinghong Lin, Linjie Li, Difei Gao, Zhengyuan Yang, Shiwei Wu, Zechen Bai, Weixian Lei, Lijuan Wang, and Mike~Zheng Shou.
\newblock Showui: One vision-language-action model for gui visual agent, 2024.

\bibitem{wu2024foundations}
Biao Wu, Yanda Li, Meng Fang, Zirui Song, Zhiwei Zhang, Yunchao Wei, and Ling Chen.
\newblock Foundations and recent trends in multimodal mobile agents: A survey.
\newblock {\em arXiv preprint arXiv:2411.02006}, 2024.

\bibitem{wang2024mobileagentautonomousmultimodalmobile}
Junyang Wang, Haiyang Xu, Jiabo Ye, Ming Yan, Weizhou Shen, Ji~Zhang, Fei Huang, and Jitao Sang.
\newblock Mobile-agent: Autonomous multi-modal mobile device agent with visual perception, 2024.

\bibitem{tan2024towards}
Weihao Tan, Ziluo Ding, Wentao Zhang, Boyu Li, Bohan Zhou, Junpeng Yue, Haochong Xia, Jiechuan Jiang, Longtao Zheng, Xinrun Xu, Yifei Bi, Pengjie Gu, Xinrun Wang, B{\"o}rje~F. Karlsson, Bo~An, and Zongqing Lu.
\newblock Towards general computer control: A multimodal agent for red dead redemption {II} as a case study.
\newblock In {\em ICLR 2024 Workshop on Large Language Model (LLM) Agents}, 2024.

\bibitem{bai2024digirltraininginthewilddevicecontrol}
Hao Bai, Yifei Zhou, Mert Cemri, Jiayi Pan, Alane Suhr, Sergey Levine, and Aviral Kumar.
\newblock Digirl: Training in-the-wild device-control agents with autonomous reinforcement learning, 2024.

\bibitem{pan2024autonomousevaluationrefinementdigital}
Jiayi Pan, Yichi Zhang, Nicholas Tomlin, Yifei Zhou, Sergey Levine, and Alane Suhr.
\newblock Autonomous evaluation and refinement of digital agents, 2024.

\bibitem{wang2025distrlasynchronousdistributedreinforcement}
Taiyi Wang, Zhihao Wu, Jianheng Liu, Jianye Hao, Jun Wang, and Kun Shao.
\newblock Distrl: An asynchronous distributed reinforcement learning framework for on-device control agents, 2025.

\bibitem{shao2024deepseekmathpushinglimitsmathematical}
Zhihong Shao, Peiyi Wang, Qihao Zhu, Runxin Xu, Junxiao Song, Xiao Bi, Haowei Zhang, Mingchuan Zhang, Y.~K. Li, Y.~Wu, and Daya Guo.
\newblock Deepseekmath: Pushing the limits of mathematical reasoning in open language models, 2024.

\bibitem{razin2025makesrewardmodelgood}
Noam Razin, Zixuan Wang, Hubert Strauss, Stanley Wei, Jason~D. Lee, and Sanjeev Arora.
\newblock What makes a reward model a good teacher? an optimization perspective, 2025.

\bibitem{mavorparker2024staycuriousavoidingnoisy}
Augustine~N. Mavor-Parker, Kimberly~A. Young, Caswell Barry, and Lewis~D. Griffin.
\newblock How to stay curious while avoiding noisy tvs using aleatoric uncertainty estimation, 2024.

\bibitem{kwon2023efficient}
Woosuk Kwon, Zhuohan Li, Siyuan Zhuang, Ying Sheng, Lianmin Zheng, Cody~Hao Yu, Joseph~E. Gonzalez, Hao Zhang, and Ion Stoica.
\newblock Efficient memory management for large language model serving with pagedattention.
\newblock In {\em Proceedings of the ACM SIGOPS 29th Symposium on Operating Systems Principles}, 2023.

\end{thebibliography}

\newpage
\newpage
\appendix


\section{Technical Details of GUI Environment}\label{appendix:environment}

Our GUI environment is constructed on a Linux desktop system and wrapped as a Gymnasium \footnote{\url{https://gymnasium.farama.org/}} environment for reinforcement learning training. The observation space consists of RGB screenshots with configurable resolution. The action space is structured as a dictionary space, supporting a variety of mouse and keyboard interactions, with detailed action types and examples shown in Table \ref{tab:action_space}.

\begin{table}[h]
    \centering
    \caption{GUI Environment Action Space}\label{tab:action_space}
    \scalebox{0.8}{
        \begin{tblr}{
          column{1} = {c},
          cell{1}{1} = {c=2}{},
          cell{2}{1} = {r=7}{},
          cell{9}{1} = {r=2}{},
          cell{11}{1} = {c=2}{},
        }
        \hline
        Action Type &  & Attributes & Action Function Call Example\\\hline
        Mouse 
         & Move & (x:int, y:int) & {\ttfamily Move(960, 540)} \\\hline
         & Click & (x:int, y:int) & {\ttfamily Click(960, 540)}\\\hline
         & Right Click & (x:int, y:int) & {\ttfamily RightClick(960, 540)} \\\hline
         & Double Click & (x:int, y:int) & {\ttfamily DoubleClick(960, 540)}  \\\hline
         & Scroll Up & (x:int, y:int) & {\ttfamily ScrollUp(960, 540)} \\\hline
         & Scroll Down & (x:int, y:int) & {\ttfamily ScrollDown(960, 540)} \\\hline
         & Drag To & (x:int, y:int) & {\ttfamily DragTo(960, 540)} \\\hline
        Keyboard 
         & Key or Combined-keys & (key:string) & {\ttfamily Key("Space") Key("Shift+K")} \\\hline
         & Text & (x:int, y:int, text:string)& {\ttfamily Text(960, 540, "Hello World!")}\\\hline
         None & & - & {\ttfamily None()} \\\hline
        \end{tblr}
    }
\end{table}

The environment validates the action strings VLM agent's output, verifying whether they conform to valid function call formats and whether the attributes values fall within acceptable ranges. If the format is incorrect, the action type will be set to None. Additionally, the environment invokes an OCR module to parse both pre-action and post-action screenshots, which provides input for text embedding generation in the world model, reward computation, and metric calculation. The GUI environment in this paper employs the configurations specified in Table \ref{tab:GUI-environment}.

\begin{table}[h]
\centering
\caption{Configuration of GUI Environment}\label{tab:GUI-environment}
\begin{tabular}{ll}
\hline
\multicolumn{2}{c}{\textbf{Environment Configuration}} \\
\hline
Screen width & 1920 \\
Screen height & 1080 \\
Wait after action execution & 1.0 seconds \\
Maximum steps per episode & 10 \\
Parallel environments & 8 \\
OCR model & PP-OCRv4-mobile-det + PP-OCRv4-mobile-rec \\
\hline
\end{tabular}
\end{table}

\section{Technical Details of VLM Agent}\label{appendix:vlm-agent}
We employ the Qwen2.5-VL series as our base VLM, where the input prompts consist of a fixed textual prompt and a dynamically changing image of current screenshot. The screenshots maintain their original resolution. The following presents the textual components of the prompt:

\begin{Verbatim}[frame=single, formatcom=\color{black}]
You are exploring a computer desktop environment with a screen size of 
{{video_width}}x{{video_height}}. You can interact with it using the 
keyboard and mouse. Your goal is to explore this environment as much as 
possible within a limited number of steps.

Available action format:
- Move(x, y): Move the mouse to coordinates (x, y)
- Click(x, y): Left-click at coordinates (x, y)
- RightClick(x, y): Right-click at coordinates (x, y)
- DoubleClick(x, y): Double left-click at coordinates (x, y)
- ScrollUp(x, y): Scroll up at coordinates (x, y)
- ScrollDown(x, y): Scroll down at coordinates (x, y)
- Text(x, y, "text"): Enter text "text" at coordinates (x, y)
- Key("key"): Press a single key
- Key("Shift+K"): Combination key

Note that opening icons on the desktop requires a double click.
Please select a meaningful action to continue exploring. Each action 
consumes steps, so please choose the most valuable operation.

Please reply in the following JSON format:

{
  "intent": "Explanation of why this action was chosen and what goal it 
  aims to achieve",
  "action": "Specific action, for example Click(123, 456)"
}
\end{Verbatim}

We utilize vLLM's Structured Outputs feature\footnote{\url{https://docs.vllm.ai/en/latest/features/structured_outputs.html}} to enforce model outputs as JSON strings containing "intent" and "action" fields. After updating the VLM weights in each episode, the parameters are immediately synchronized to vLLM before beginning sampling for the next episode, maintaining an online training paradigm.

\section{Technical Details of World Model}\label{appendix:worldmodel}

\begin{figure}[h]
    \centering
    \includegraphics[width=0.7\linewidth]{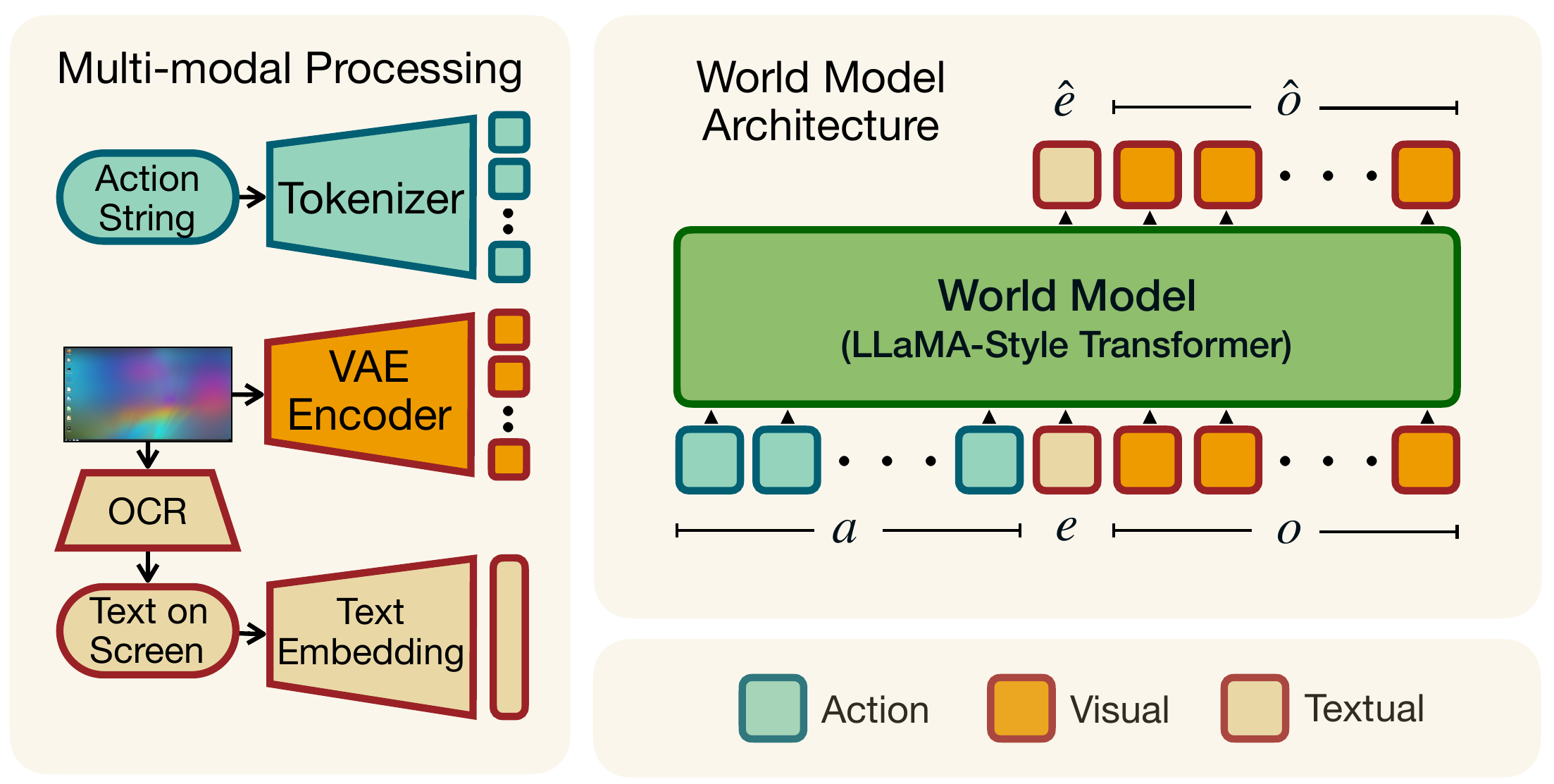}
    \caption{World Model Architecture}
    \label{fig:world-model-architecture}
\end{figure}

We employ a LLaMA-style Transformer as the backbone of our world model, the architectural design is illustrated in Figure \ref{fig:world-model-architecture}.
The model predicts the next environment states by given the current action and state. 

For action $a$, we use the LLaMA tokenizer to encode the action's function-call-style string into word tokens. For all text on the screen extracted by the OCR module, we use a pre-trained text embedding model to project all text into one dense vector, denoted as $e$. For image $o$, we utilize a pre-trained VAE encoder to map the screenshot to visual tokens. The world model is trained with state reconstruction loss, which minimizes the mean squared error between predicted and target image tokens and text embeddings. 

\section{Training Process}\label{appendix:optim-step}

Algorithm \ref{alg:rl-training-process} outlines the RL training procedure. Table \ref{tab:VLM-config} lists the VLM training configurations and hyperparameters used for GRPO, and Table \ref{tab:world-model-config} summarizes the world model’s training settings.
\begin{algorithm}[h]
\caption{RL Training Process of ScreenExplorer}\label{alg:rl-training-process}
\KwIn{Maximum episodes $N$, environment $\mathcal{E}$, actor model $\pi_\theta$, world model $f_\phi$, buffer $\mathcal{B}$}
\KwOut{Trained actor model $\pi_\theta$ and world model $f_\phi$}

\For{episode $=1$ \KwTo $N$}{
    Initialize rollout buffer $\mathcal{B}$\;
    Reset environment $\mathcal{E}$ to obtain initial observation $s_0$\;
    
    \tcp{Collect rollouts}
    \For{step $t=0$ \KwTo max\_steps}{
        Select action by VLM agent: $a_t \sim \pi_\theta(a|o_t)$\;  
        Execute $a_t$ in $\mathcal{E}$, obtain $s'_{t}$ and $ r_{format}$ \;
        World model predicts next state: $\hat{s}_{t} = f_\phi(s_t, a_t)$ \;
        Store $(s_t, a_t, s'_t, \hat{s}_{t}, r_{format})$ in $\mathcal{B}$\;
    }
    
    \tcp{Reward and advantage computation}
    \ForEach{trajectory $\tau$ in $\mathcal{B}$}{
        Compute overall reward $r$ based on trajectory $\tau$, store in $\mathcal{B}$.
    }
    Compute GRPO advantages for all samples in $\mathcal{B}$\;
    
    \tcp{World model training}
    \For{epoch $=1$ \KwTo world\_model\_training\_epoch}{
        \ForEach{batch in $\mathcal{B}$}{
         Update $\phi$ to minimize reconstruction loss $\mathcal{L}_{\text{WM}}(\phi)$ \;
         }
    }

    \tcp{VLM optimization}
    \ForEach{batch in $\mathcal{B}$}{
        Update actor model $\pi_\theta$ using GRPO with computed advantages\;
    }
}
\end{algorithm}

\begin{table}[h]
  \centering
  \begin{minipage}[t]{0.4\textwidth}
        \centering
        \caption{VLM Training Configuration and GRPO Hyperparameters}\label{tab:VLM-config}
        \begin{tabular}{ll}
        \hline
        \multicolumn{2}{c}{\textbf{VLM}} \\
        \hline
        Batch size & 16 \\
        Mixed precision & bf16 \\
        Maximum gradient norm & 1.0 \\
        LoRA rank & 16 \\
        Learning rate & 4e-5 \\
        Training batch size & 16 \\
        Maximum completion length & 128 \\
        Rollout temperature & 1.0 \\
        \hline
        \multicolumn{2}{c}{\textbf{GRPO Parameters}} \\
        \hline
        KL divergence coefficient($\beta$)  & 0.04 \\
        PPO lower bound($\varepsilon_{low}$) & 0.2 \\
        PPO upper bound($\varepsilon_{high}$) & 0.28 \\
        \hline
        \end{tabular}
  \end{minipage}%
  \hfill 
  \begin{minipage}[t]{0.55\textwidth}
    \centering
    \caption{World Model Training Configuration}\label{tab:world-model-config}
    \begin{tabular}{ll}
    \hline
    \multicolumn{2}{c}{\textbf{World Model}} \\
    \hline
    World model base & Llama-3.2-1B \\
    Image tokenizer model & Cosmos-Tokenizer-CI16x16 \\
    Text embedding model & BAAI/bge-m3 \\
    Training epoch  & 3 \\
    Training batch size & 32 \\
    Learning rate & 4e-5 \\
    Mixed precision & bf16 \\
    Maximum gradient norm & 1.0 \\
    \hline
    \end{tabular}
  \end{minipage}
\end{table}

The experiment with the 3B model was conducted using a single Nvidia A100 GPU, requiring approximately 26 hours to complete 200 steps. During stable training, each parallel episode took approximately 400-500 seconds, with rollout time accounting for the majority of the duration (about 250-300 seconds). Within the rollout time, the most time-consuming operations were vLLM decoding for VLM and OCR steps. The OCR processing time was directly correlated with the quantity of text present on the screen, with processing time increasing in scenarios containing higher text density.

\FloatBarrier
\section{Experience Stream Distillation}\label{appendix:distillation}

The purpose of Experience Stream Distillation is to identify valuable exploratory behaviors from historical exploration trajectories generated during RL training. The model distilled using these actions can directly inherit the effective environmental interaction capabilities from its predecessor generation, while maintaining diversity in exploration capabilities through data filtering and balancing.

\begin{figure}[t]
    \centering
    \subfigure[Experience stream collection through RL training and distillation for next-generation models]{
            \includegraphics[width=0.55\textwidth]{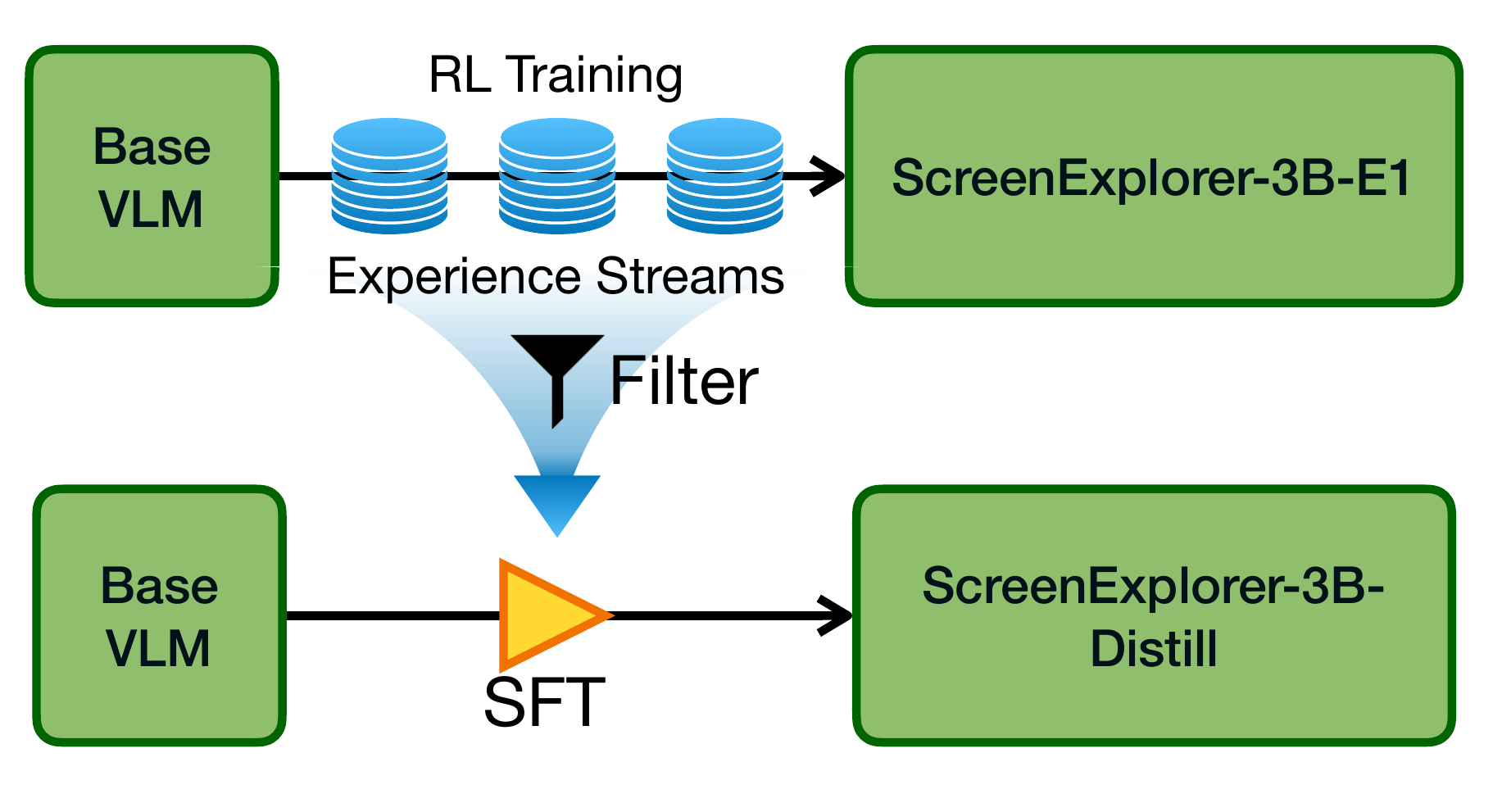}
            \label{fig:distill-pipeline}
    }
    \quad
    \subfigure[Data cycle for sustainable, self-fueling learning paradigm.]{
            \includegraphics[width=0.35\textwidth]{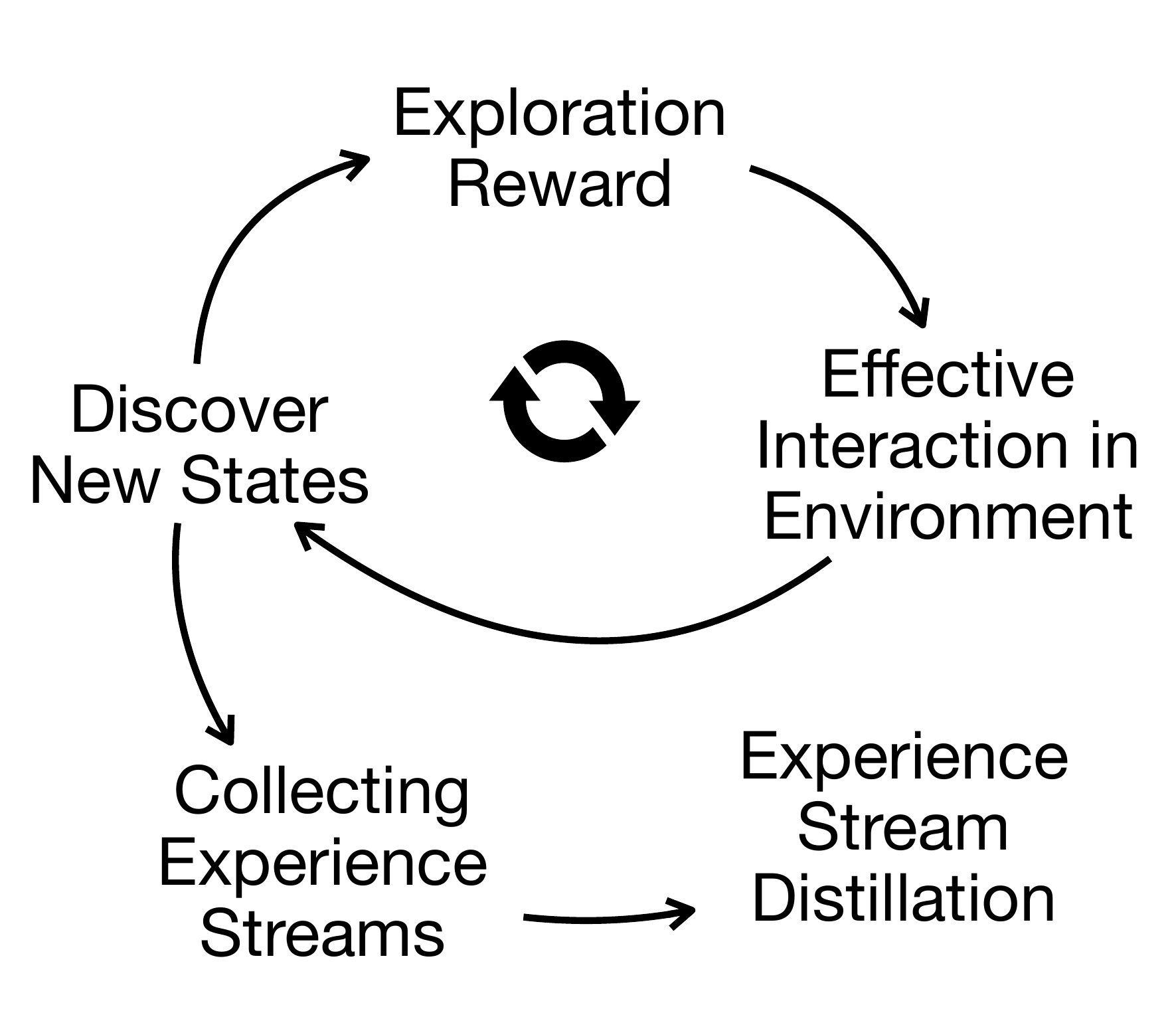}
            \label{fig:cycle}
    }
    \caption{Training Pipeline Integrating RL and Experience Stream Distillation.}
\end{figure}

We first filter the trajectories generated during RL training, identifying and retaining diverse exploration steps that successfully complete specific tasks. These single-step actions are then organized into datasets for Supervised Fine-Tuning (SFT) on the base model. Figure \ref{fig:distill-pipeline} illustrates this process.

We employed two filtering strategies: manual filtering and automated filtering. For manual filtering of experience streams, we established the following criteria:

\begin{enumerate}
    \item Begin from the 30-th episode.
    \item The intent text description should clearly indicate a specific action (e.g., click, type, scroll) and specify a target (e.g., an icon, button, field, or screen location).
    \item The executed action should accomplish the goal described in the intent.
    \item The language in the intent should flow smoothly and contain no word repetitions.
\end{enumerate}

In addition to manual filtering, we implemented an automated filtering process without human intervention, based on the following criteria:

\begin{enumerate}
    \item Begin from the 30-th episode.
    \item The output format is correct, where $r_{format}=1$.
    \item Advantages greater than $0$.
    \item The intent text should clearly indicate a specific action, evaluated by \textit{gpt-4o-mini-2024-07-18}.
\end{enumerate}
The automated filtering employs the following prompts:
\begin{Verbatim}[frame=single, formatcom=\color{black}]
You are evaluating whether an intent string clearly specifies a computer 
operation instruction.

A good intent should:
1. Clearly indicate a specific action (e.g., click, type, scroll)
2. Specify a target (e.g., an icon, button, field, or screen location)
3. Be unambiguous about what the user wants to accomplish
4. The language flows smoothly and there are no words repeated.

Intent to evaluate: "{{intent}}"

If the intent meets the criteria above, rewrite it as a clear task 
objective.
If the intent does not meet the criteria, mark it as not good and use an 
empty string as the task.
Keep using the same language as the input intents.

Provide your evaluation in the following JSON format:
```json
{
  "is_good_intent": true/false,
  "reasoning": "Your detailed explanation for why this intent is good or 
  not",
  "task": "Rewritten clear task objective if good, or empty string if not"
}
```
\end{Verbatim}

During both filtering processes, we did not introduce or construct additional data. 
Table \ref{tab:distill-performance} presents the exploration metrics of models after SFT using both manual filtering and automated filtering datasets. The results indicate that, compared to the baseline model, SFT training on either datasets effectively enhances the model's exploration capabilities and diversity. Compared to automatic filtering, manual filtering provides higher quality data for distillation by additionally verifying the consistency between intent descriptions and actually executed actions.

\begin{table}[h]
  \centering
  \small
  \caption{Impact of Different Filtering Strategies in Experience Stream Distillation on Model Exploration Performance}
  \label{tab:distill-performance}
  \resizebox{\textwidth}{!}{
  \begin{tabular}{@{} l c c c c c c c c @{}}
    \toprule
      \multirow{2}{*}{\textbf{Model}}
      & \multirow{2}{*}{\textbf{\makecell{\# SFT Training Set}}}
      & \multirow{2}{*}{\textbf{Setting}}
      & \multirow{2}{*}{\textbf{\makecell{Correct \\ Format}}}
      & \multicolumn{2}{c}{\textbf{Trajectory-level}}
      & 
      \multicolumn{2}{c}{\textbf{Group-level}}
      & 
      \multirow{2}{*}{\textbf{\makecell{Avg. \\ Diversity}}} \\
      \cmidrule(lr){5-6} \cmidrule(lr){7-8}
      & & & &
      $d_{\mathrm{seq}}^{\mathrm{vis}}(\tau)$ 
      & $d_{\mathrm{seq}}^{\mathrm{text}}(\tau)$ 
      & $D_{\mathrm{grp}}^{\mathrm{vis}}(\mathcal{G})$ 
      & $D_{\mathrm{grp}}^{\mathrm{text}}(\mathcal{G})$ 
      & \\
    \midrule
    \rowcolor{gray!15}\multicolumn{9}{l}{\textit{Base Model}} \\
    \multirow{2}{*}{Qwen2.5-VL-3B}
      & \multirow{2}{*}{-} & $t=1.0$ & 0.62 & 0.16 & 0.10 & 0.40 & 0.19 & 0.21 \\
      &                    & $t=0.5$ & 0.84 & 0.15 & 0.08 & 0.31 & 0.14 & 0.17 \\
    \midrule
    \rowcolor{gray!15}\multicolumn{9}{l}{\textit{Ours}} \\
     \multirow{2}{*}{\makecell[l]{ScreenExplorer-3B-E1 \\ (RL training from scratch)}}
        & \multirow{2}{*}{-} & $t=1.0$ & 0.99 & 0.57  & 0.33 & 0.68 & 0.45 & 0.51 \\
        &                    & $t=0.5$ & 1.00 & 0.57  & 0.33 & 0.72 & 0.46 & 0.52\\
     \midrule 
     \multirow{2}{*}{\makecell[l]{ScreenExplorer-3B-Distill\\(Manual Filtering)} } 
     & \multirow{2}{*}{216} & $t=1.0$ & 0.93 & 0.64 & 0.37 & 0.68 & 0.43 & 0.53 \\
     &                      & $t=0.5$ & 0.99 & 0.66 & 0.41 & 0.67 & 0.44 & 0.55 \\ 
    \midrule
    \multirow{2}{*}{\makecell[l]{ScreenExplorer-3B-Distill \\ (Automated Filtering)}}
    & \multirow{2}{*}{199} & $t=1.0$ & 0.94 & 0.64 & 0.39 & 0.71 & 0.46 & 0.55 \\
    &                      & $t=0.5$ & 1.0  & 0.62 & 0.42 & 0.62 & 0.45 & 0.53 \\
    \bottomrule
  \end{tabular}
  }
\end{table}

Figure \ref{fig:cycle} illustrates the data cycle process where an agent explores and collection experience stream data in an open-world environment, ultimately achieving continuous improvement through experience stream distillation. This data cycle mechanism enables the model to transcend the limitations of "static corpus + offline training," advancing toward a sustainable, self-fueling learning paradigm—representing a viable pathway for enhancing agent capabilities when human-generated data becomes exhausted in the future.

\FloatBarrier
\section{Baseline Settings}\label{appendix:baseline-settings}
For the General Used Models OpenAI gpt-4o and Qwen2.5-VL variants, we employed a fixed prompt designed to encourage free exploration, as same in Appendix \ref{appendix:vlm-agent}.
For OpenAI Computer Use, we utilize the officially recommended invocation method, with prompts that are not publicly disclosed. For doubao-1.5-ui-tars, we employ the following system prompts:
\begin{Verbatim}[frame=single, formatcom=\color{black}]
You are a GUI agent. You are given a task and your action history, with
screenshots. You need to perform the next action to complete the task.
## Output Format
```
Thought: ...
Action: ...
```
## Action Space
click(start_box='[x1, y1, x2, y2]')
left_double(start_box='[x1, y1, x2, y2]')
right_single(start_box='[x1, y1, x2, y2]')
drag(start_box='[x1, y1, x2, y2]', end_box='[x3, y3, x4, y4]')
hotkey(key='')
type(content='') #If you want to submit your input, use "\n" at the end of 
`content`.
scroll(start_box='[x1, y1, x2, y2]', direction='down or up or right or 
left')
wait() #Sleep for 5s and take a screenshot to check for any changes.
finished(content='xxx') # Use escape characters \\', \\", and \\n in 
content part to ensure we can parse the content in normal python string 
format.
## Note
- Use Chinese in `Thought` part.
- Write a small plan and finally summarize your next action (with its 
target element) in one sentence in `Thought` part.
## User Instruction
\end{Verbatim}

We employ the following task prompts:

\begin{Verbatim}[frame=single, formatcom=\color{black}]
Your goal is to explore this environment as much as possible within a 
limited number of steps. Please select a meaningful action to continue 
exploring. Note that opening icons on the desktop requires a double 
click. You must only use mouse and keyboard inputs. No other tools or 
input devices are permitted.
\end{Verbatim}

\FloatBarrier
\section{Original Capacity in VLM}\label{appendix:llms-capability}
\begin{figure}[h]
  \centering
  \begin{tabular}{@{} m{0.5\textwidth}  m{0.45\textwidth} @{}}
    \includegraphics[width=\linewidth]{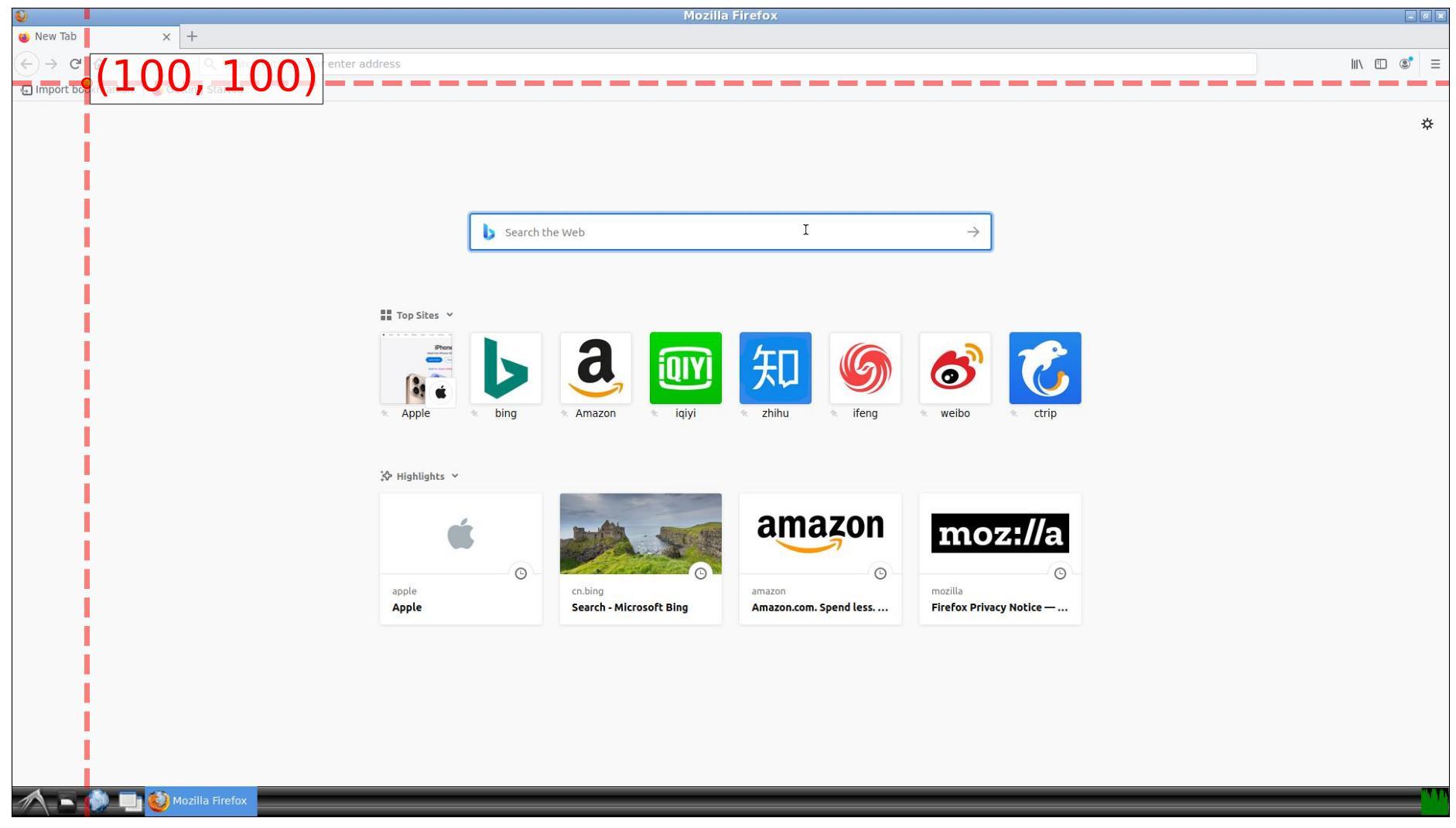} &
    \shortstack[l]{%
      \textbf{Intent:} “Move the cursor to the top-left \\ 
      corner of the screen to begin navigation.”\\[0.5ex]
      \textbf{Action:} “Move(100, 100)”
    }%
  \end{tabular}
  \caption{The \textit{Qwen2.5-VL-3B} model demonstrates spatial awareness capabilities. When the intent specifies "top-left corner of the screen", the model outputs coordinates (100, 100), which accurately corresponds to the screen's top-left position.}\label{fig:original-capacity-1}
\end{figure}

\begin{figure}[h]
  \centering
  \begin{tabular}{@{} m{0.5\textwidth}  m{0.45\textwidth} @{}}
    \includegraphics[width=\linewidth]{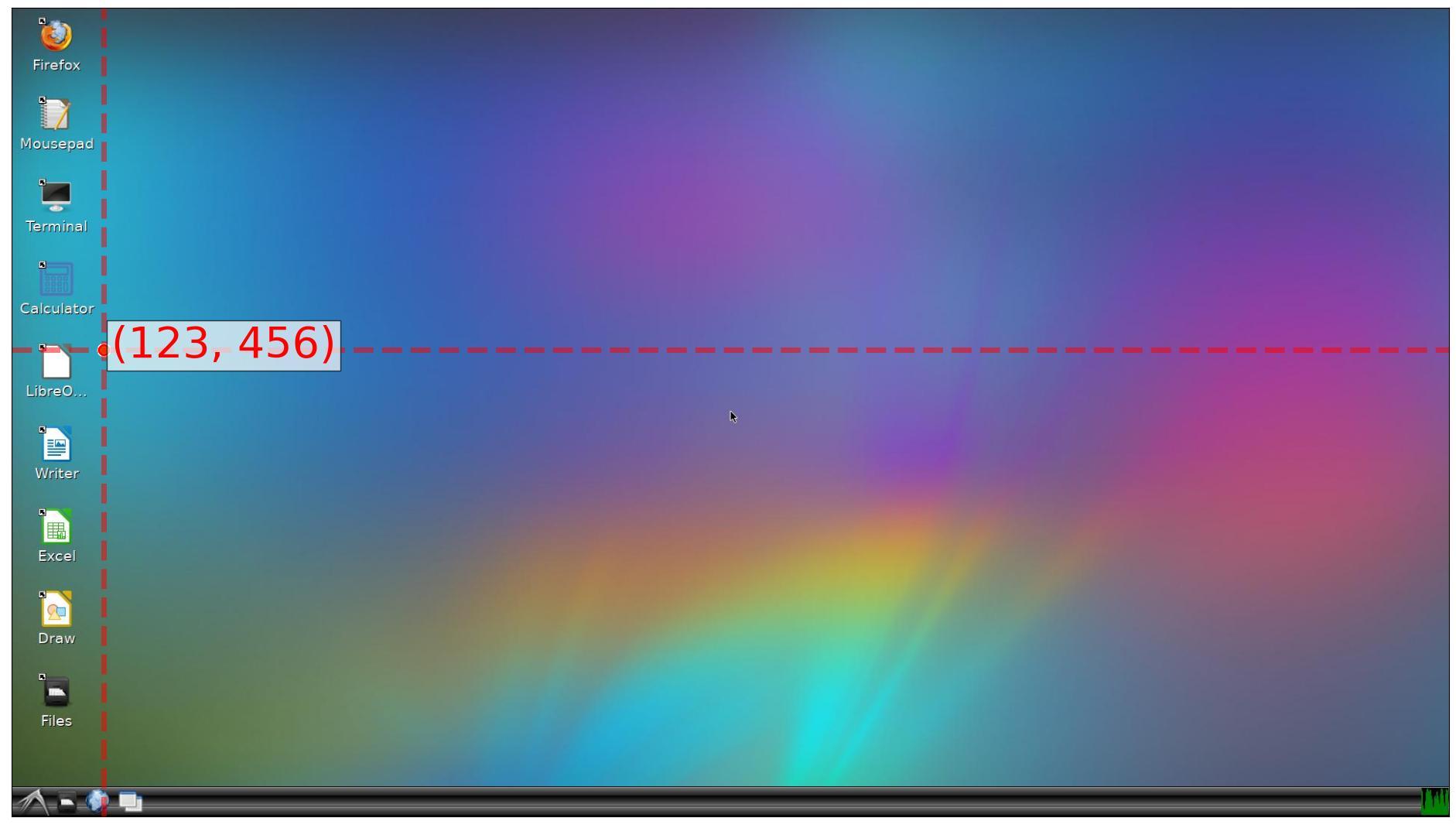} &
    \shortstack[l]{%
      \textbf{Intent:} “To start the process of accessing \\ something specific on this presumably not \\ 
      fully explored environment.”\\[0.5ex]
      \textbf{Action:} “DoubleClick(123, 456)”
    }%
  \end{tabular}
  \caption{The \textit{Qwen2.5-VL-3B} model followed the instructions in the prompt, attempting to utilize double click to interact with desktop icons. However, the specific coordinates were directly copied from the prompt text, indicating that, the language processing capabilities dominated over the visual localization abilities.}\label{fig:original-capacity-2}
\end{figure}

\begin{figure}[h]
  \centering
  \begin{tabular}{@{} m{0.5\textwidth}  m{0.45\textwidth} @{}}
    \includegraphics[width=\linewidth]{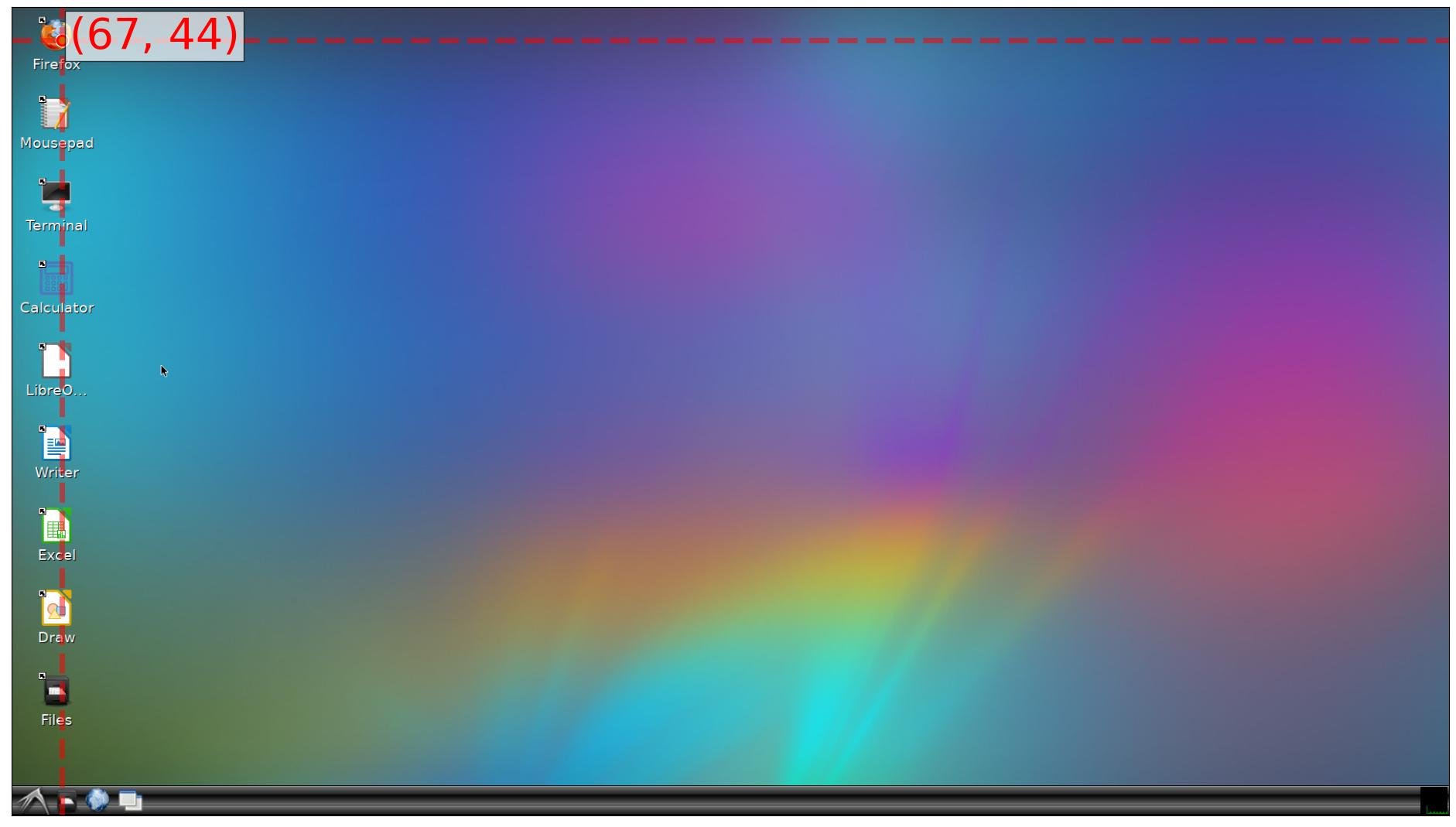} &
    \shortstack[l]{%
      \textbf{Intent:} “To open an application, specifically \\
      Firefox for web browsing.”\\[0.5ex]
      \textbf{Action:} “DoubleClick(67, 44)”
    }%
  \end{tabular}
  \caption{The original \textit{Qwen2.5-VL-3B} model demonstrates the capability to correctly launch applications in a limited number of cases, with the intent descriptions aligning with the target actions. RL training can facilitate the emergence of such effective interactions.}\label{fig:original-capacity-3}
\end{figure}

\begin{figure}[h]
  \centering
  \begin{tabular}{@{} m{0.5\textwidth}  m{0.45\textwidth} @{}}
    \includegraphics[width=\linewidth]{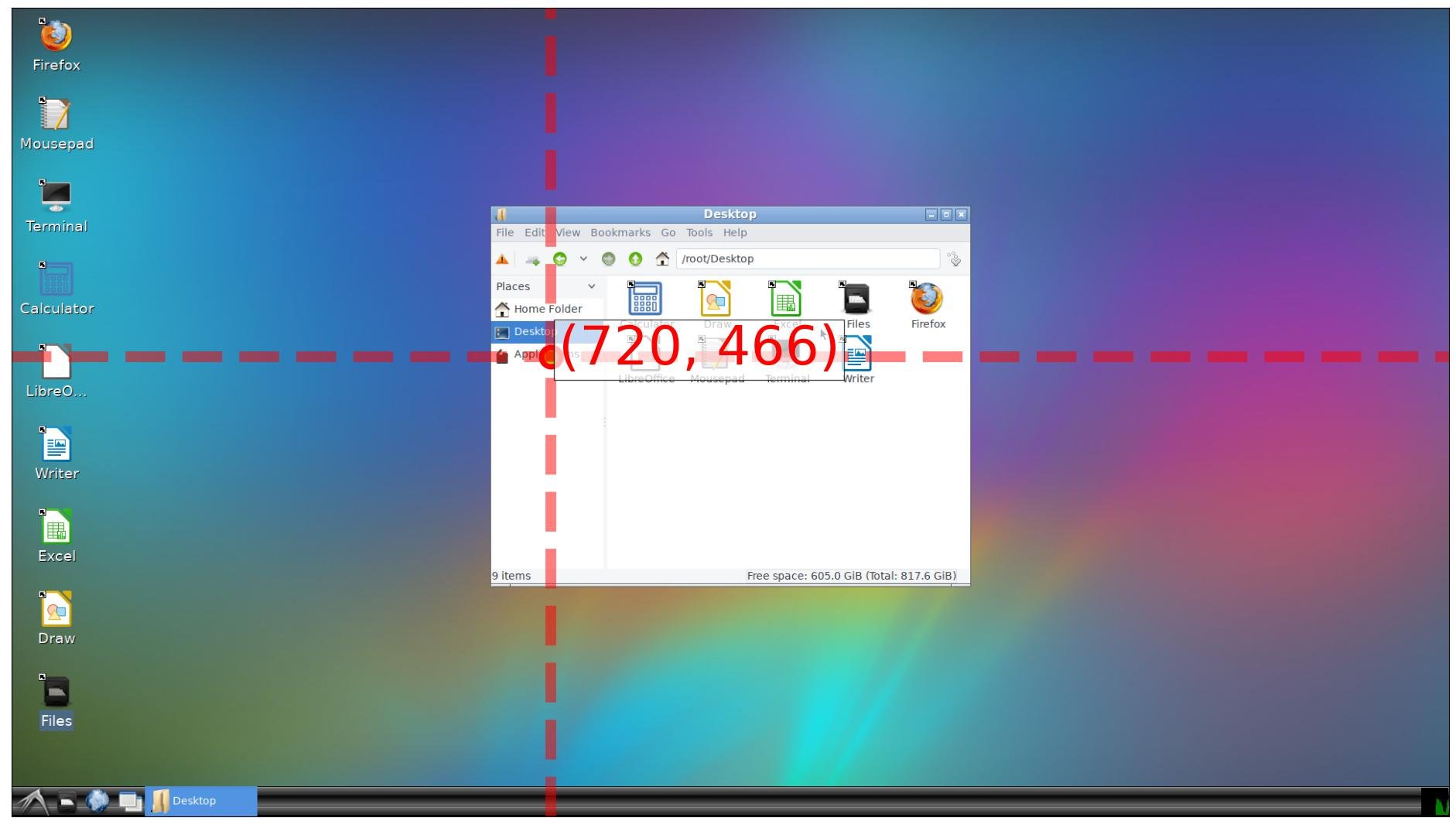} &
    \shortstack[l]{%
      \textbf{Intent:} “I chose to open the 'Applications' \\ 
      folder because it may contain additional \\ 
      software and utilities that aren't immediately \\
      visible on the desktop, allowing me to explore \\
      more functionalities available in this environment.”\\[0.5ex]
      \textbf{Action:} “DoubleClick(720, 466)”
    }%
  \end{tabular}
  \caption{The original \textit{Qwen2.5-VL-72B} model, due to its larger model scale, demonstrates enhanced capability in interpreting action intentions, recognizing UI elements within images and generates more precise coordinate outputs.}\label{fig:original-capacity-4}
\end{figure}

We find the \textit{Qwen2.5-VL} model has demonstrated some fundamental capabilities in GUI exploration, as evidenced by the following representative cases in Figure \ref{fig:original-capacity-1} to \ref{fig:original-capacity-4}.

\FloatBarrier
\section{Ablation on Exploration Reward Components}\label{appendix:reward-ablation}

\begin{figure}[h]
    \centering
    \includegraphics[width=1\linewidth]{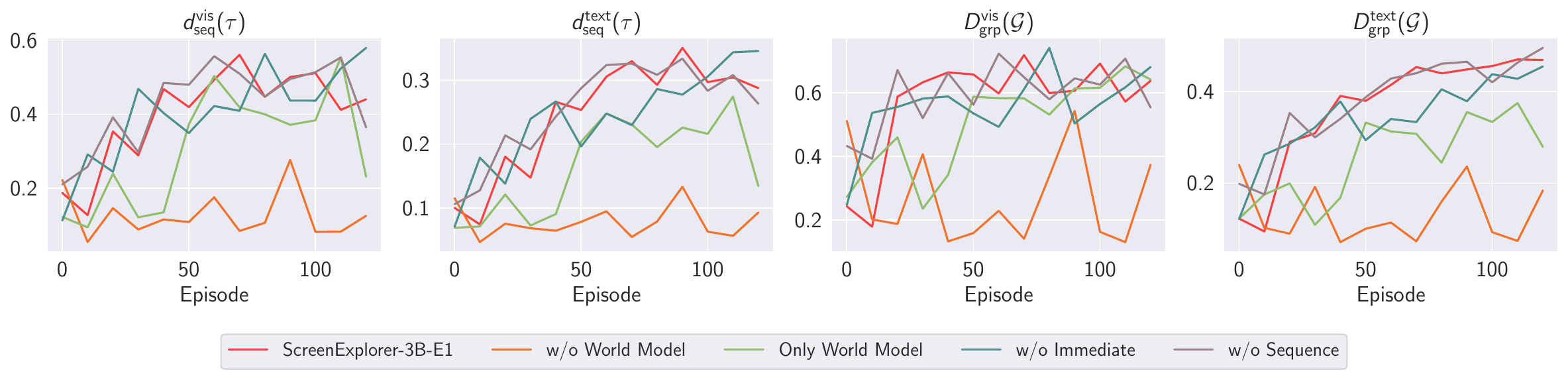}
    \caption{Impact of ablation studies on exploration metrics across different exploration reward components.}
    \label{fig:rewards-ablation}
\end{figure}

Our Exploration Reward designed comprises three categorical components: Instantaneous Change Reward, Subsequent Change Reward, and World Model Curiosity Reward. We examine how these three exploration reward components influence the training of exploration capabilities. We conducted ablation studies by removing each of these rewards individually, denoted as \textbf{ w/o Instant}, \textbf{w/o Sequence}, and \textbf{w/o World Model}. Additionally, we included a ablation setting using only world model reward, labeled as \textbf{Only World Model}. Figure \ref{fig:rewards-ablation} presents a comparison of exploration metrics between these four ablation settings and \textit{ScreenExplorer-3B-E1}. The results demonstrate that the ablation setting \textbf{w/o World Model} exhibits the poorest performance in exploration metrics, with this group experiencing stagnation in the early exploration phase and showing limited potential for future exploration growth. Similarly, the group using \textbf{Only World Model} also showed suboptimal performance.

\begin{figure}[h]
  \centering
 \begin{tabular}{
    @{}
    m{0.195\textwidth}@{\hspace{1pt}}
    m{0.195\textwidth}@{\hspace{1pt}}
    m{0.195\textwidth}@{\hspace{1pt}}
    m{0.195\textwidth}@{\hspace{1pt}}
    m{0.195\textwidth}@{}
  }
    \includegraphics[width=\linewidth]{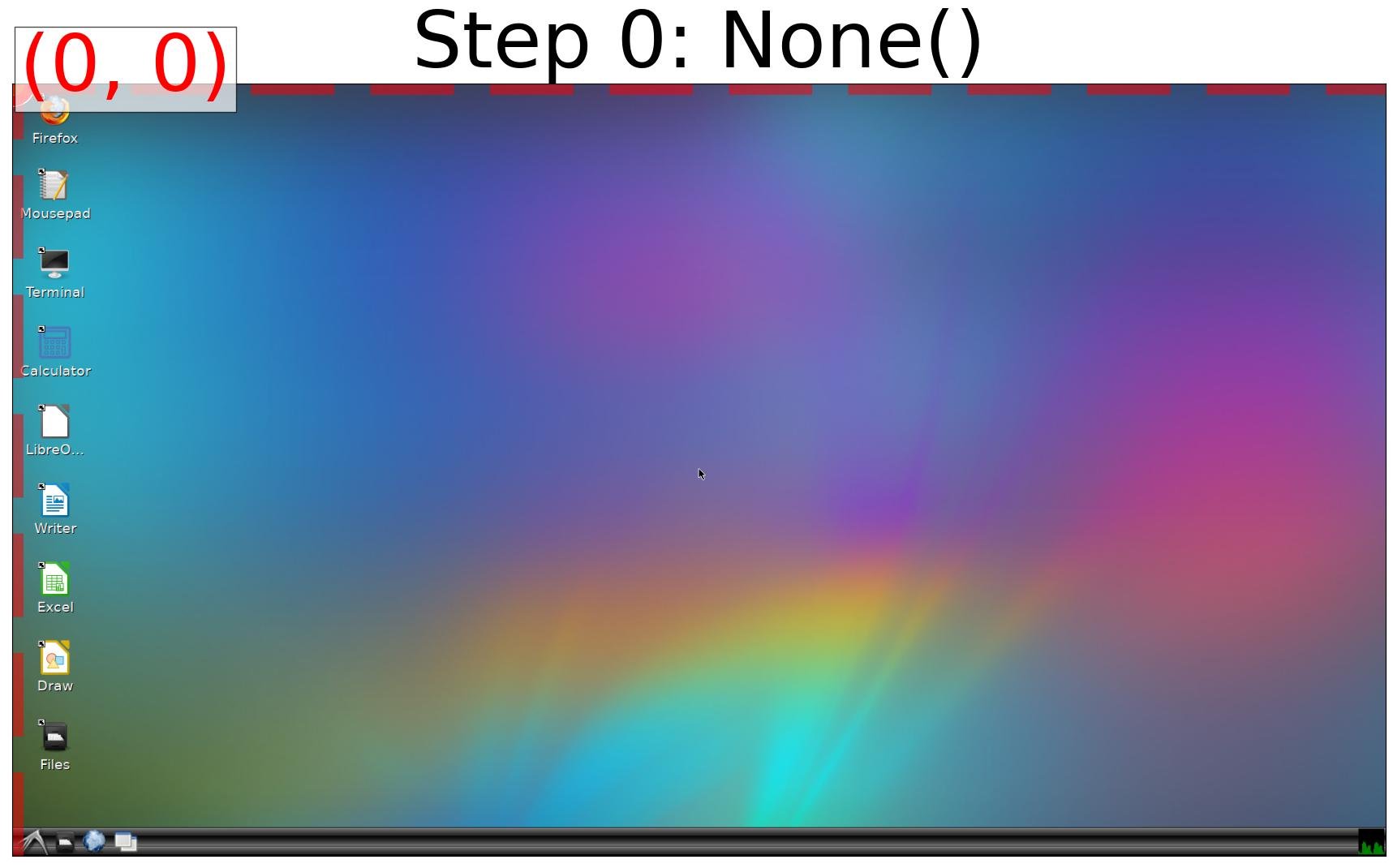} &
    \includegraphics[width=\linewidth]{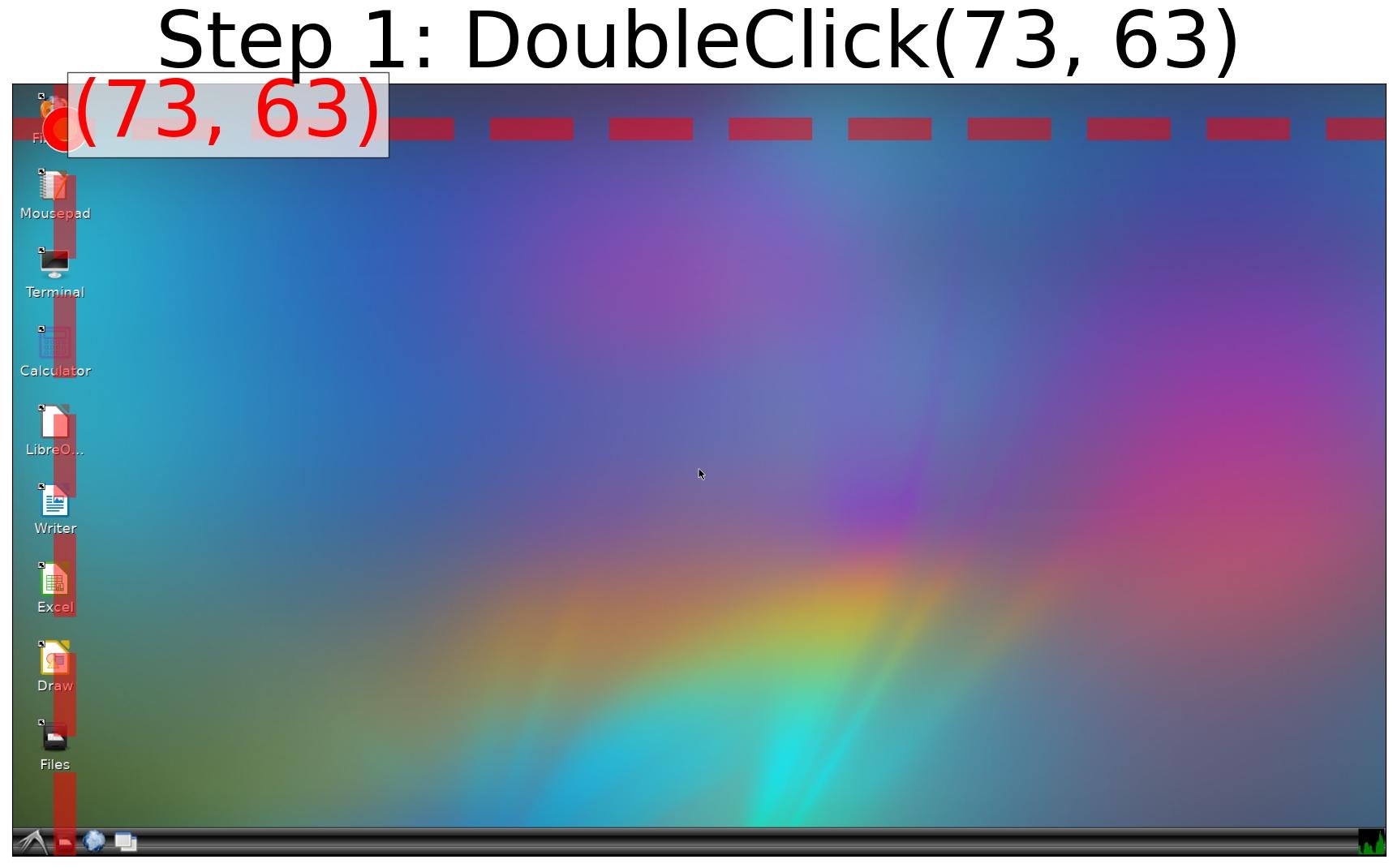} &
    \includegraphics[width=\linewidth]{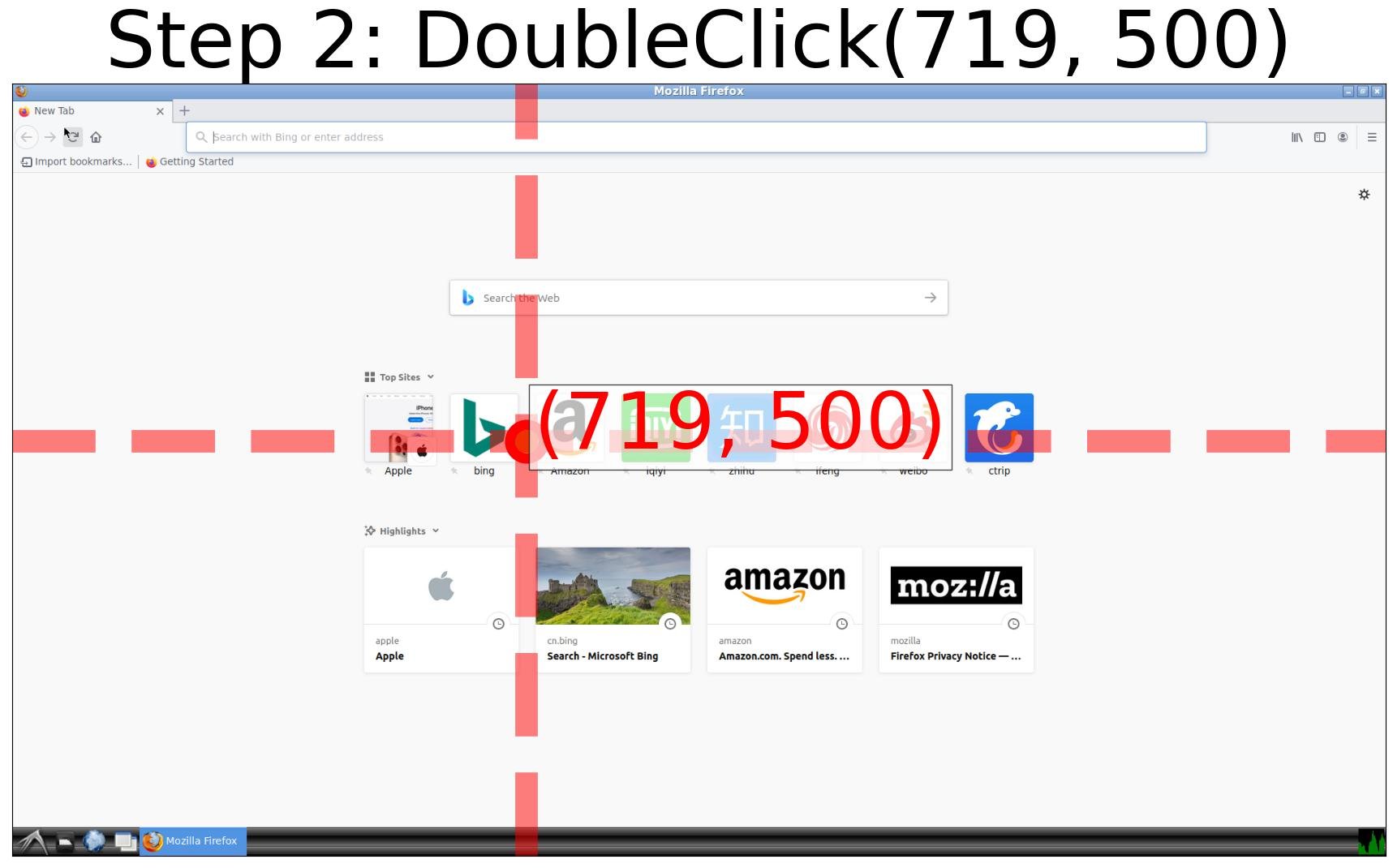} &
    \includegraphics[width=\linewidth]{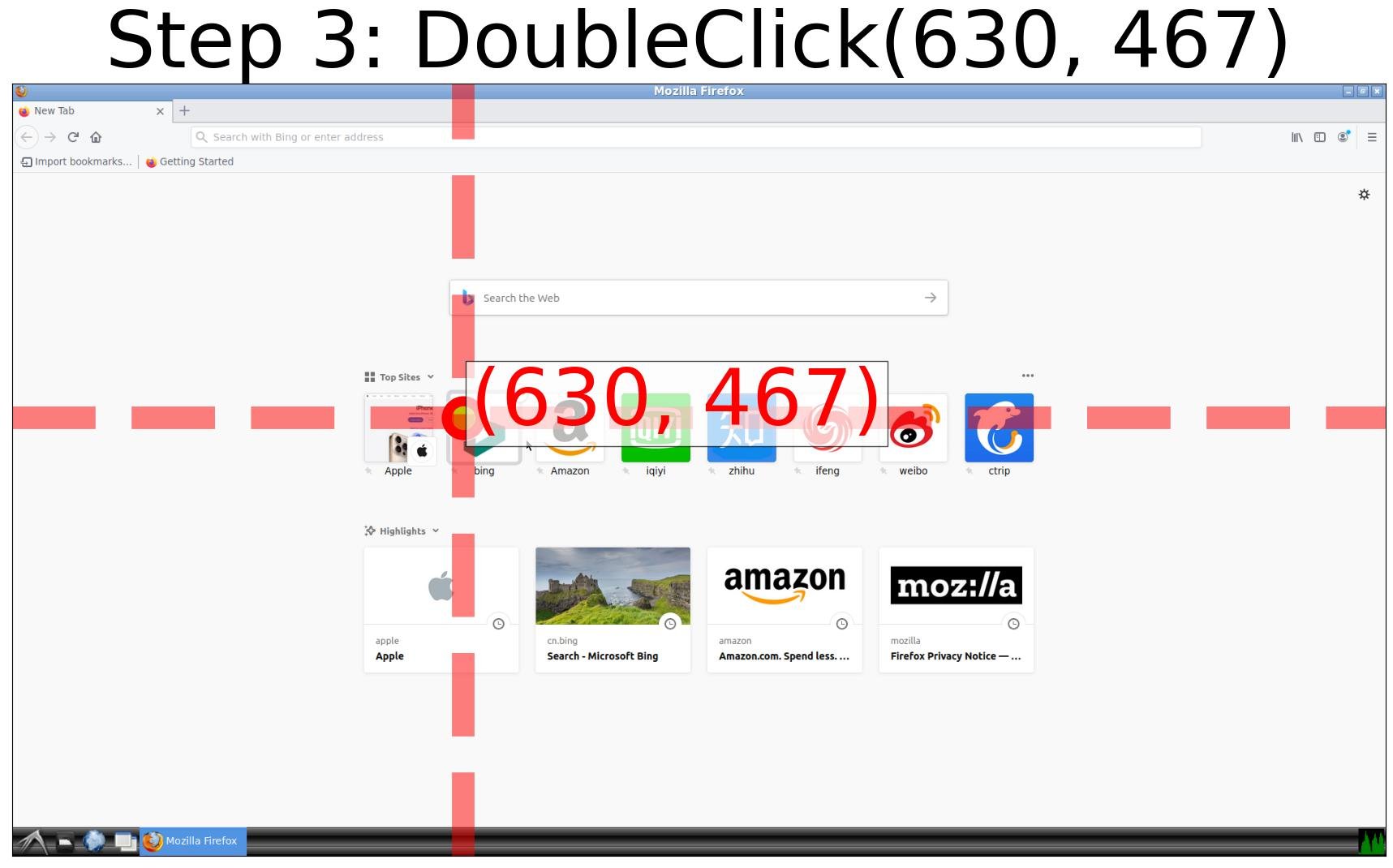} &
    \includegraphics[width=\linewidth]{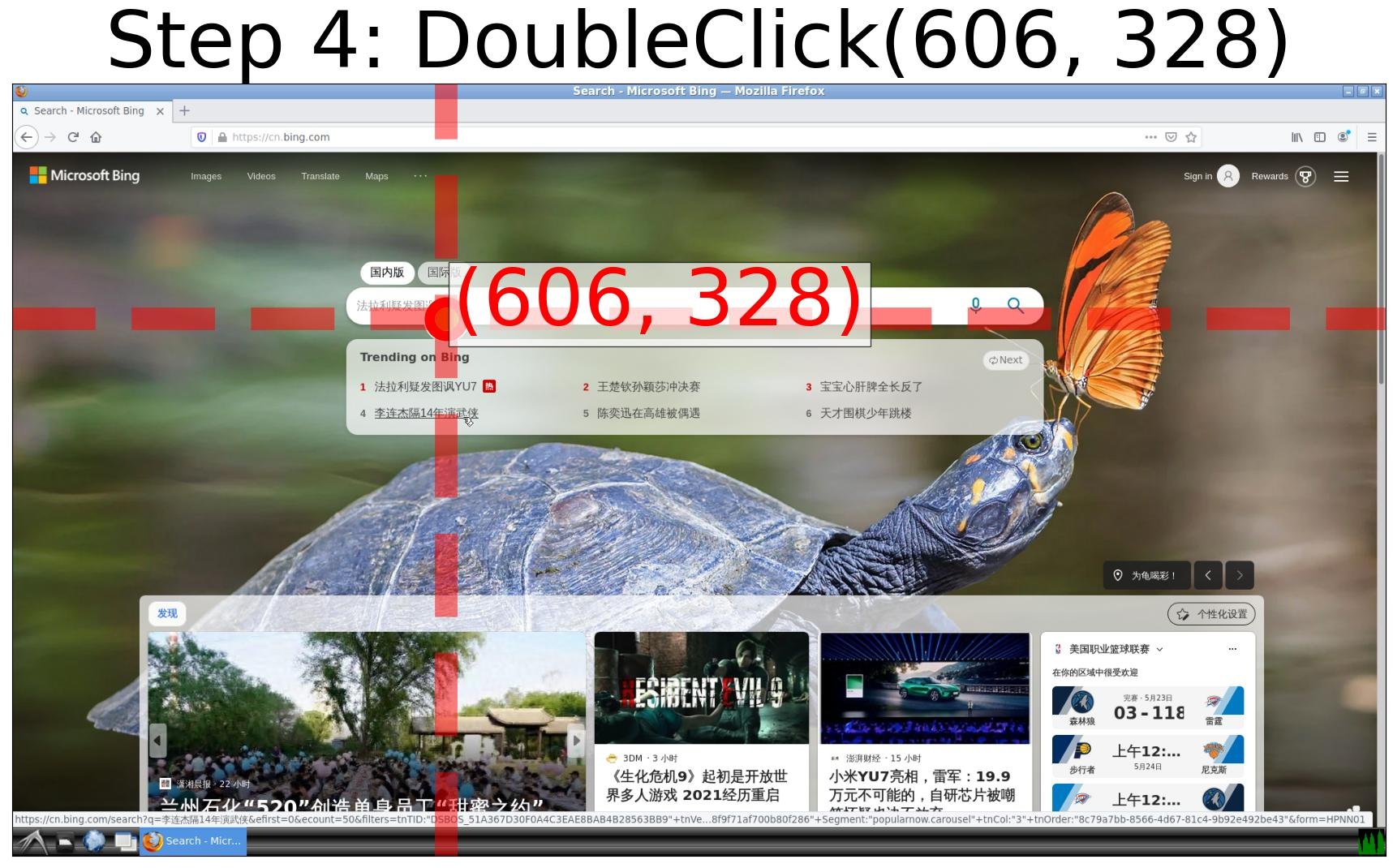} \\ 
    \includegraphics[width=\linewidth]{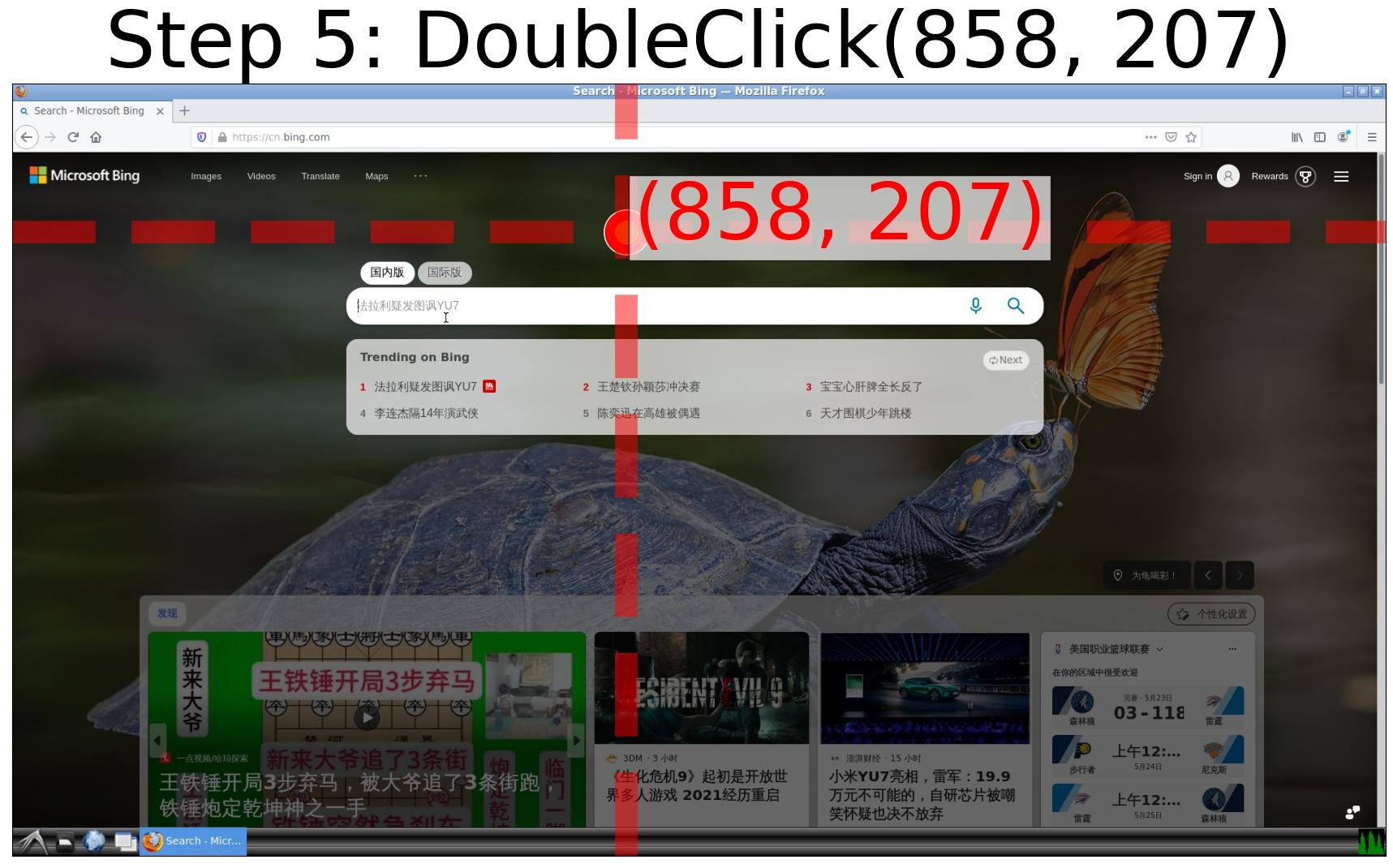} &
    \includegraphics[width=\linewidth]{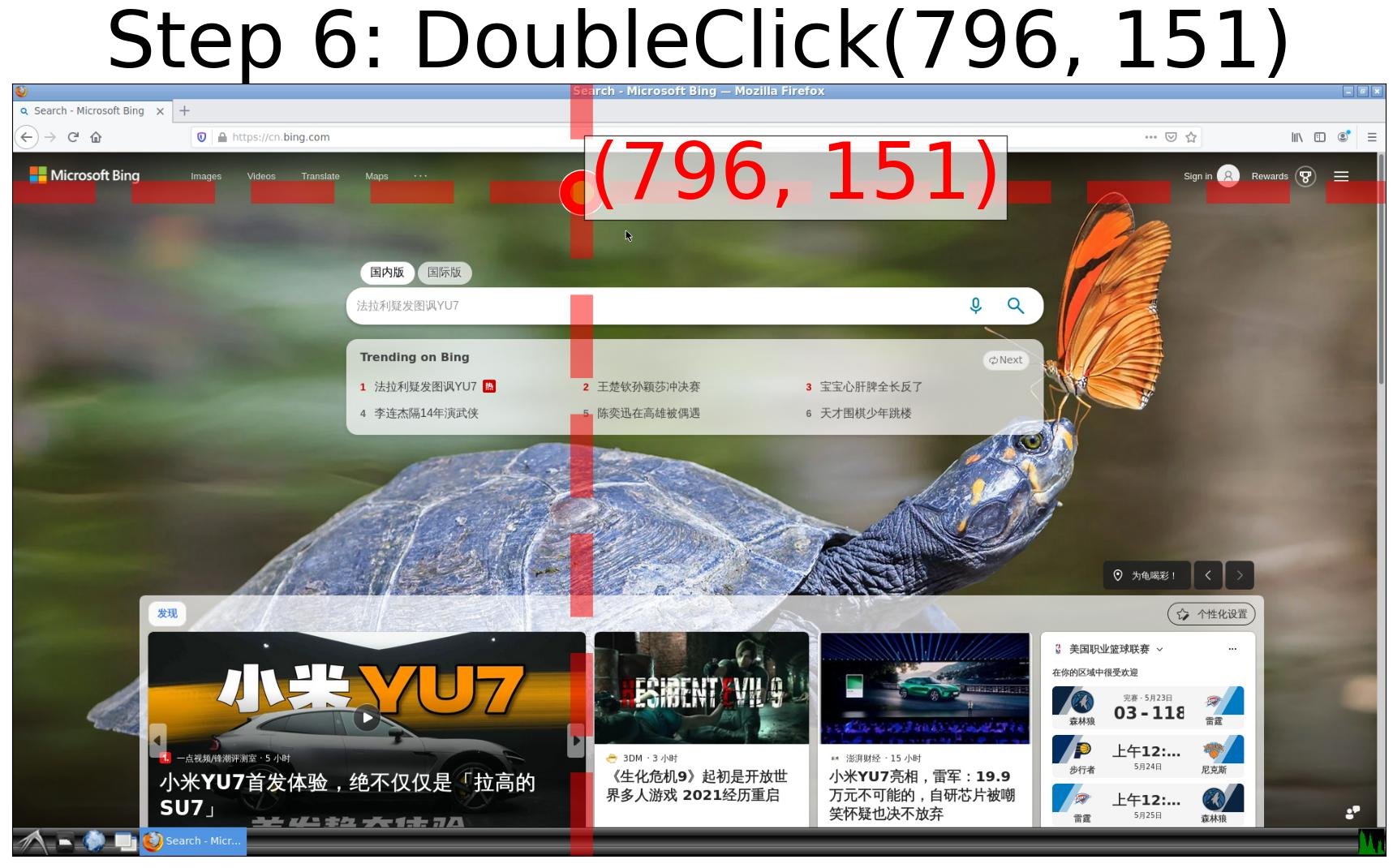} &
    \includegraphics[width=\linewidth]{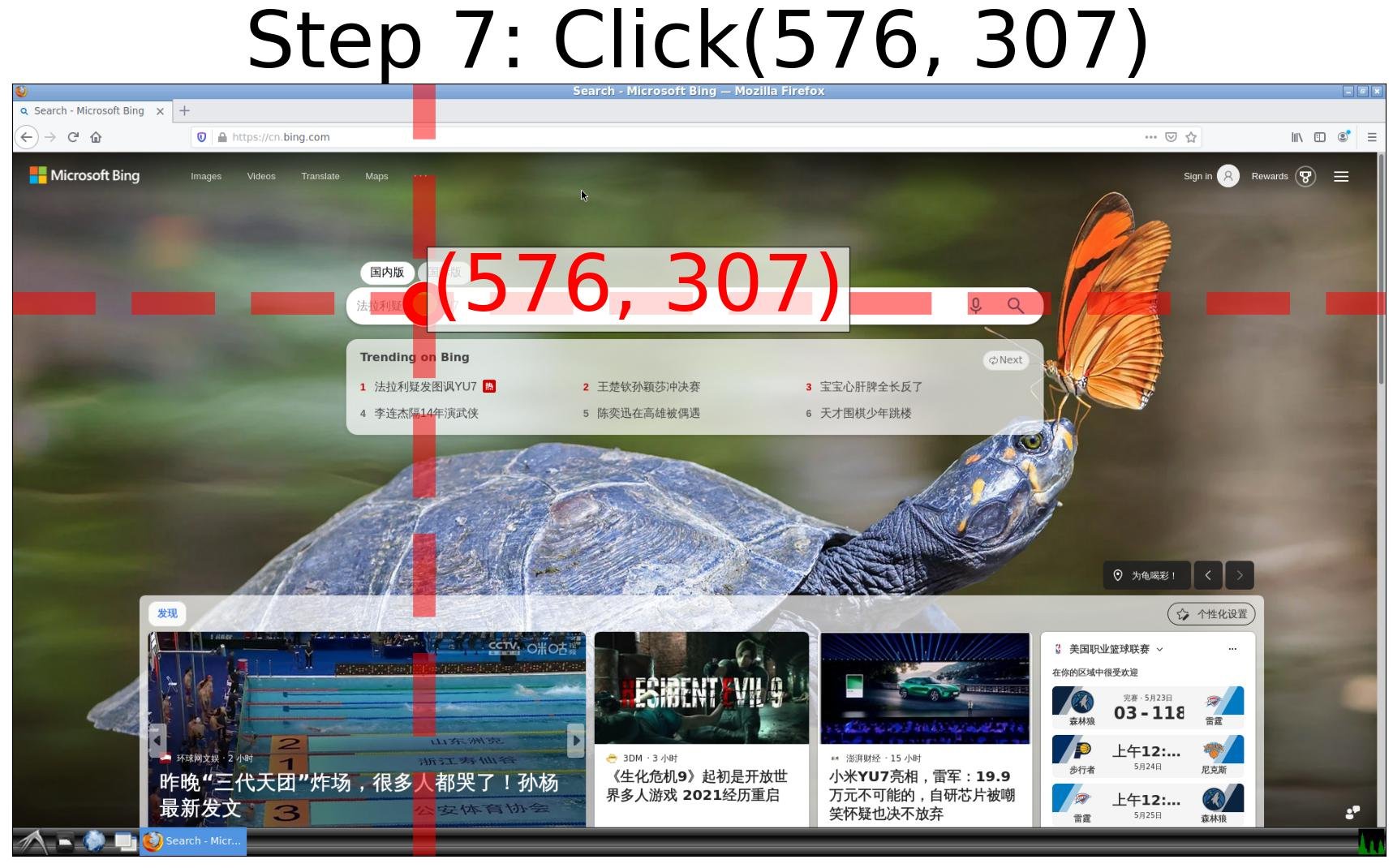} &
    \includegraphics[width=\linewidth]{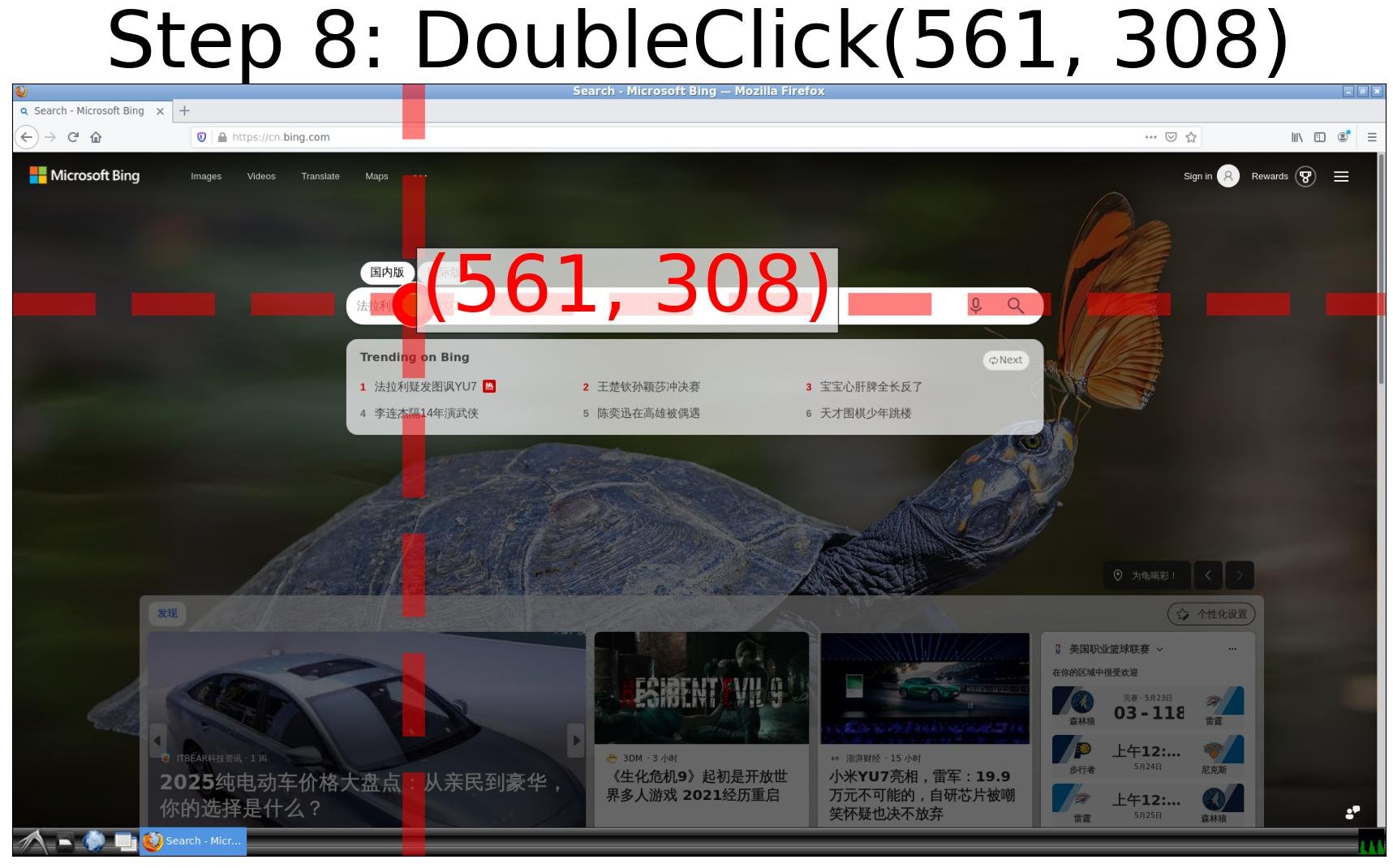} &
    \includegraphics[width=\linewidth]{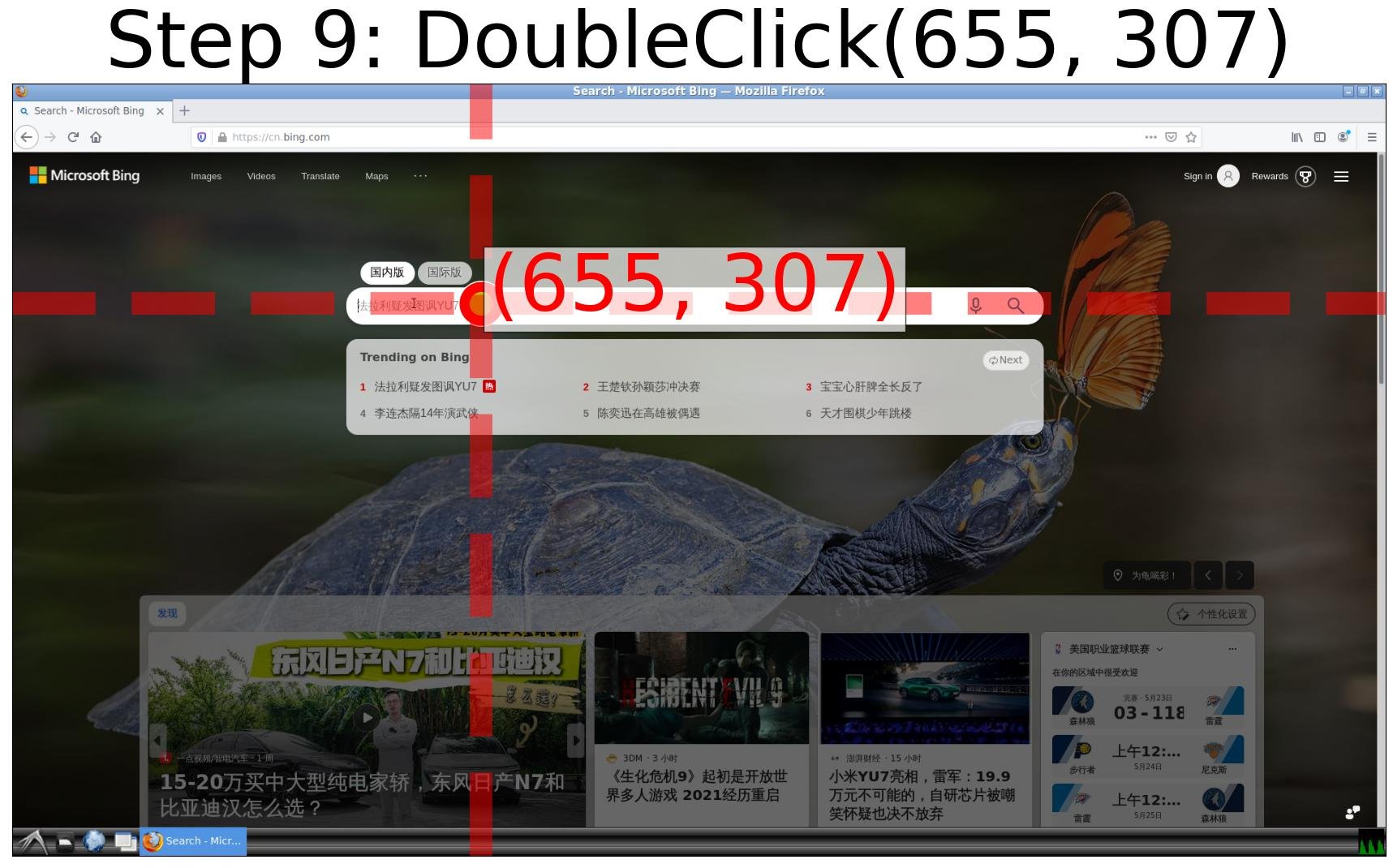}
  \end{tabular}
  \caption{In a case study from the ablation group \textbf{w/o Instant}, while the agent successfully learned to open the browser, it demonstrated limited capability for deeper exploration.}\label{fig:ablation-no-instantaneous}
\end{figure}

Comparing ablation settings \textbf{w/o Instant} and \textbf{w/o Sequence}, \textbf{w/o Instant} exhibited inferior performance in exploration metrics. Through analysis of training trajectories, we discovered that in the absence of immediate rewards, the model frequently became trapped in ineffective attempts, as illustrated by a typical case in Figure \ref{fig:ablation-no-instantaneous}.

We conducted two additional interesting ablation studies:

In the ablation setting \textbf{w/o Visual}, we removed all visual reward signals, specifically eliminating the $r^{vis}_{inst}$, $r^{vis}_{seq}$, and $r^{vis}_{world}$ reward components. This means that the model's exploration was driven solely by changes in on-screen text. Interestingly, even without rewards from visual information and relying exclusively on text-based reward signals, the agent was still capable of effective exploration in the GUI environment. We attribute this to the fact that the GUI is a complex environment containing abundant textual information.

\begin{figure}[h]
    \centering
    \includegraphics[width=0.9\linewidth]{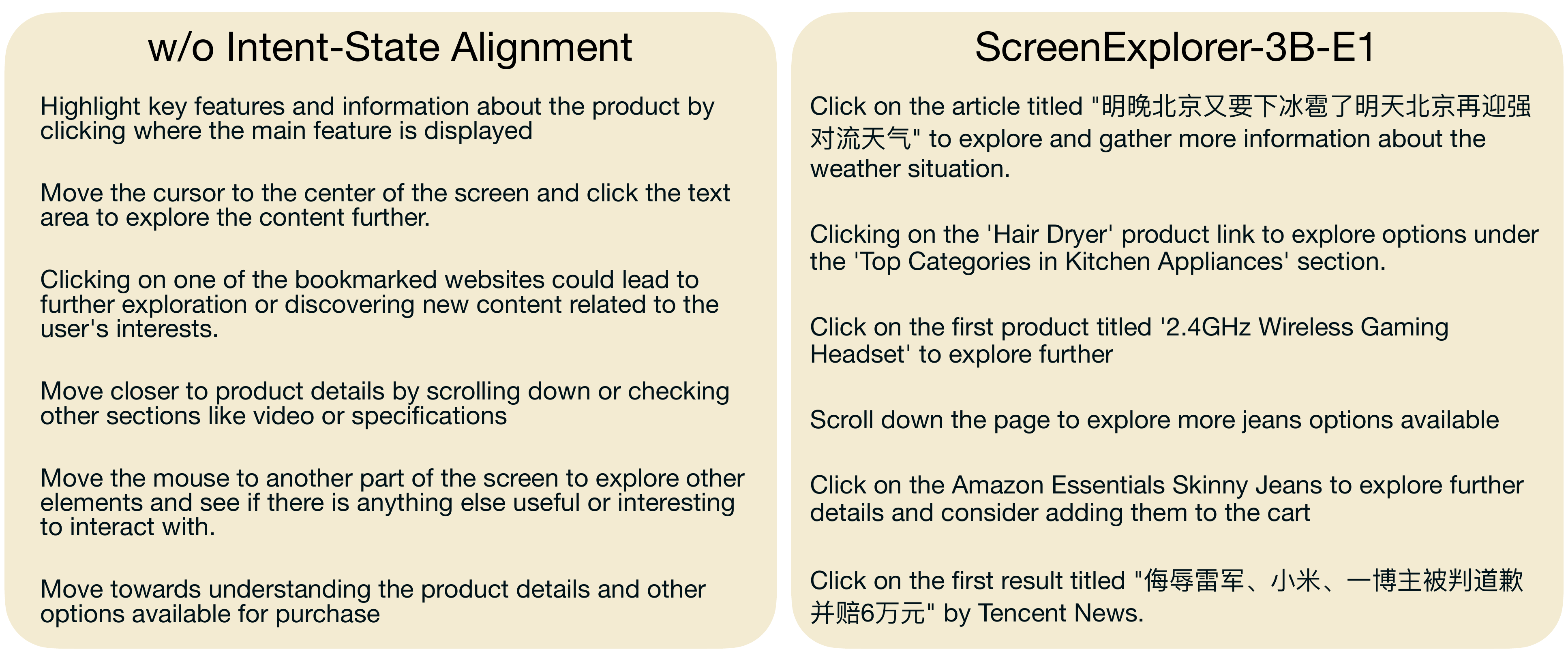}
    \caption{Intent descriptions generated by \textit{ScreenExplorer-3B-E1} and ablation \textbf{w/o Intent-State Alignment} group. the \textit{ScreenExplorer-3B-E1} model demonstrates a stronger propensity to reference existing on-screen text when generating intents, resulting in more specific and contextually grounded intentions. In contrast, the \textbf{w/o Intent-State Alignment} group generates more generalized intents with less specific task objectives and minimal reference to on-screen textual content.}
    \label{fig:intent-example}
\end{figure}

Another notable finding is that when Intent-State Alignment rewards were incorporated during training, the model exhibited a stronger tendency to reference existing on-screen text in its intents, thereby enhancing the correlation between intent descriptions and screen content. Figure \ref{fig:intent-example} presents a comparative analysis of intent descriptions generated by \textit{ScreenExplorer-3B-E1} and the model in \textbf{w/o Intent-State Alignment} ablation setting, where the latter excludes both $r_{des}$ and $r_{inter}$ reward components.

The ablation studies indicate that among all exploration rewards, the World Model reward is the most crucial, followed by the Instantaneous Change Reward. Furthermore, the incorporation of additional rewards can effectively modify the model's exploration preferences and Chain-of-Thought output patterns, providing valuable insights for future reward design and the construction of task-specific GUI datasets.

\FloatBarrier
\section{Case Study}\label{appendix:case-study}
\begin{figure}[h]
  \centering
 \begin{tabular}{
    @{}
    m{0.195\textwidth}@{\hspace{1pt}}
    m{0.195\textwidth}@{\hspace{1pt}}
    m{0.195\textwidth}@{\hspace{1pt}}
    m{0.195\textwidth}@{\hspace{1pt}}
    m{0.195\textwidth}@{}
  }
    \includegraphics[width=\linewidth]{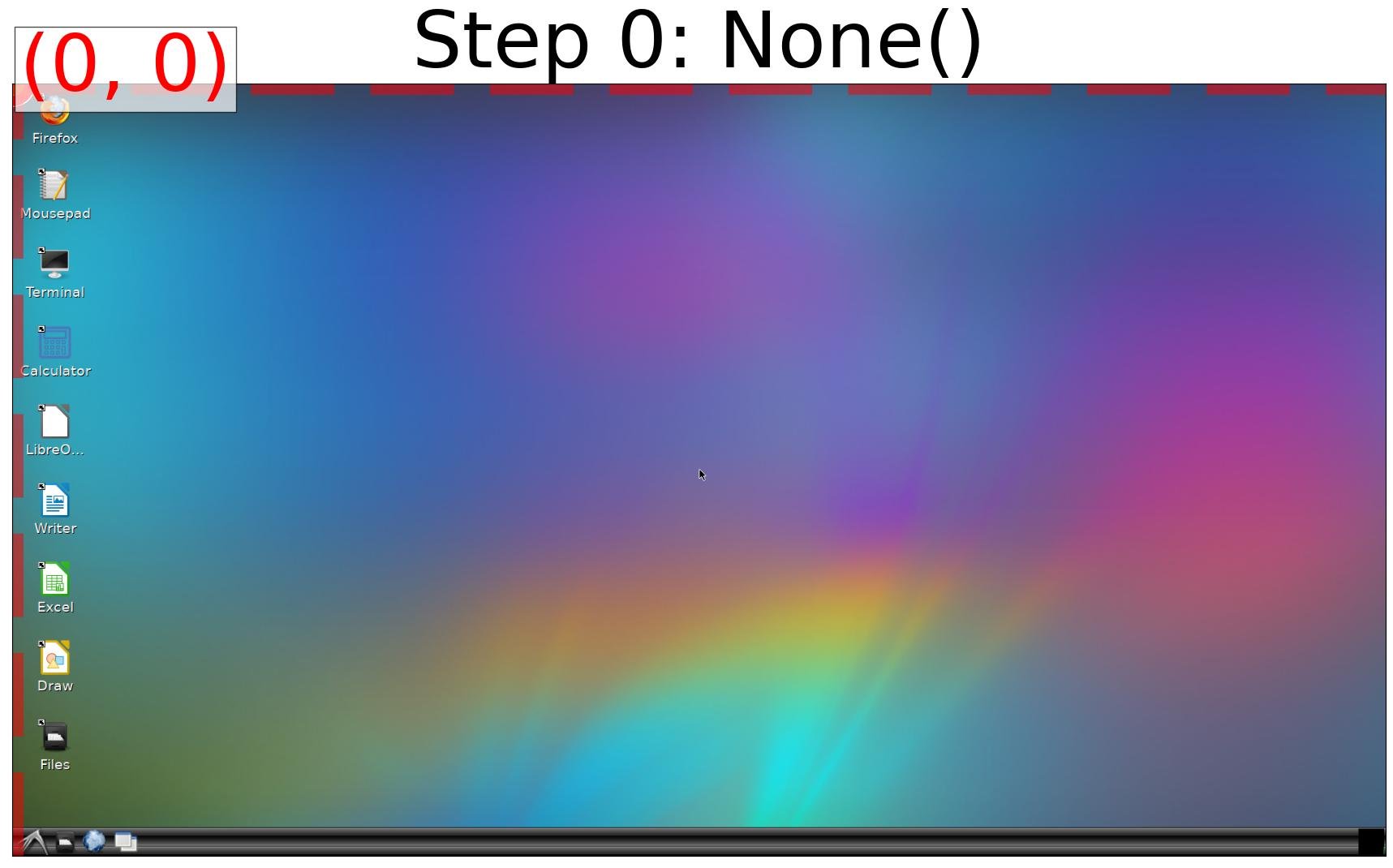} &
    \includegraphics[width=\linewidth]{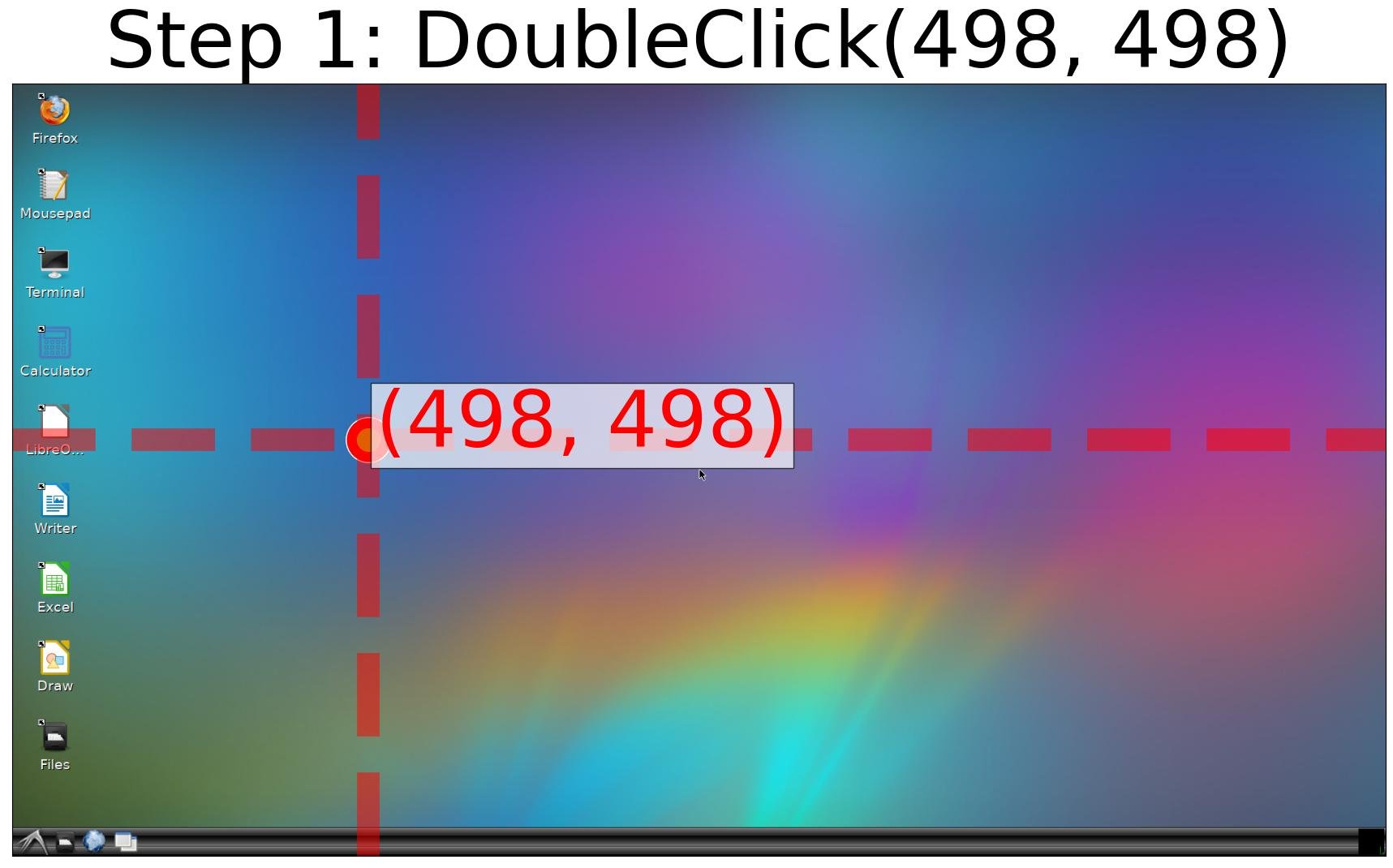} &
    \includegraphics[width=\linewidth]{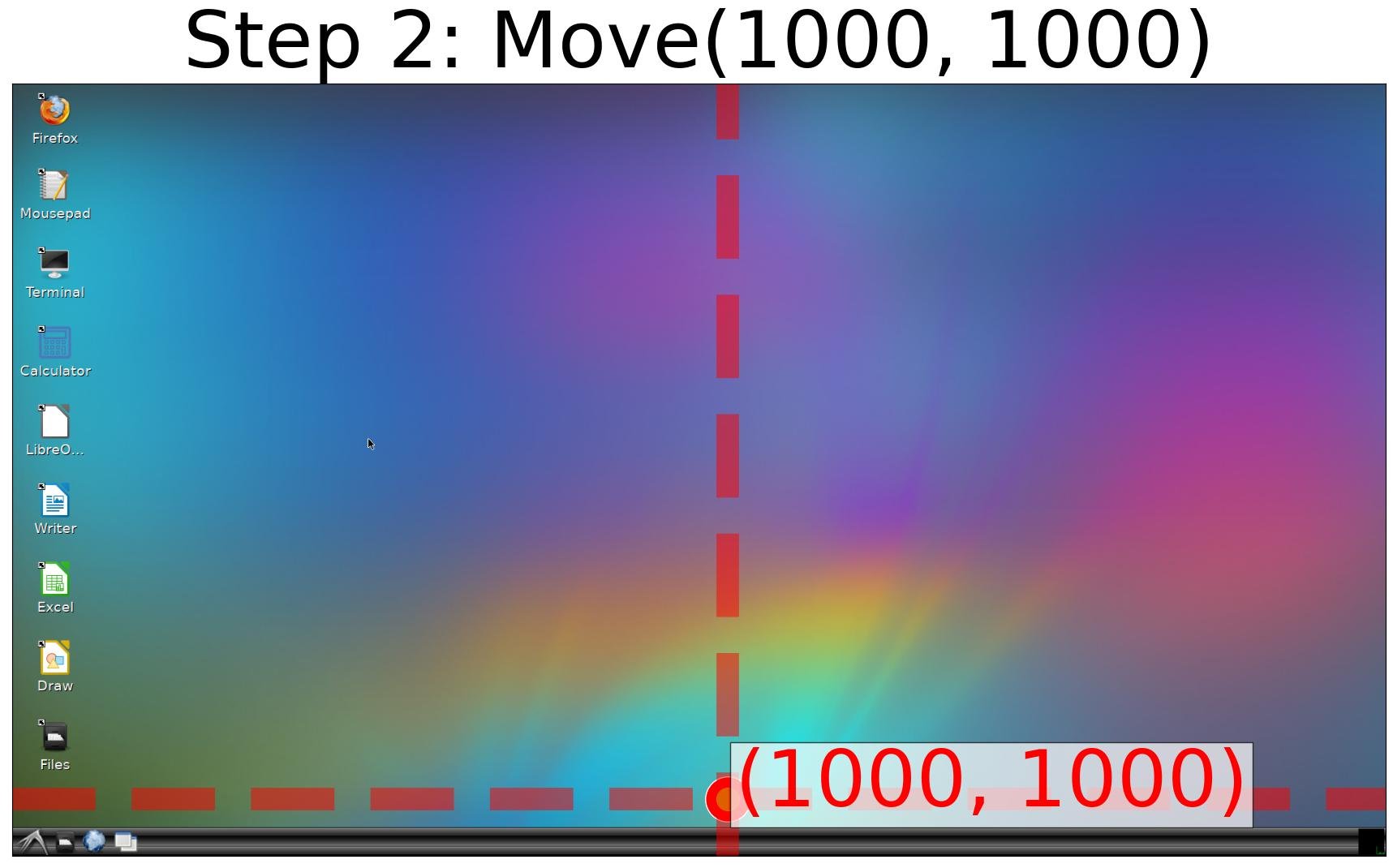} &
    \includegraphics[width=\linewidth]{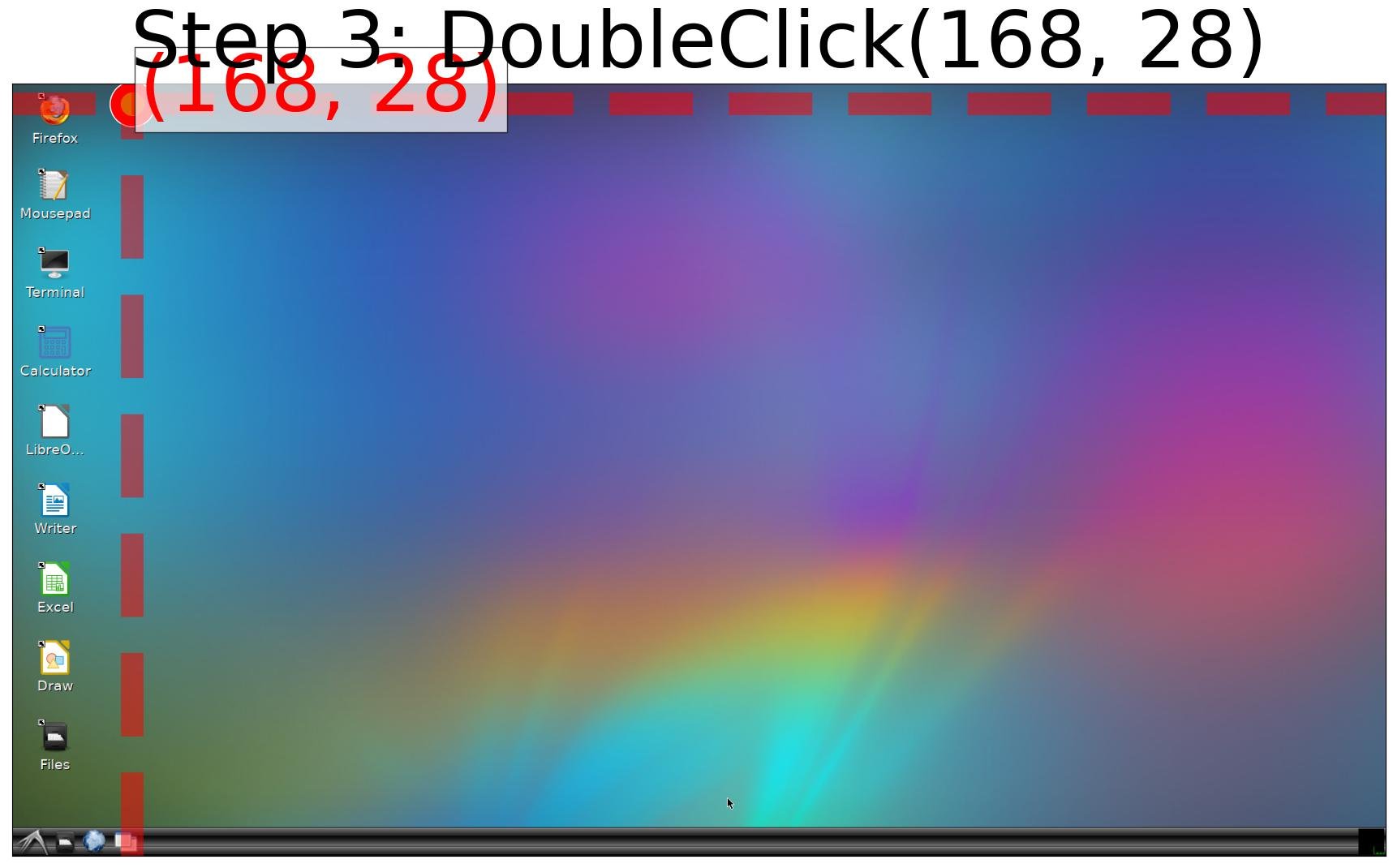} &
    \includegraphics[width=\linewidth]{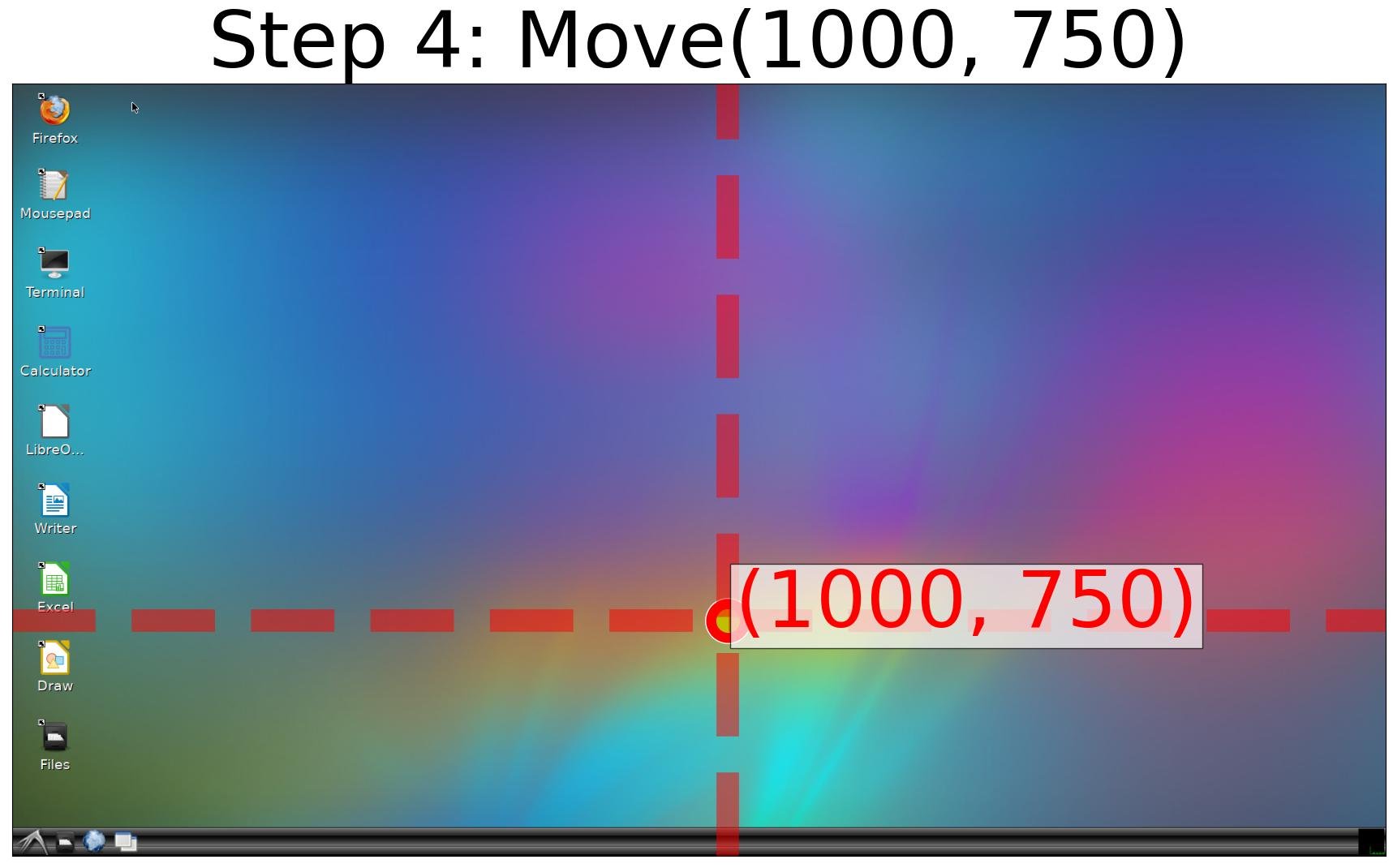} \\
    \includegraphics[width=\linewidth]{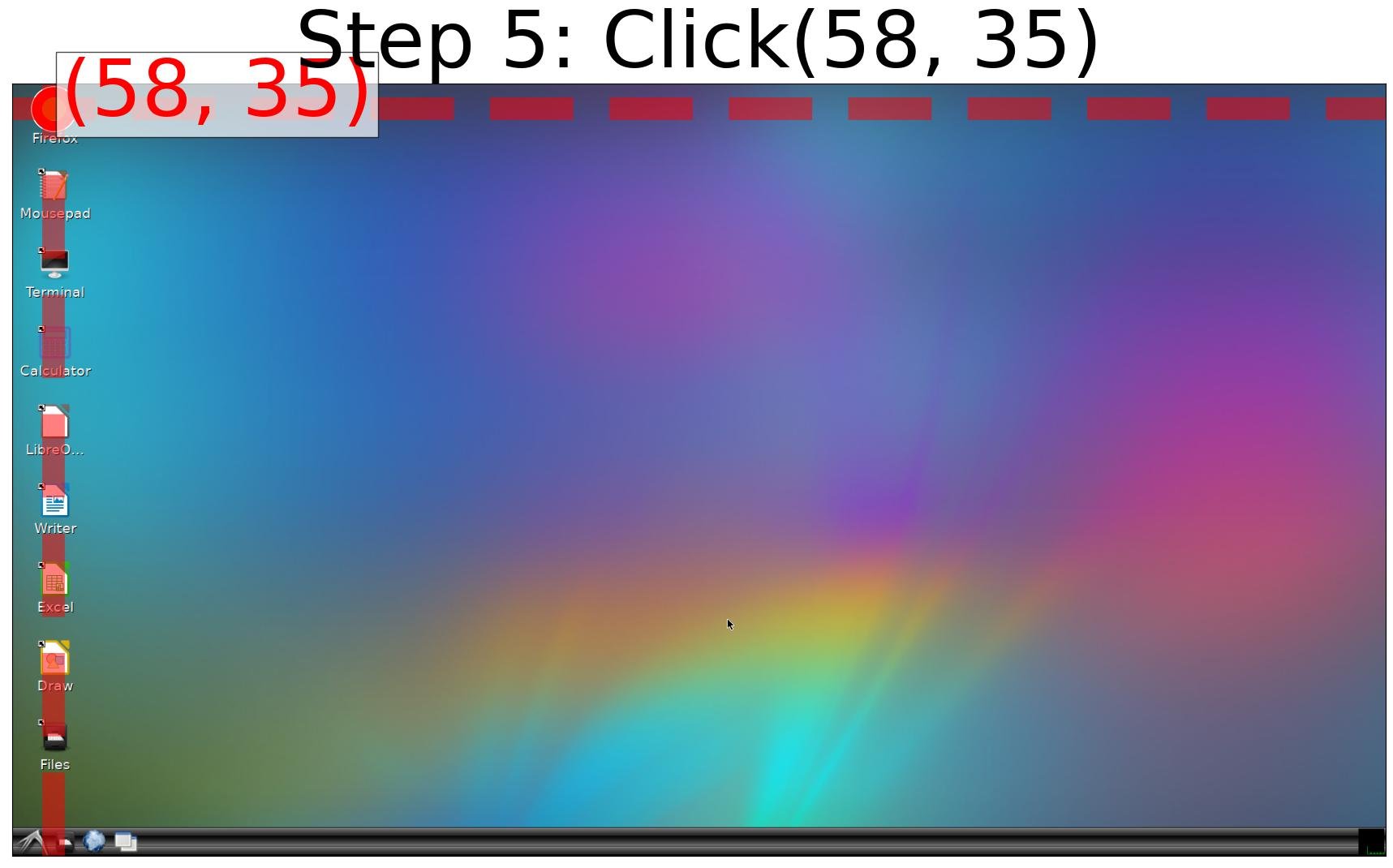} &
    \includegraphics[width=\linewidth]{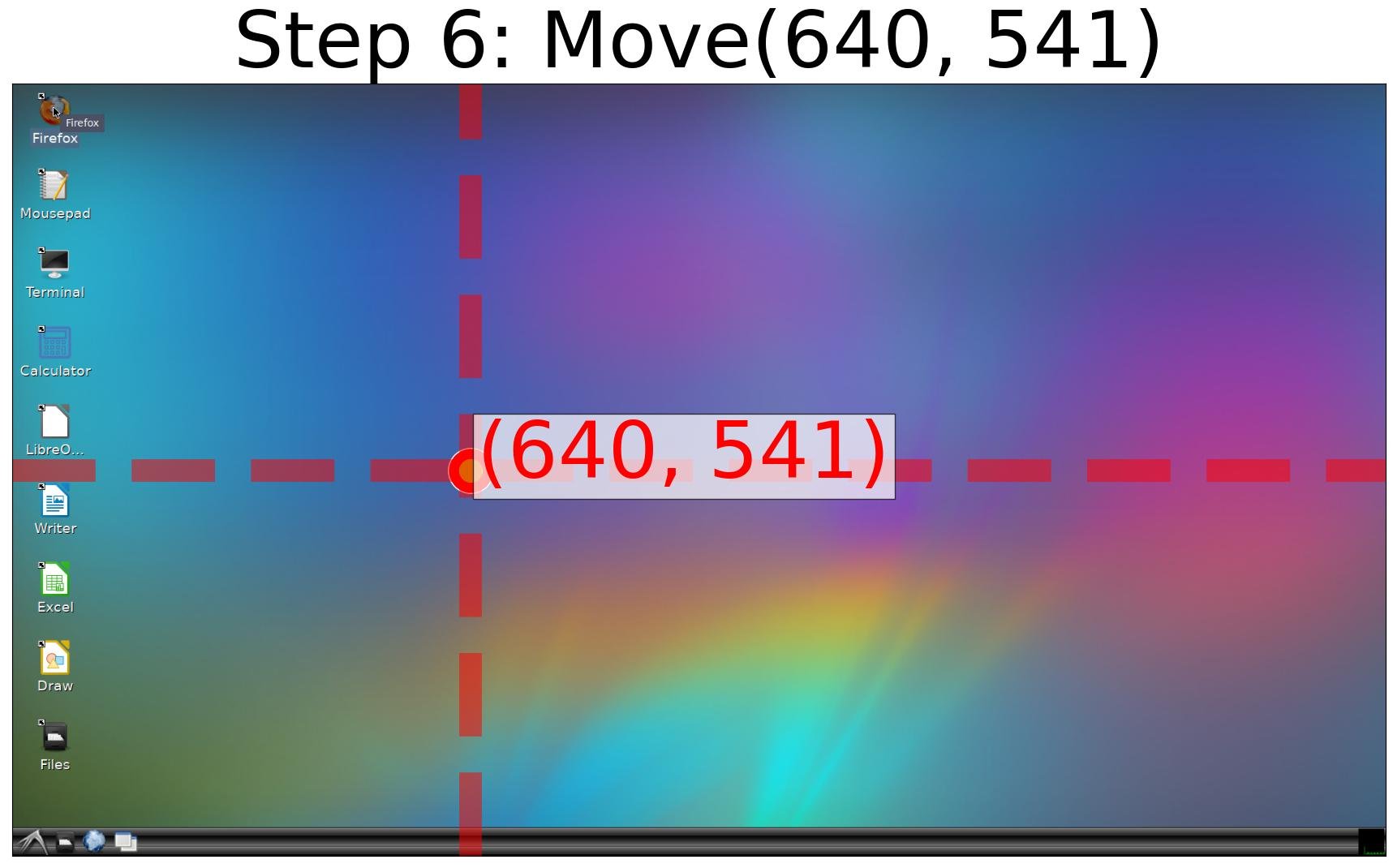} &
    \includegraphics[width=\linewidth]{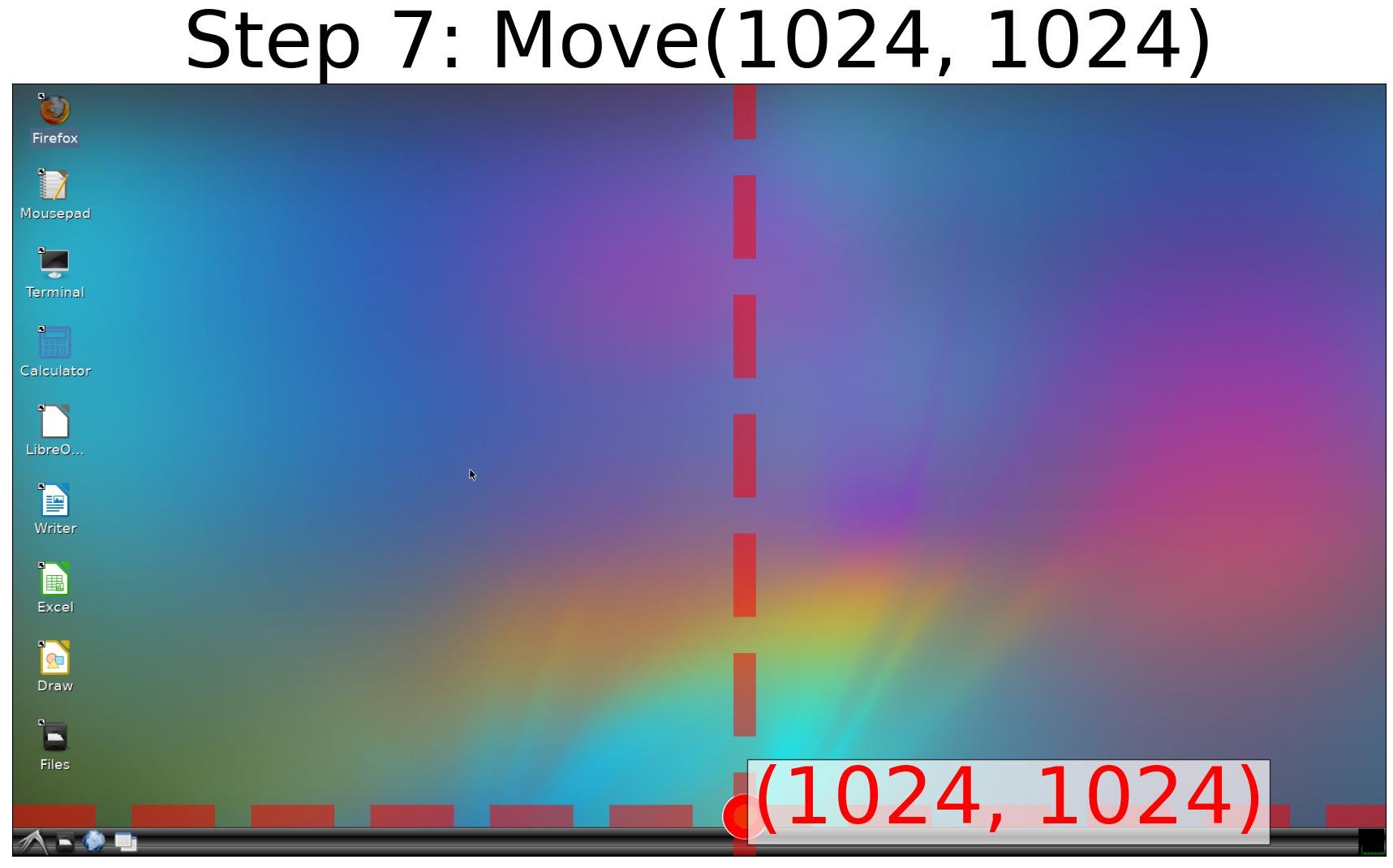} &
    \includegraphics[width=\linewidth]{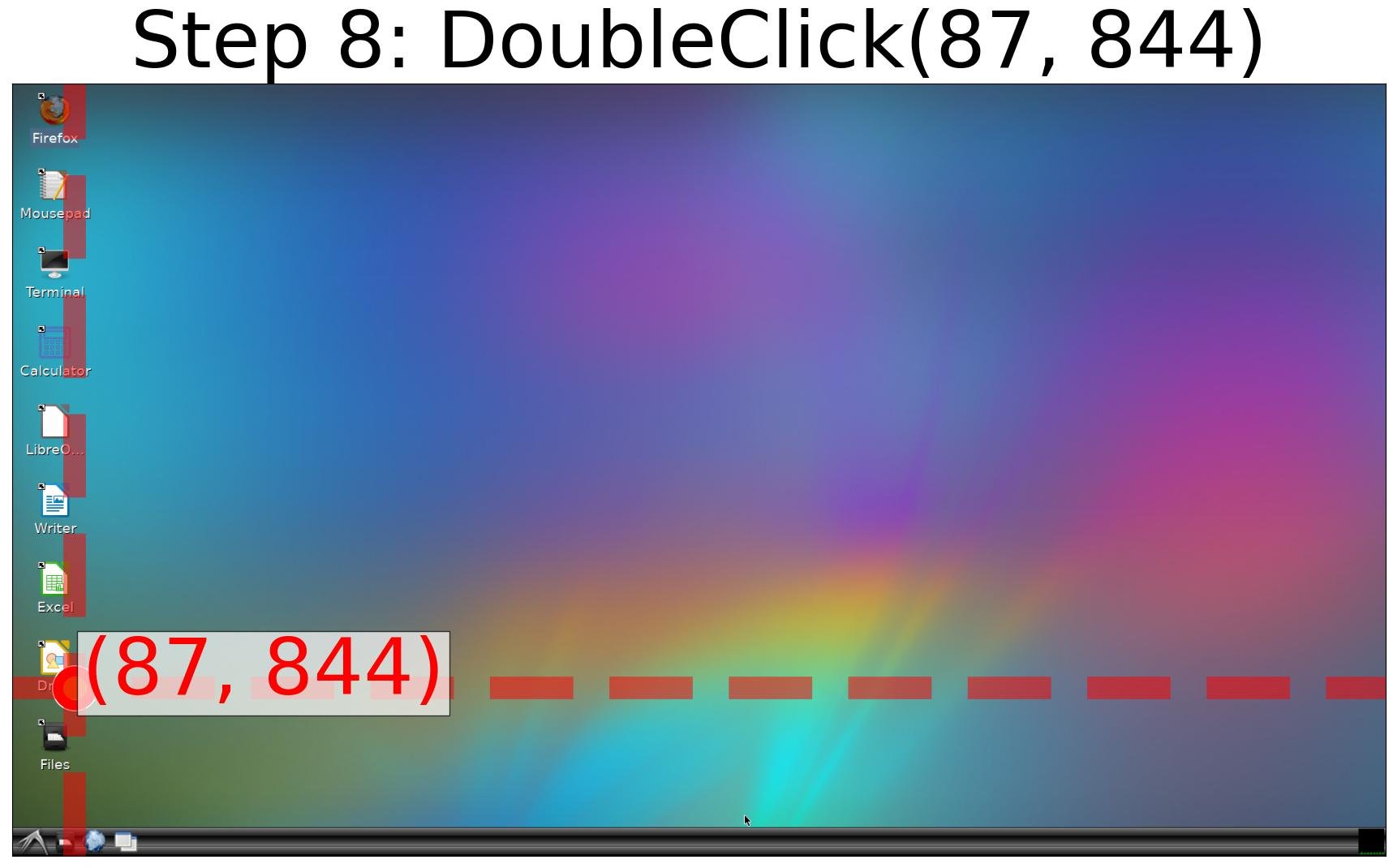} &
    \includegraphics[width=\linewidth]{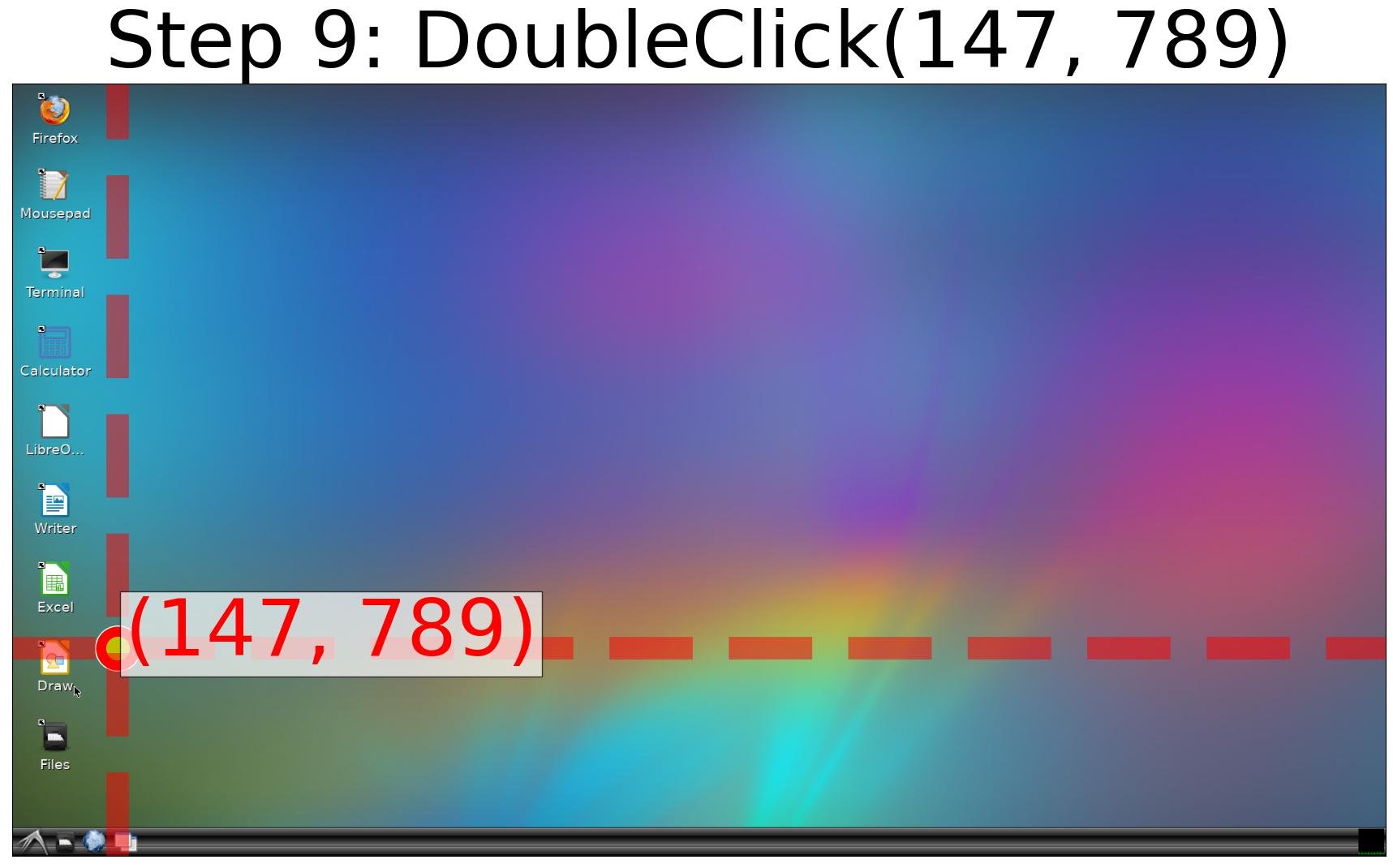}
  \end{tabular}
  \caption{The original \textit{Qwen2.5-VL-3B} model: The initial Qwen2.5-VL-3B model generated near-random coordinates, making it difficult to accurately click on icons and effectively interact with the environment, resulting in extremely low exploration rewards and diversity metrics.}\label{fig:case-study-1}
\end{figure}

\begin{figure}[h]
  \centering
 \begin{tabular}{
    @{}
    m{0.195\textwidth}@{\hspace{1pt}}
    m{0.195\textwidth}@{\hspace{1pt}}
    m{0.195\textwidth}@{\hspace{1pt}}
    m{0.195\textwidth}@{\hspace{1pt}}
    m{0.195\textwidth}@{}
  }
    \includegraphics[width=\linewidth]{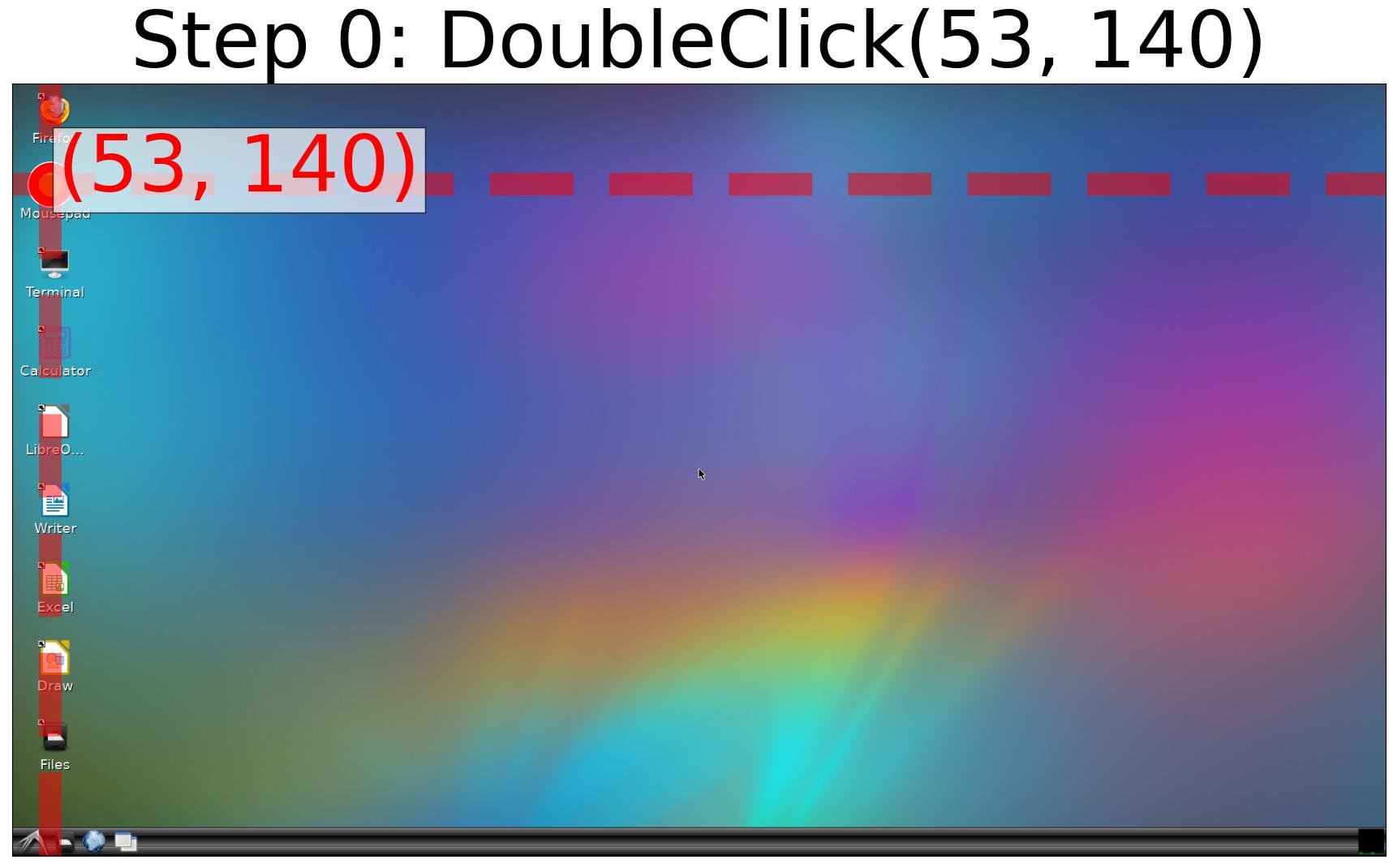} &
    \includegraphics[width=\linewidth]{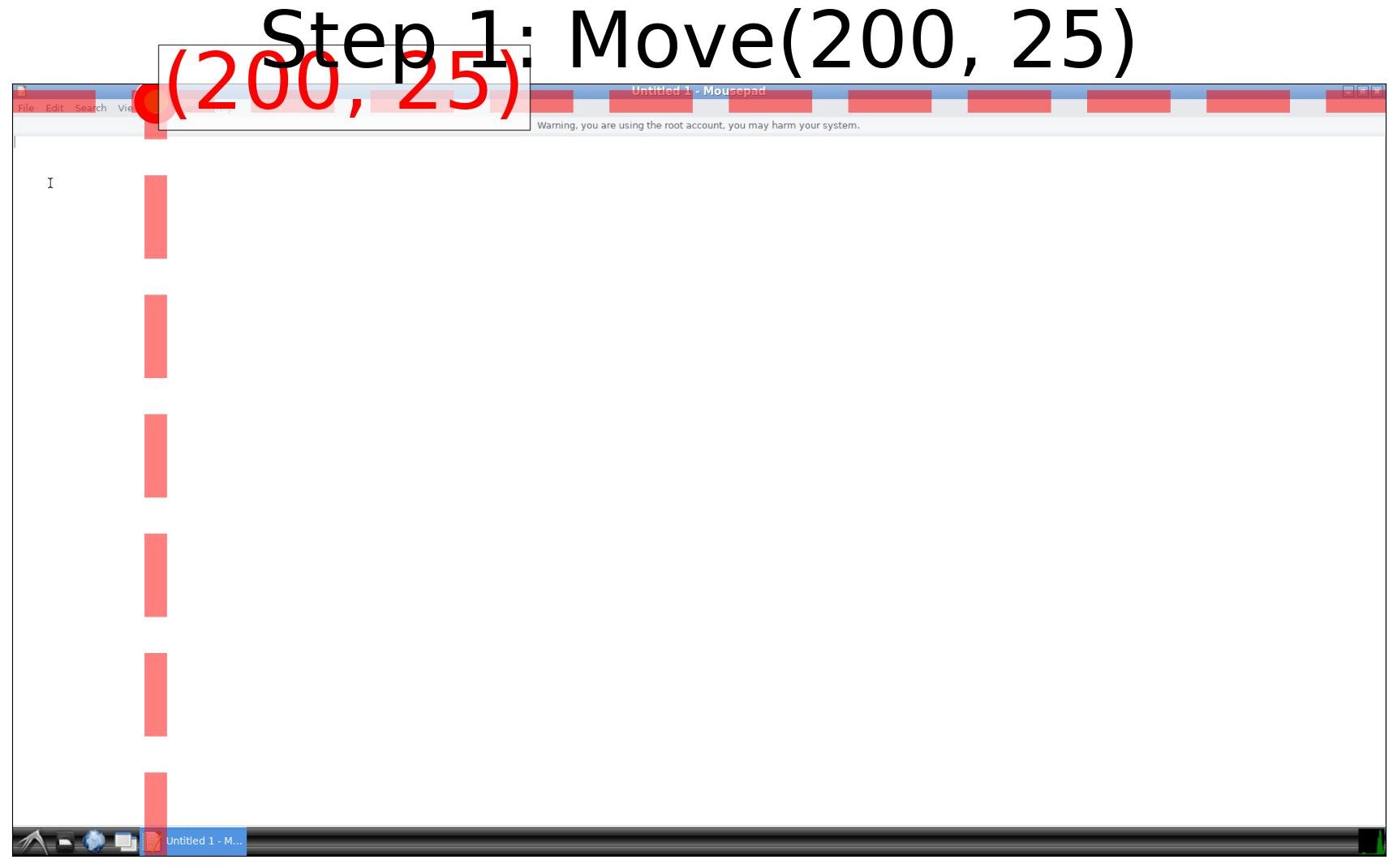} &
    \includegraphics[width=\linewidth]{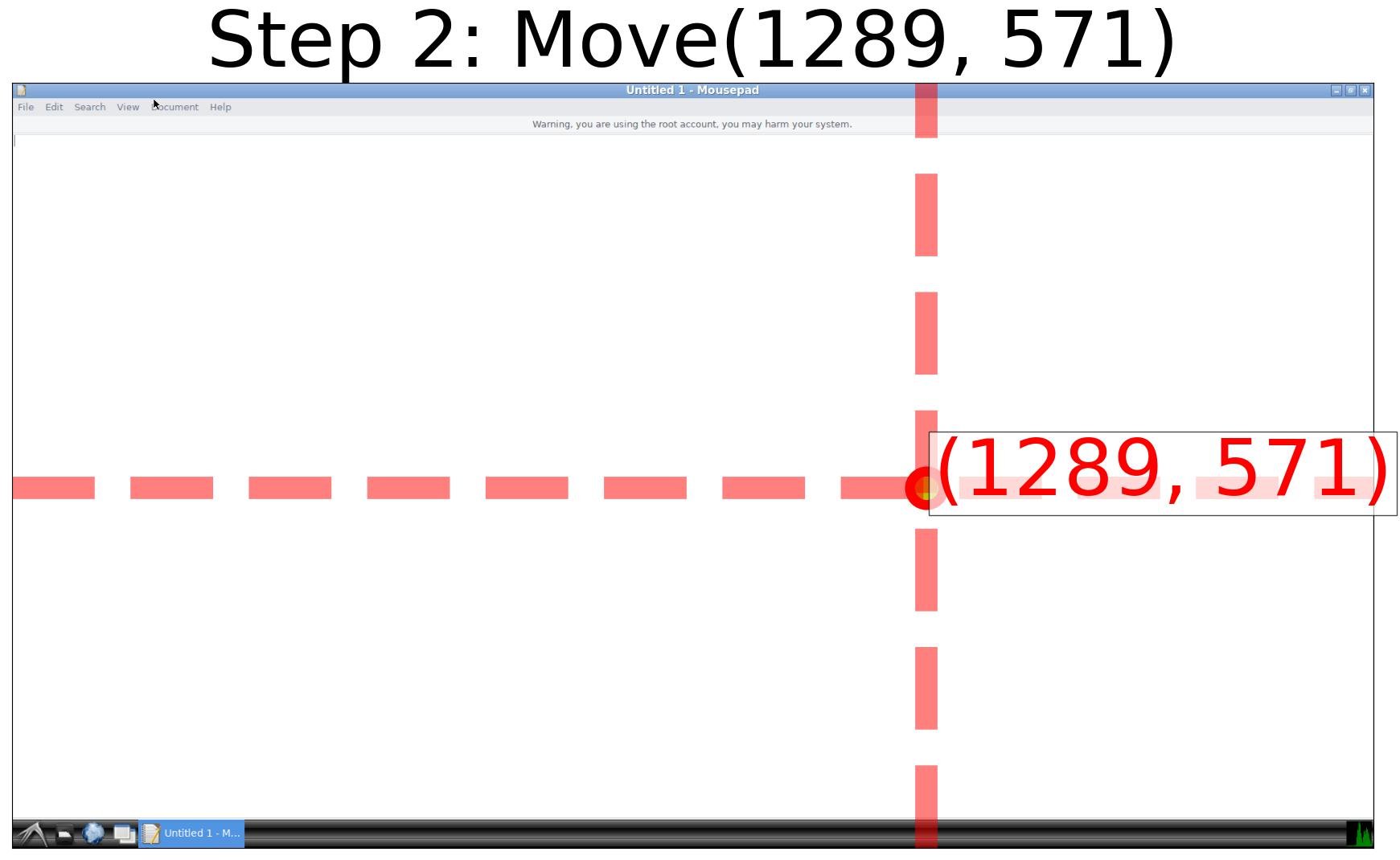} &
    \includegraphics[width=\linewidth]{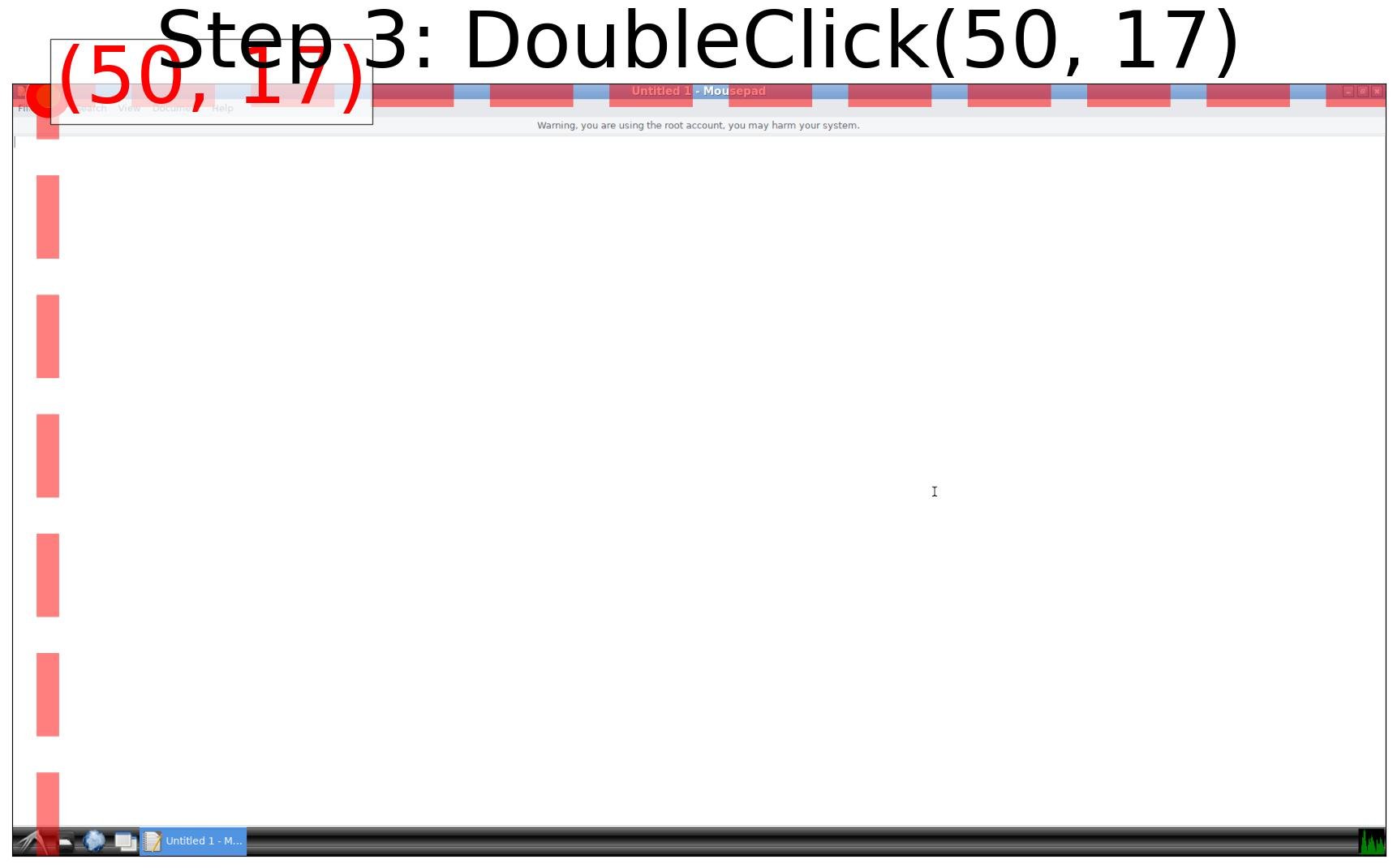} &
    \includegraphics[width=\linewidth]{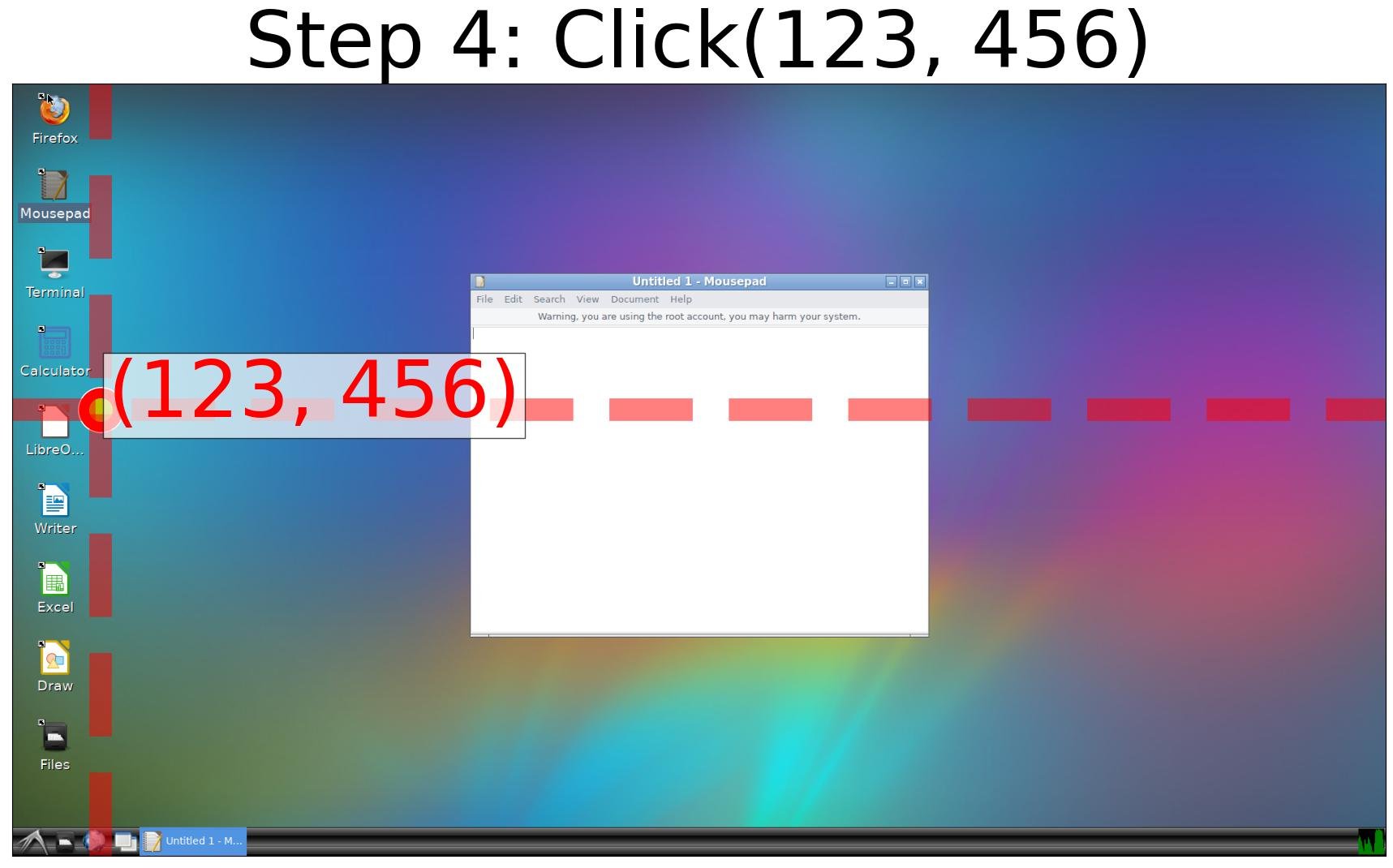} \\
    \includegraphics[width=\linewidth]{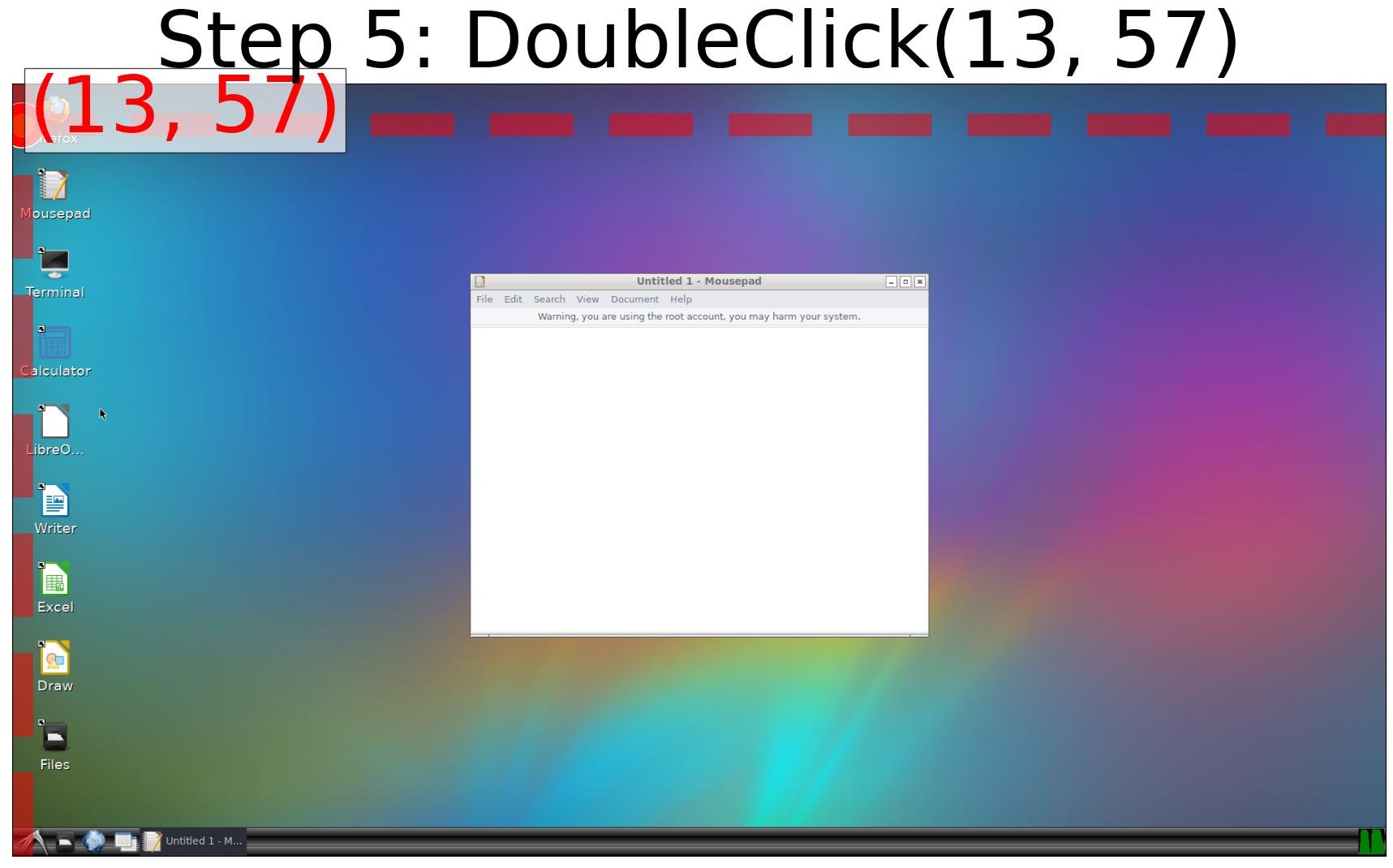} &
    \includegraphics[width=\linewidth]{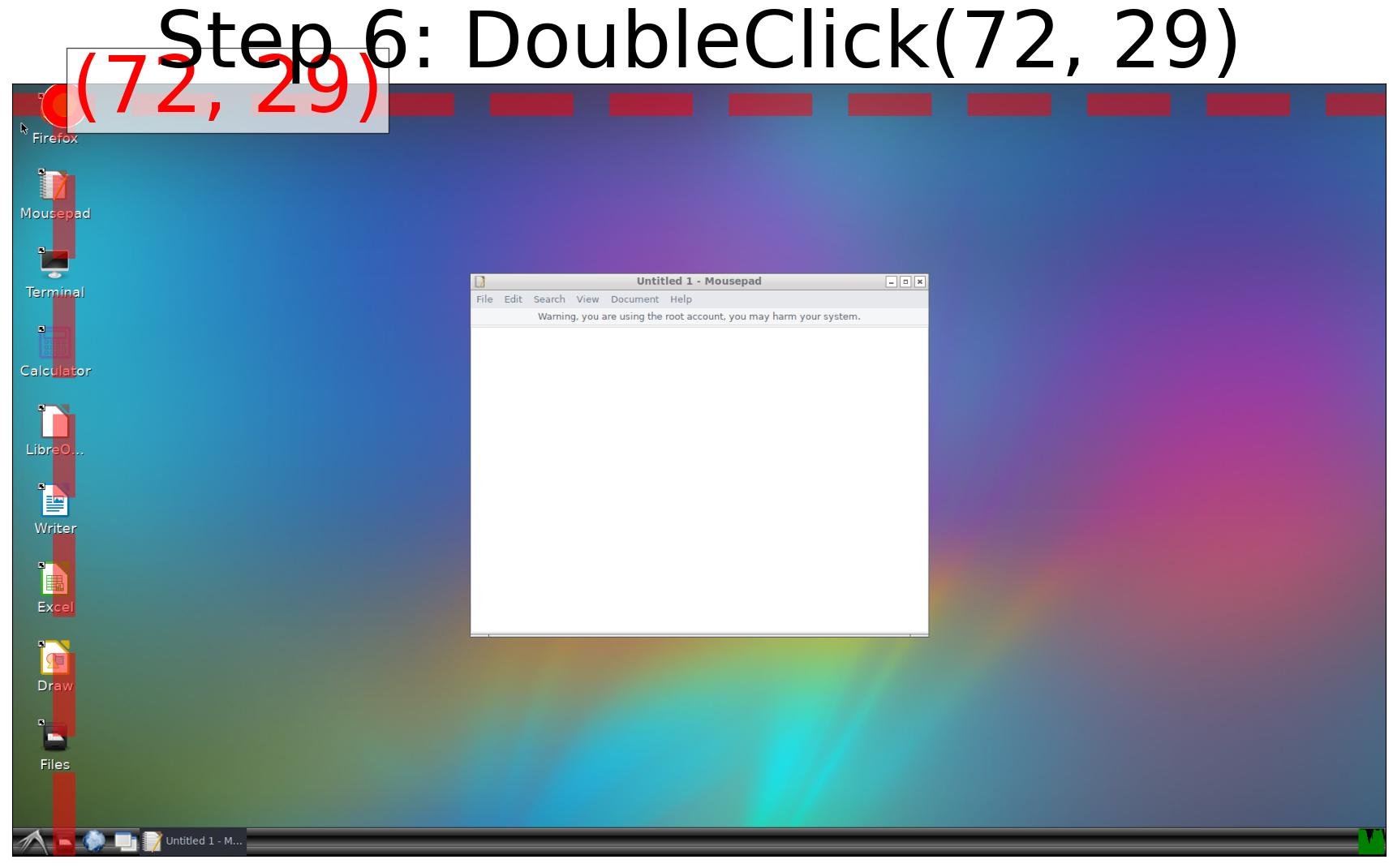} &
    \includegraphics[width=\linewidth]{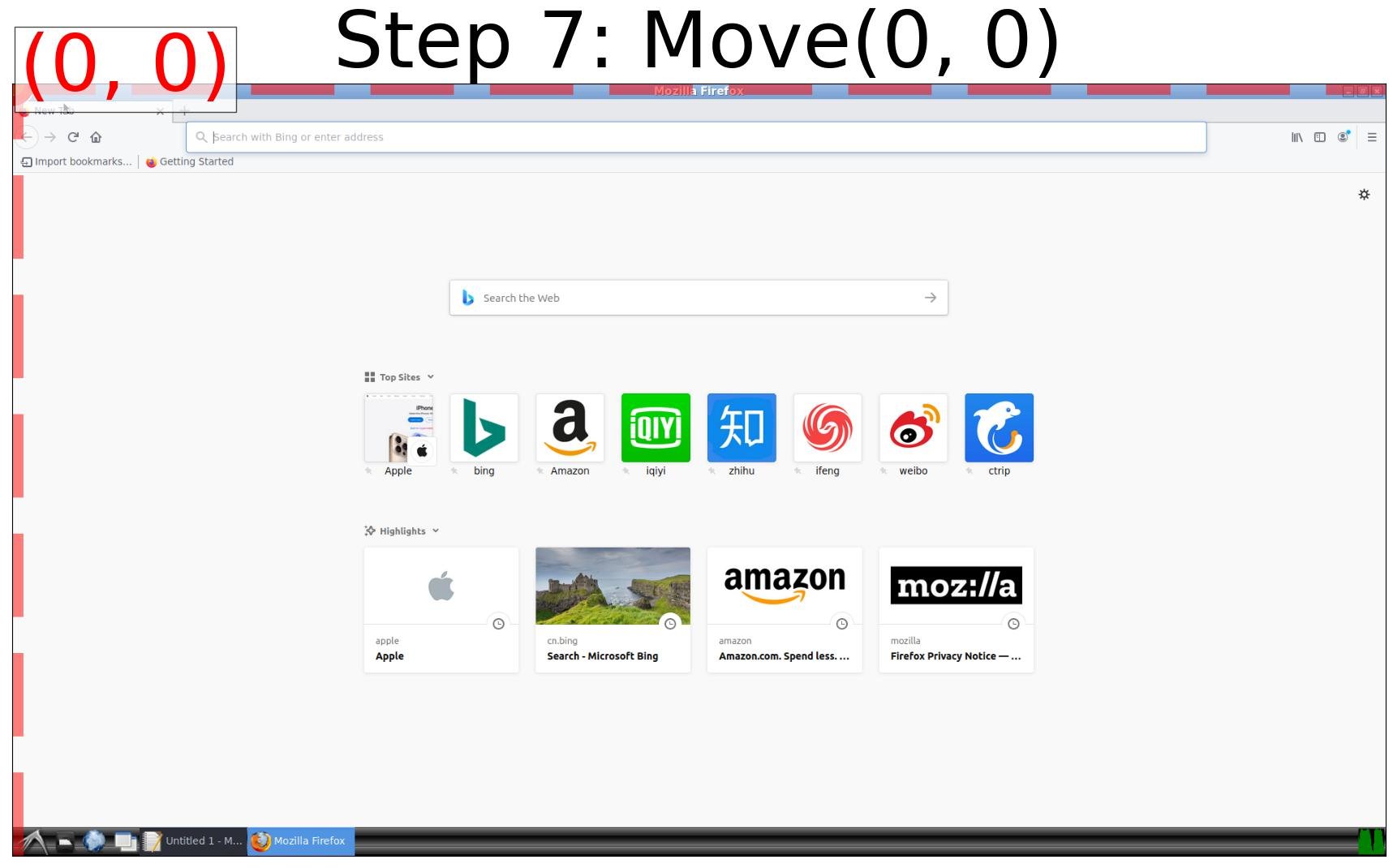} &
    \includegraphics[width=\linewidth]{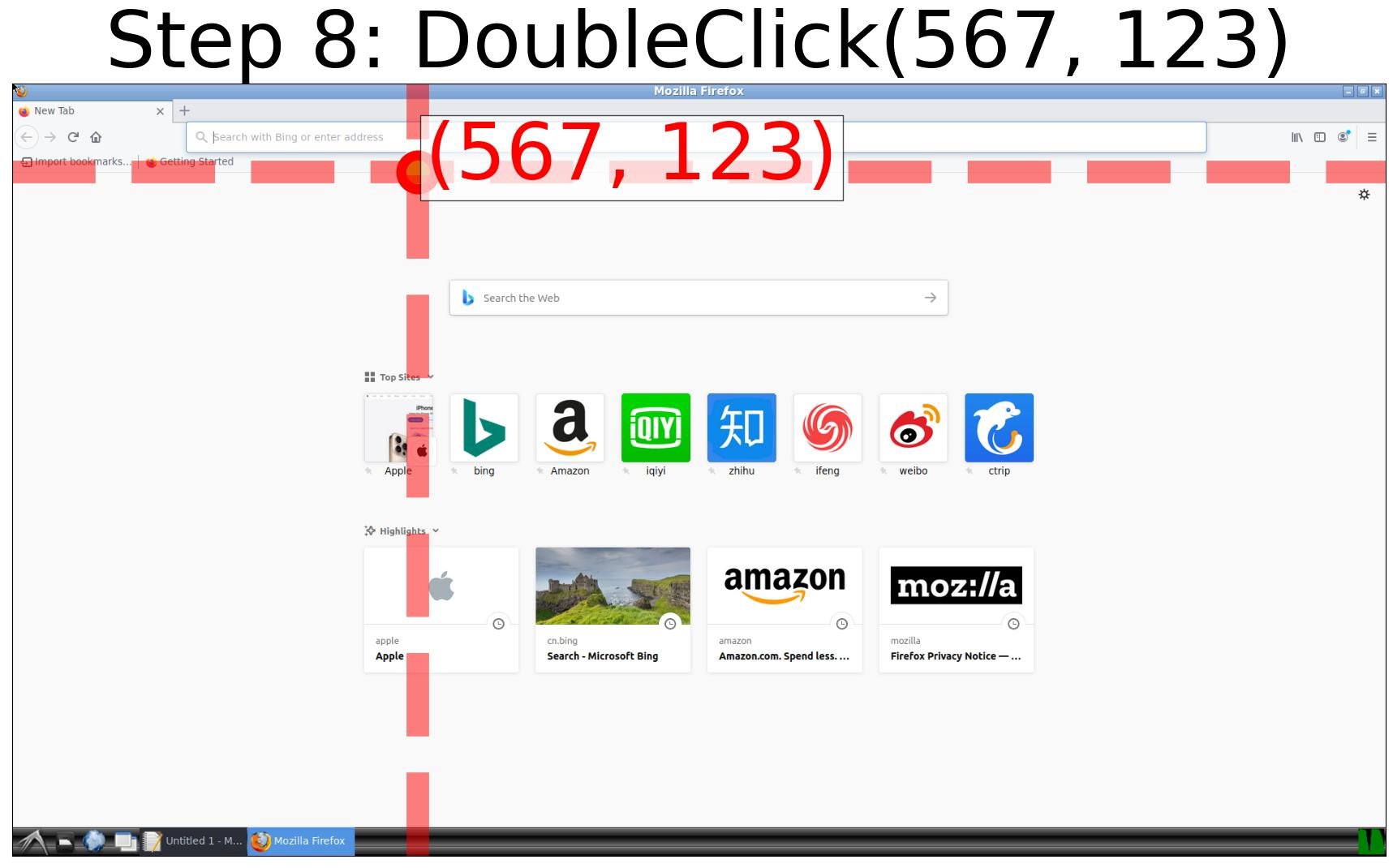} &
    \includegraphics[width=\linewidth]{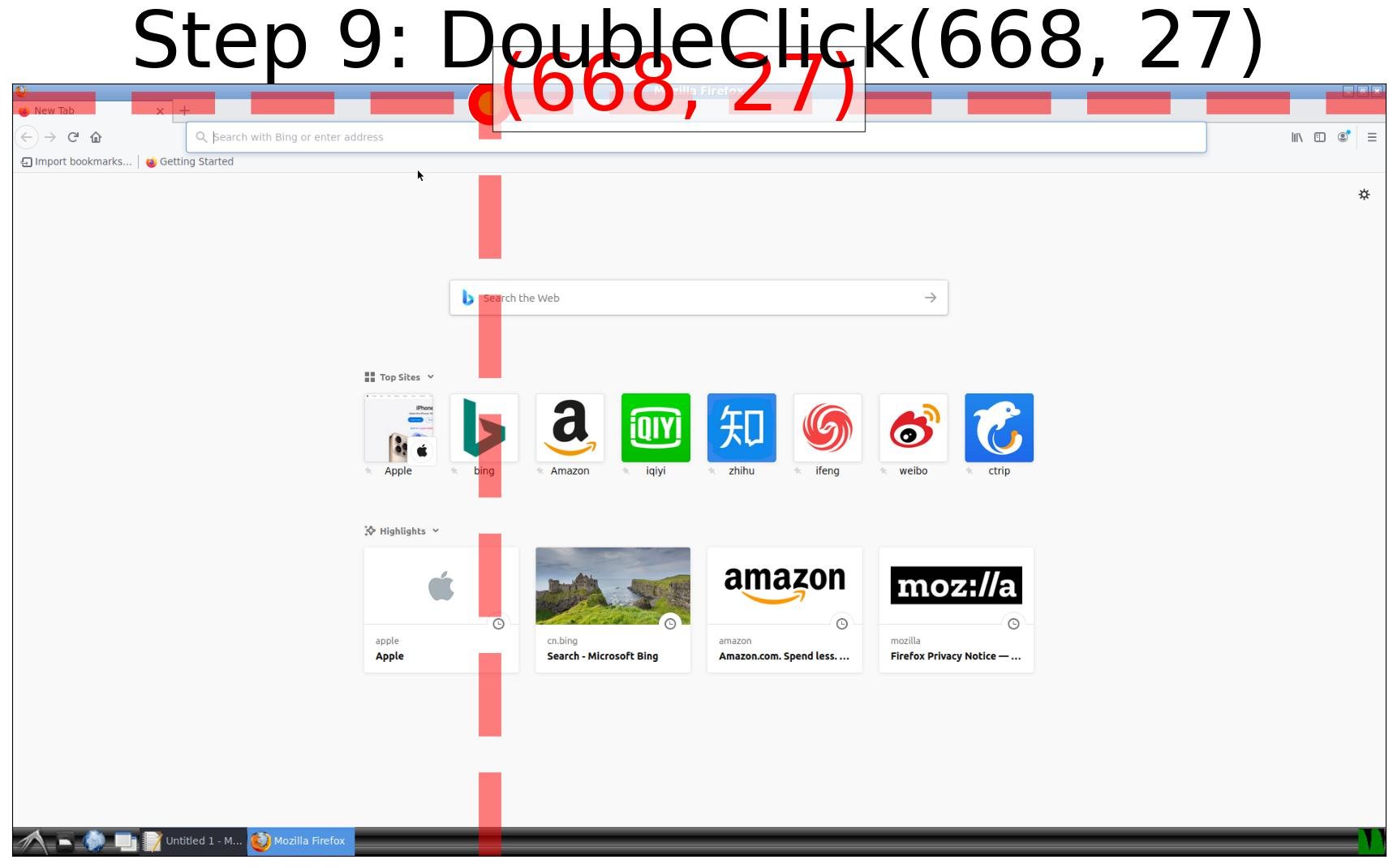}
  \end{tabular}
  \caption{Episode-50 of \textit{ScreenExplorer-3B-E1}: The model demonstrated the capability to output commands for launching desktop applications, resulting in significant screen state changes, though it had not yet developed exploration behaviors in specific apps.}\label{fig:case-study-2}
\end{figure}

\begin{figure}[h]
  \centering
 \begin{tabular}{
    @{}
    m{0.195\textwidth}@{\hspace{1pt}}
    m{0.195\textwidth}@{\hspace{1pt}}
    m{0.195\textwidth}@{\hspace{1pt}}
    m{0.195\textwidth}@{\hspace{1pt}}
    m{0.195\textwidth}@{}
  }
    \includegraphics[width=\linewidth]{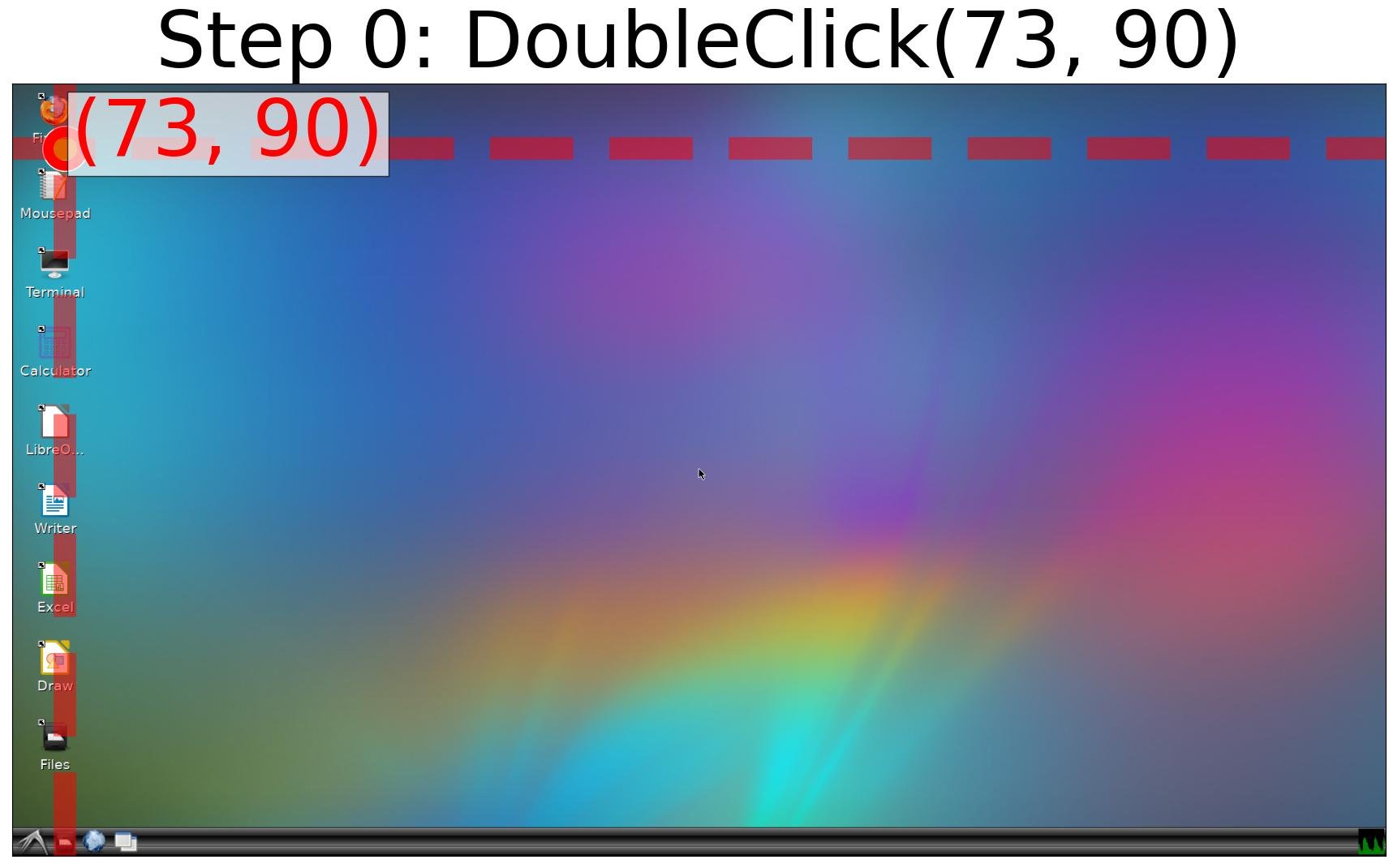} &
    \includegraphics[width=\linewidth]{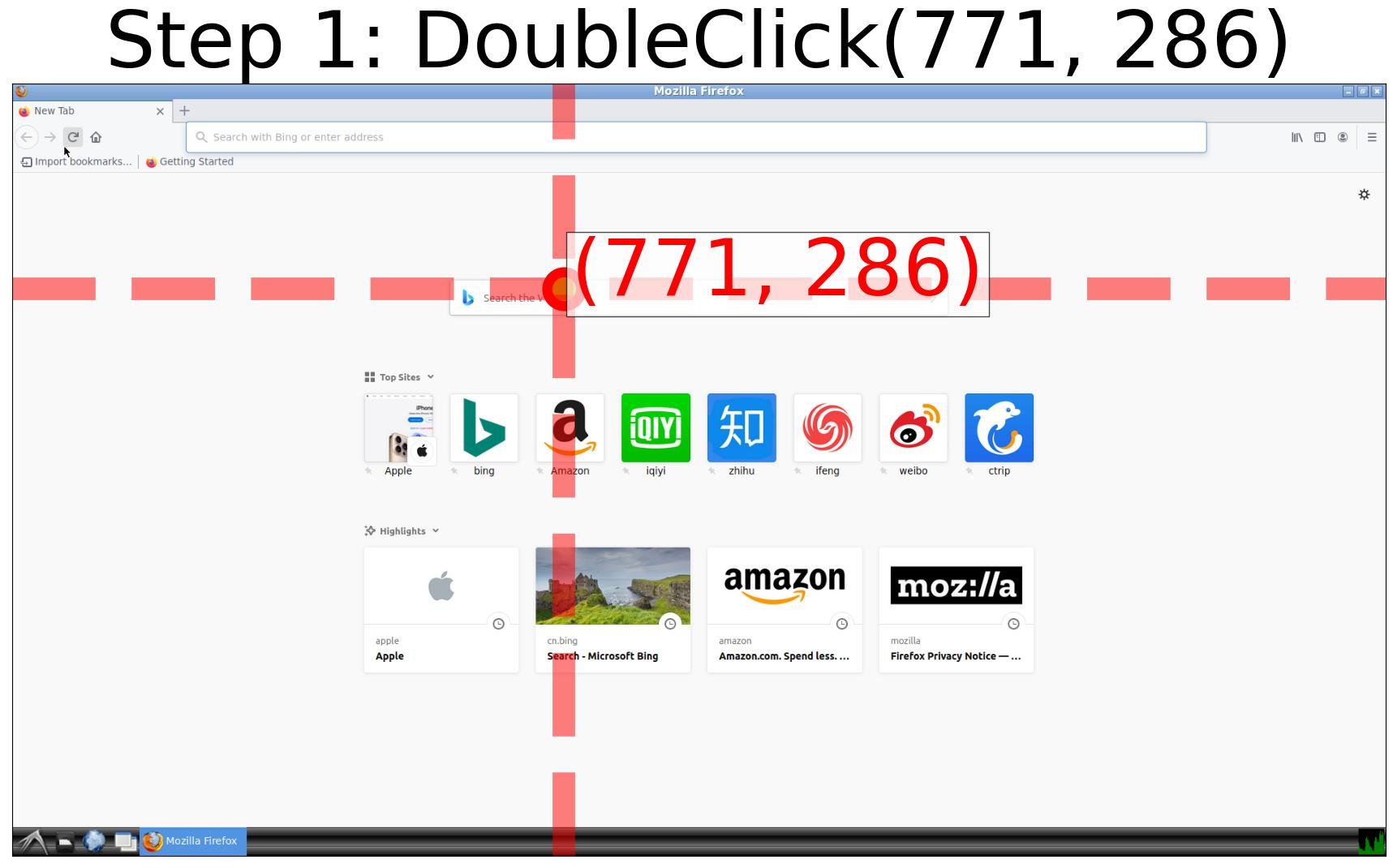} &
    \includegraphics[width=\linewidth]{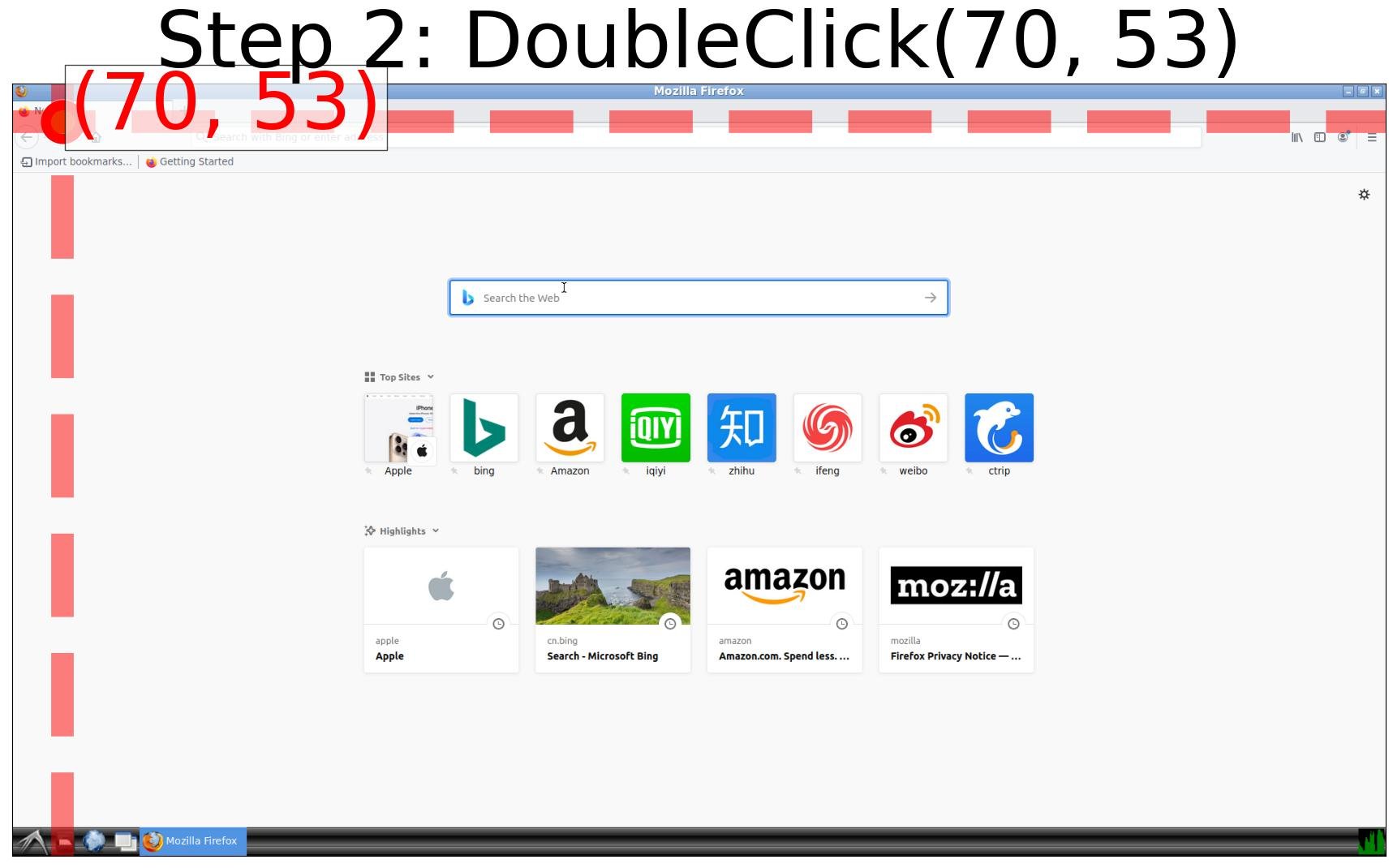} &
    \includegraphics[width=\linewidth]{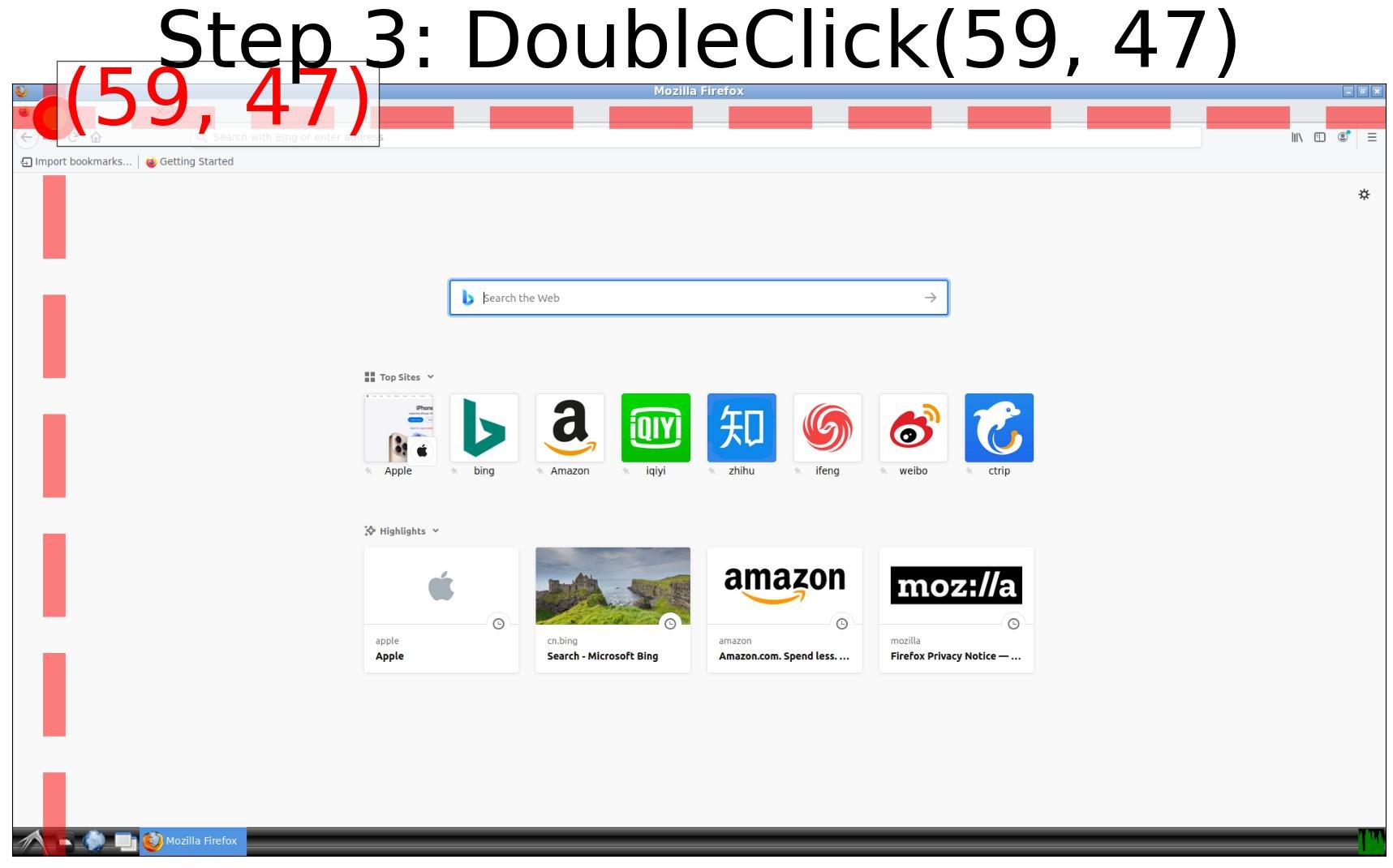} &
    \includegraphics[width=\linewidth]{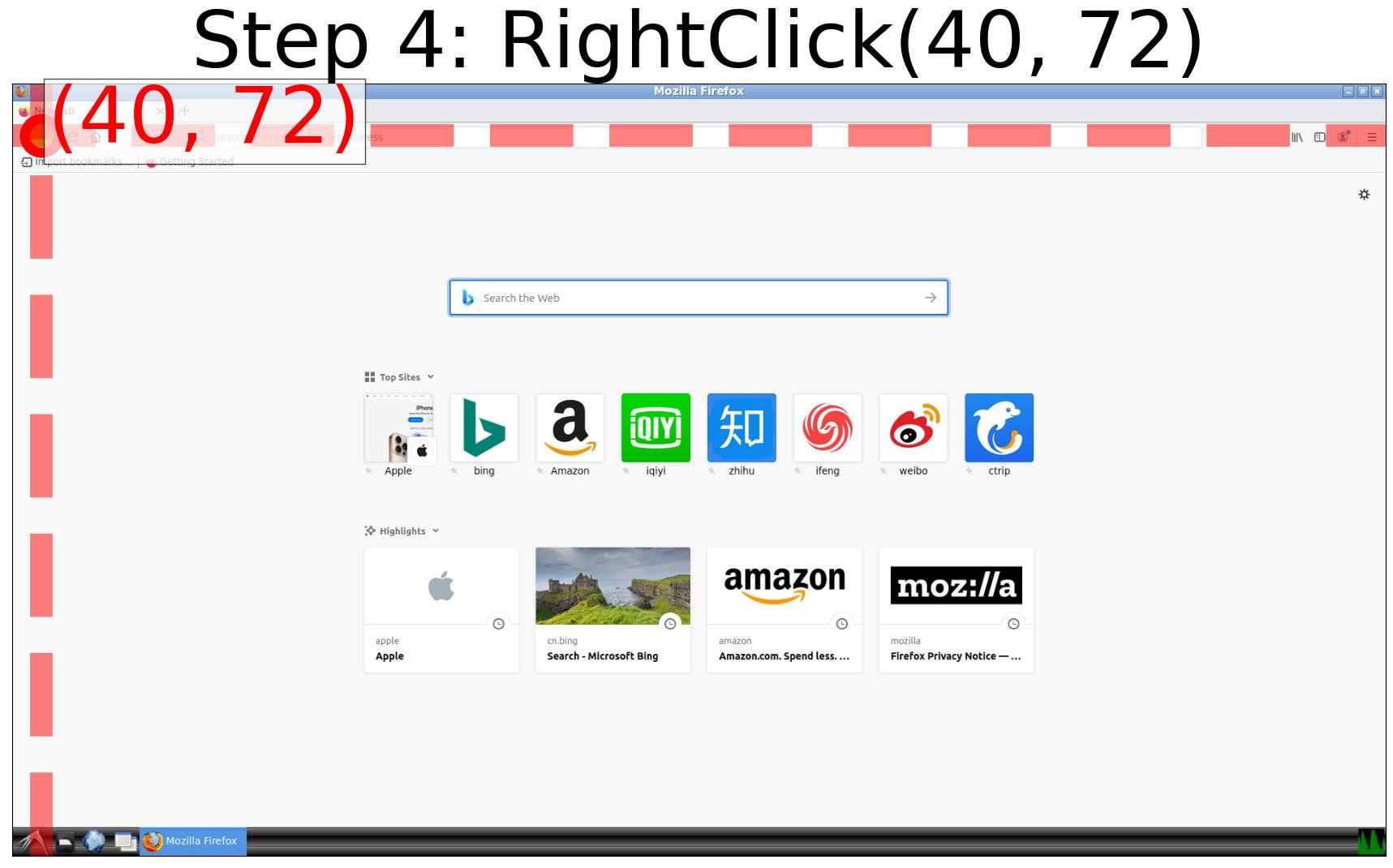} \\
    \includegraphics[width=\linewidth]{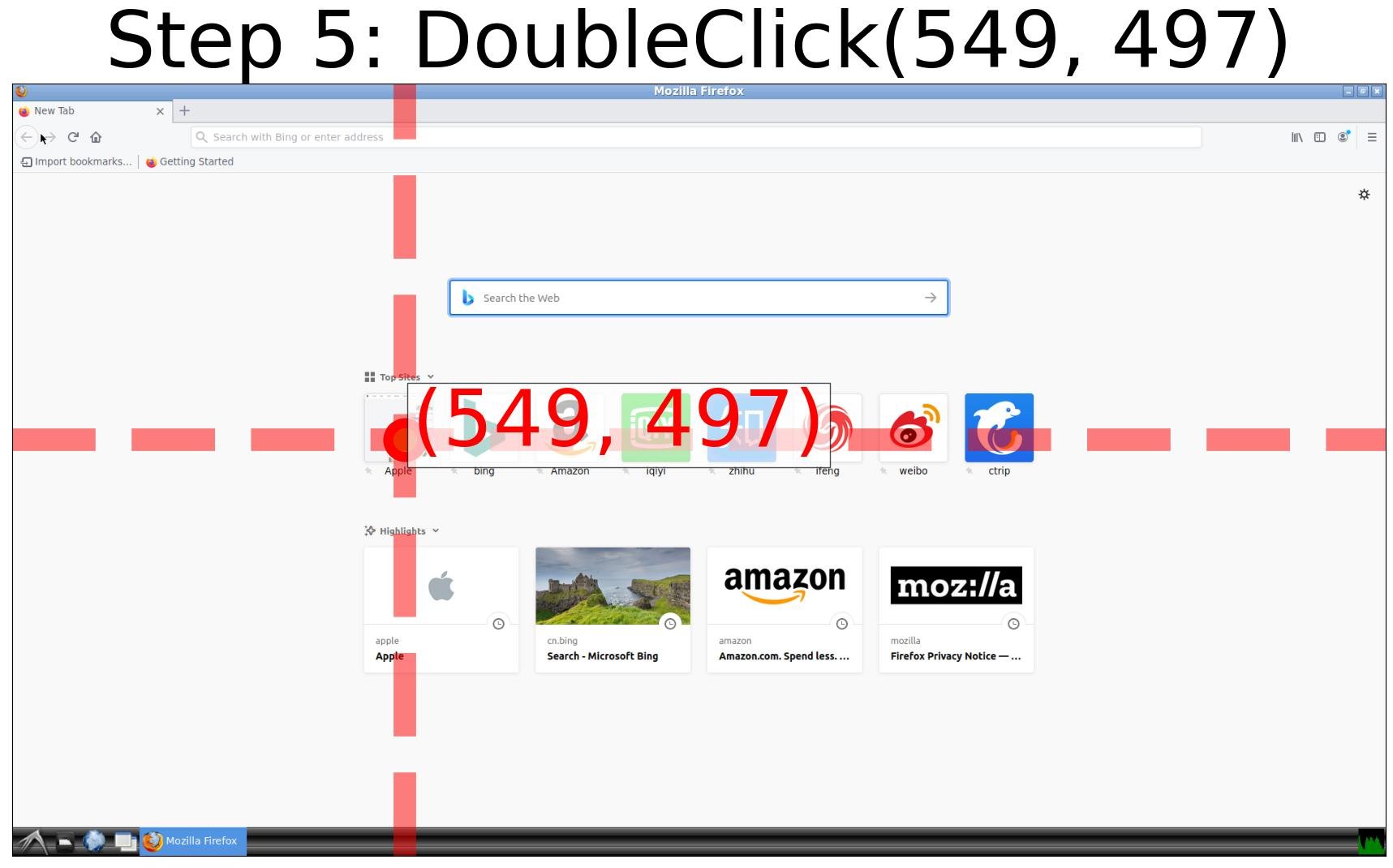} &
    \includegraphics[width=\linewidth]{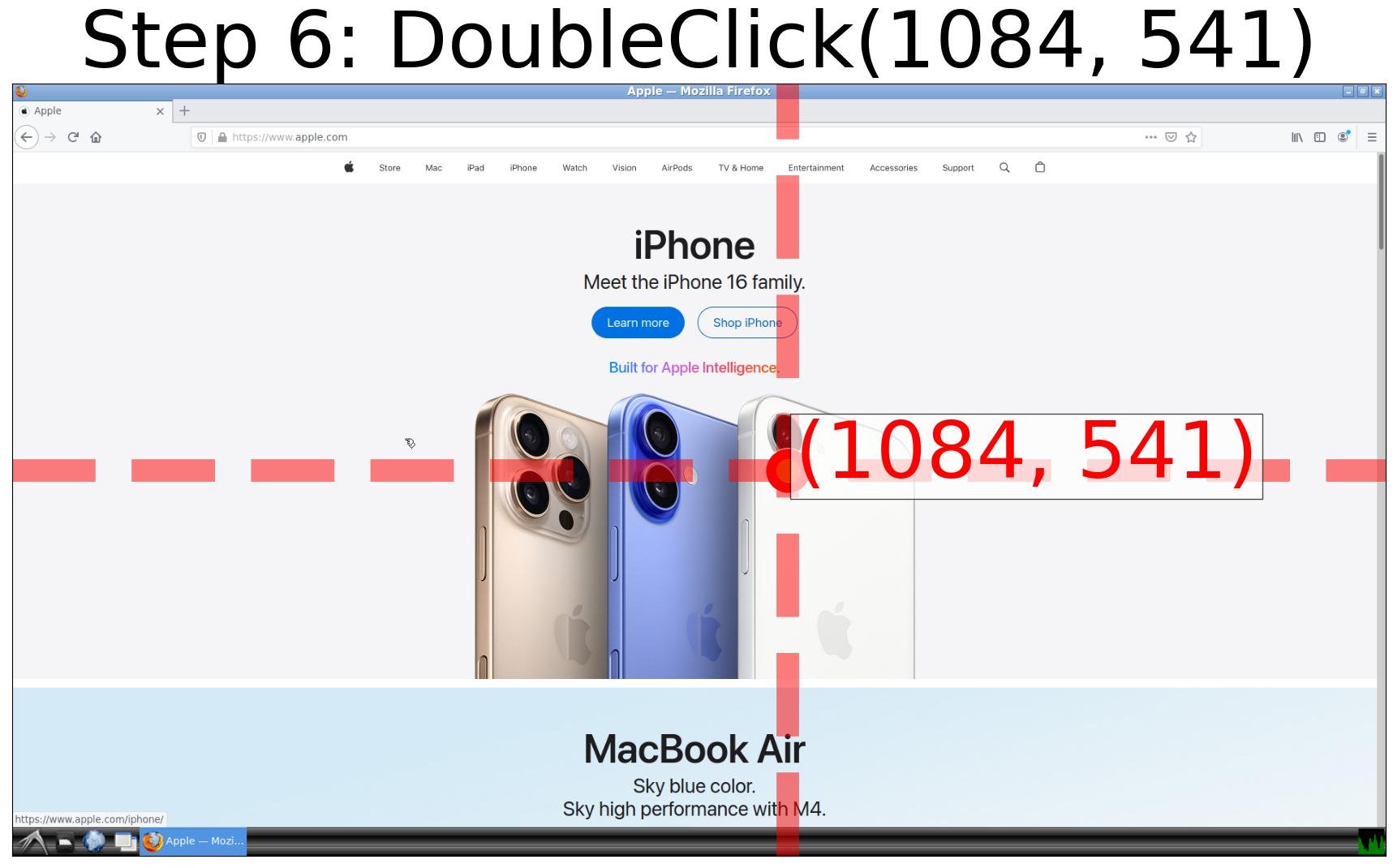} &
    \includegraphics[width=\linewidth]{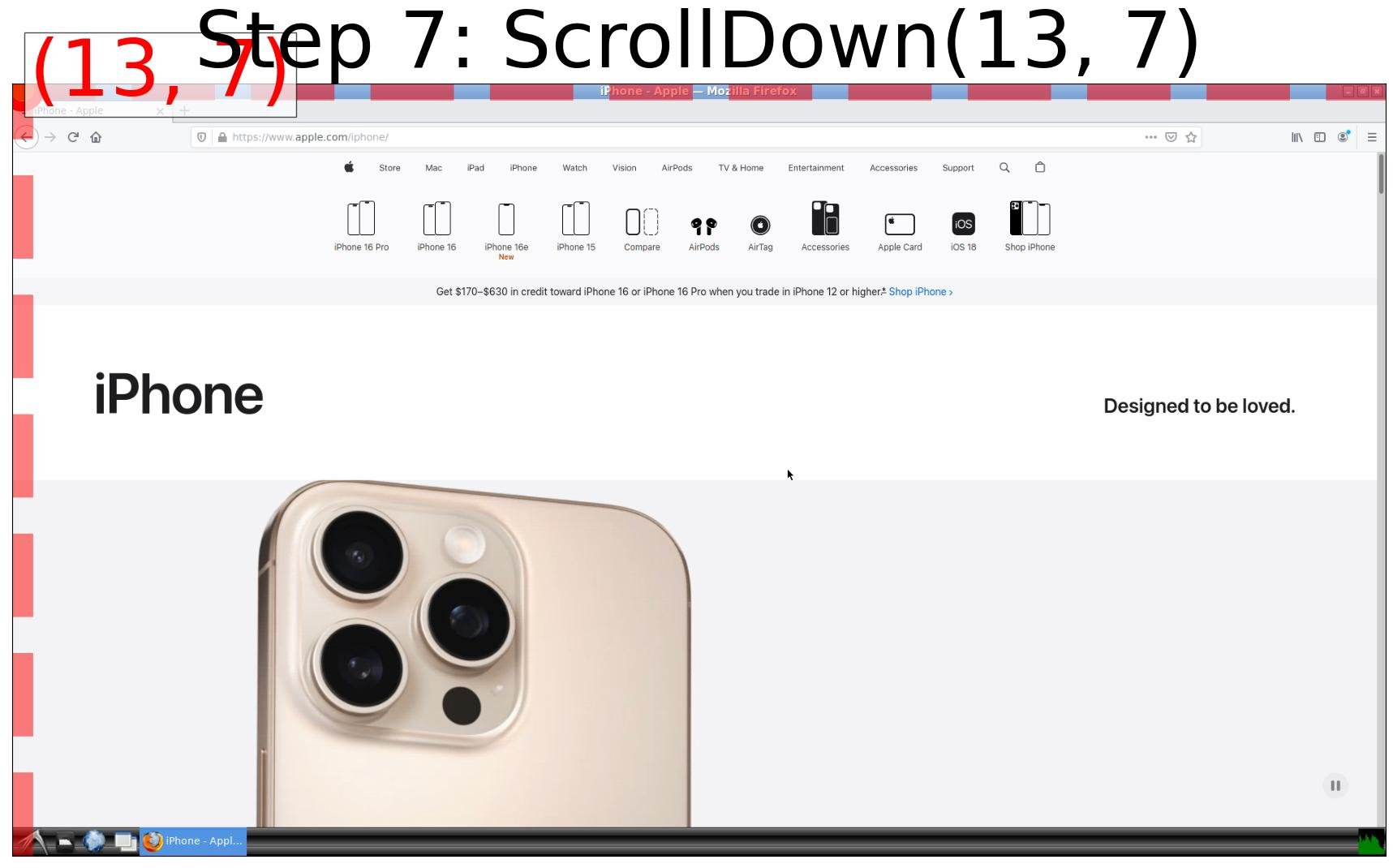} &
    \includegraphics[width=\linewidth]{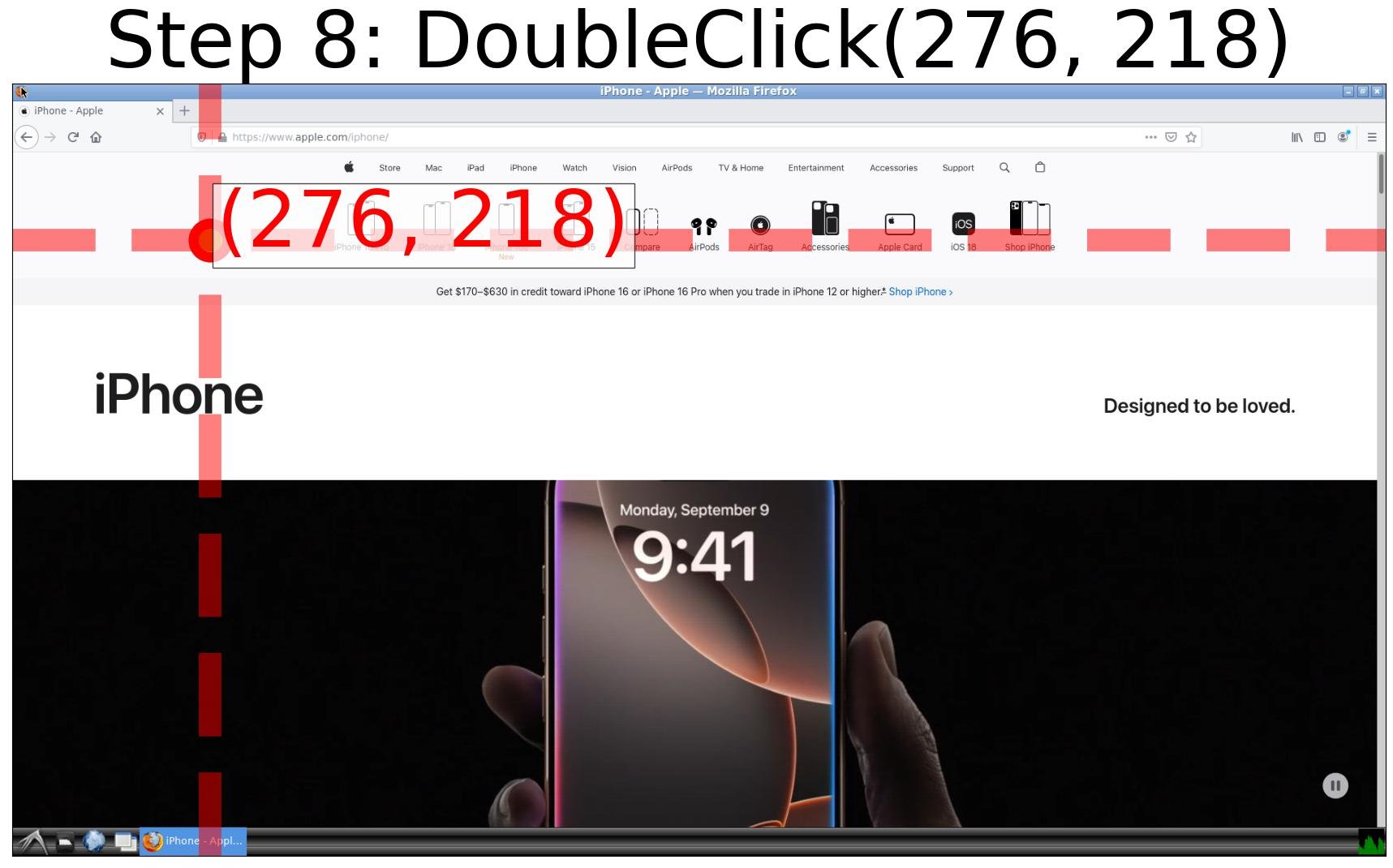} &
    \includegraphics[width=\linewidth]{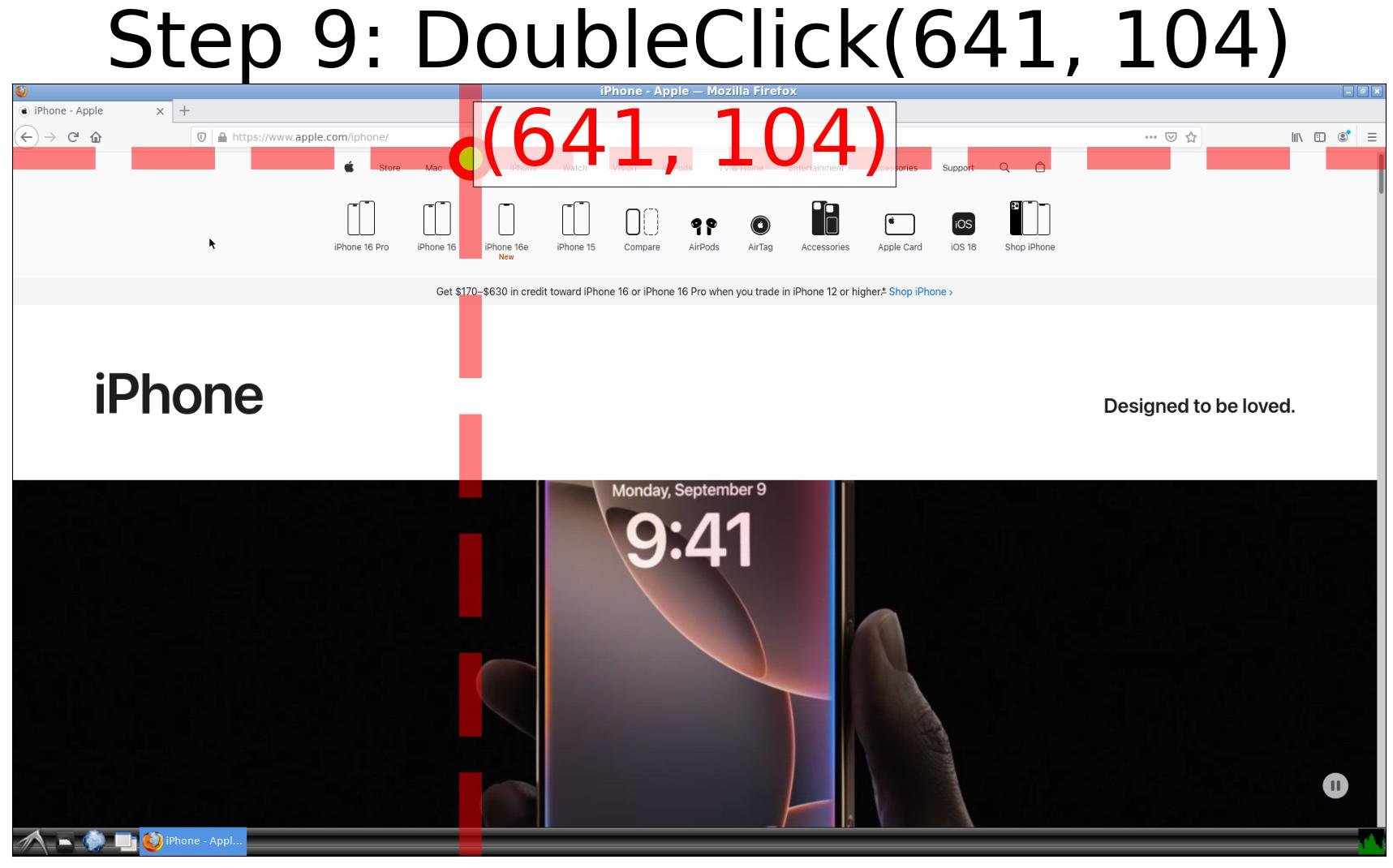}
  \end{tabular}
  \caption{Episode-80 of \textit{ScreenExplorer-3B-E1}: The model demonstrated the ability to navigate from Firefox's homepage to explore a specific webpage.}\label{fig:case-study-3}
\end{figure}

\begin{figure}[h]
  \centering
 \begin{tabular}{
    @{}
    m{0.195\textwidth}@{\hspace{1pt}}
    m{0.195\textwidth}@{\hspace{1pt}}
    m{0.195\textwidth}@{\hspace{1pt}}
    m{0.195\textwidth}@{\hspace{1pt}}
    m{0.195\textwidth}@{}
  }
    \includegraphics[width=\linewidth]{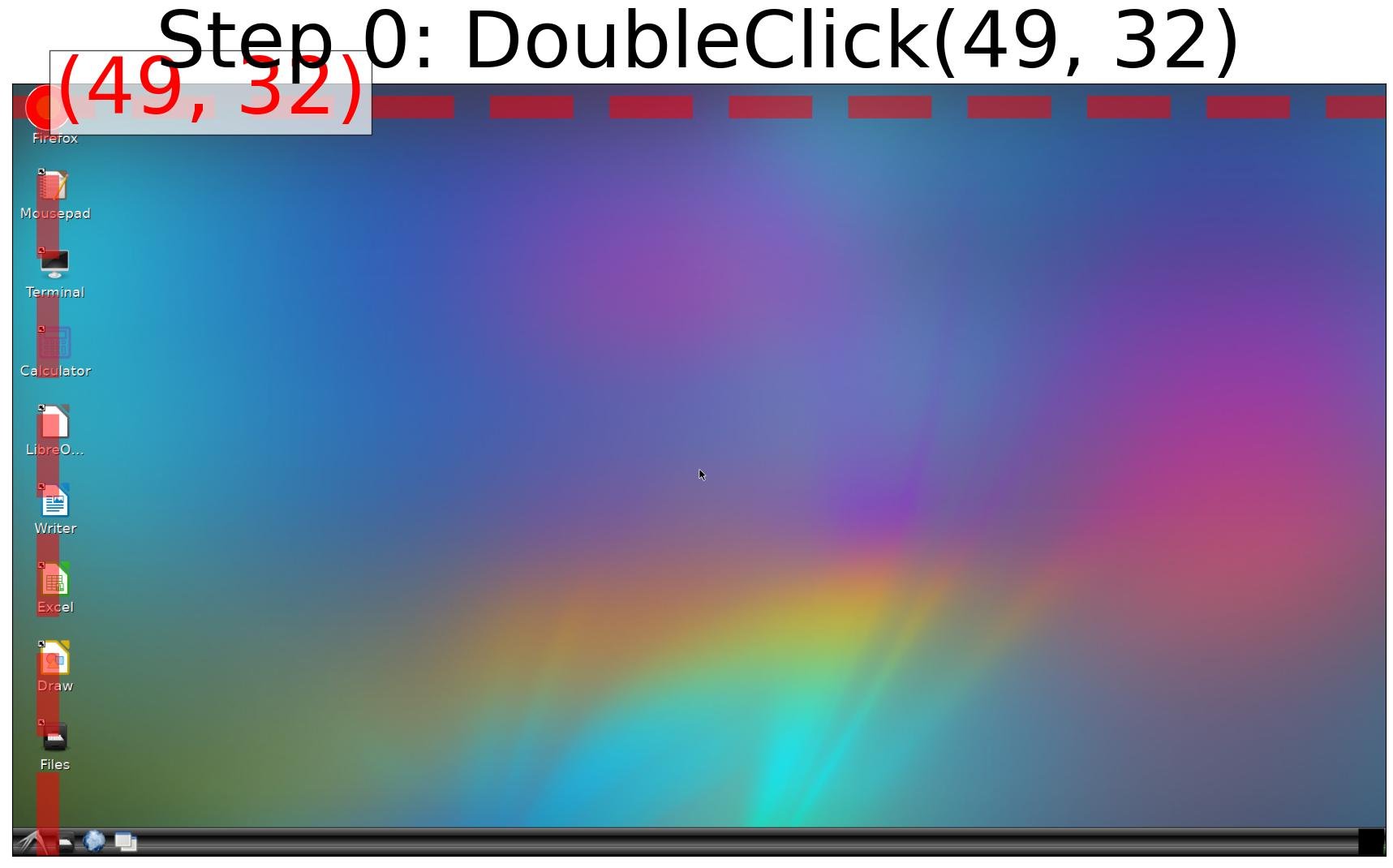} &
    \includegraphics[width=\linewidth]{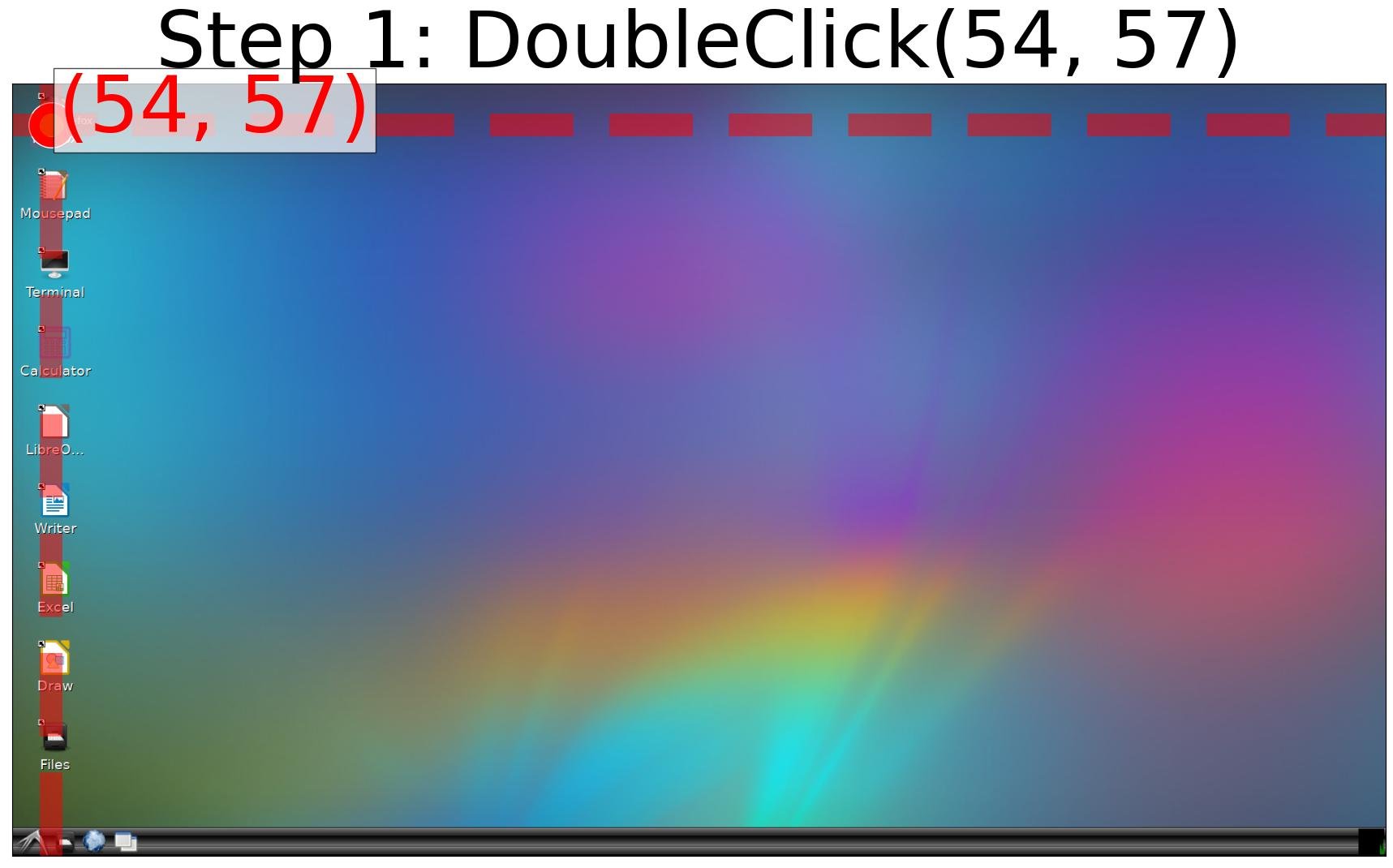} &
    \includegraphics[width=\linewidth]{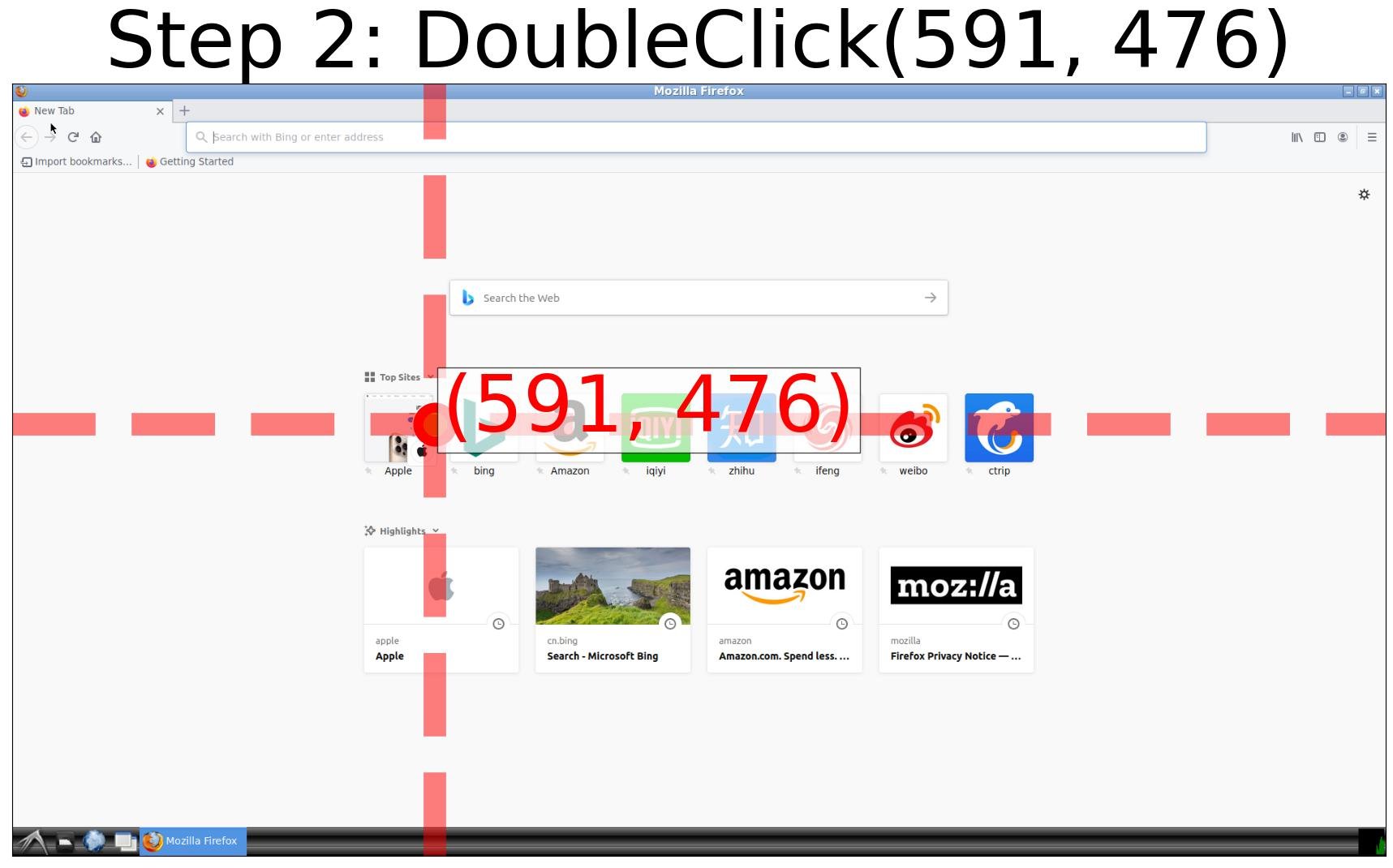} &
    \includegraphics[width=\linewidth]{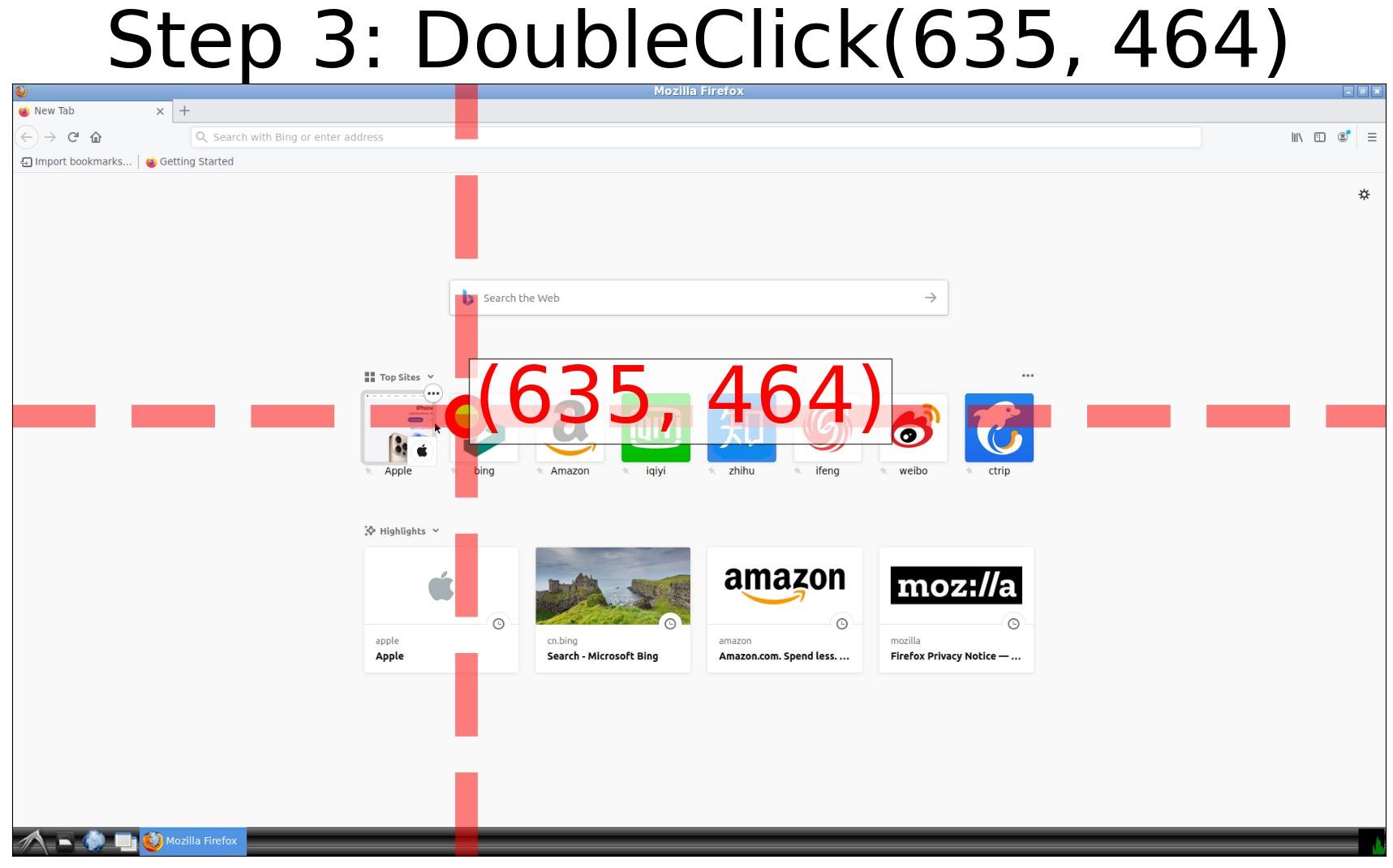} &
    \includegraphics[width=\linewidth]{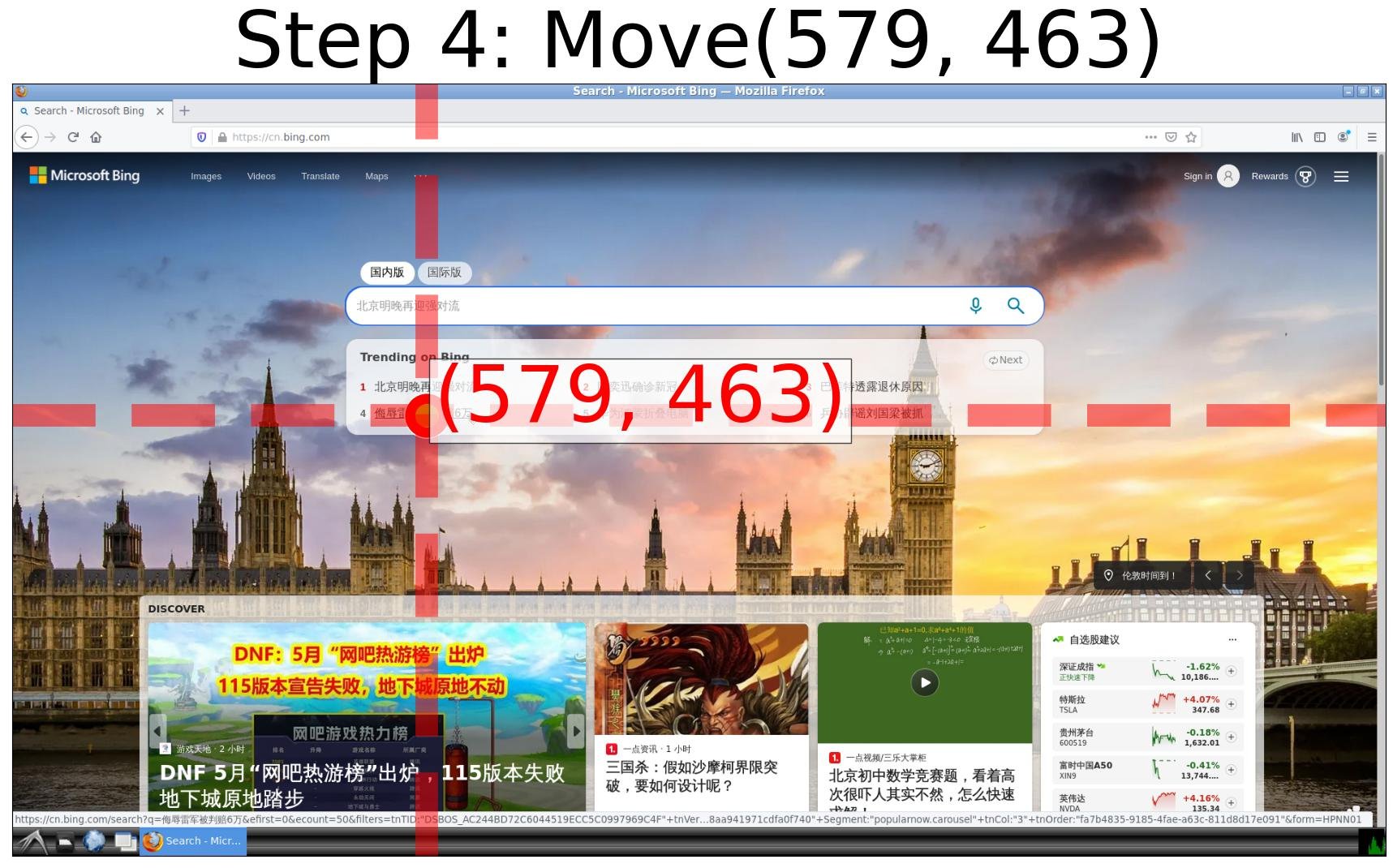} \\
    \includegraphics[width=\linewidth]{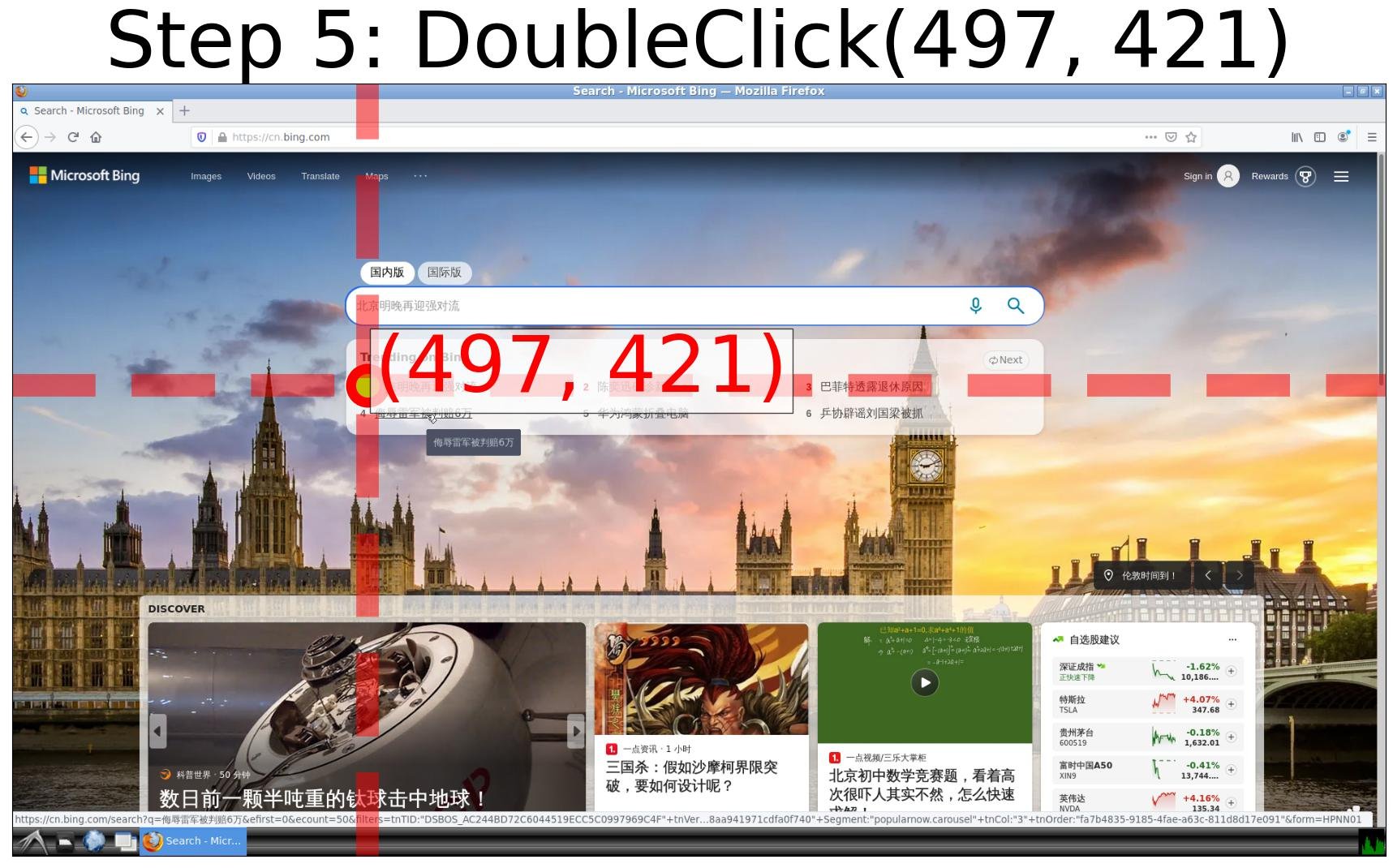} &
    \includegraphics[width=\linewidth]{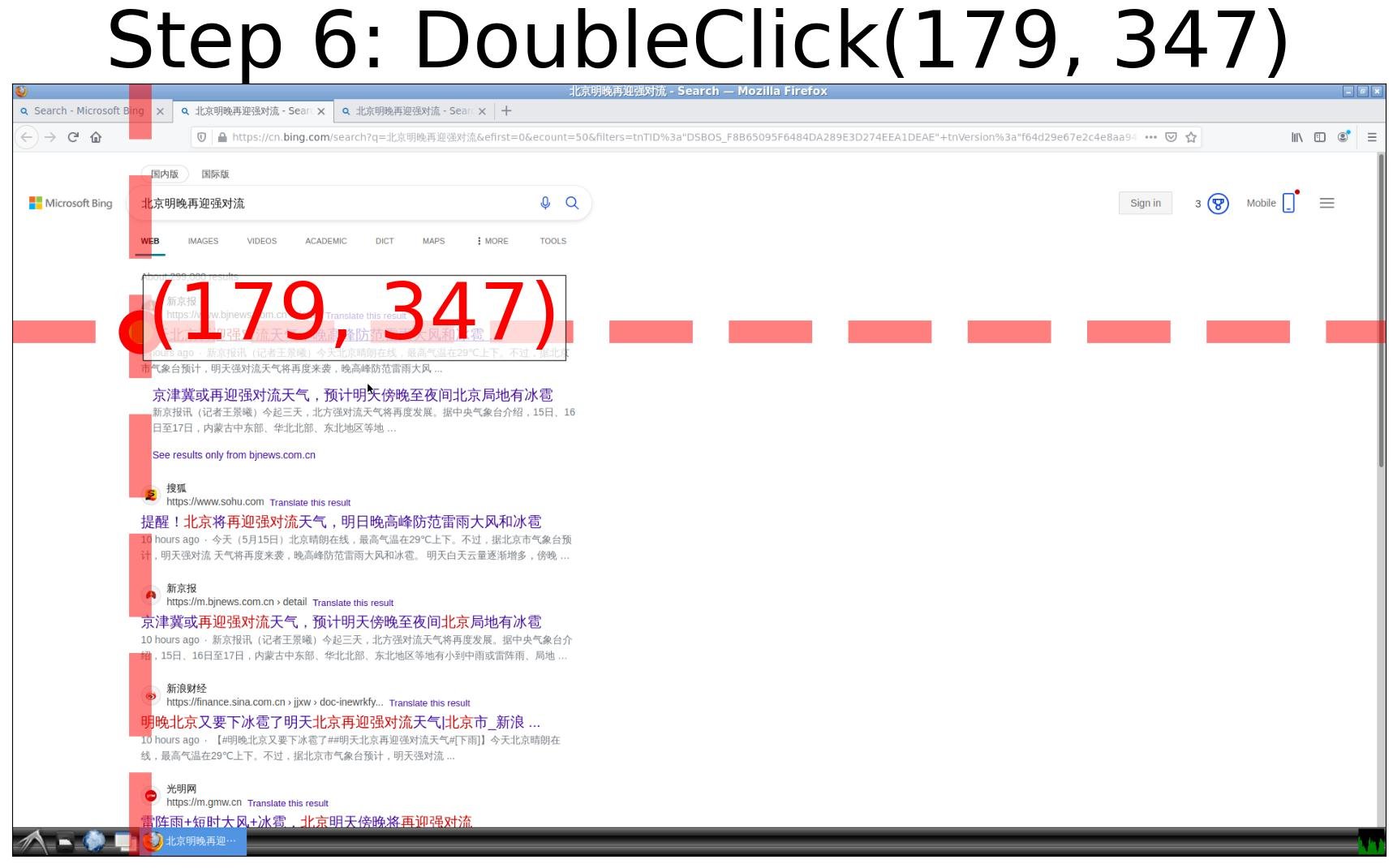} &
    \includegraphics[width=\linewidth]{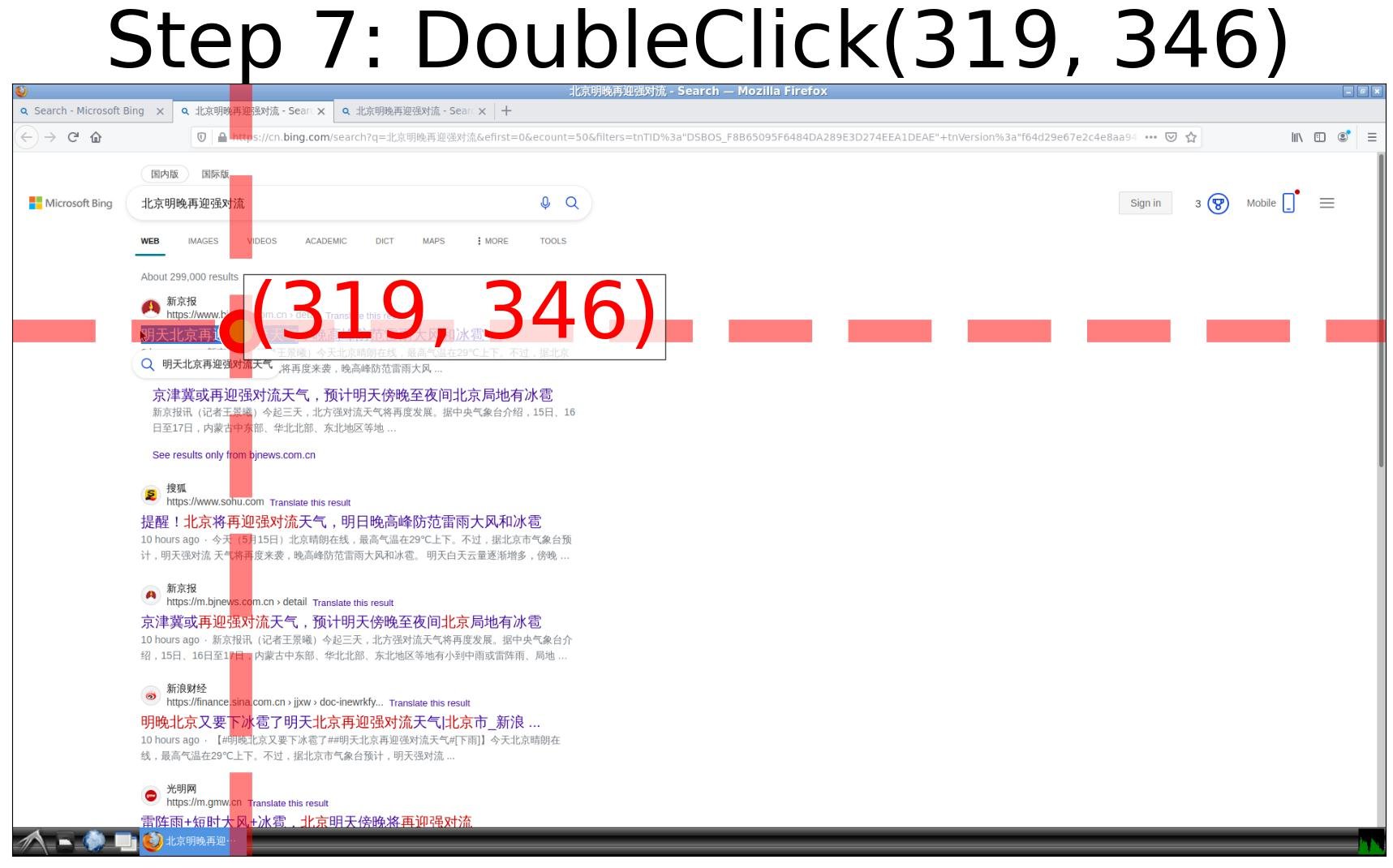} &
    \includegraphics[width=\linewidth]{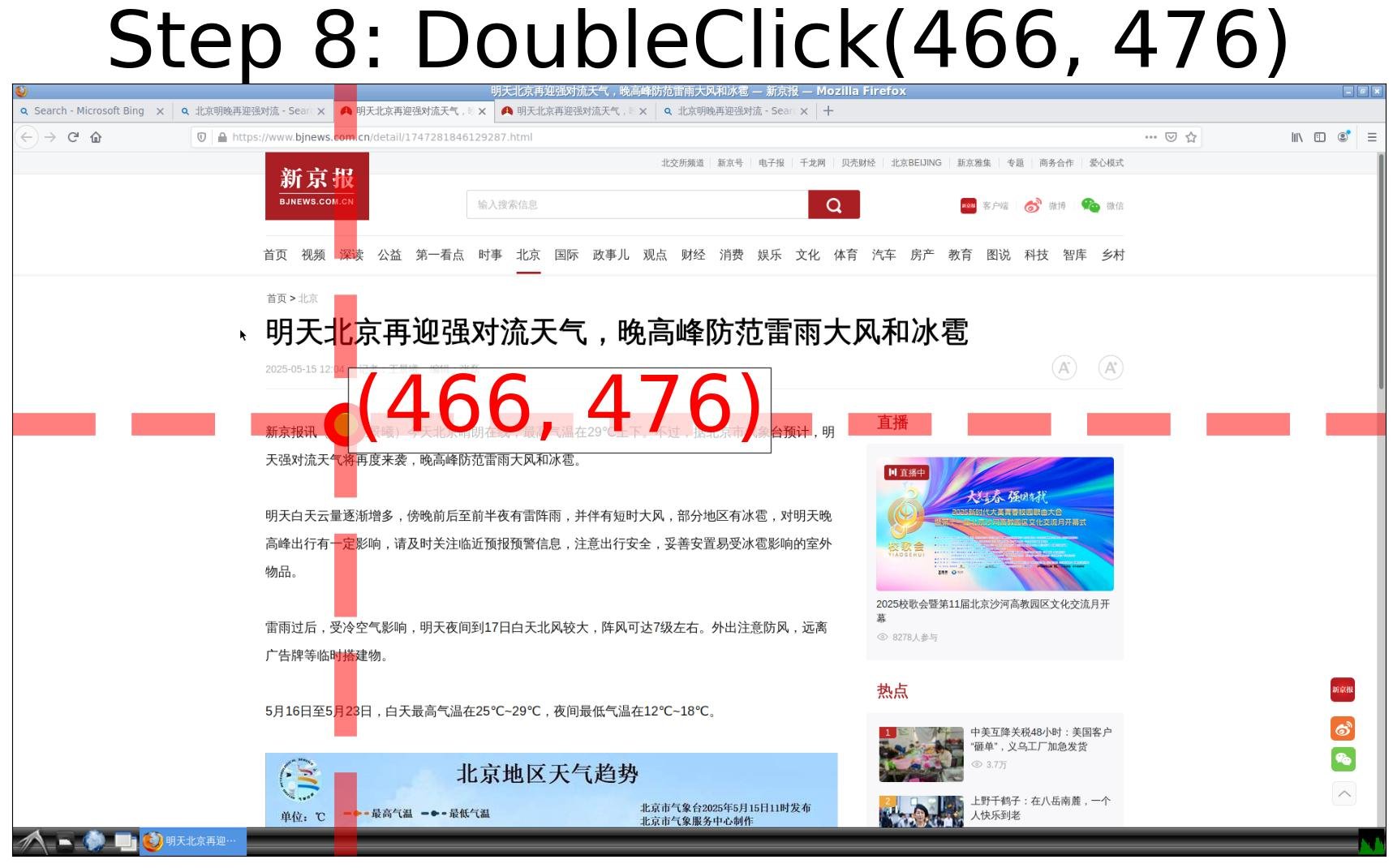} &
    \includegraphics[width=\linewidth]{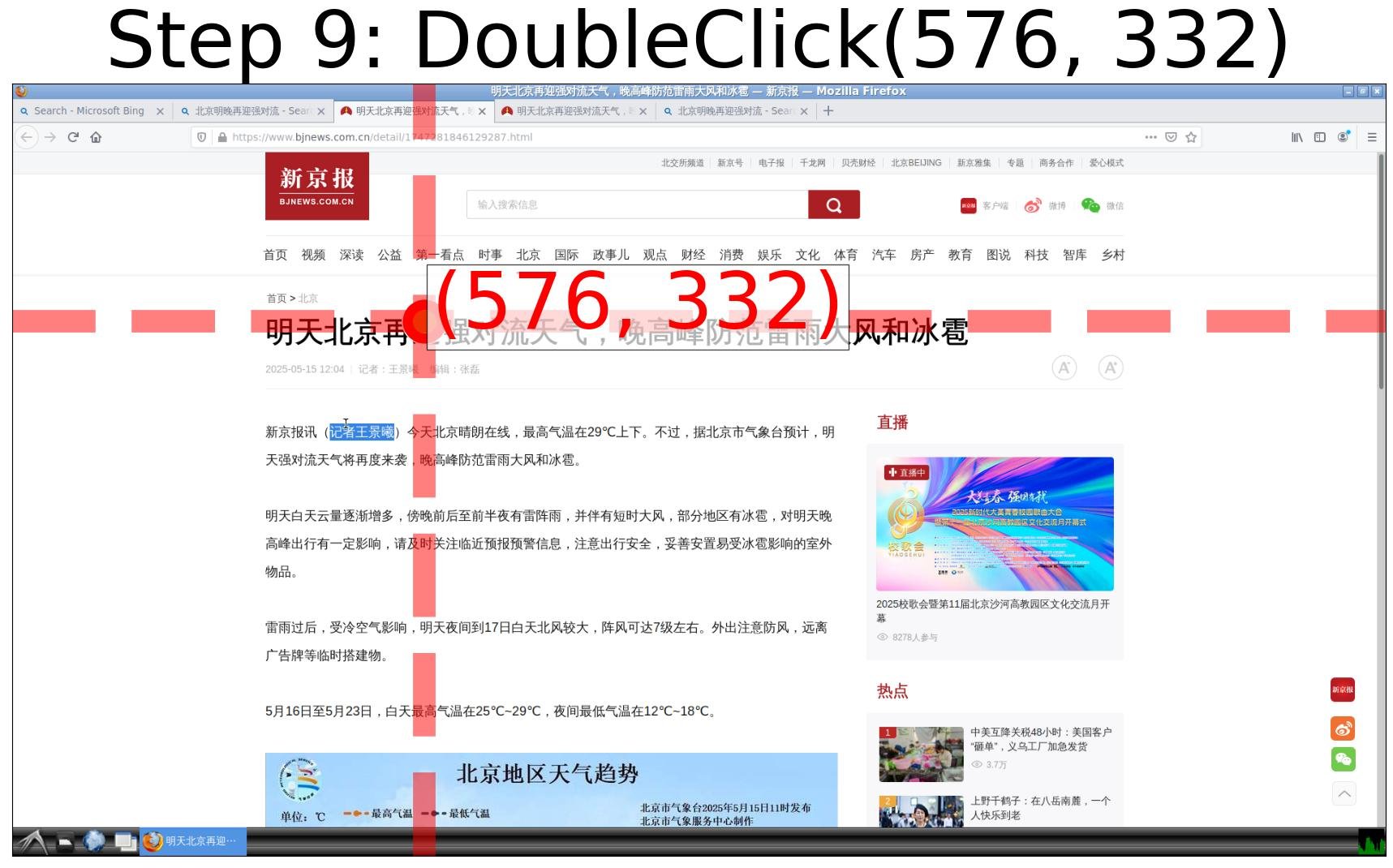}
  \end{tabular}
  \caption{Episode-100 of \textit{ScreenExplorer-3B-E1}: The model demonstrated the capability to navigate through news feeds and access articles for reading.}\label{fig:case-study-4}
\end{figure}

\begin{figure}[h]
  \centering
 \begin{tabular}{
    @{}
    m{0.195\textwidth}@{\hspace{1pt}}
    m{0.195\textwidth}@{\hspace{1pt}}
    m{0.195\textwidth}@{\hspace{1pt}}
    m{0.195\textwidth}@{\hspace{1pt}}
    m{0.195\textwidth}@{}
  }
    \includegraphics[width=\linewidth]{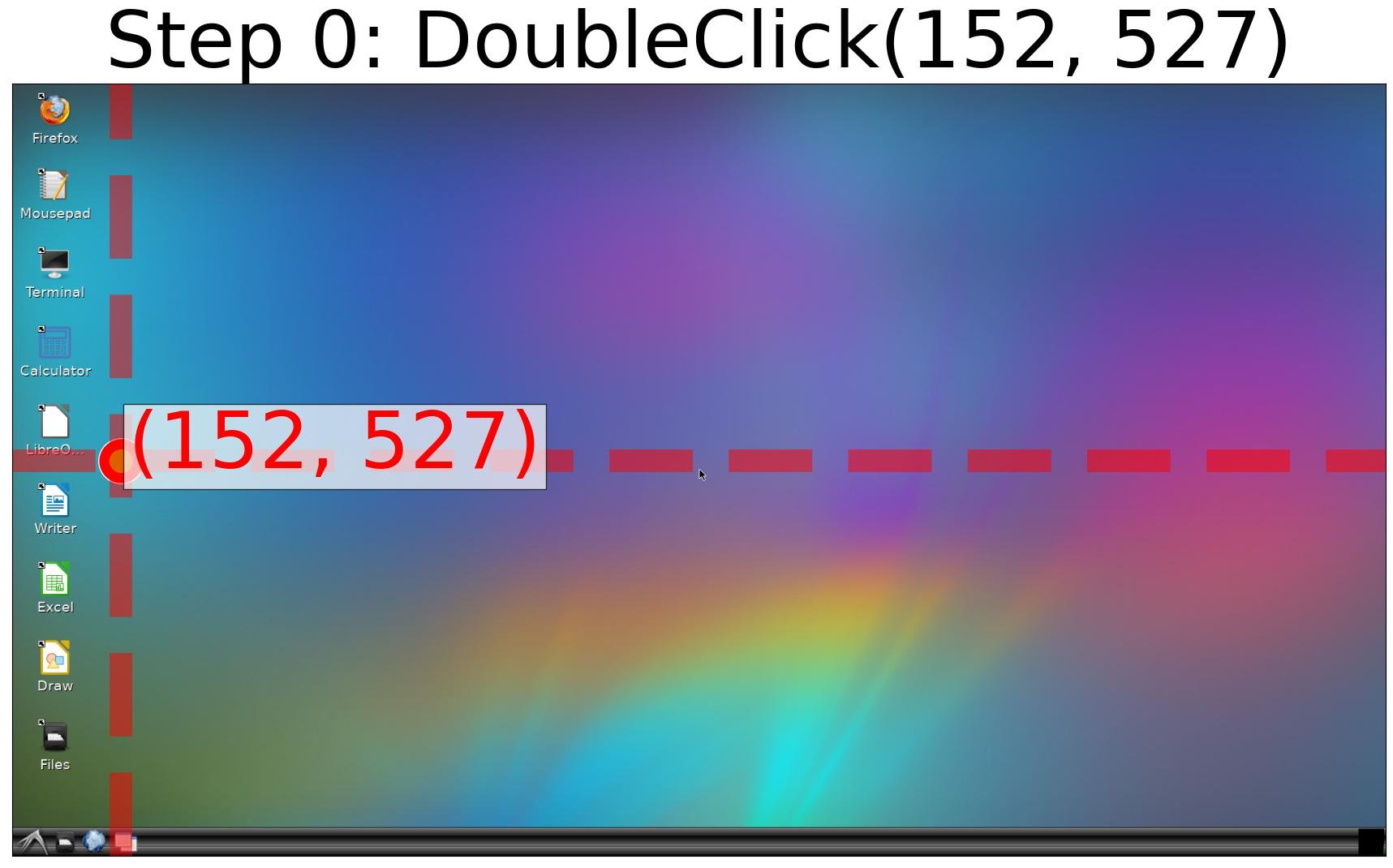} &
    \includegraphics[width=\linewidth]{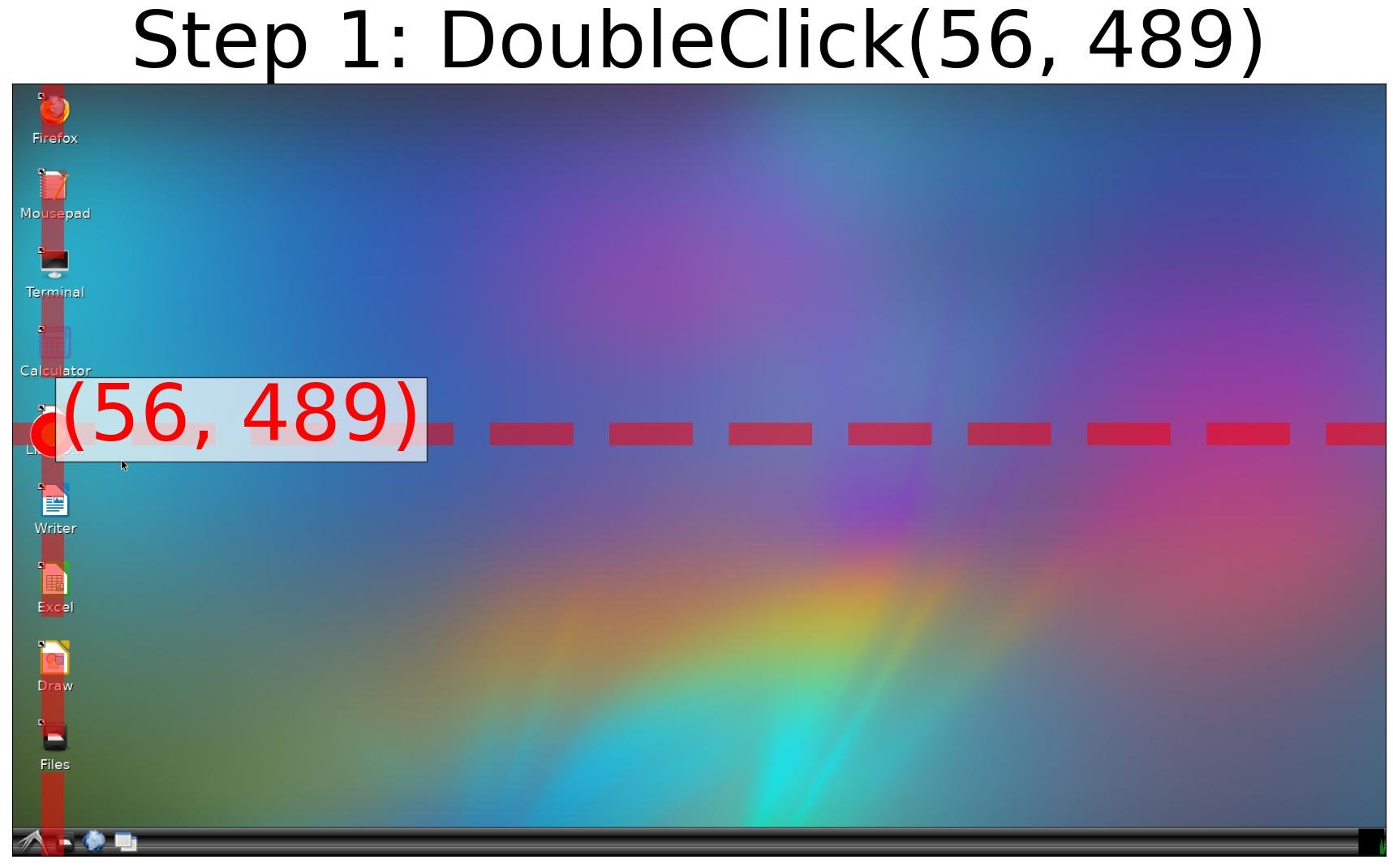} &
    \includegraphics[width=\linewidth]{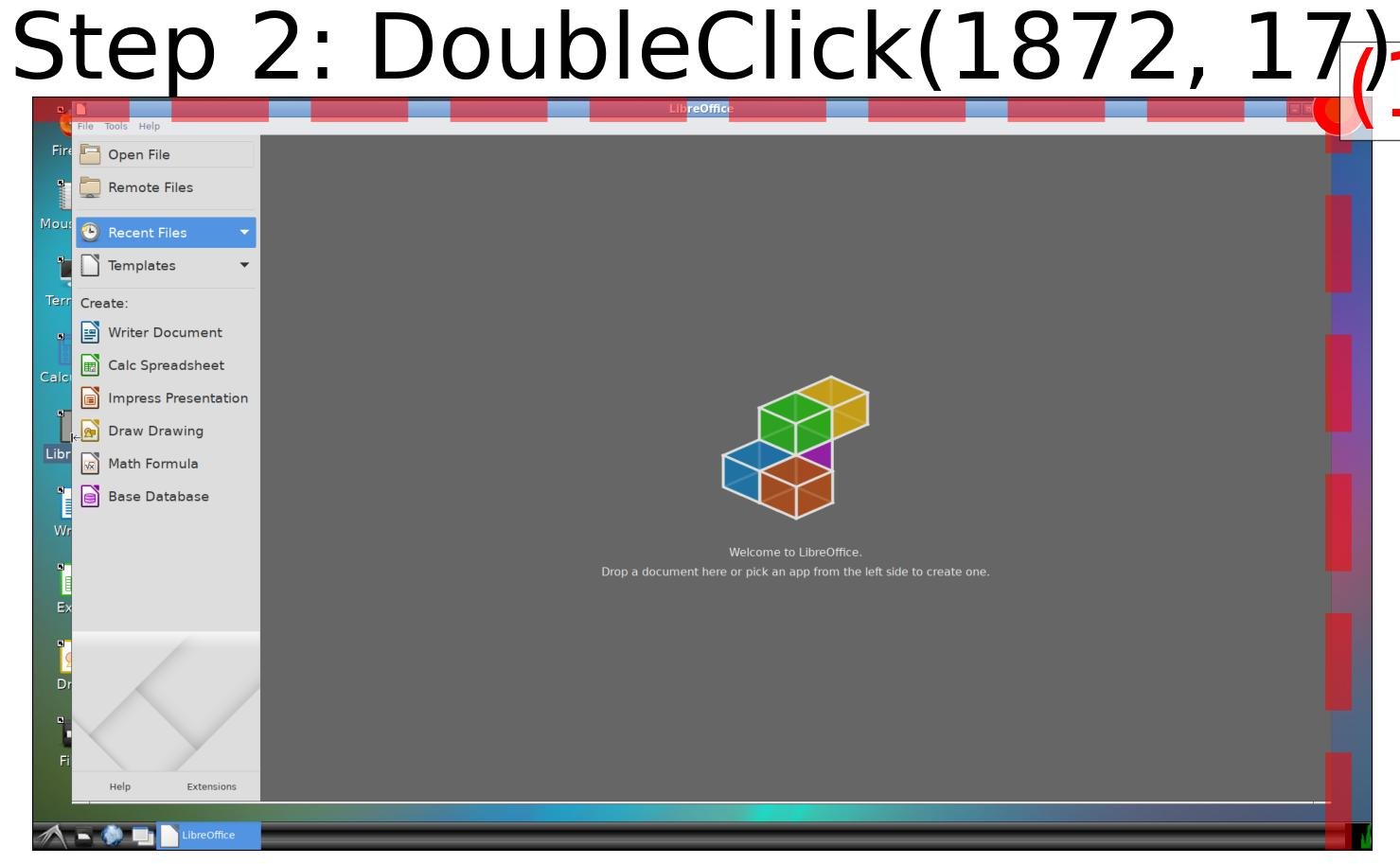} &
    \includegraphics[width=\linewidth]{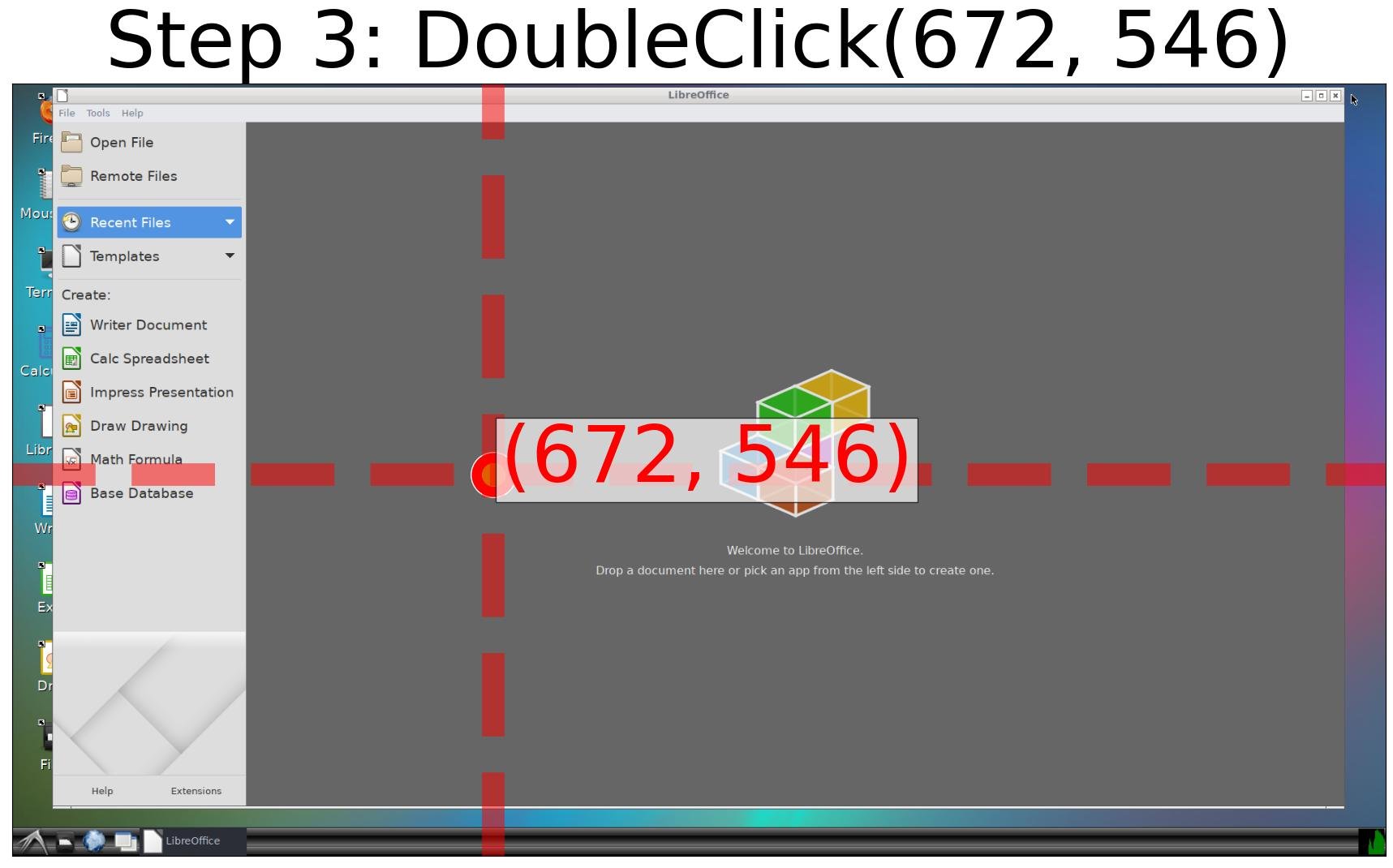} &
    \includegraphics[width=\linewidth]{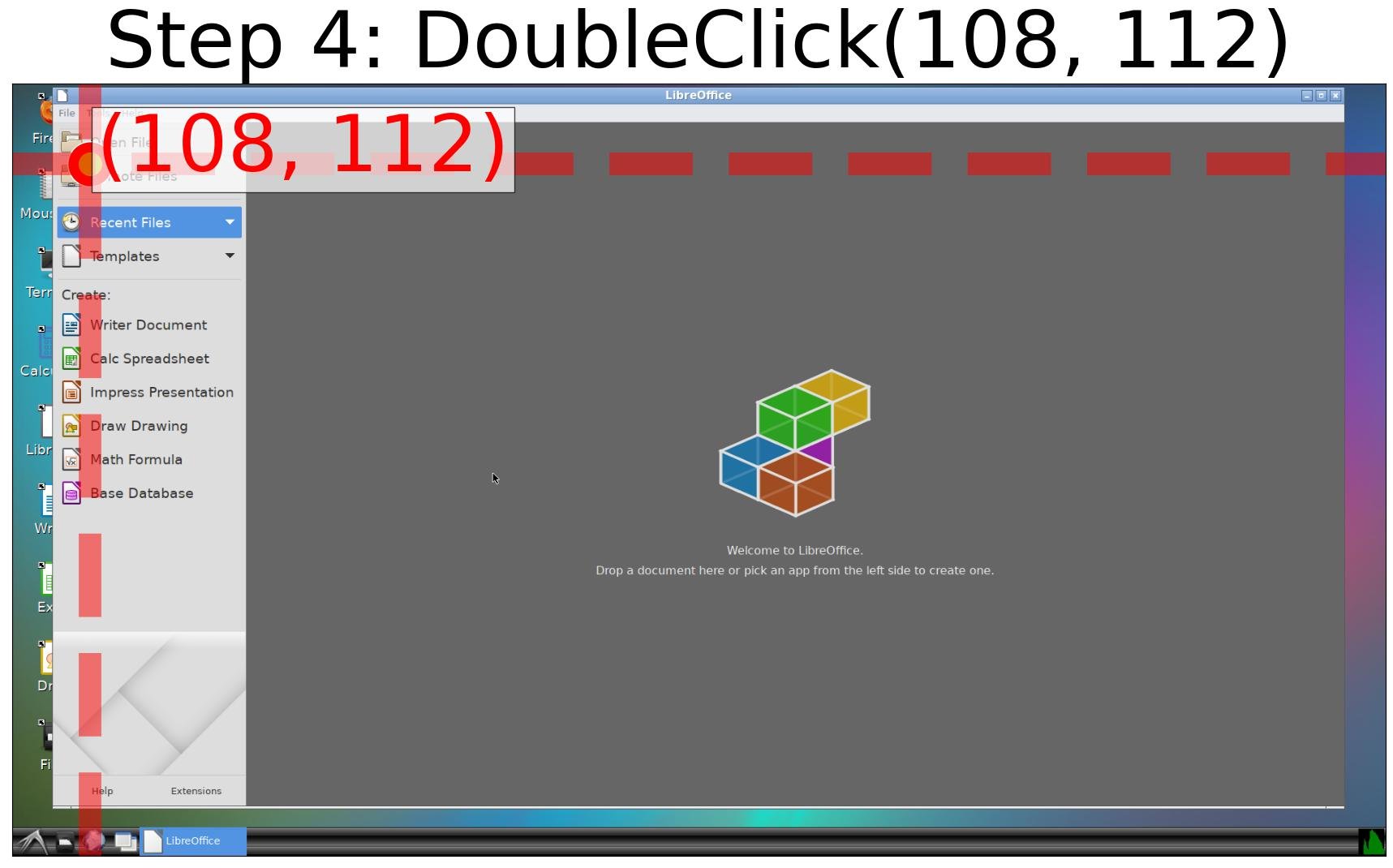} \\
    \includegraphics[width=\linewidth]{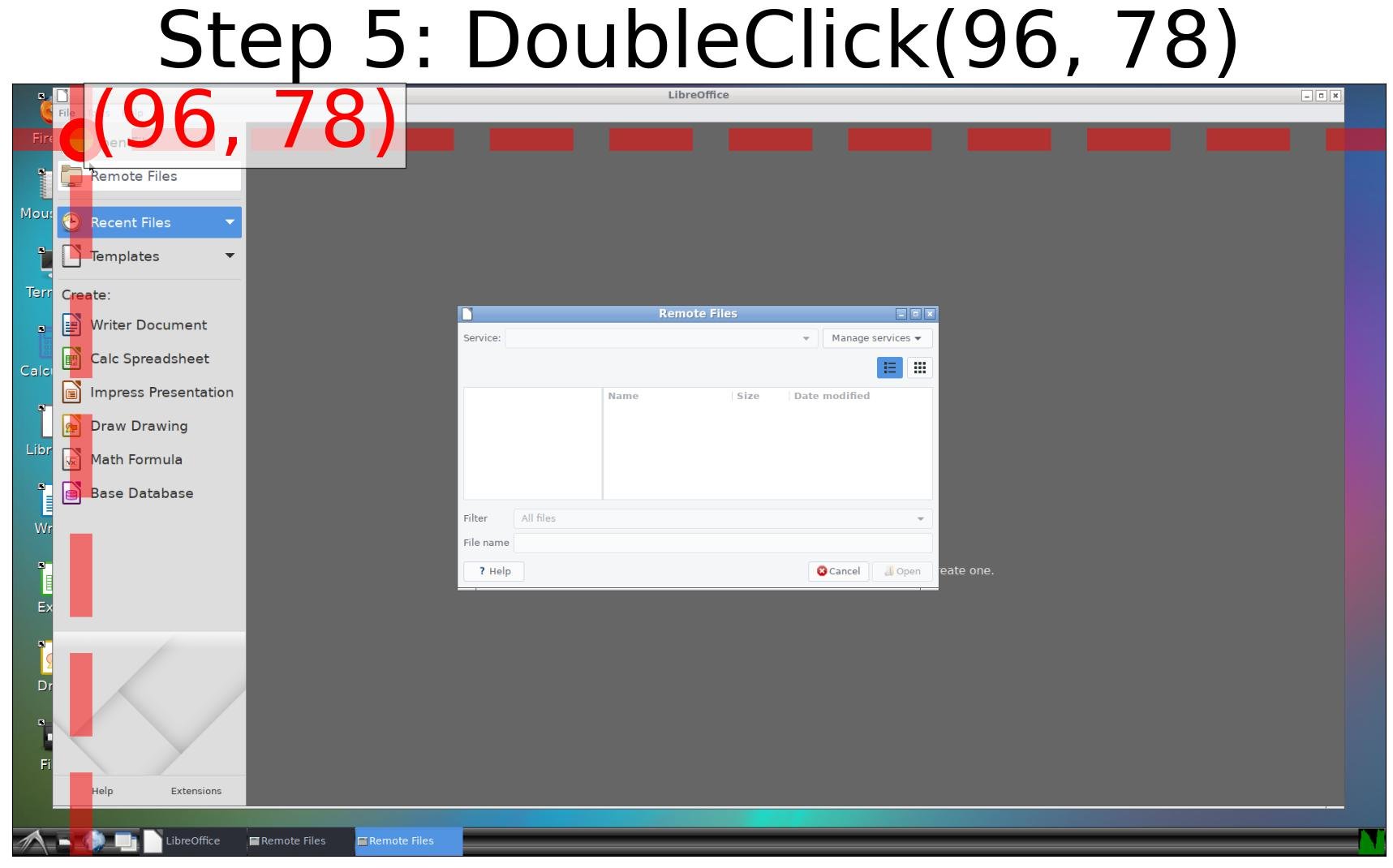} &
    \includegraphics[width=\linewidth]{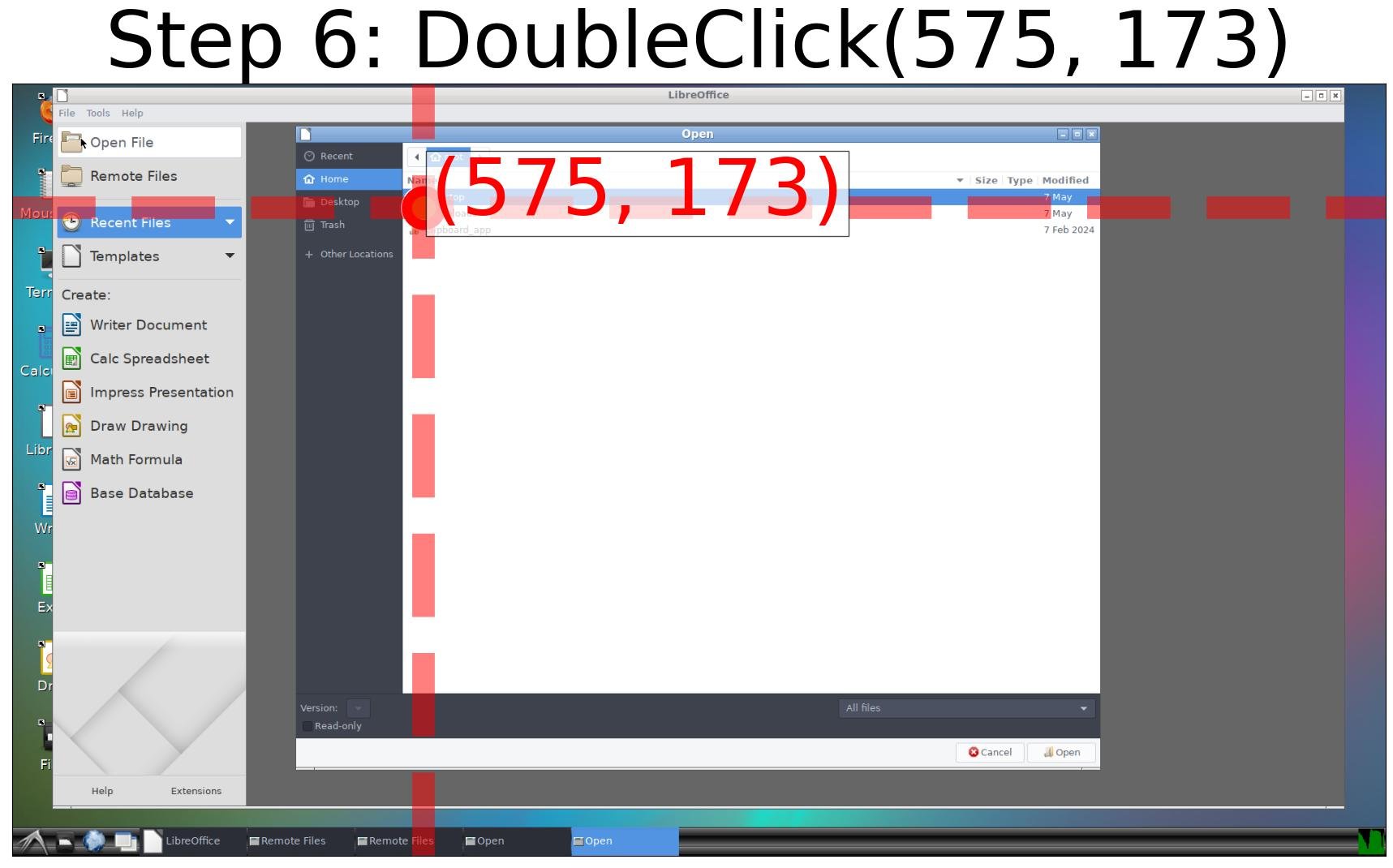} &
    \includegraphics[width=\linewidth]{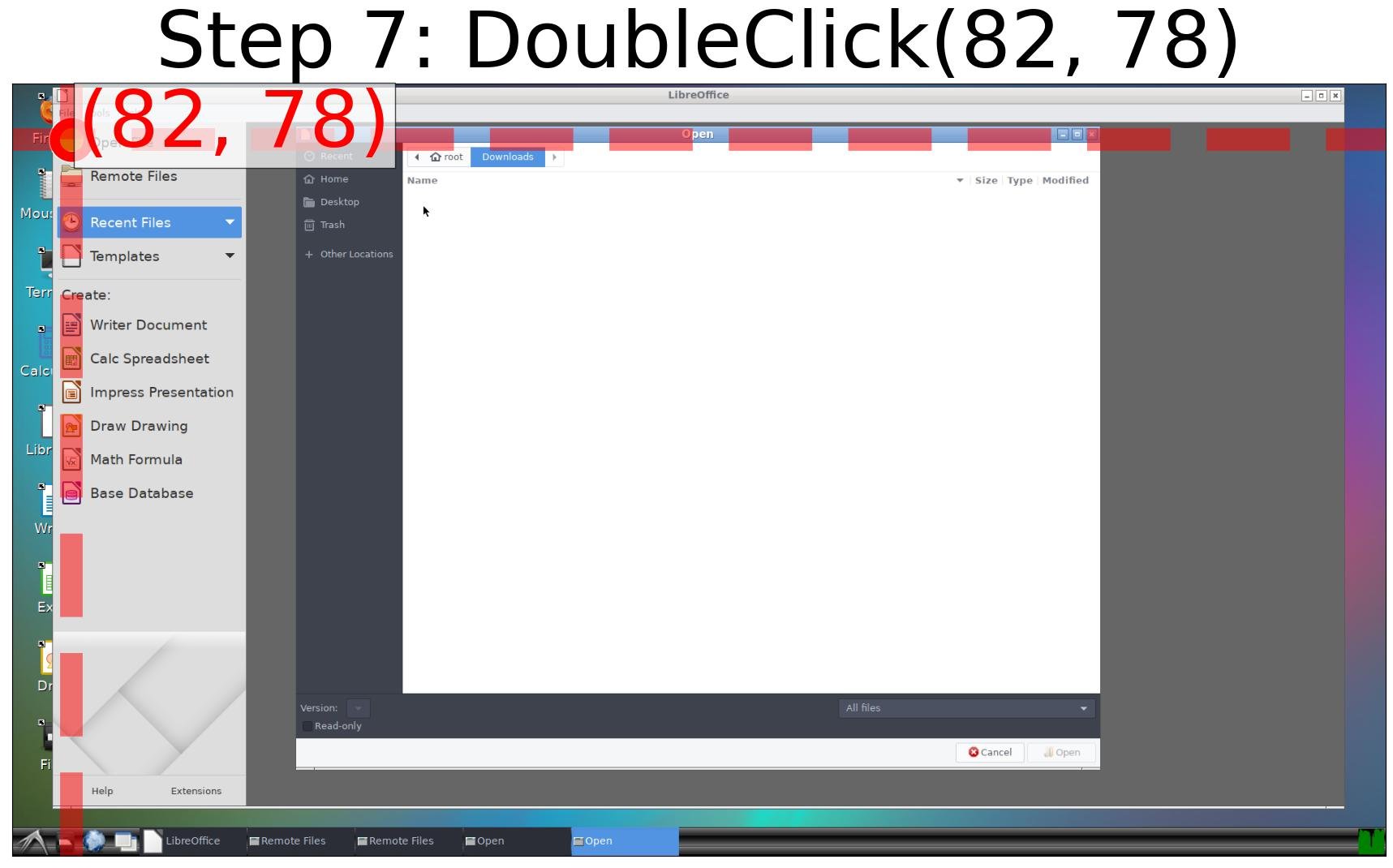} &
    \includegraphics[width=\linewidth]{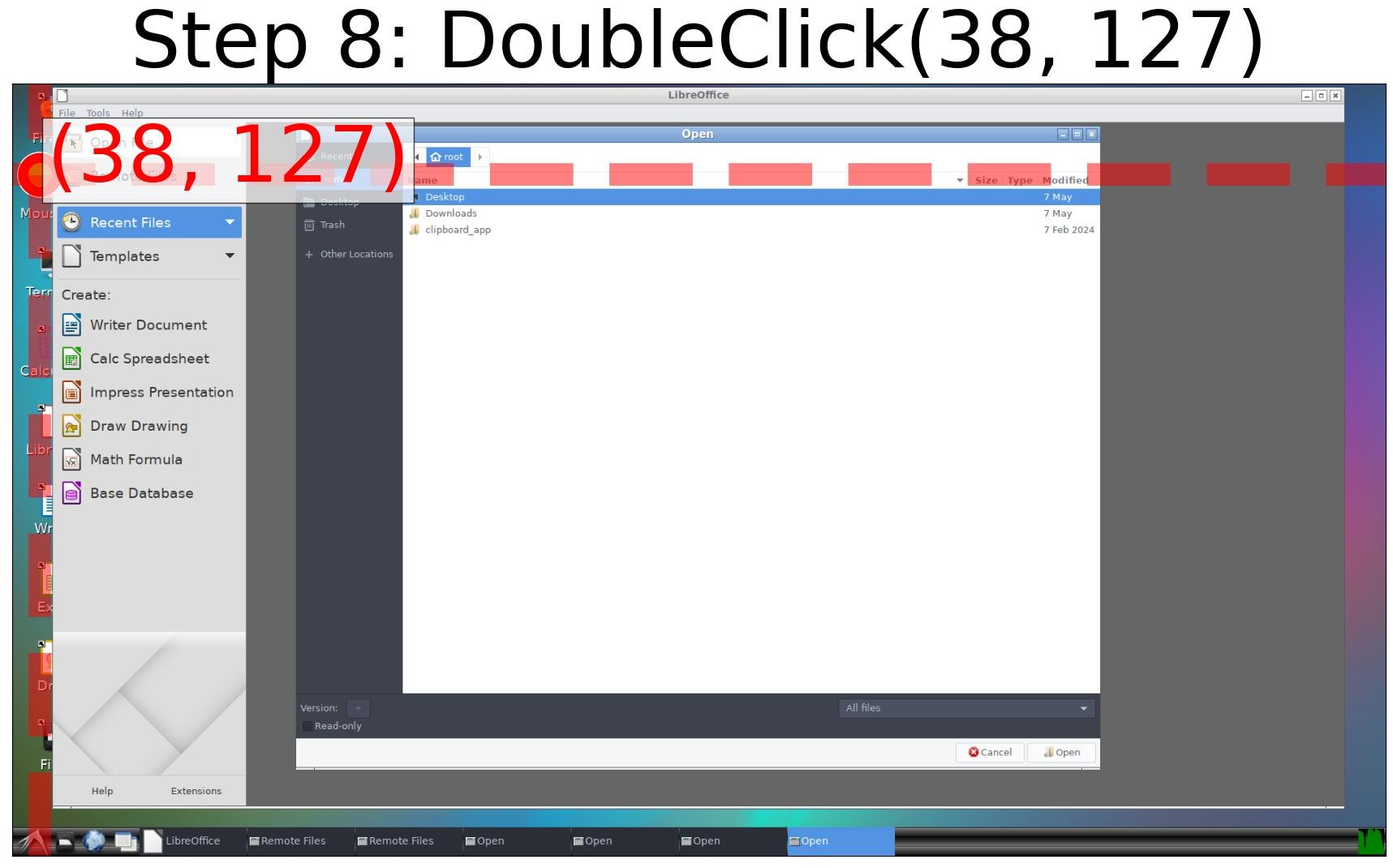} &
    \includegraphics[width=\linewidth]{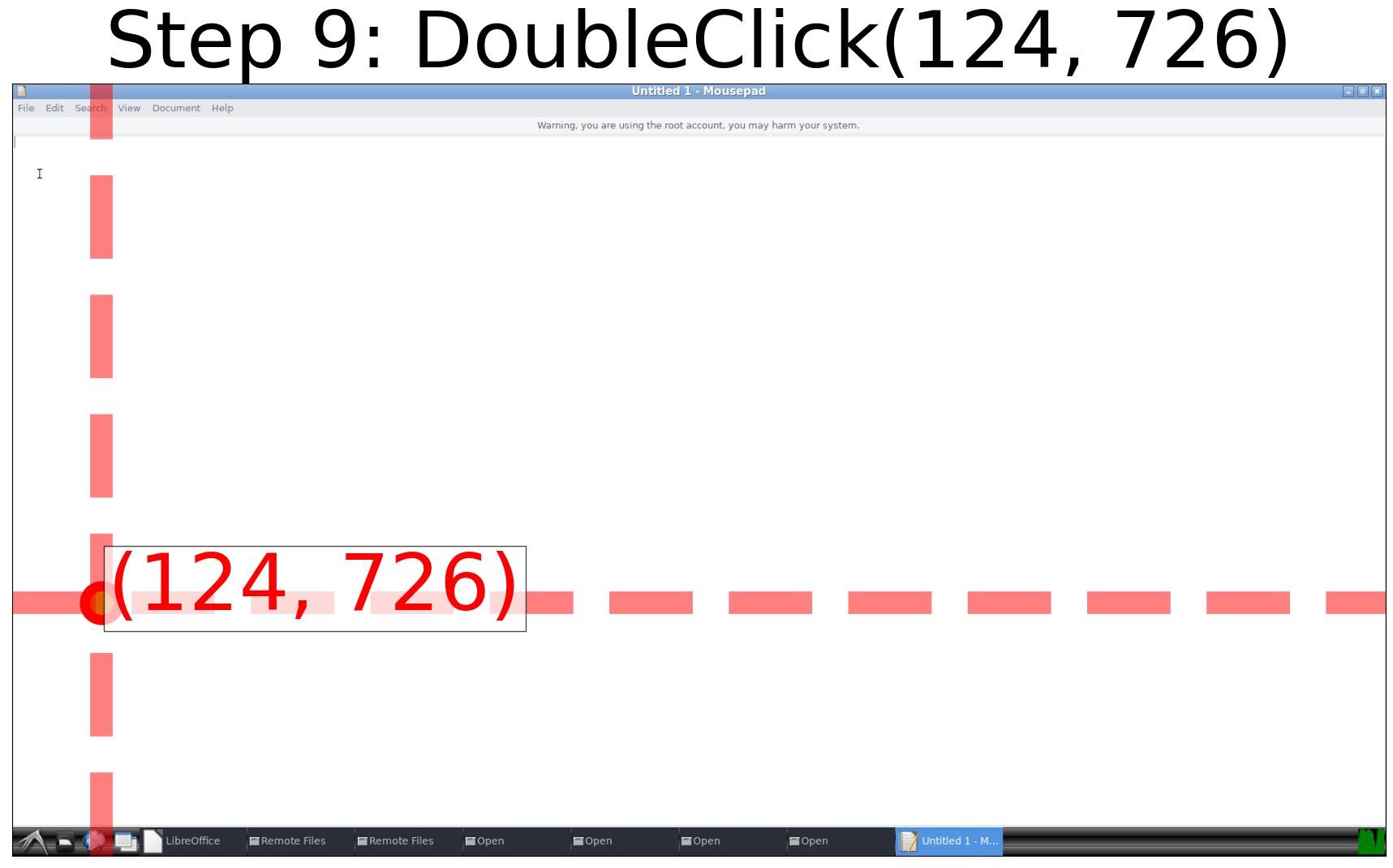}
  \end{tabular}
  \caption{Episode-100 of \textit{ScreenExplorer-3B-E1}: The model was exploring the LibreOffice software interface.}\label{fig:case-study-5}
\end{figure}

\begin{figure}[h]
  \centering
  \begin{tabular}{@{} m{0.5\textwidth}  m{0.45\textwidth} @{}}
    \includegraphics[width=\linewidth]{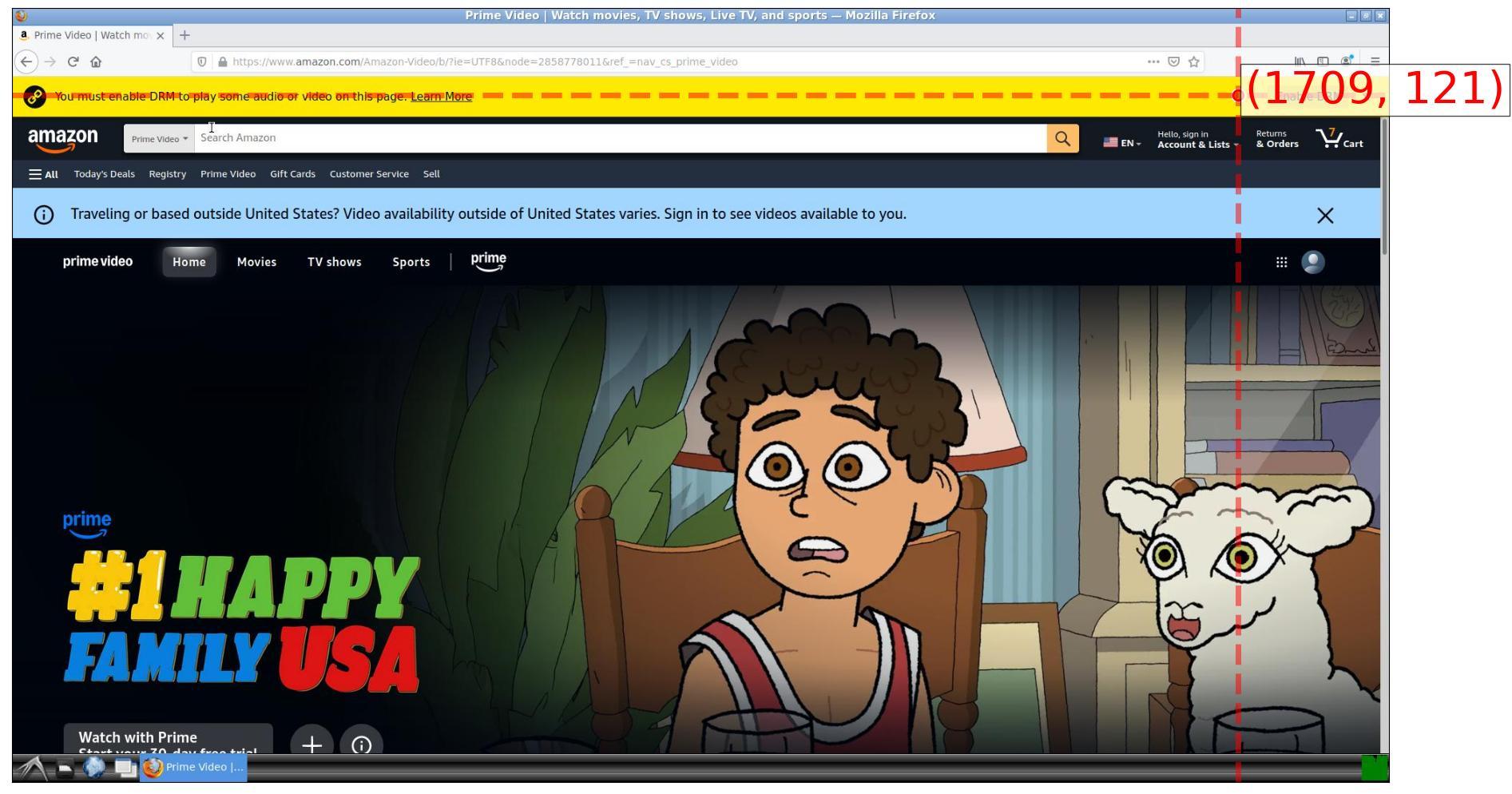} &
    \shortstack[l]{%
      \textbf{Intent:} “First, I need to enable DRM to watch \\
      the stories about the children, which might be \\ 
      a prerequisite before exploring content further.”\\[0.5ex]
      \textbf{Action:} “DoubleClick(1709, 121)”
    }%
  \end{tabular}
  \caption{At episode 70, the agent detected a DRM warning in the browser and observed that "stories about the children" were available for watching. Consequently, it opted to dismiss the DRM warning, demonstrating its ability to reason based on the screen state and make contextually appropriate decisions.}\label{fig:case-study-6}
\end{figure}

\begin{figure}[h]
  \centering
  \begin{tabular}{@{} m{0.5\textwidth}  m{0.45\textwidth} @{}}
    \includegraphics[width=\linewidth]{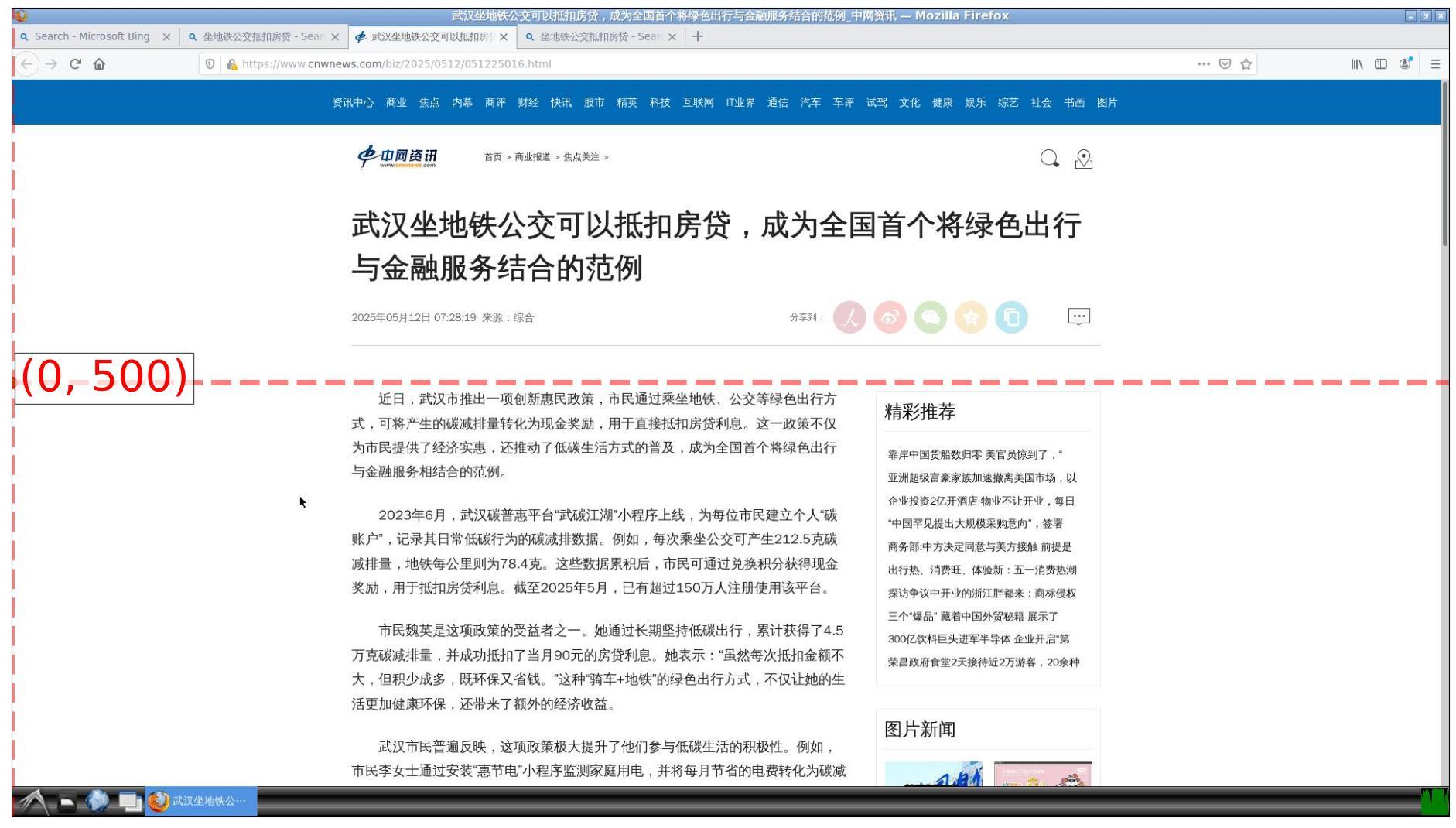} &
    \shortstack[l]{%
      \textbf{Intent:} “Scroll down to read more content \\
      on the webpage”\\[0.5ex]
      \textbf{Action:} “ScrollDown(0, 500)”
    }%
  \end{tabular}
  \caption{At episode 80, the agent scrolled down on the news page to acquire additional information, demonstrating its capacity for content exploration. }\label{fig:case-study-7}
\end{figure}

\begin{figure}[h]
  \centering
  \begin{tabular}{@{} m{0.5\textwidth}  m{0.45\textwidth} @{}}
    \includegraphics[width=\linewidth]{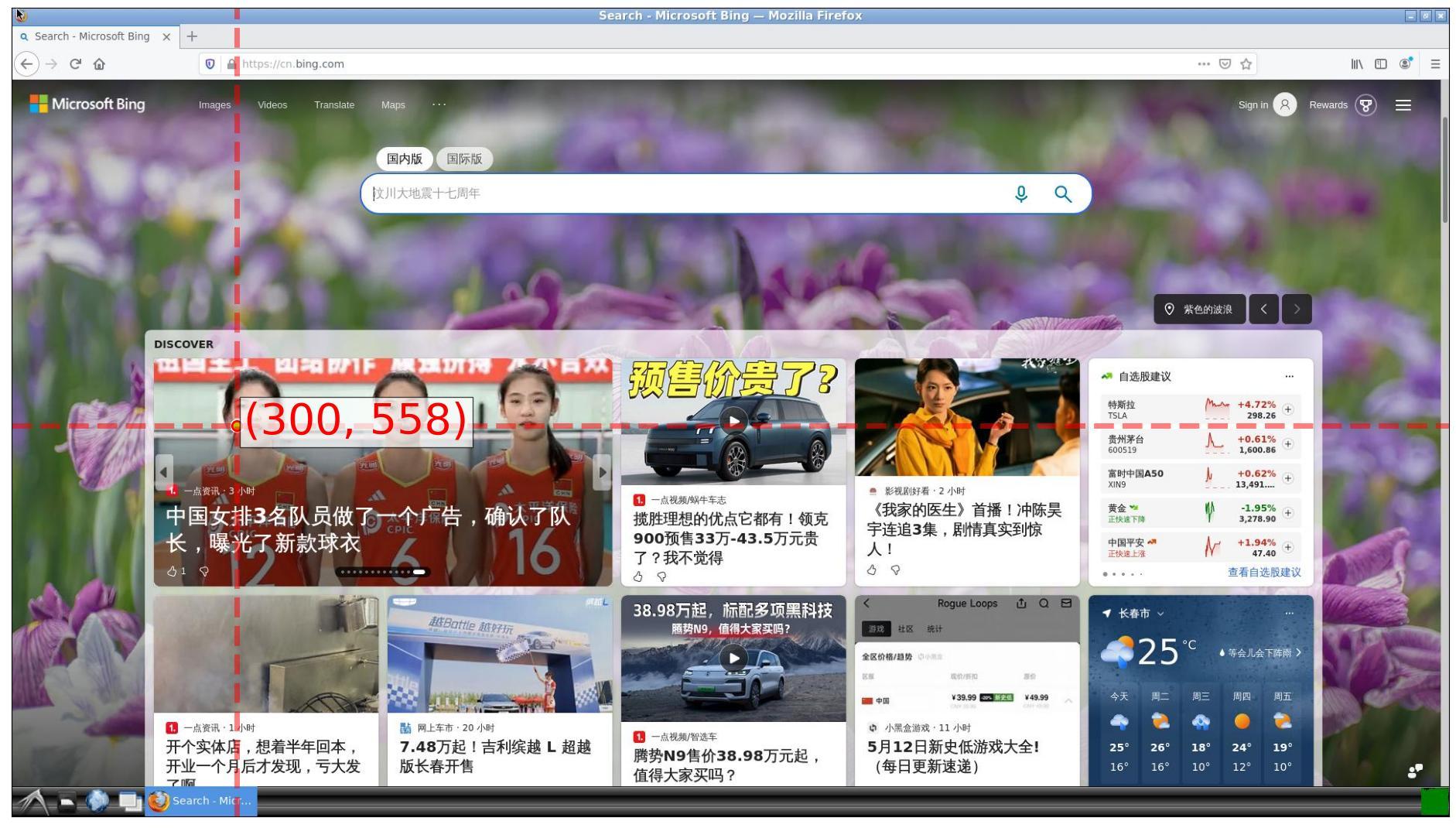} &
    \shortstack[l]{%
      \textbf{Intent:} “Select the first news item about the \\
      Chinese women's volleyball team to explore \\
      more details” \\[0.5ex]
      \textbf{Action:} “DoubleClick(300, 558)”
    }%
  \end{tabular}
  \caption{At episode 100, the agent demonstrated cross-modal translation capabilities, recognizing Chinese news headlines from the visual input and generating corresponding English outputs in its intent description.}\label{fig:case-study-8}
\end{figure}

\begin{figure}[h]
  \centering
  \begin{tabular}{@{} m{0.5\textwidth}  m{0.45\textwidth} @{}}
    \includegraphics[width=\linewidth]{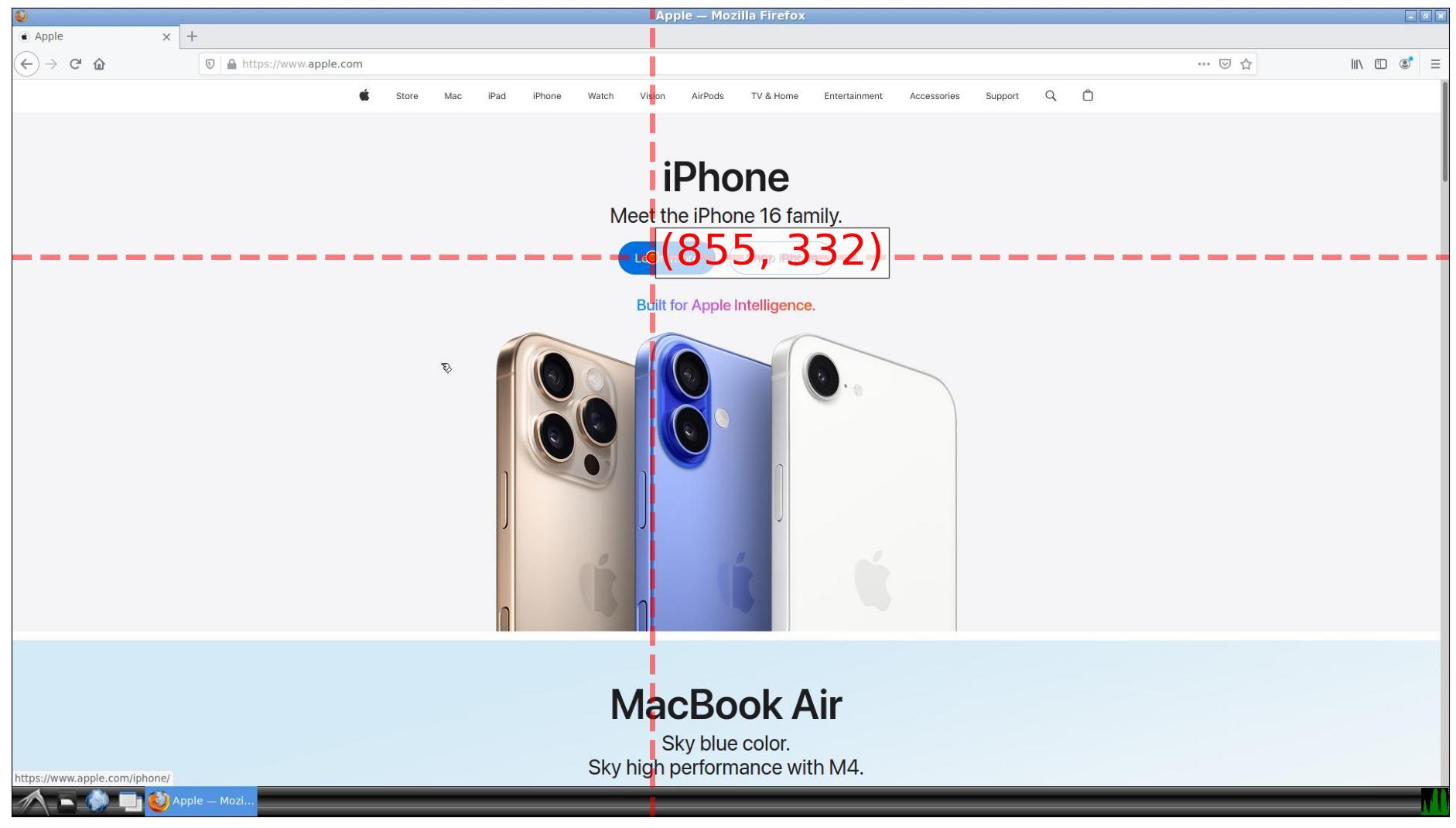} &
    \shortstack[l]{%
      \textbf{Intent:} “Having reached the iPhone page at \\
      apple.com, the next logical step would be to \\ 
      learn more about the new iPhone models.”\\[0.5ex]
      \textbf{Action:} “DoubleClick(855, 332)”
    }%
  \end{tabular}
  \caption{At episode 110, the intent description from \textit{ScreenExplorer-3B-E1} demonstrates its ability to recognize that the current page content pertains to "apple.com". Furthermore, the model specified the next operational plan in its intent, and subsequently executed actions targeting the correct button.}\label{fig:case-study-9}
\end{figure}

In this section, we will present several cases during the training process of \textit{ScreenExplorer-3B-E1} to understand the development of exploration capabilities throughout the reinforcement learning training process, from Figure \ref{fig:case-study-1} to \ref{fig:case-study-9}.

\FloatBarrier
\section{Experiments on Qwen2.5-VL-7B}\label{appendix:7B-VLM}

To investigate whether our method could be effective on larger models and validate the importance of incorporating a world model, we used \textit{Qwen2.5-VL-7B} as our base model, enabled all reward settings for RL training, and added an ablation group without the world model (\textit{Qwen2.5-VL-7B w/o World Model}) for comparison with \textit{ScreenExplorer-3B-E1} discussed in the main text. 

\begin{figure}[h]
    \centering
    \includegraphics[width=1\linewidth]{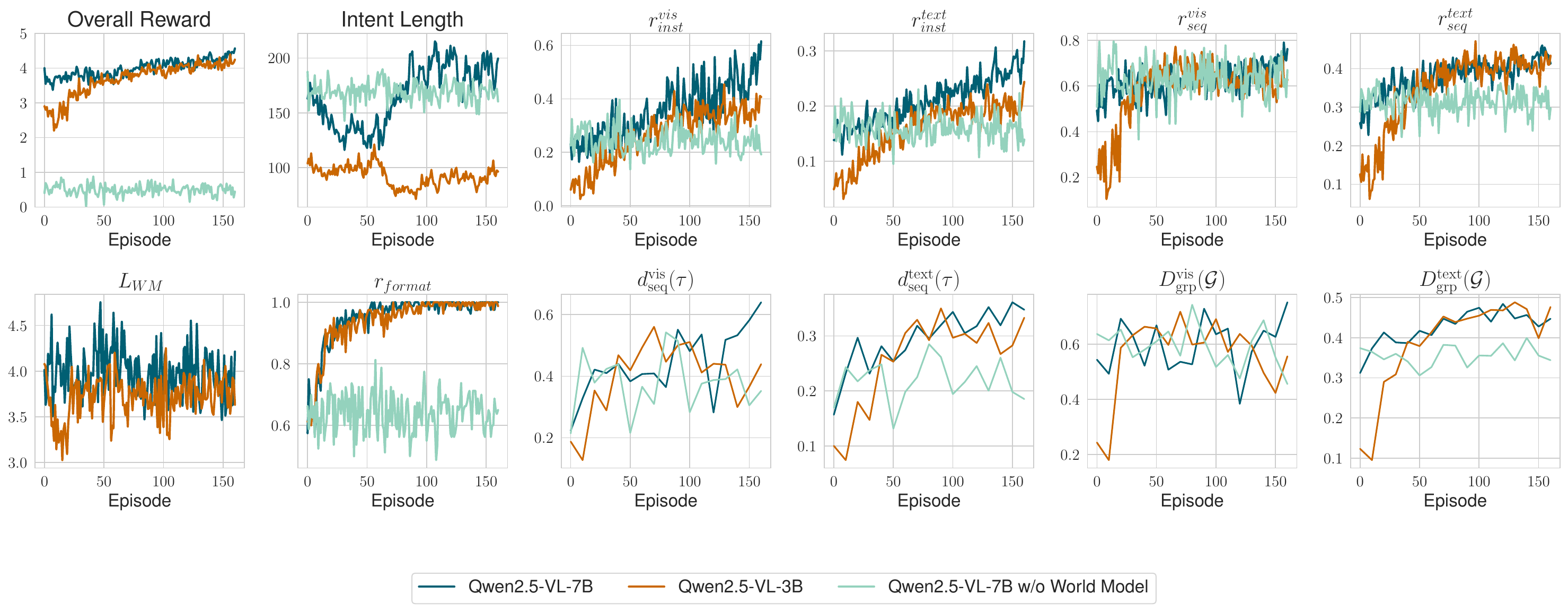}
    \caption{Comparison of rewards and metrics during RL training for free exploration tasks using \textit{Qwen2.5-VL-7B} and \textit{Qwen2.5-VL-3B} as base models.}
    \label{fig:7B-3B}
\end{figure}

Figure \ref{fig:7B-3B} shows the changes in various rewards and metrics during the training process for these three models. We found that compared to the 3B model, the larger 7B model achieved higher rewards in free exploration tasks and could explore pages with greater visual and textual diversity, as reflected in $r^{vis}_{inst}$ and $r^{text}_{inst}$. The graph shows that the 7B model obtained higher rewards from the beginning, indicating its ability to interact effectively with the environment from the start, ultimately achieving greater exploration rewards with a faster increase in reward values. The most significant performance gap between 7B and 3B models was observed in the intent output length, with the 7B model tending to generate longer text outputs.

However, for the ablation group \textit{Qwen2.5-VL-7B w/o World Model}, its exploration capabilities failed to improve beyond the initial 7B model's performance. This is evidenced by the lack of improvement in all rewards and metrics, remaining at the same capability level as the original 7B model. This demonstrates the critical importance of incorporating a world model in reinforcement learning for exploration in the open GUI world.

\begin{figure}[h]
  \centering
 \begin{tabular}{
    @{}
    m{0.195\textwidth}@{\hspace{1pt}}
    m{0.195\textwidth}@{\hspace{1pt}}
    m{0.195\textwidth}@{\hspace{1pt}}
    m{0.195\textwidth}@{\hspace{1pt}}
    m{0.195\textwidth}@{}
  }
    \includegraphics[width=\linewidth]{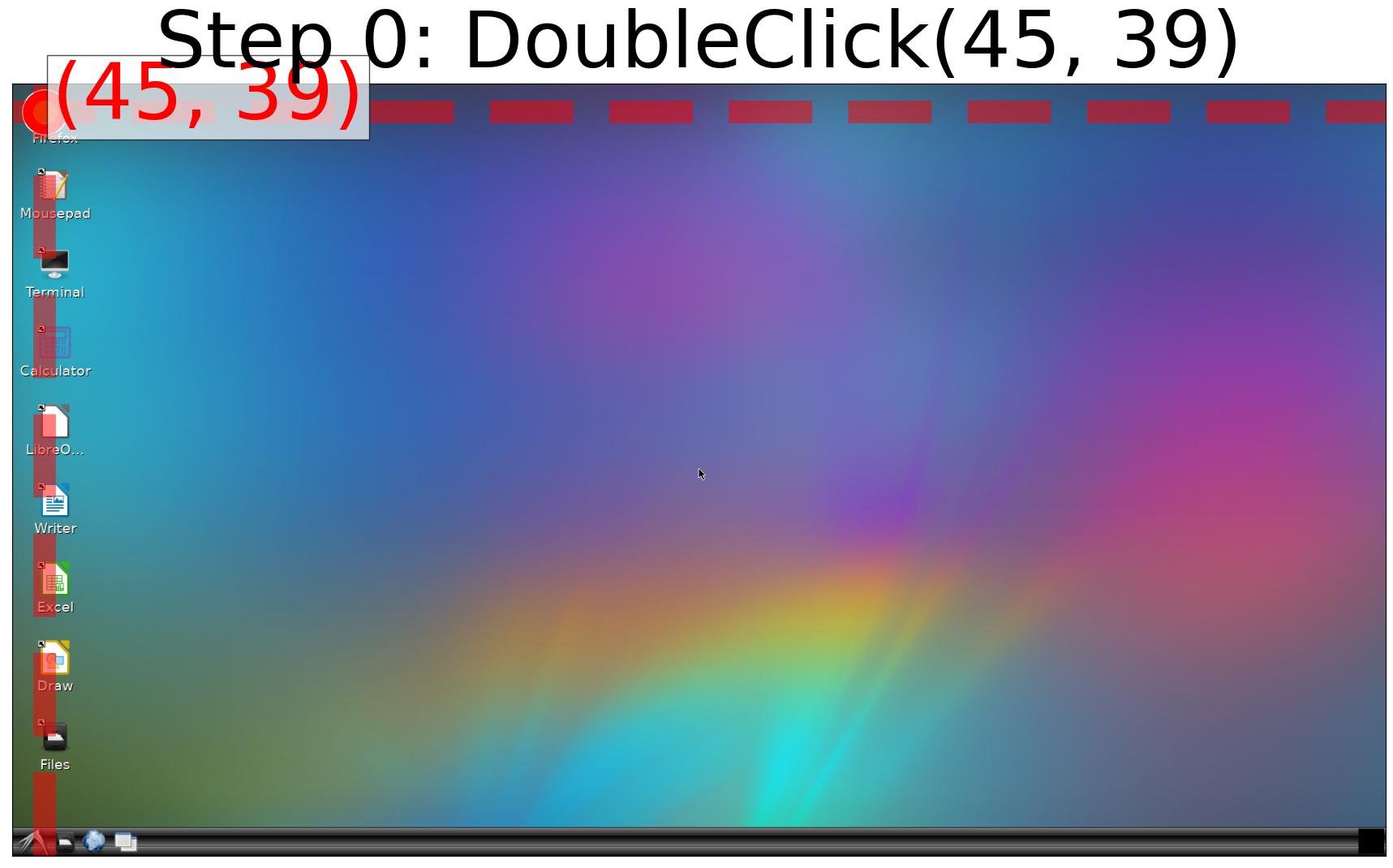} &
    \includegraphics[width=\linewidth]{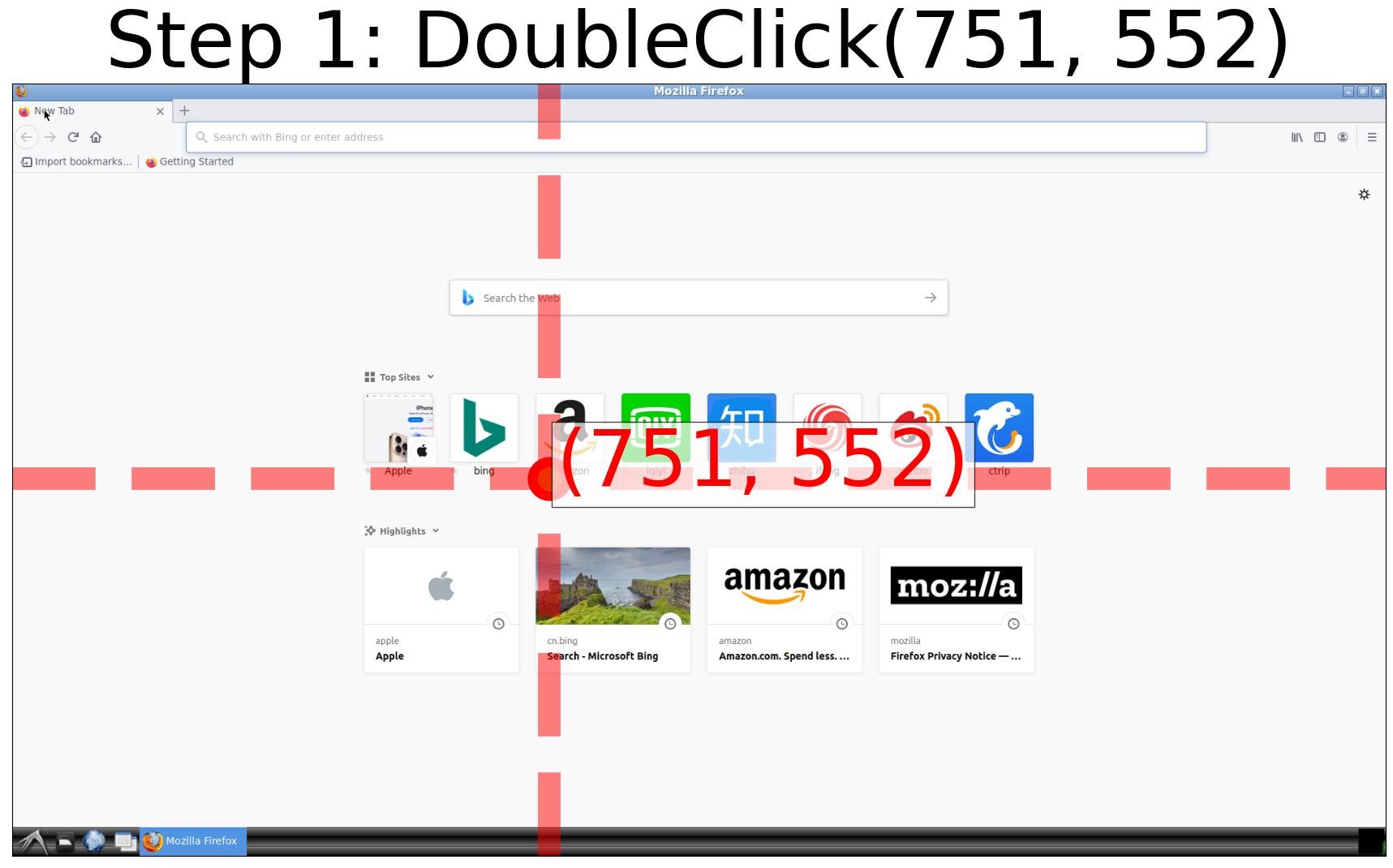} &
    \includegraphics[width=\linewidth]{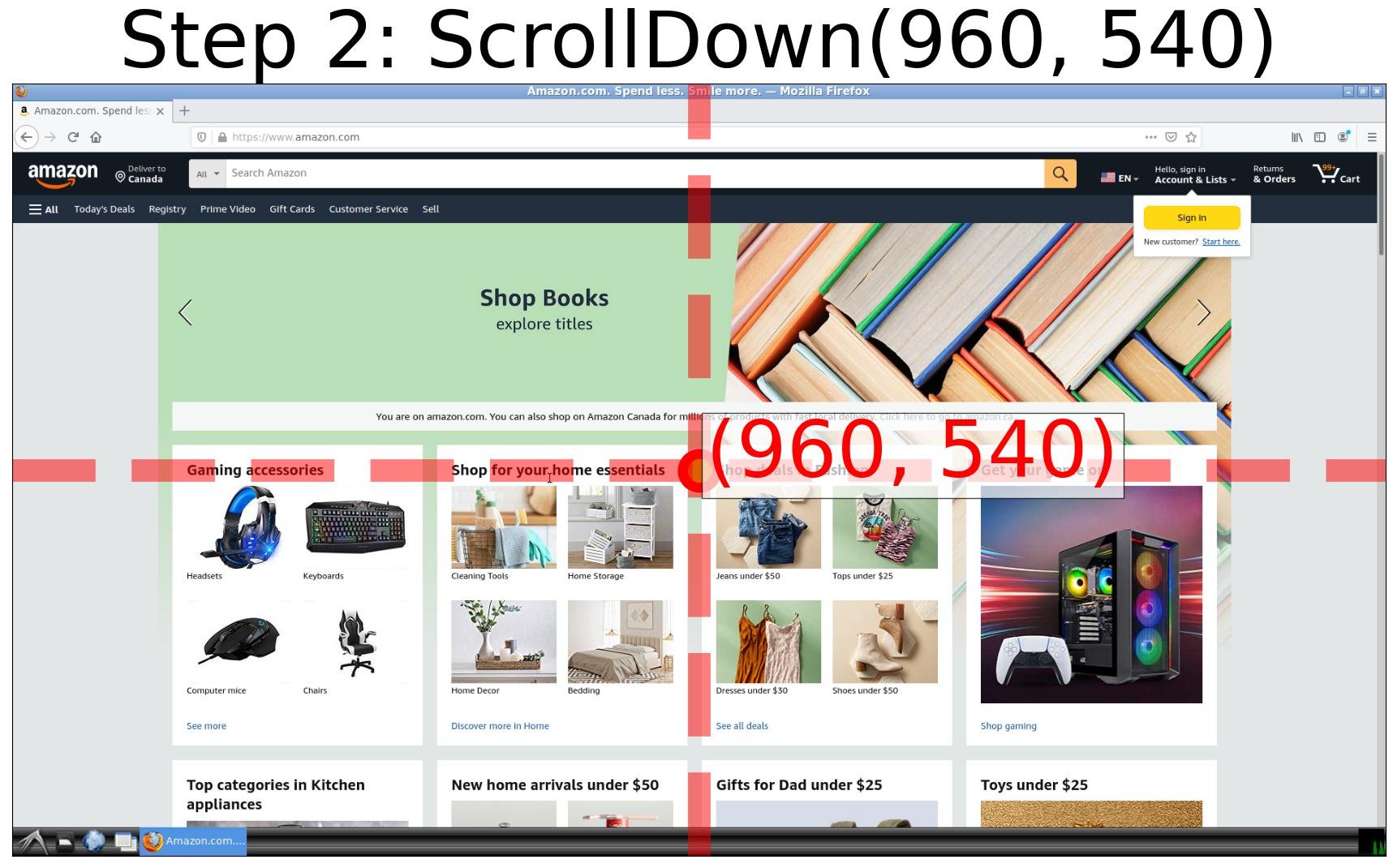} &
    \includegraphics[width=\linewidth]{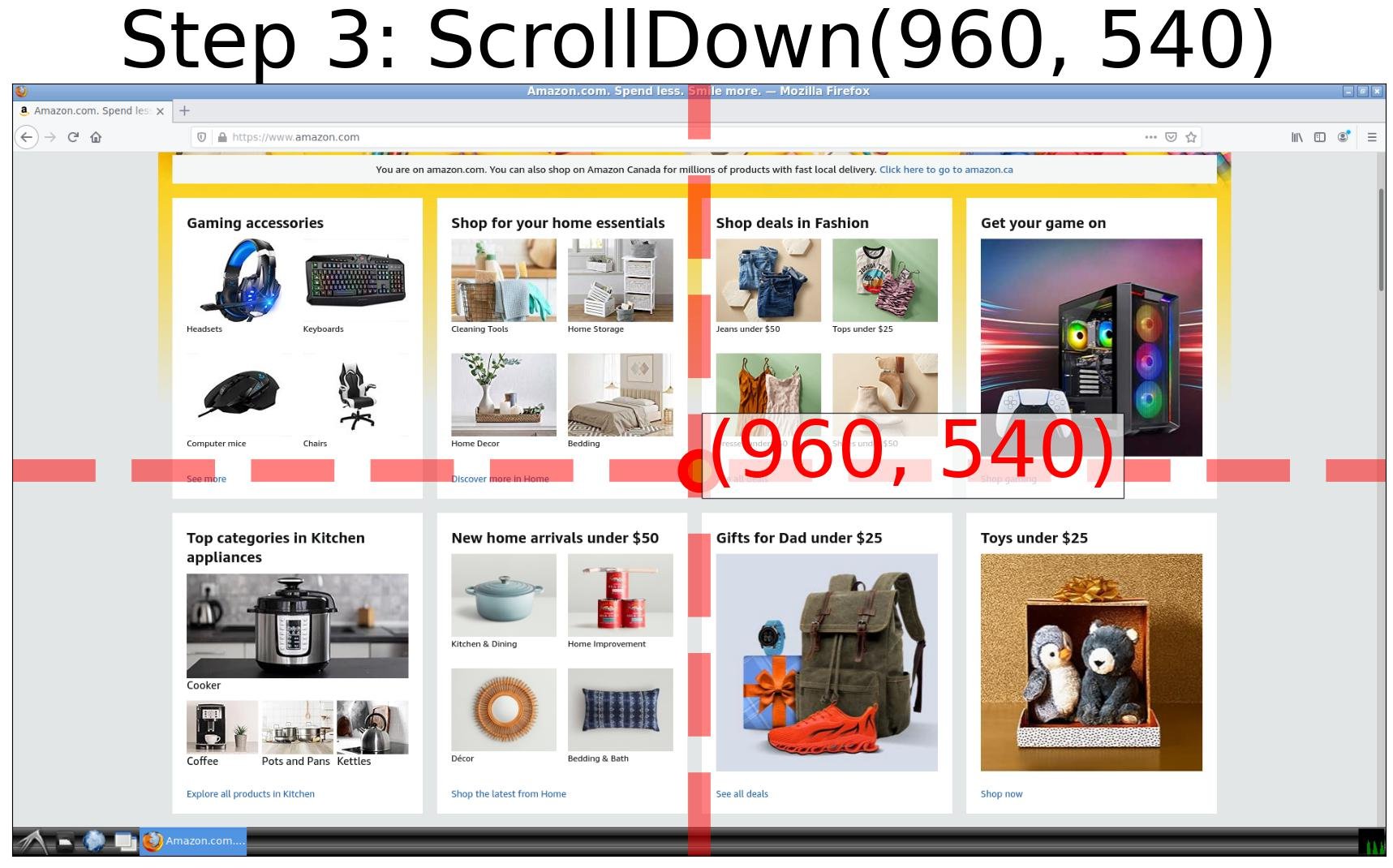} &
    \includegraphics[width=\linewidth]{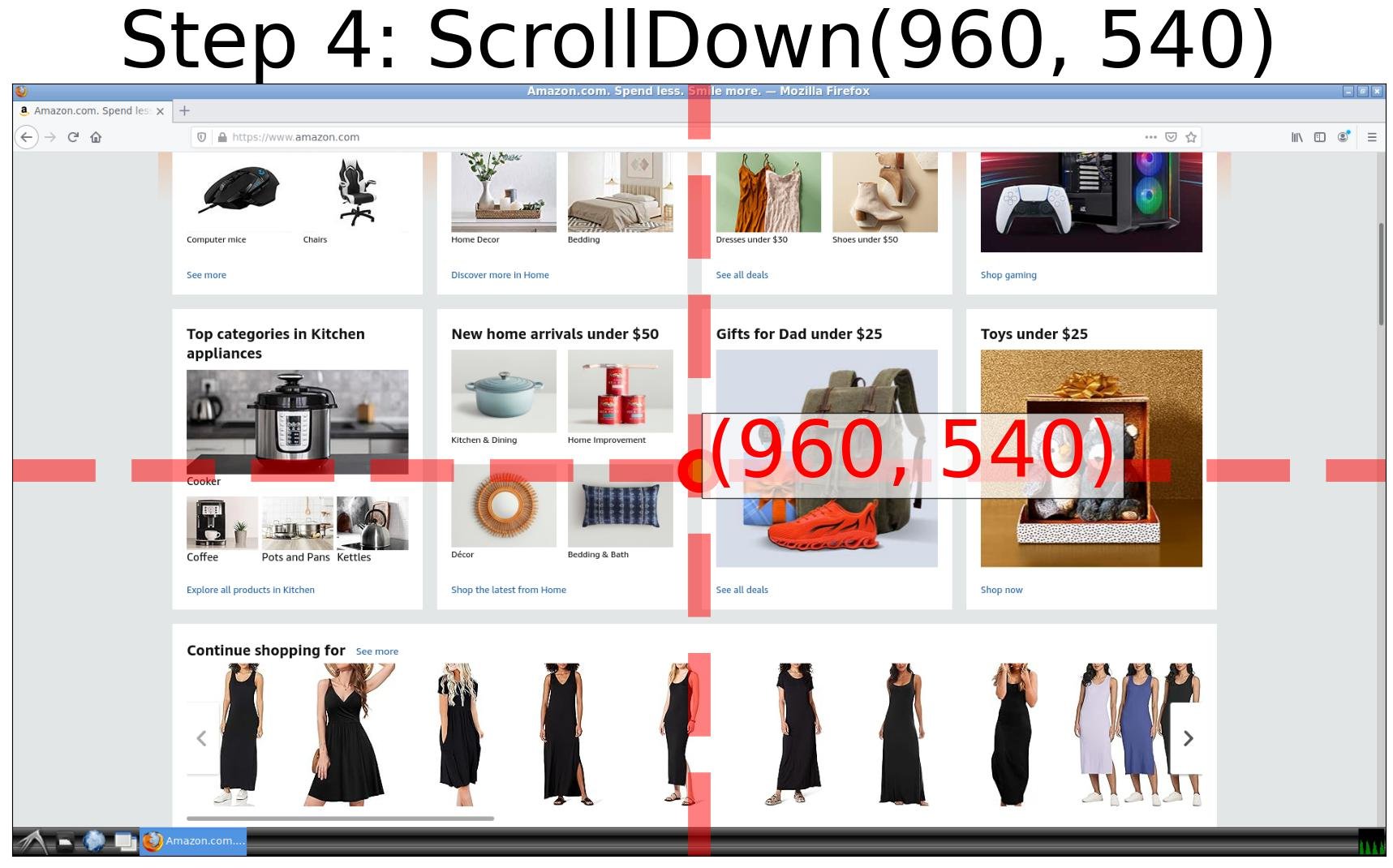} \\
    \includegraphics[width=\linewidth]{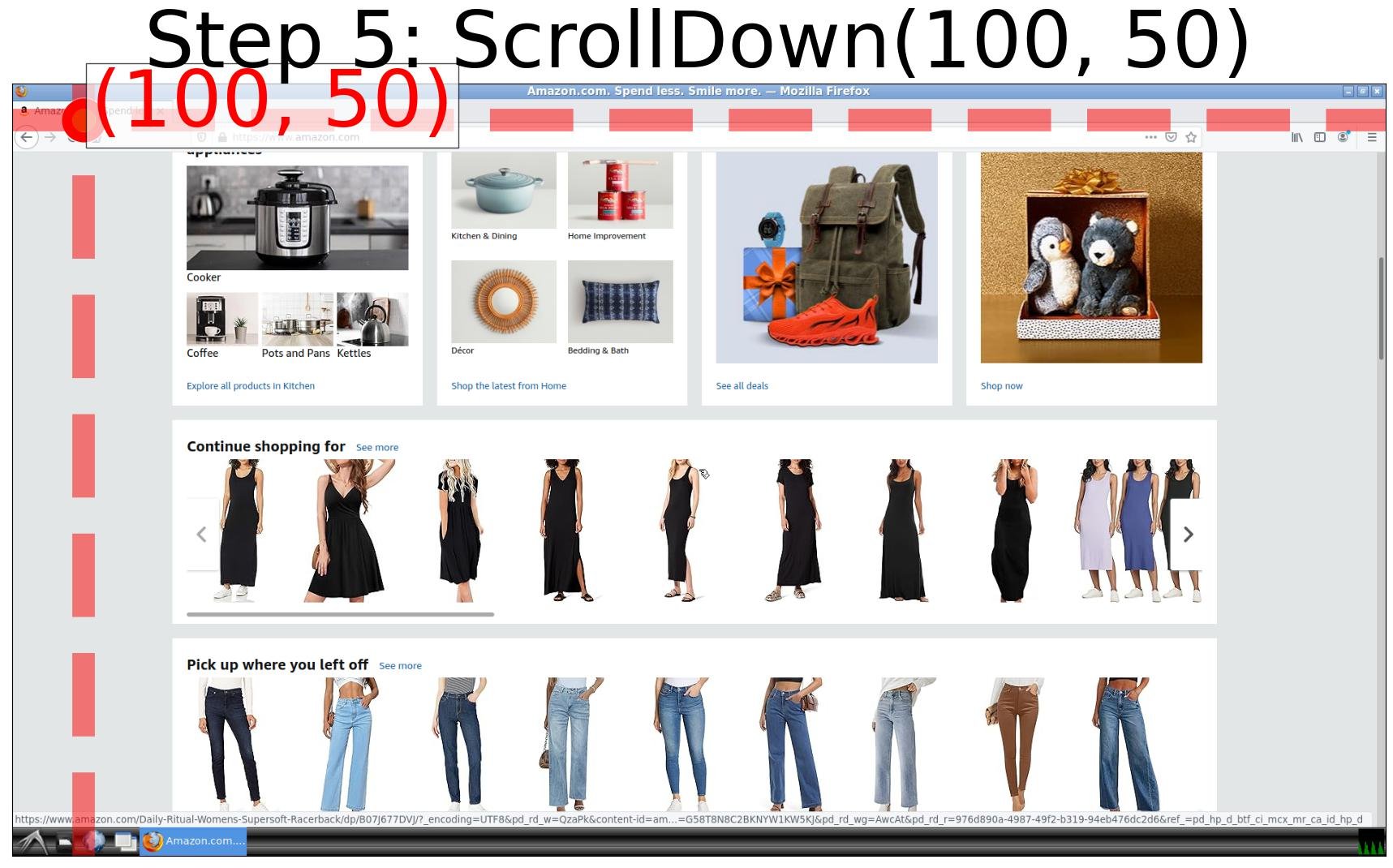} &
    \includegraphics[width=\linewidth]{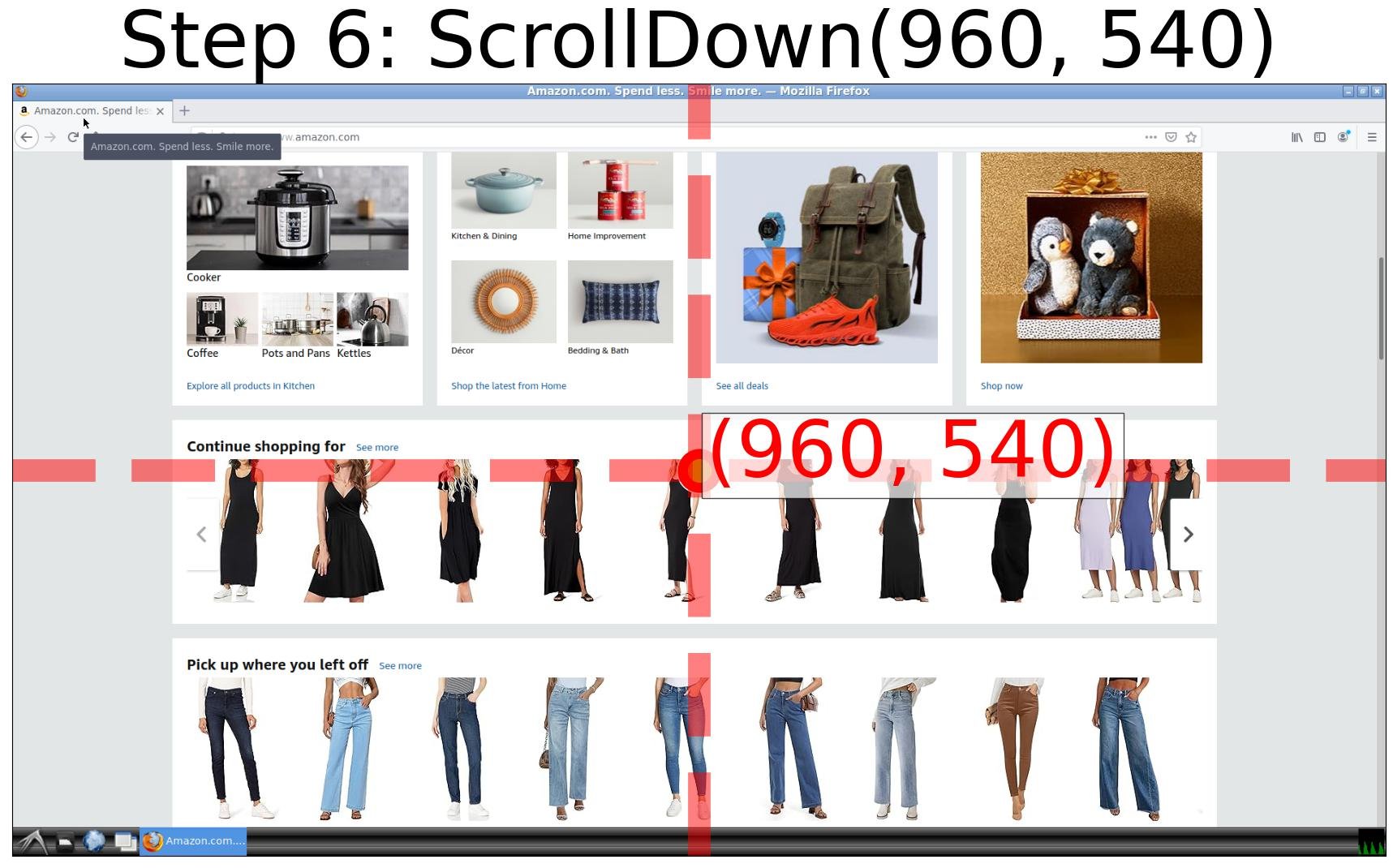} &
    \includegraphics[width=\linewidth]{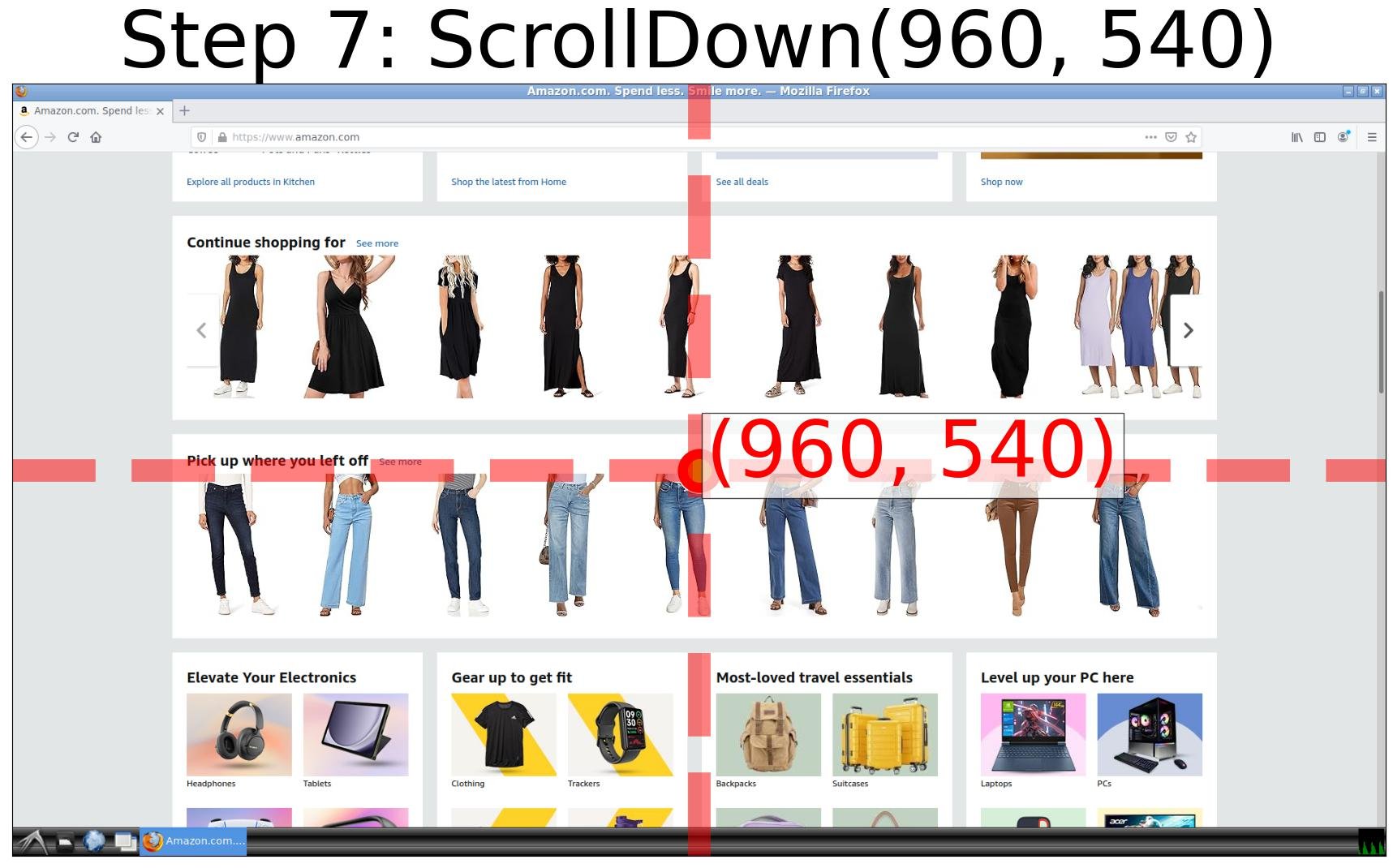} &
    \includegraphics[width=\linewidth]{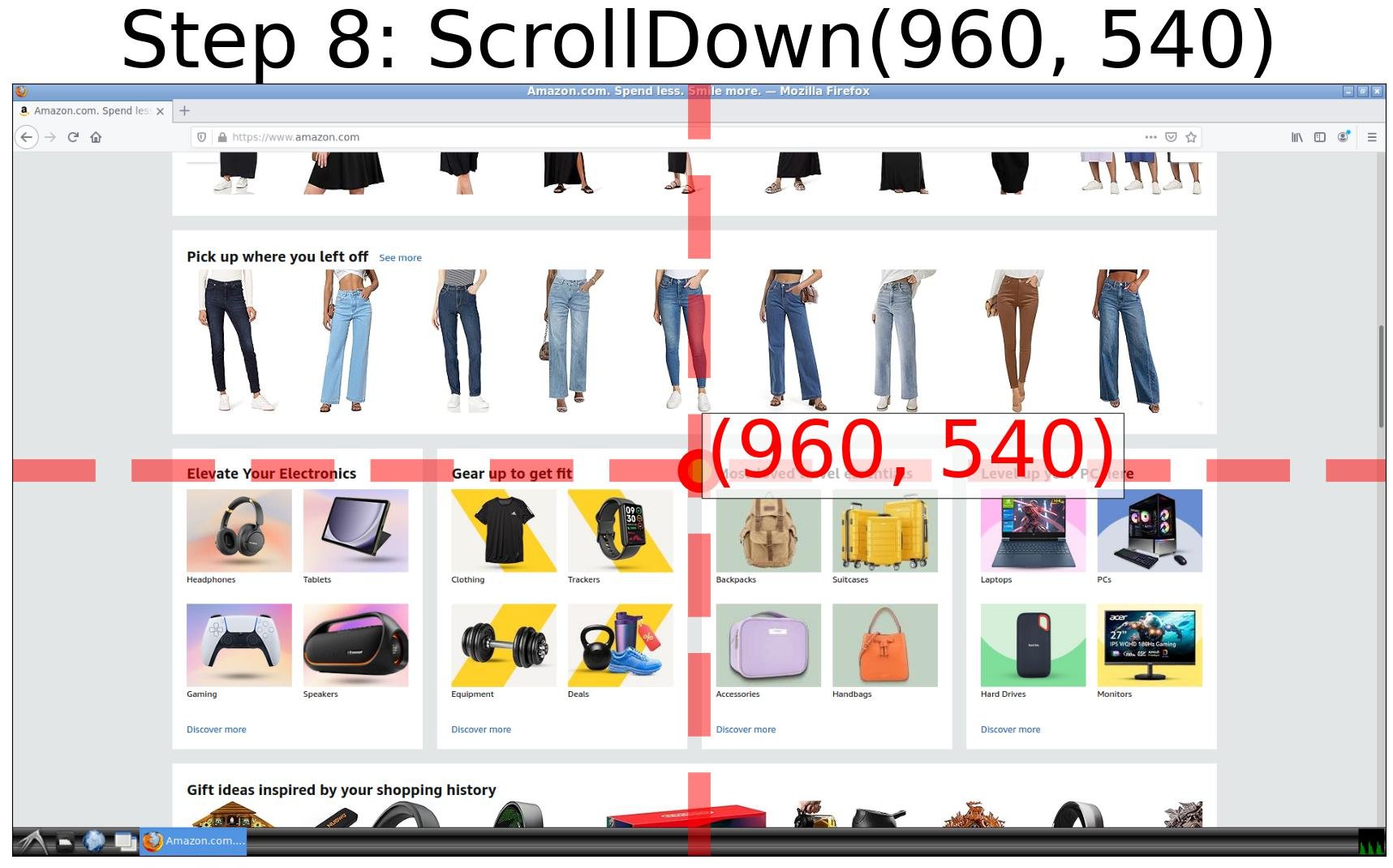} &
    \includegraphics[width=\linewidth]{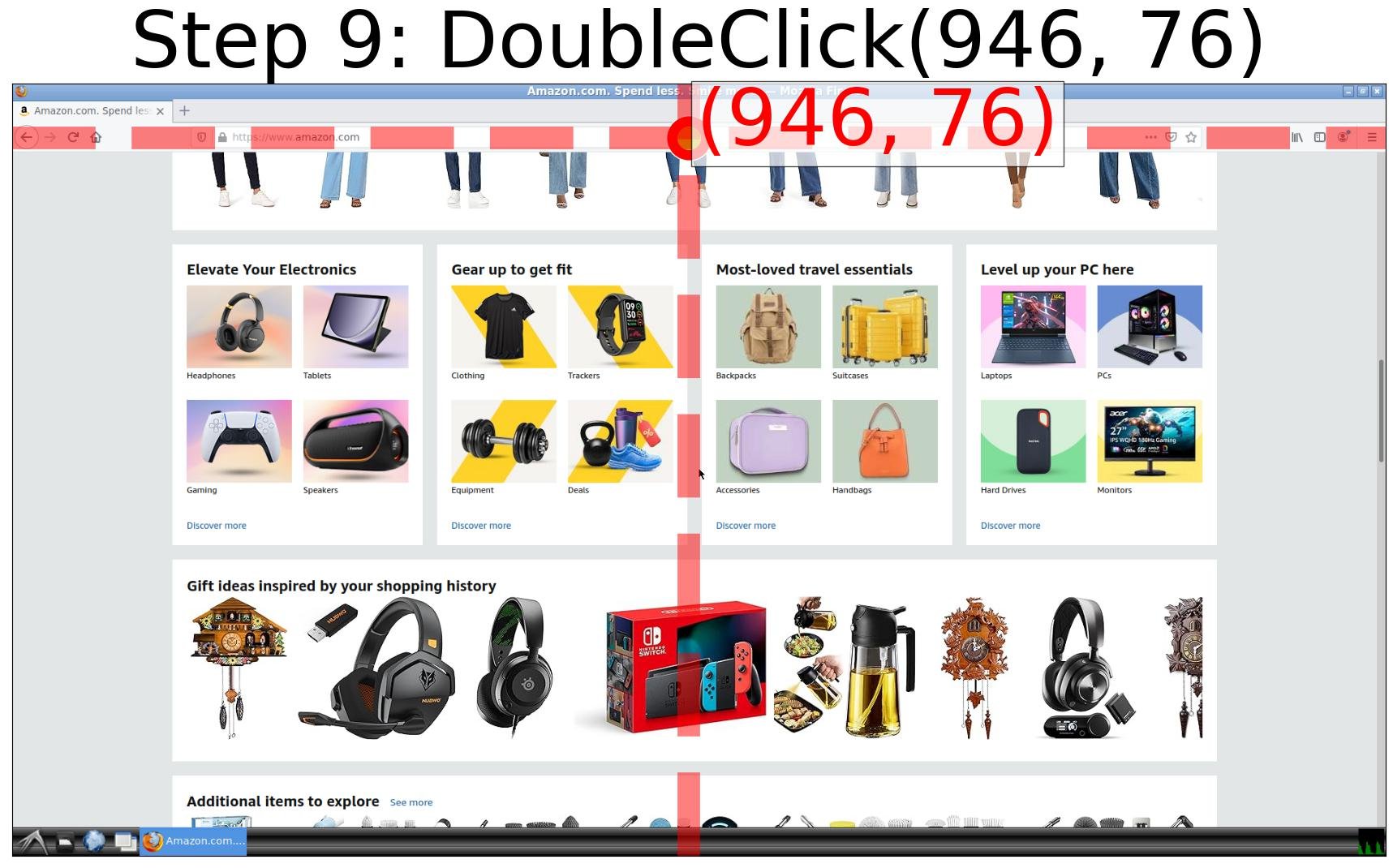}
  \end{tabular}
  \caption{Episode-50 of \textit{Qwen2.5-VL-7B}: Compared to the 3B model, the 7B model demonstrates superior proficiency in utilizing the ScrollDown command for web page navigation.}\label{fig:7B-case-study-1}
\end{figure}

\begin{figure}[h]
  \centering
 \begin{tabular}{
    @{}
    m{0.195\textwidth}@{\hspace{1pt}}
    m{0.195\textwidth}@{\hspace{1pt}}
    m{0.195\textwidth}@{\hspace{1pt}}
    m{0.195\textwidth}@{\hspace{1pt}}
    m{0.195\textwidth}@{}
  }
    \includegraphics[width=\linewidth]{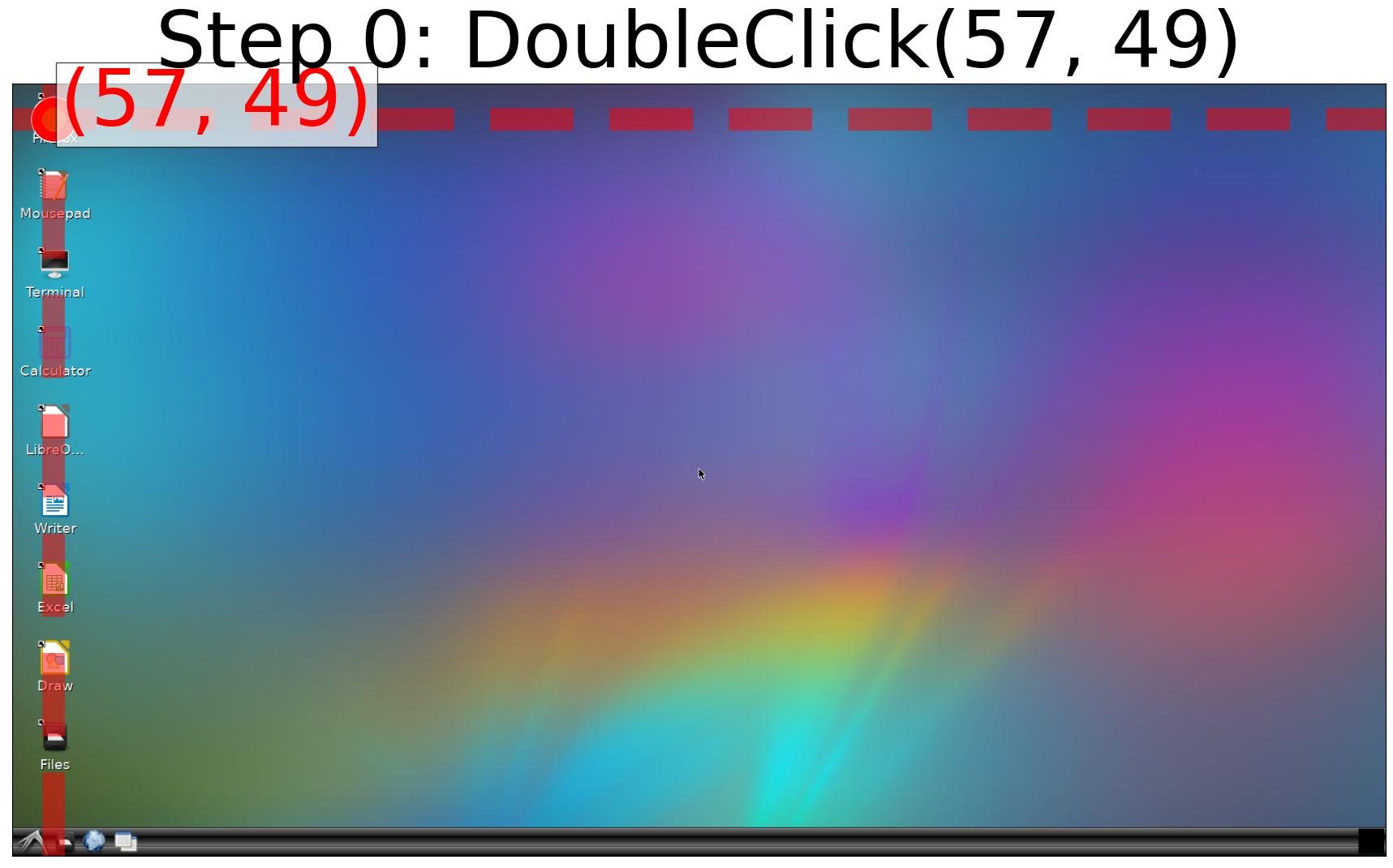} &
    \includegraphics[width=\linewidth]{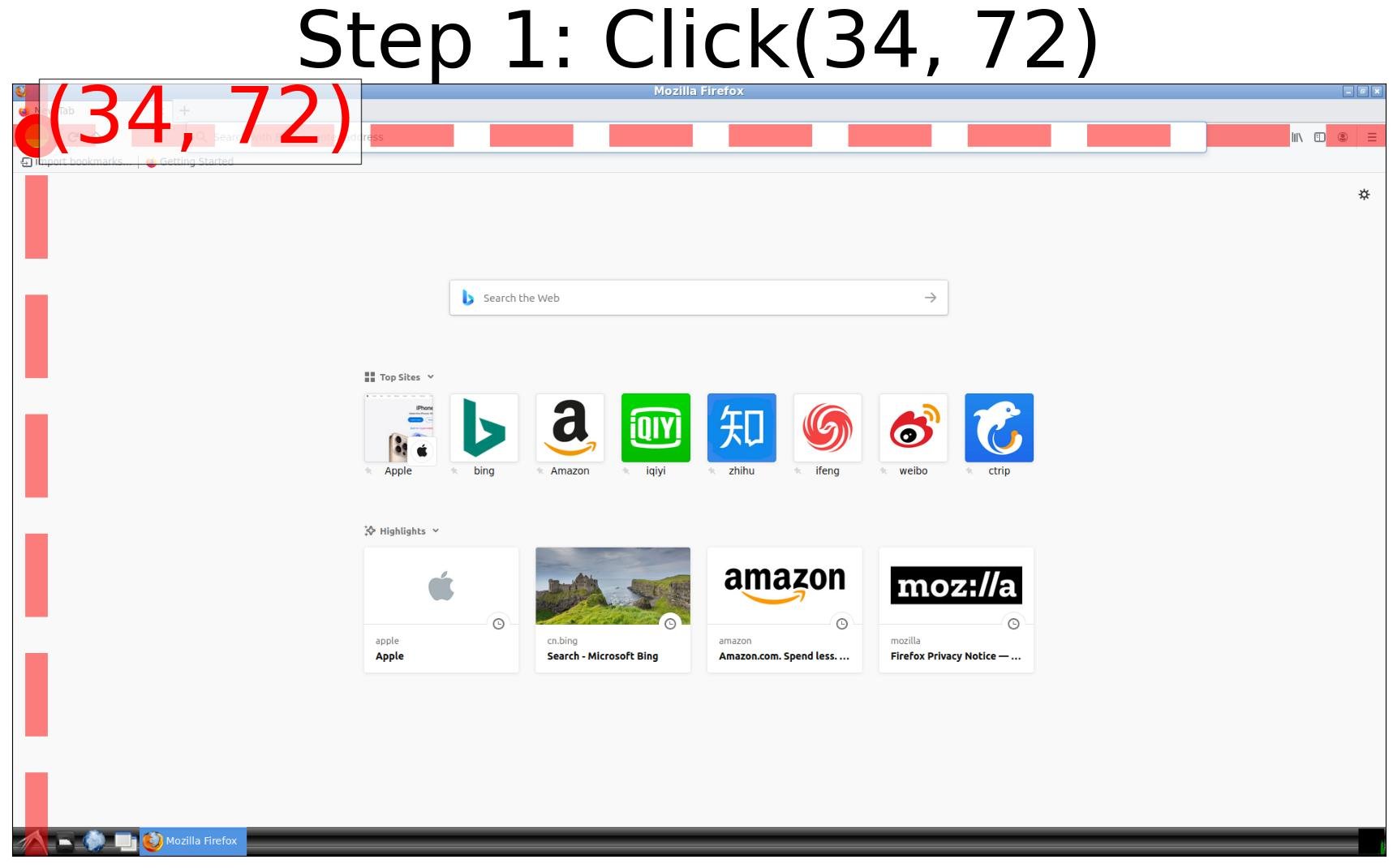} &
    \includegraphics[width=\linewidth]{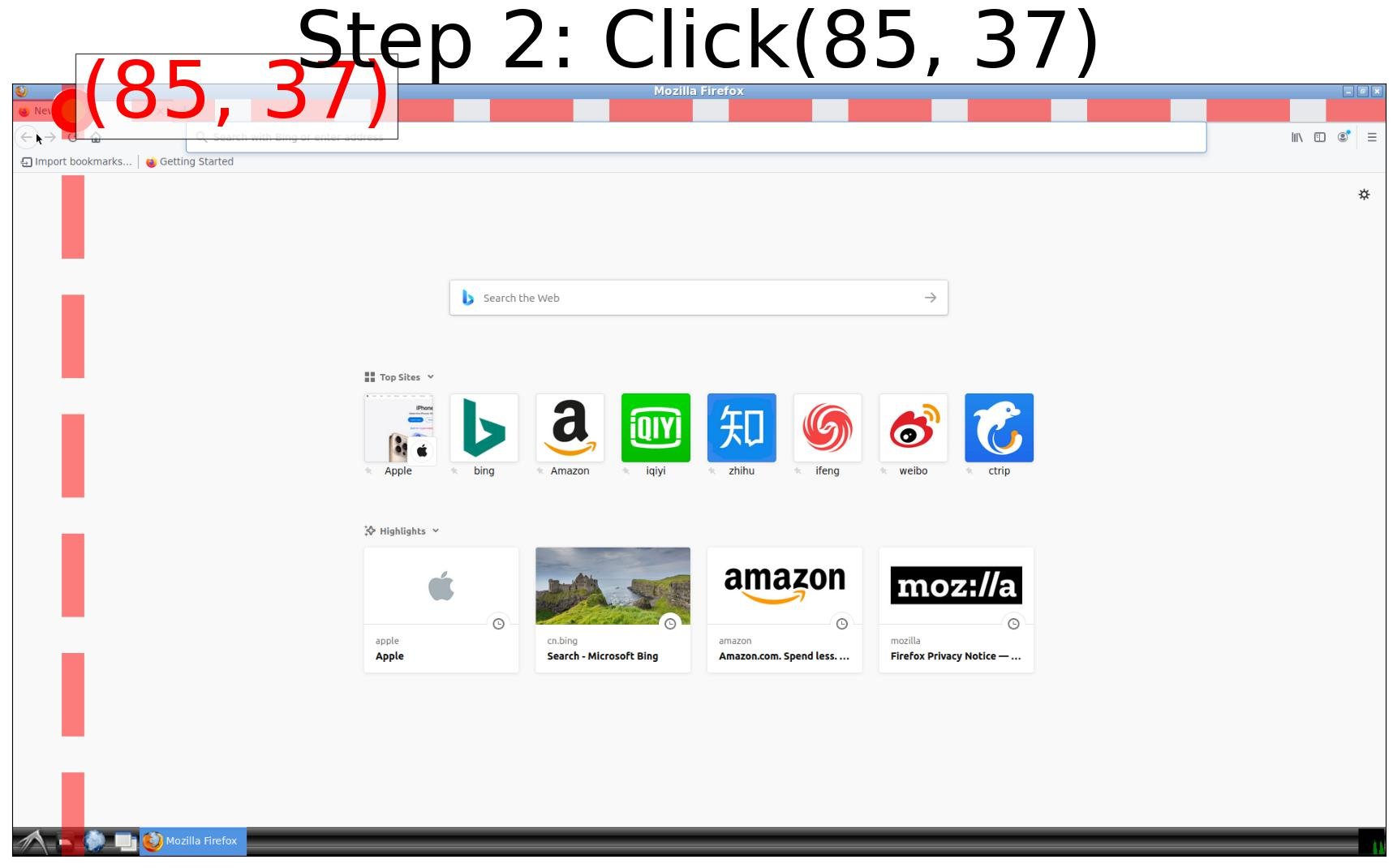} &
    \includegraphics[width=\linewidth]{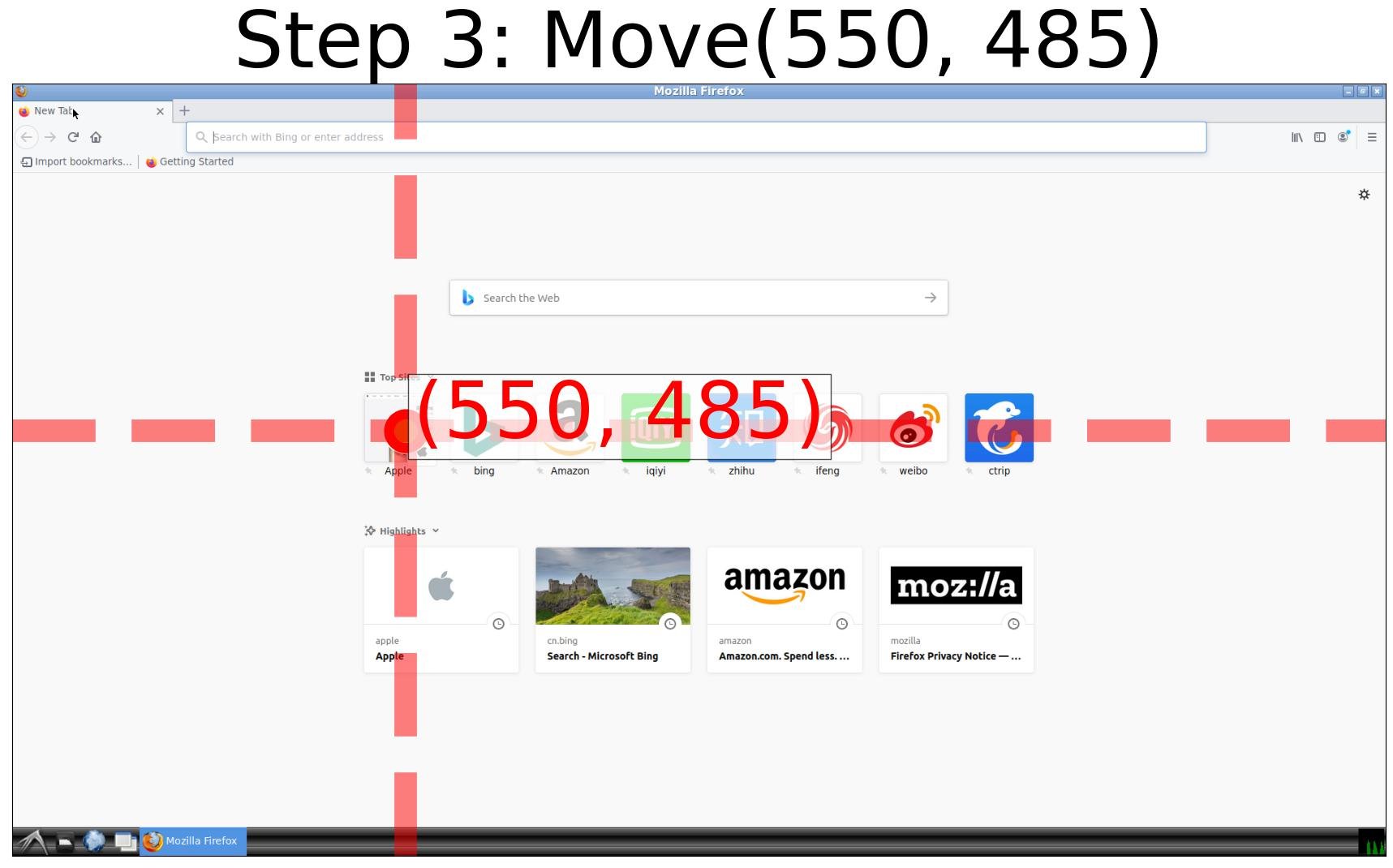} &
    \includegraphics[width=\linewidth]{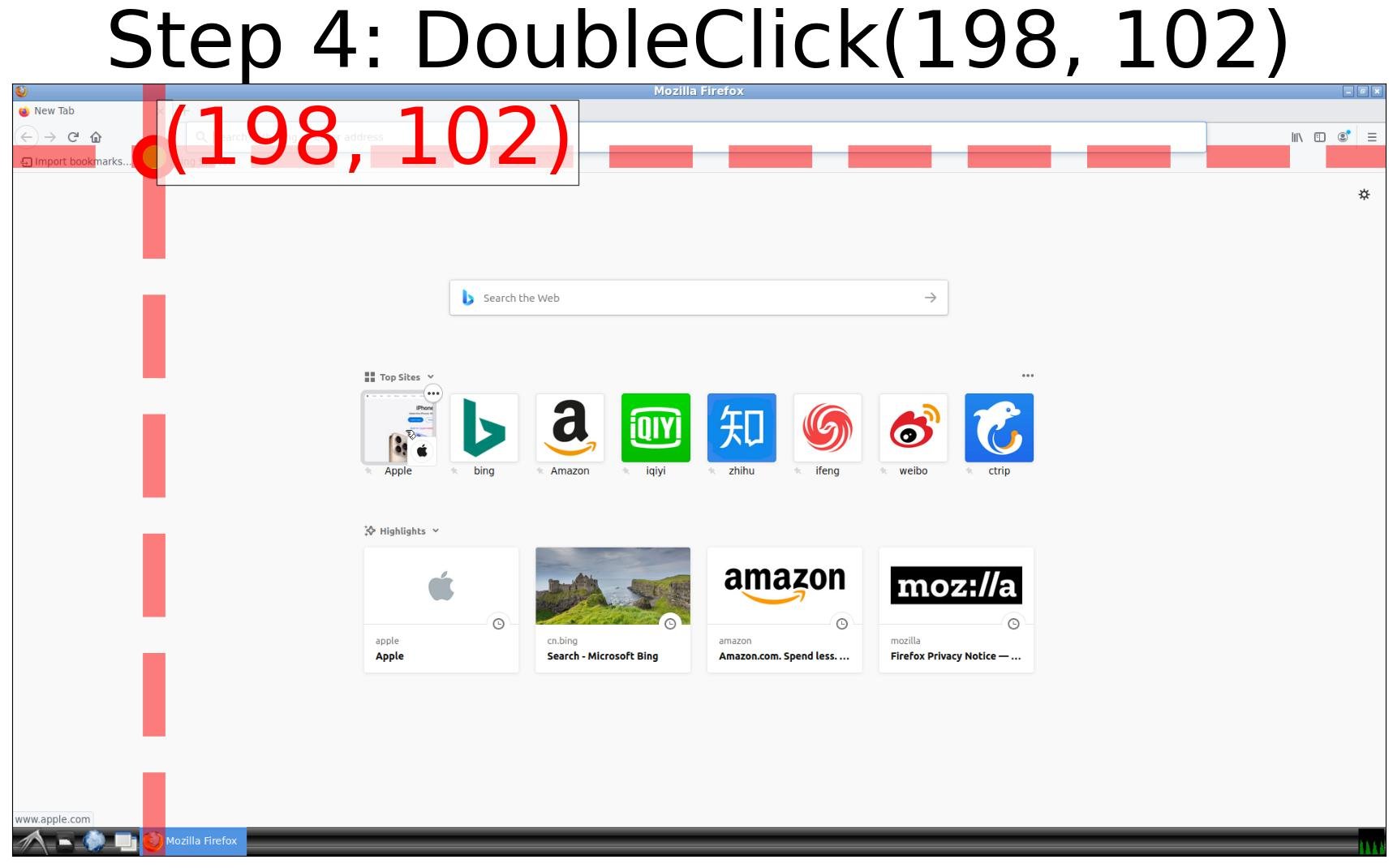} \\
    \includegraphics[width=\linewidth]{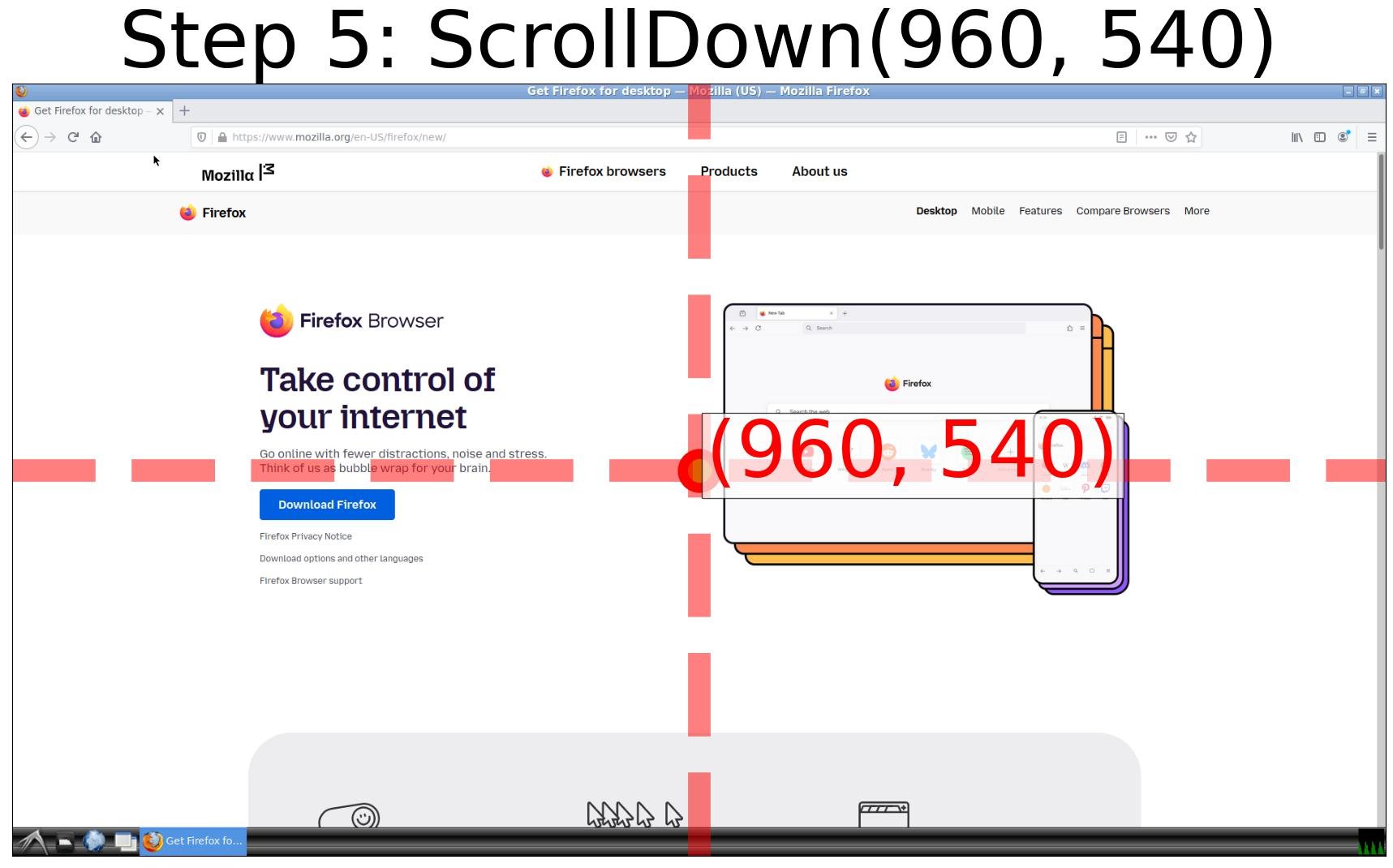} &
    \includegraphics[width=\linewidth]{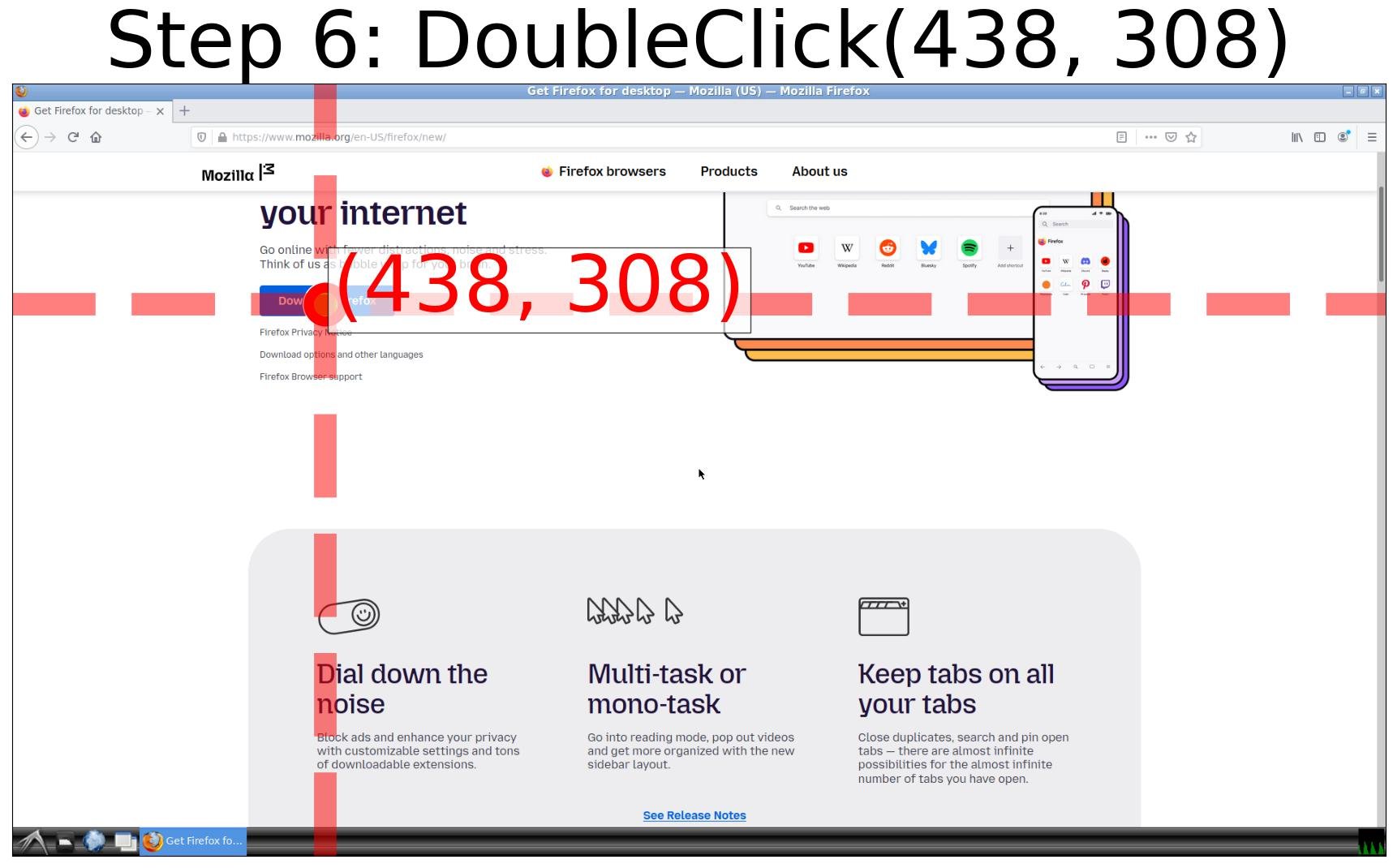} &
    \includegraphics[width=\linewidth]{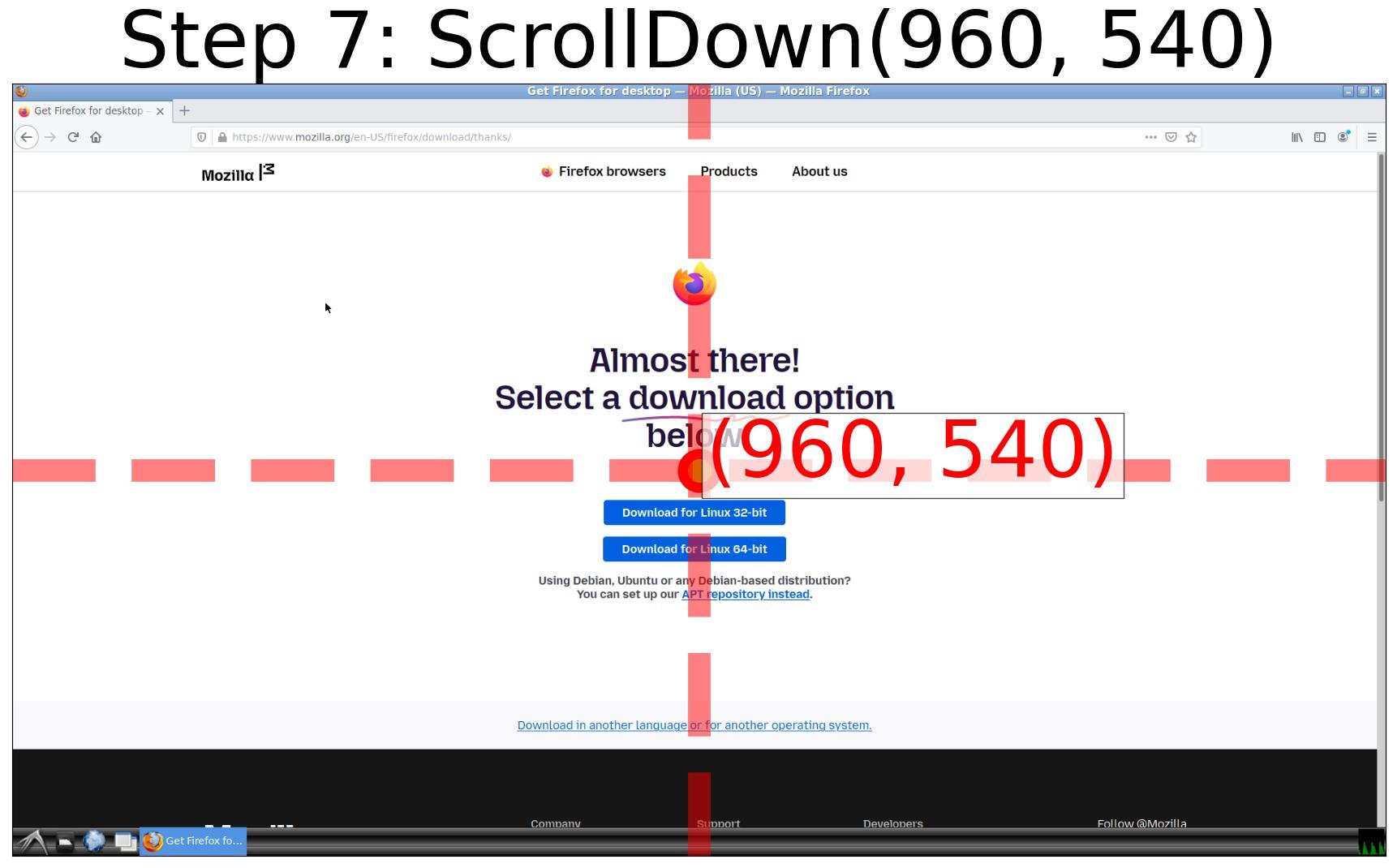} &
    \includegraphics[width=\linewidth]{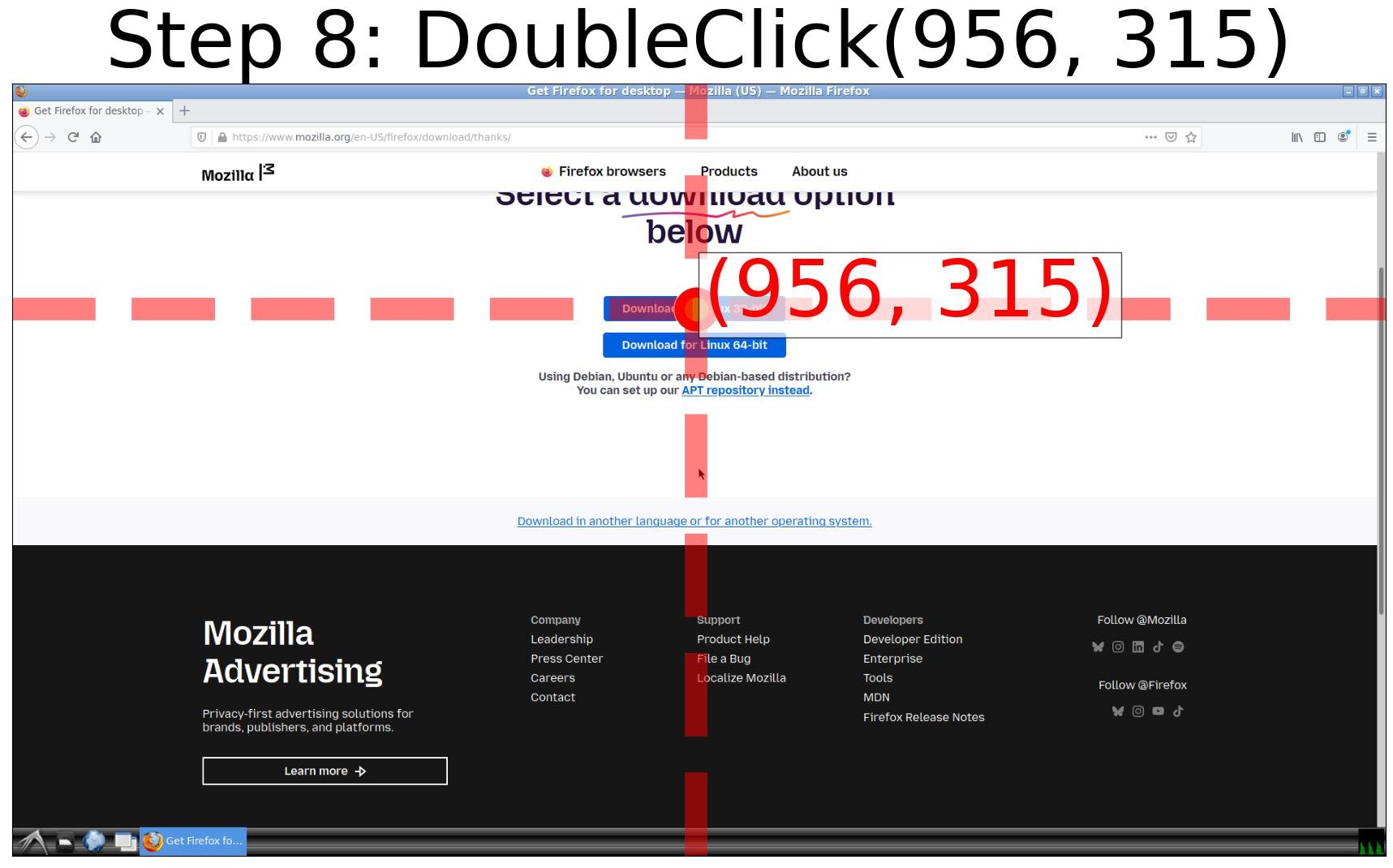} &
    \includegraphics[width=\linewidth]{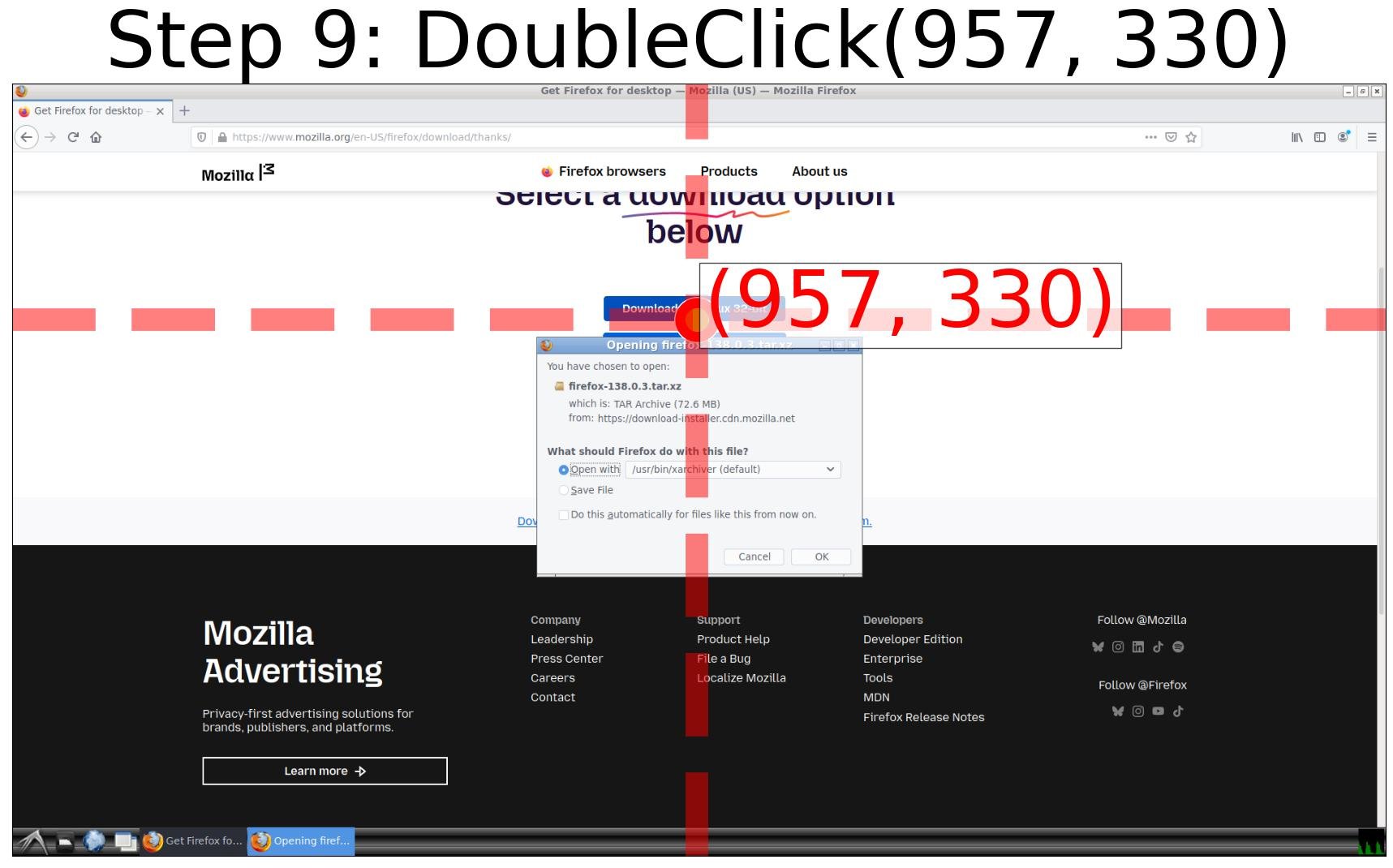}
  \end{tabular}
  \caption{Episode-50 of \textit{Qwen2.5-VL-7B}: The 7B model successfully completed a web browsing and software download sequence.}\label{fig:7B-case-study-2}
\end{figure}

\begin{figure}[h]
  \centering
 \begin{tabular}{
    @{}
    m{0.195\textwidth}@{\hspace{1pt}}
    m{0.195\textwidth}@{\hspace{1pt}}
    m{0.195\textwidth}@{\hspace{1pt}}
    m{0.195\textwidth}@{\hspace{1pt}}
    m{0.195\textwidth}@{}
  }
    \includegraphics[width=\linewidth]{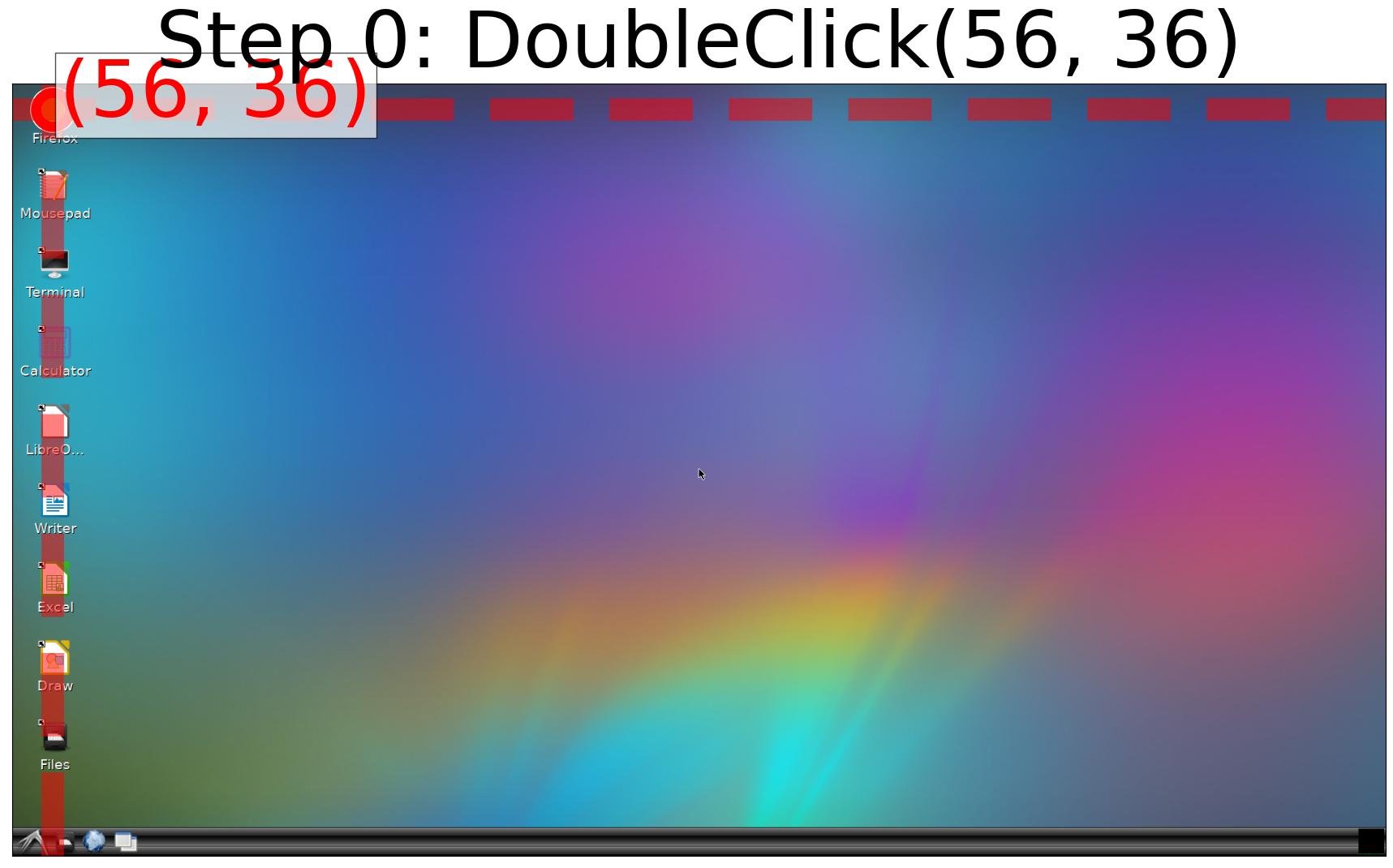} &
    \includegraphics[width=\linewidth]{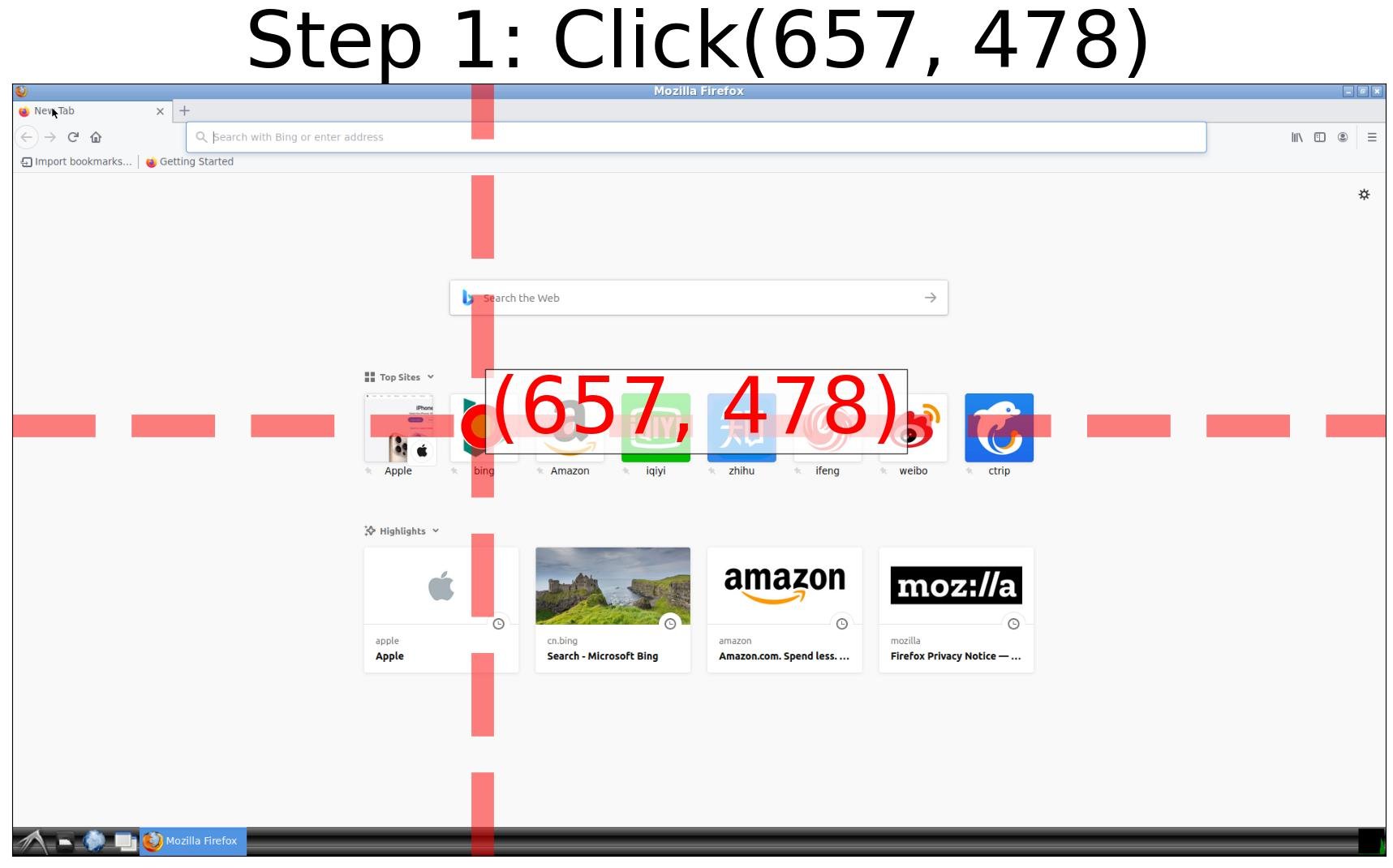} &
    \includegraphics[width=\linewidth]{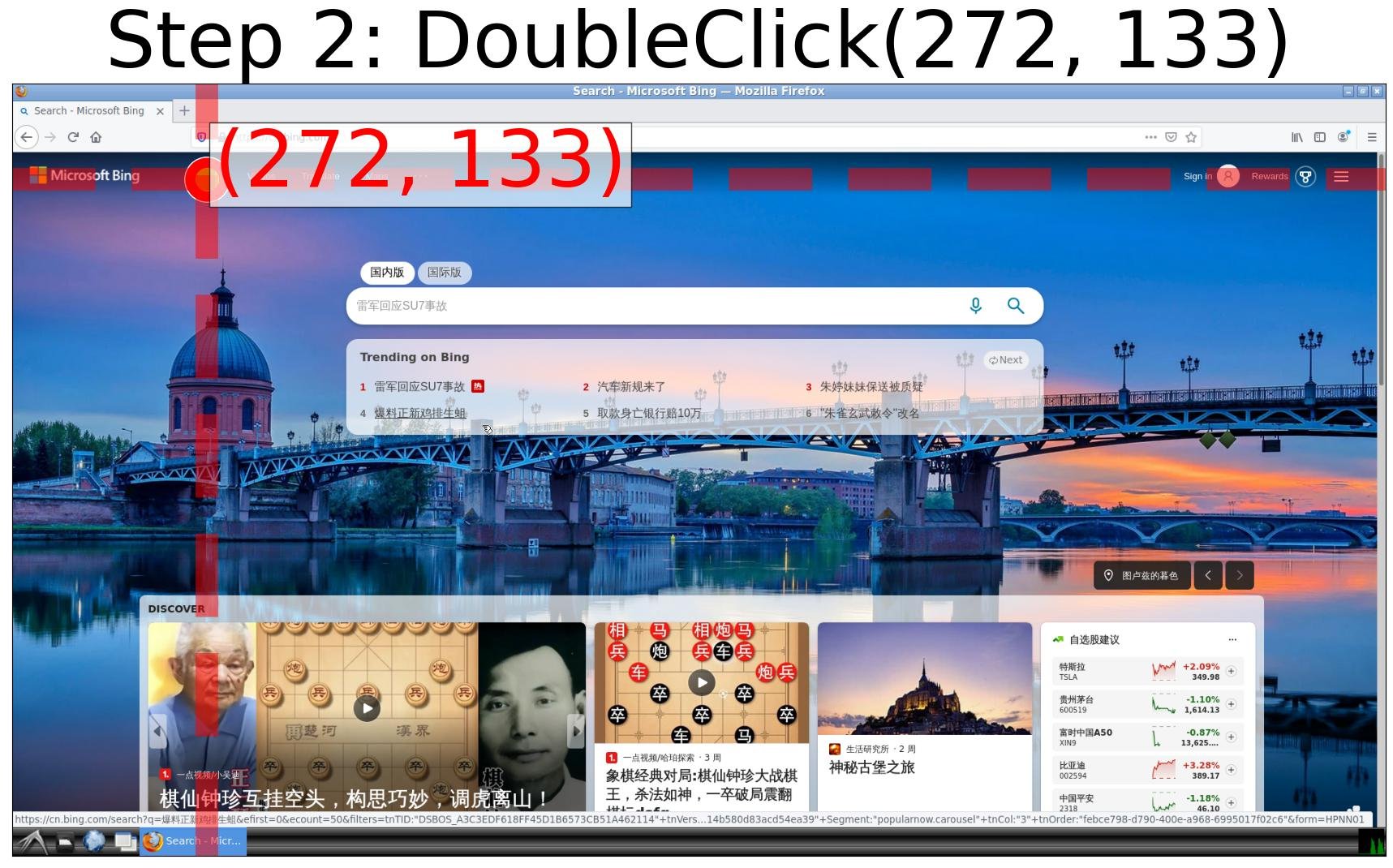} &
    \includegraphics[width=\linewidth]{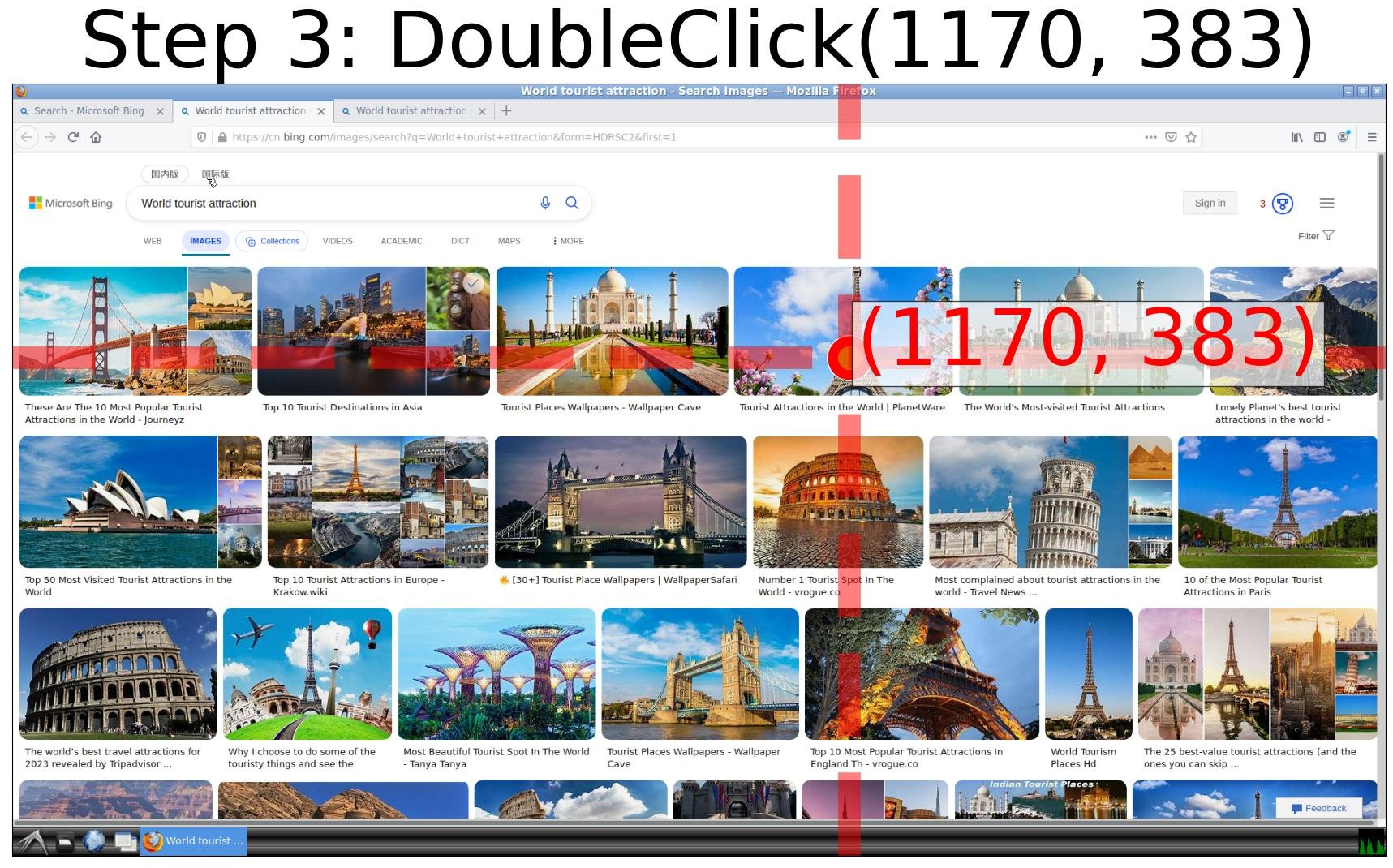} &
    \includegraphics[width=\linewidth]{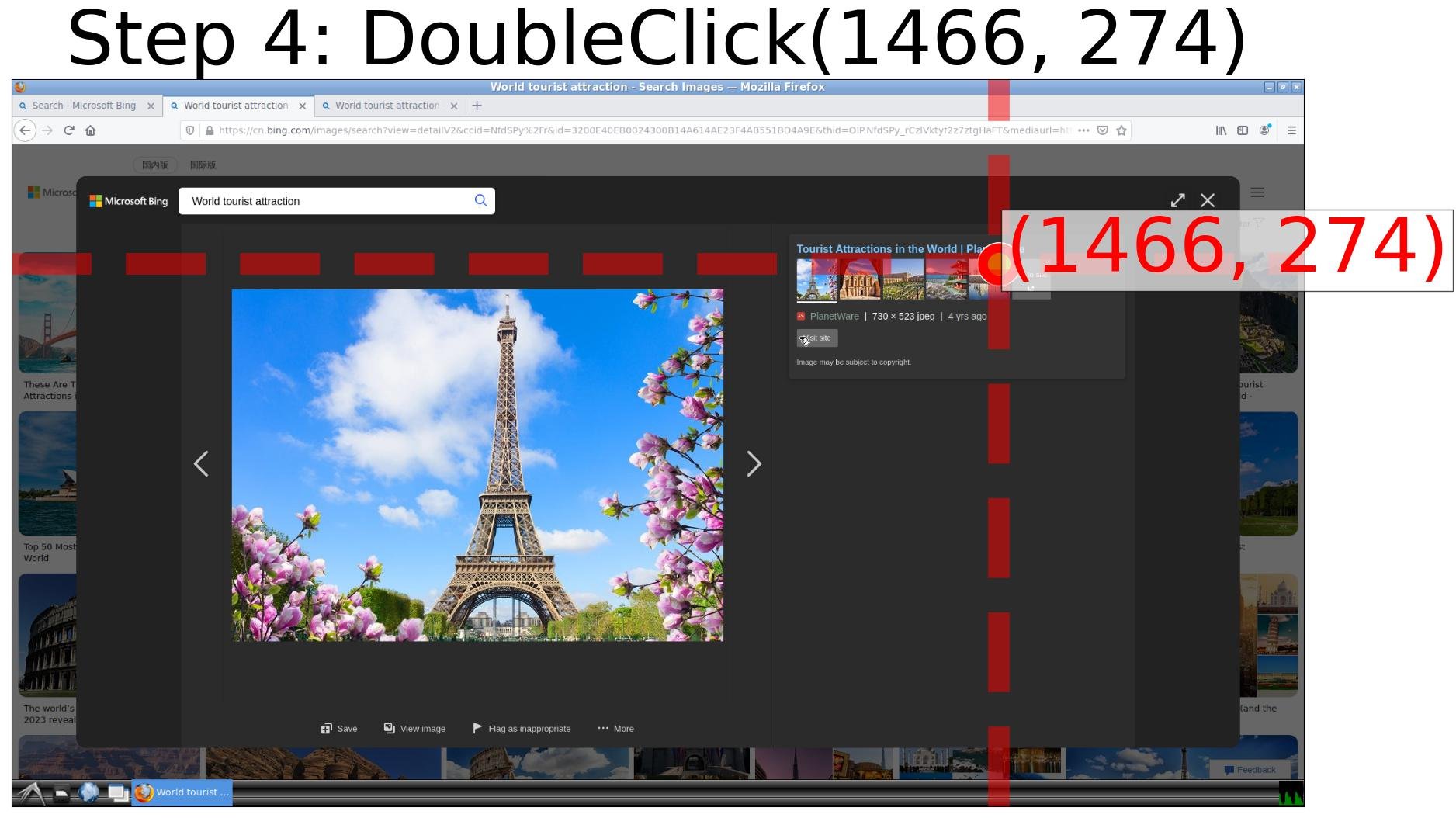} \\
    \includegraphics[width=\linewidth]{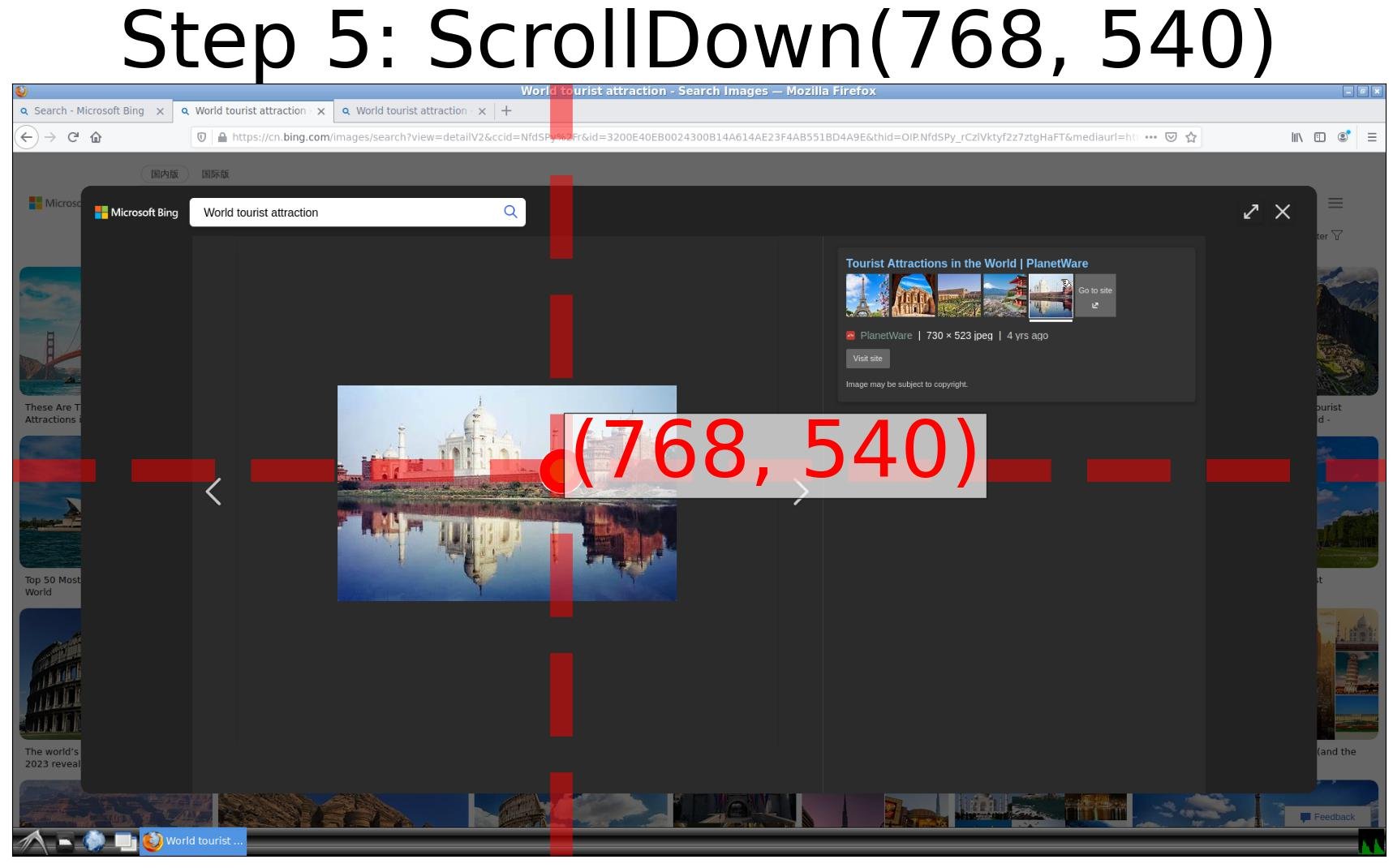} &
    \includegraphics[width=\linewidth]{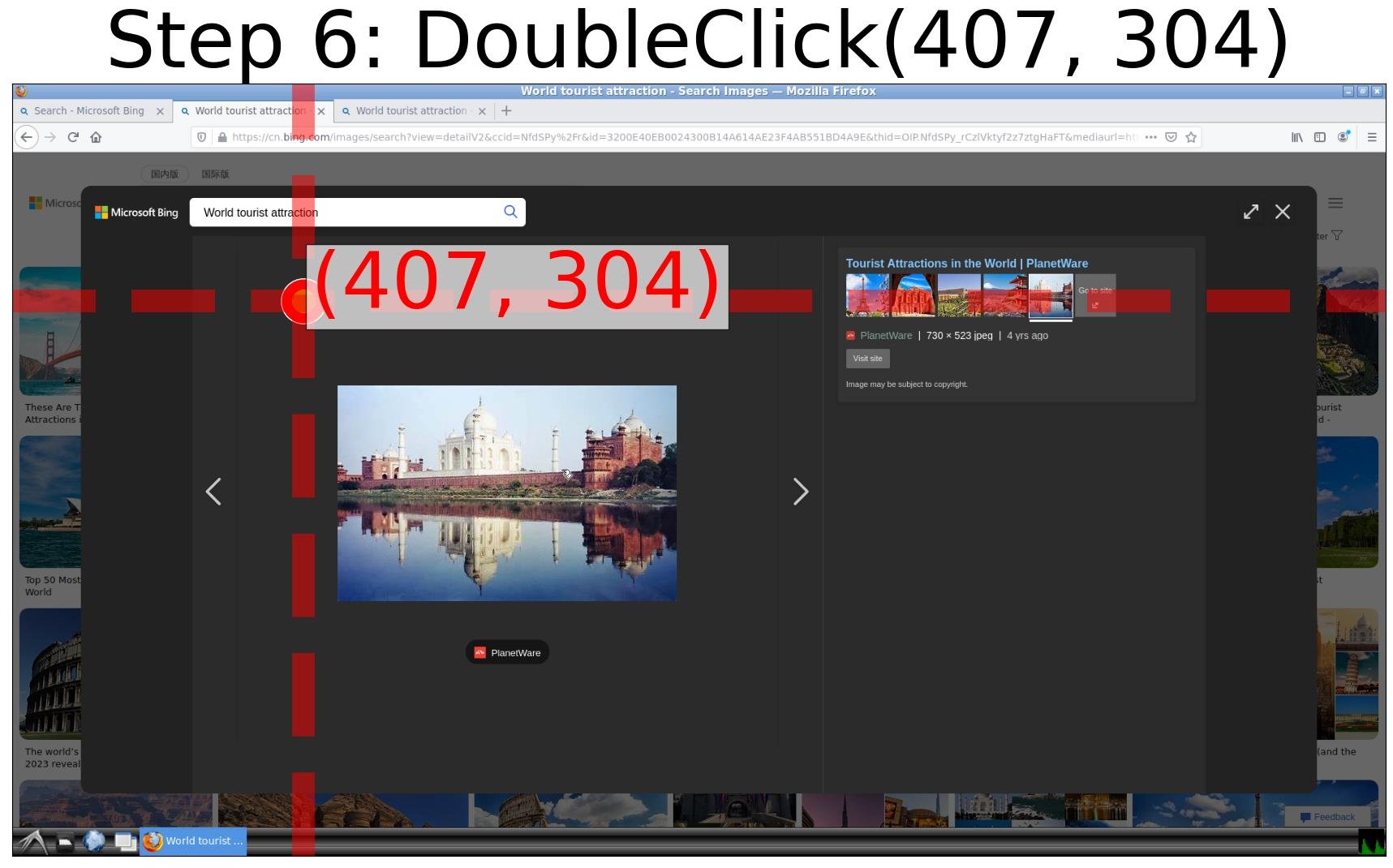} &
    \includegraphics[width=\linewidth]{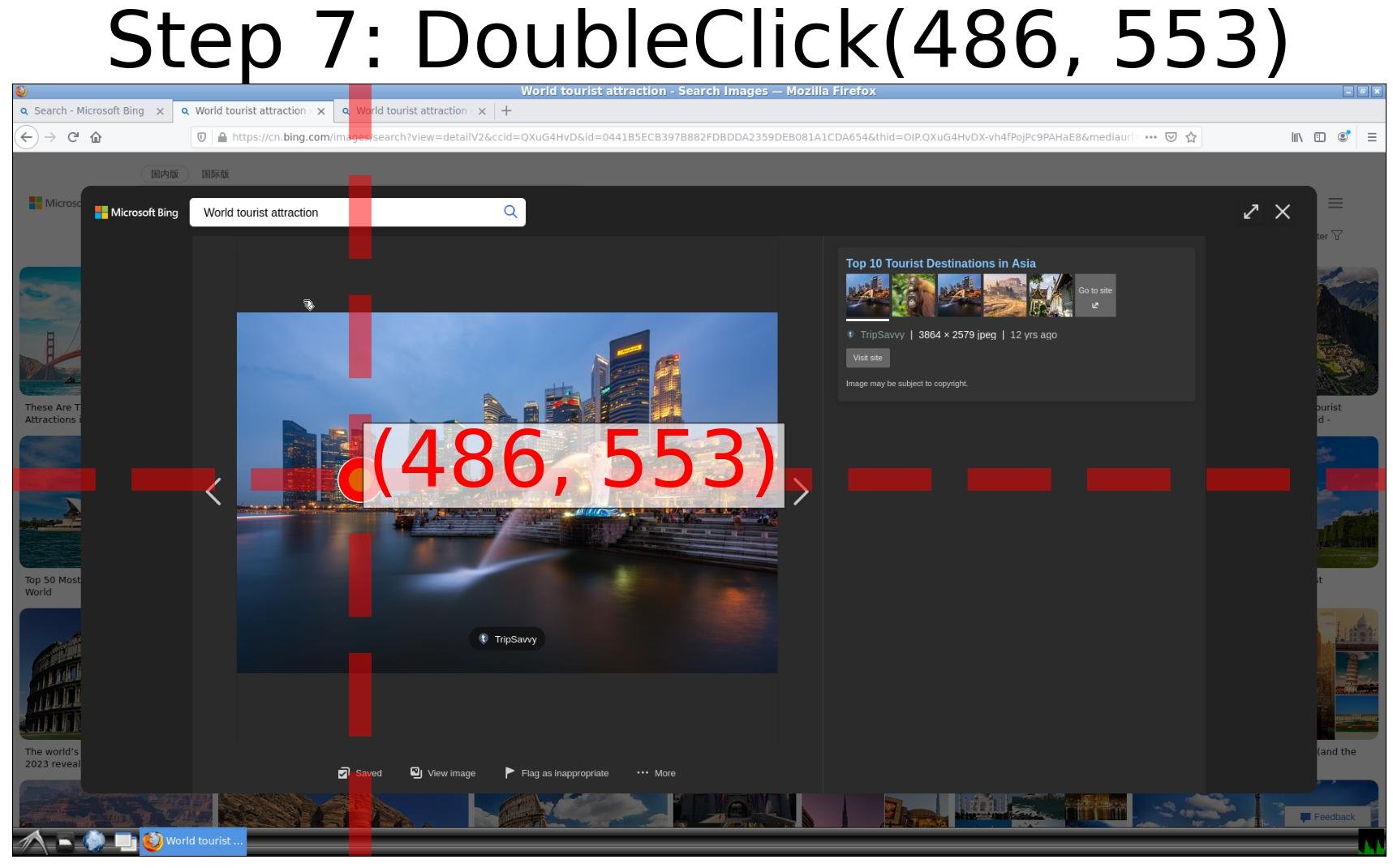} &
    \includegraphics[width=\linewidth]{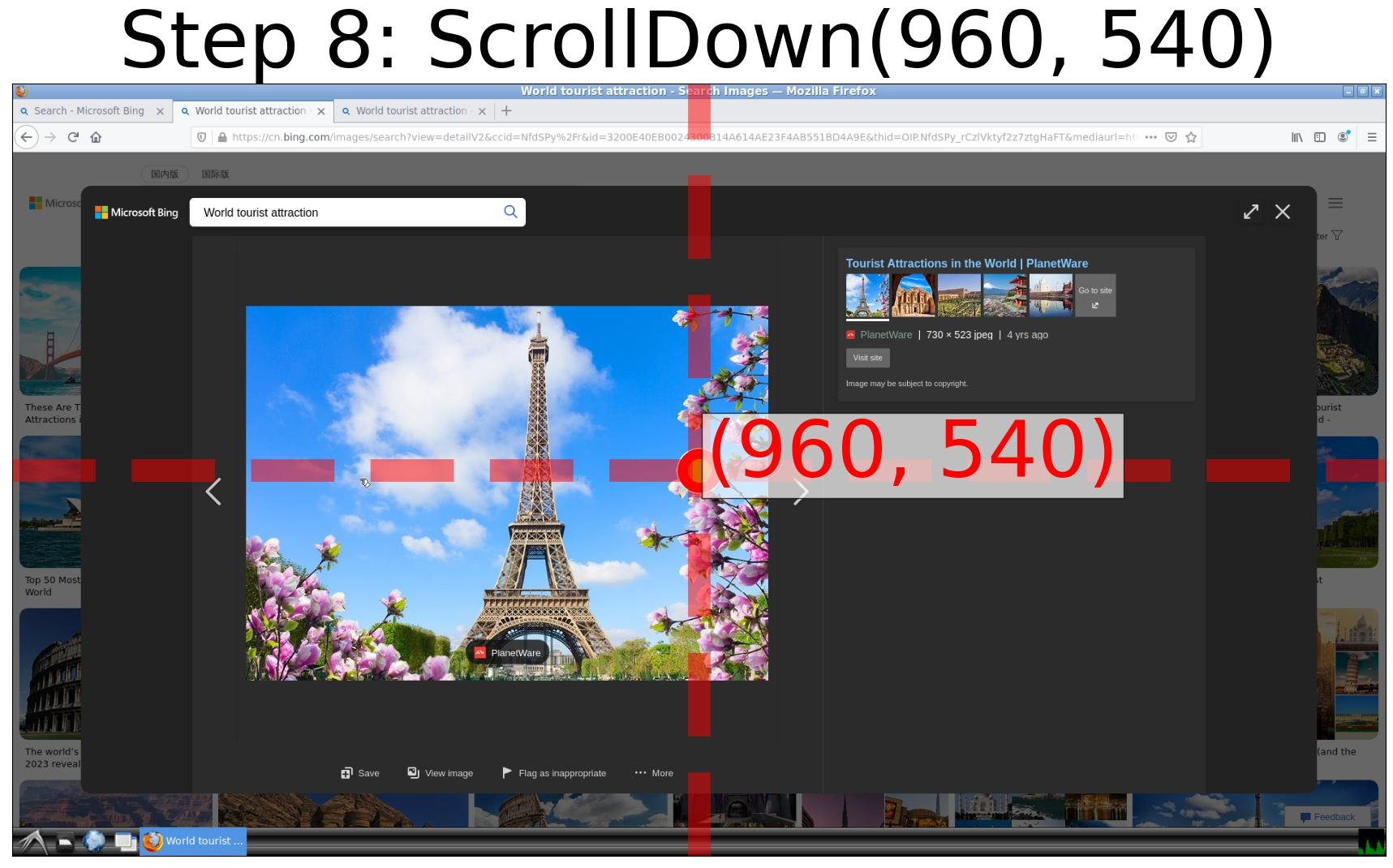} &
    \includegraphics[width=\linewidth]{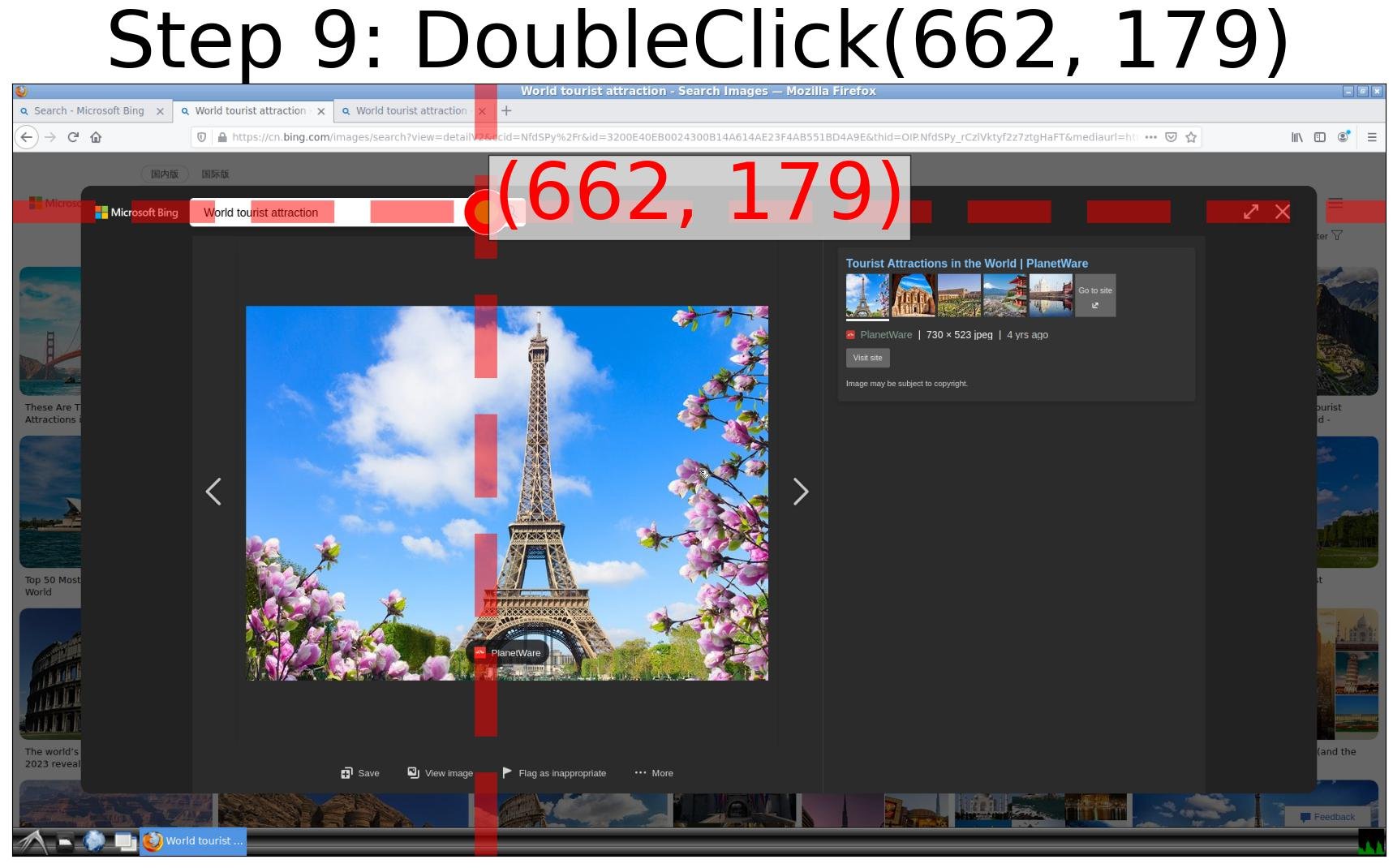}
  \end{tabular}
  \caption{Episode-90 of \textit{Qwen2.5-VL-7B}: The 7B model discovered that browsing through images yields higher exploration rewards.}\label{fig:7B-case-study-3}
\end{figure}

\begin{figure}[h]
  \centering
 \begin{tabular}{
    @{}
    m{0.195\textwidth}@{\hspace{1pt}}
    m{0.195\textwidth}@{\hspace{1pt}}
    m{0.195\textwidth}@{\hspace{1pt}}
    m{0.195\textwidth}@{\hspace{1pt}}
    m{0.195\textwidth}@{}
  }
    \includegraphics[width=\linewidth]{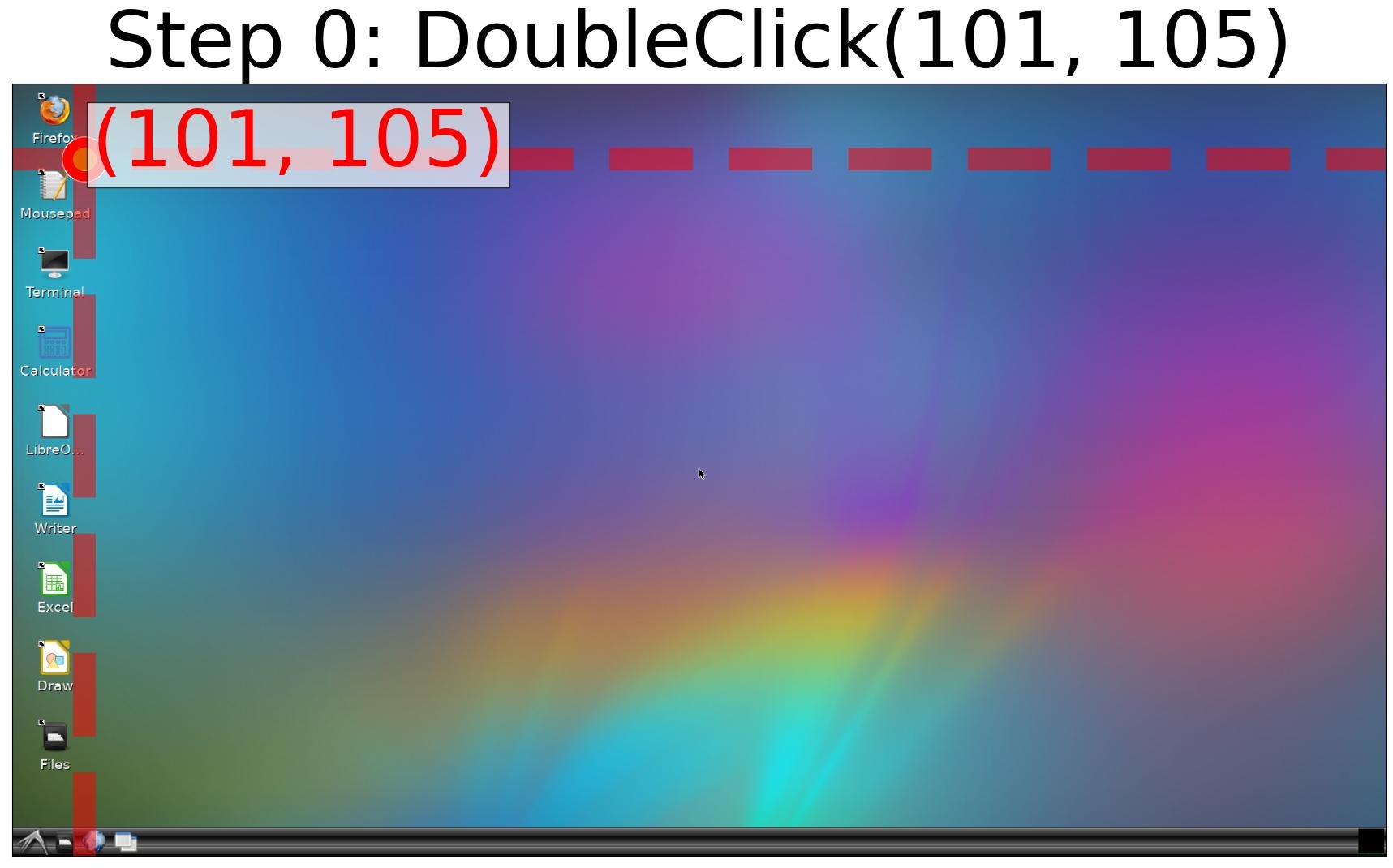} &
    \includegraphics[width=\linewidth]{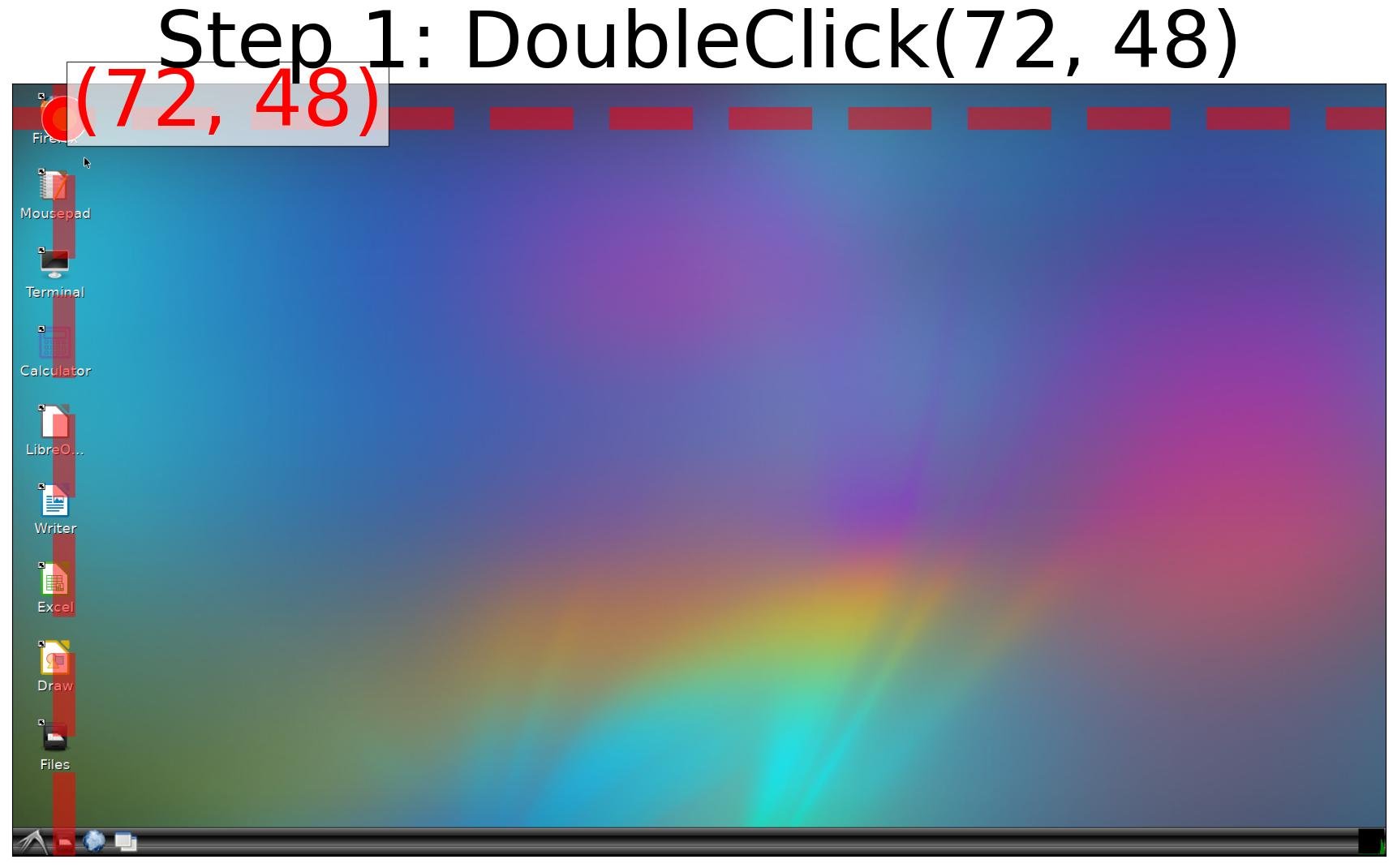} &
    \includegraphics[width=\linewidth]{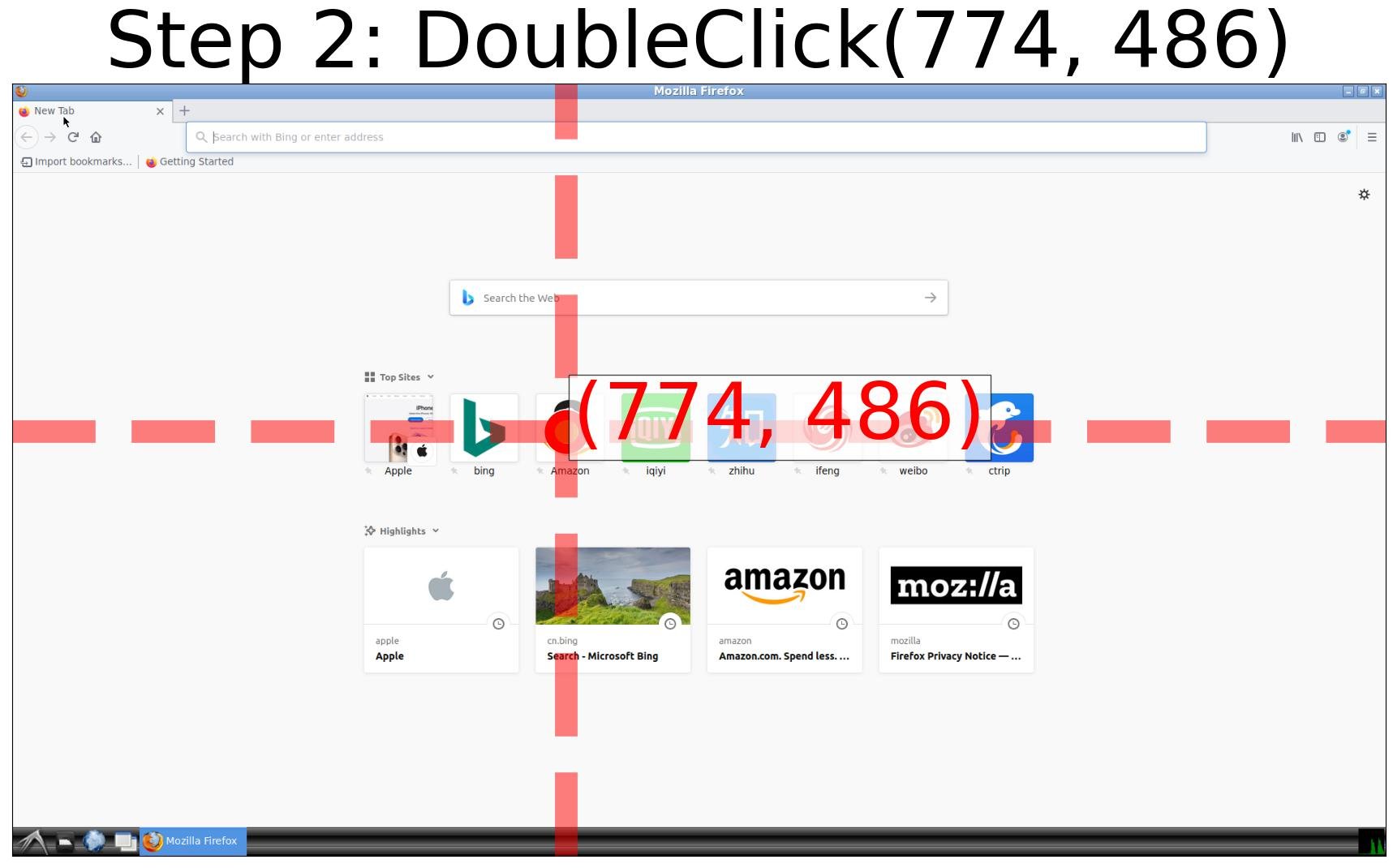} &
    \includegraphics[width=\linewidth]{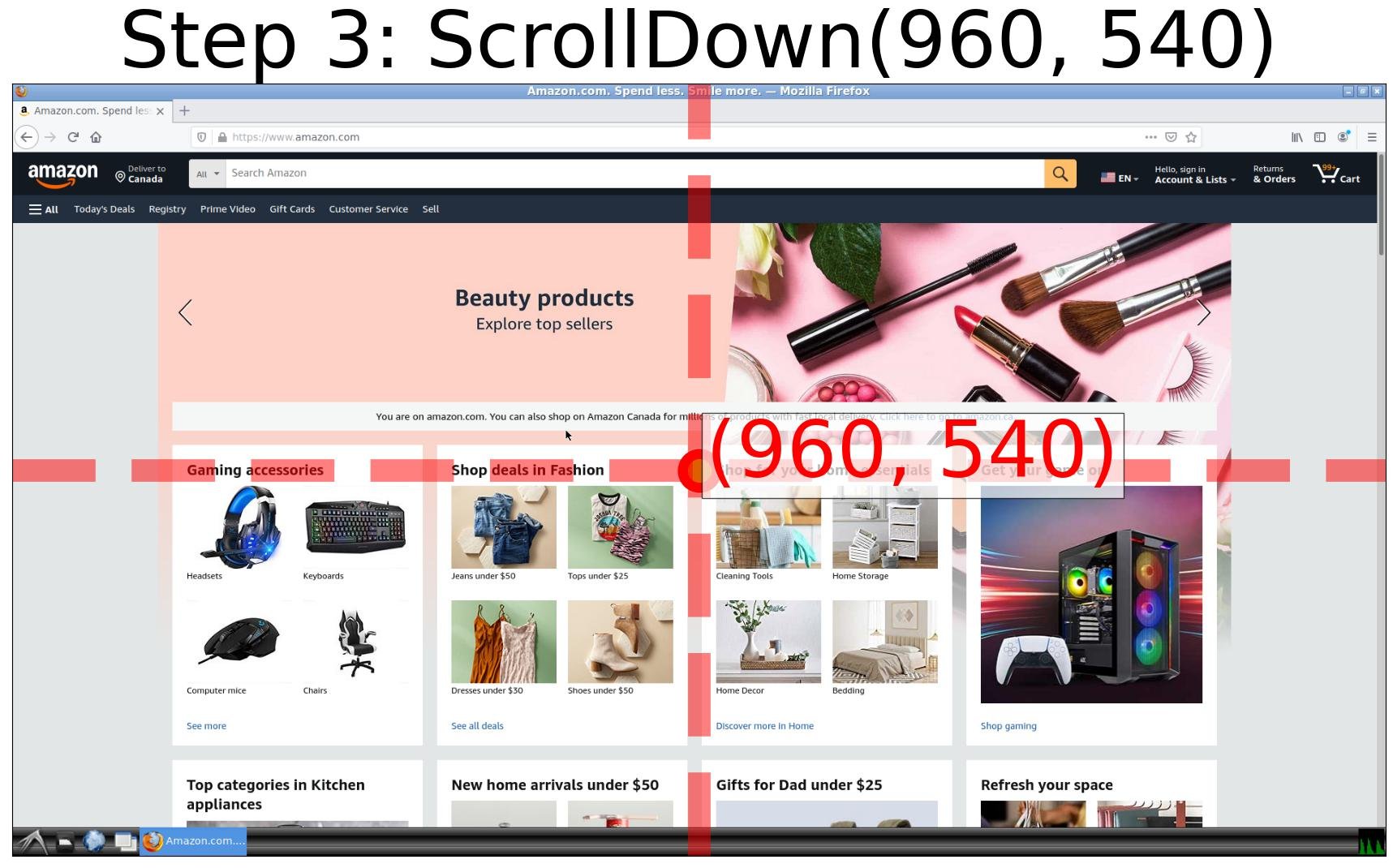} &
    \includegraphics[width=\linewidth]{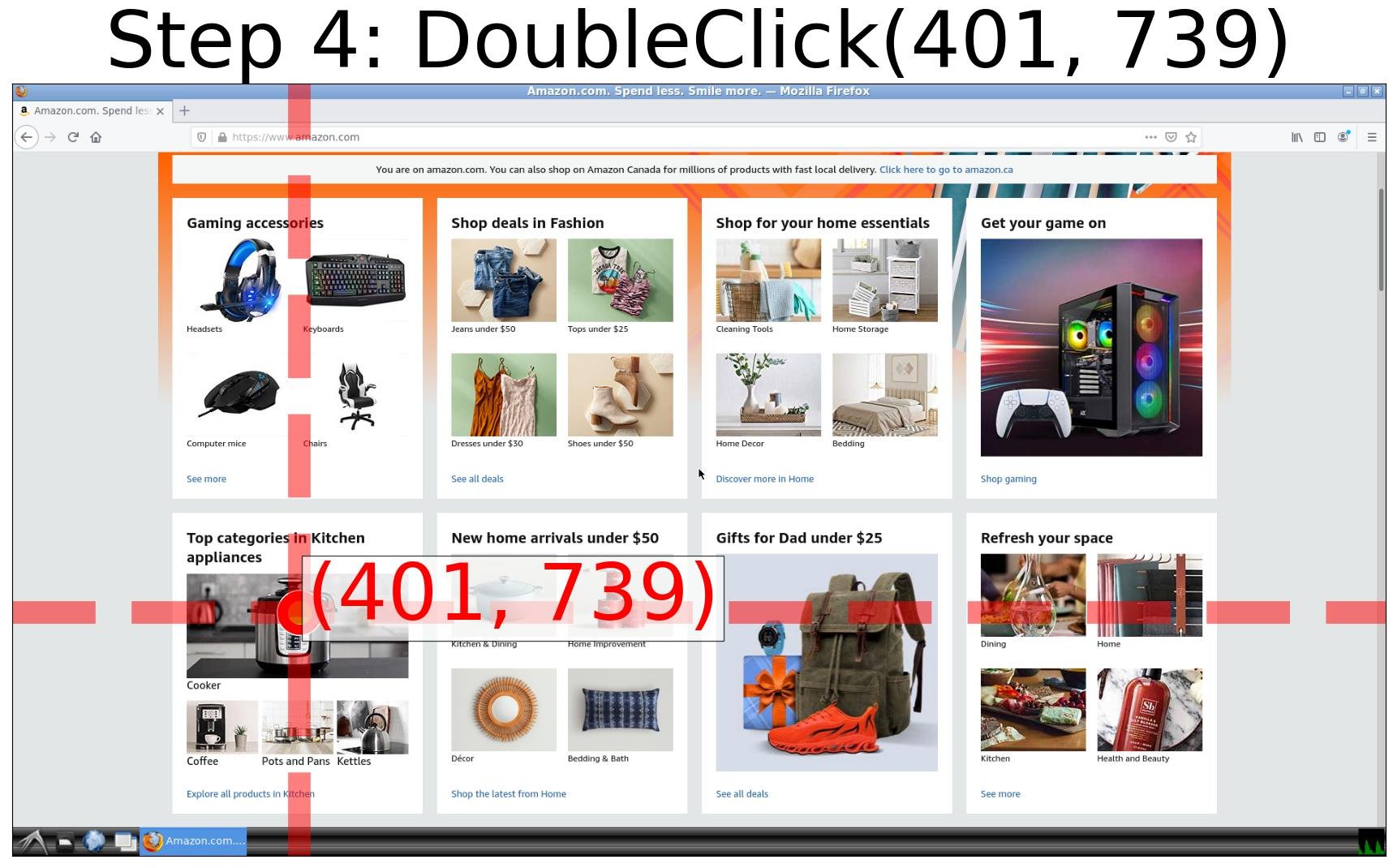} \\
    \includegraphics[width=\linewidth]{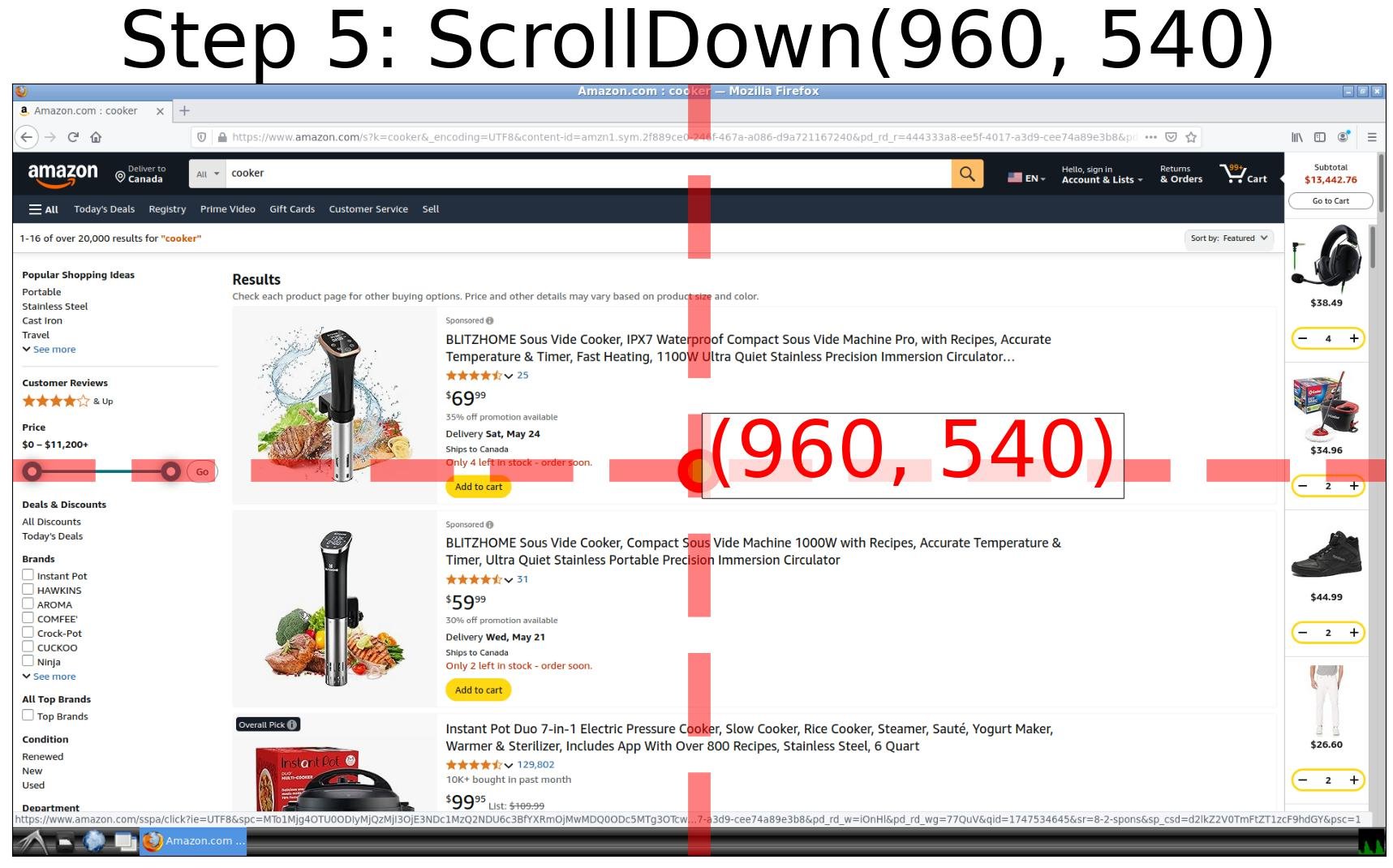} &
    \includegraphics[width=\linewidth]{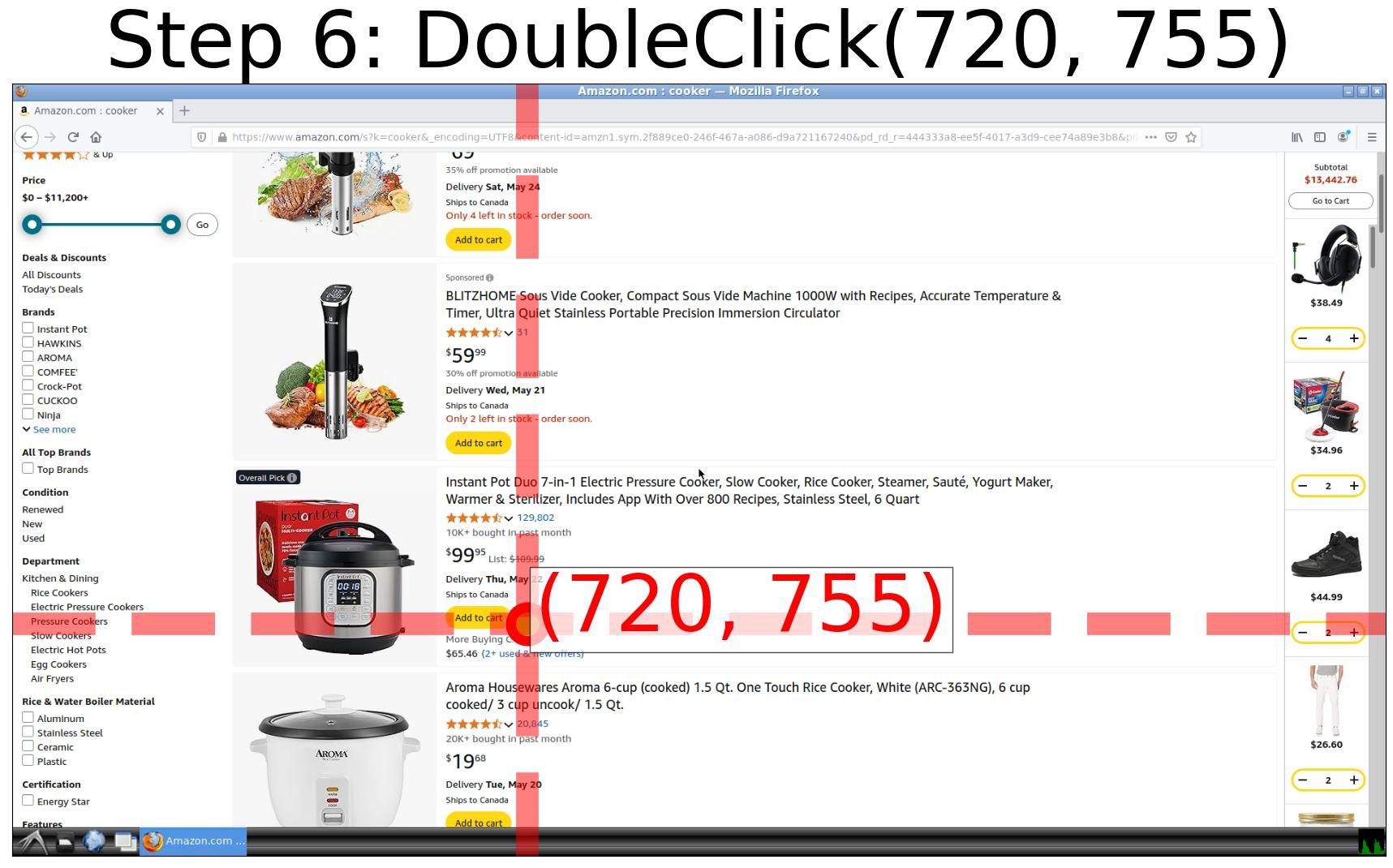} &
    \includegraphics[width=\linewidth]{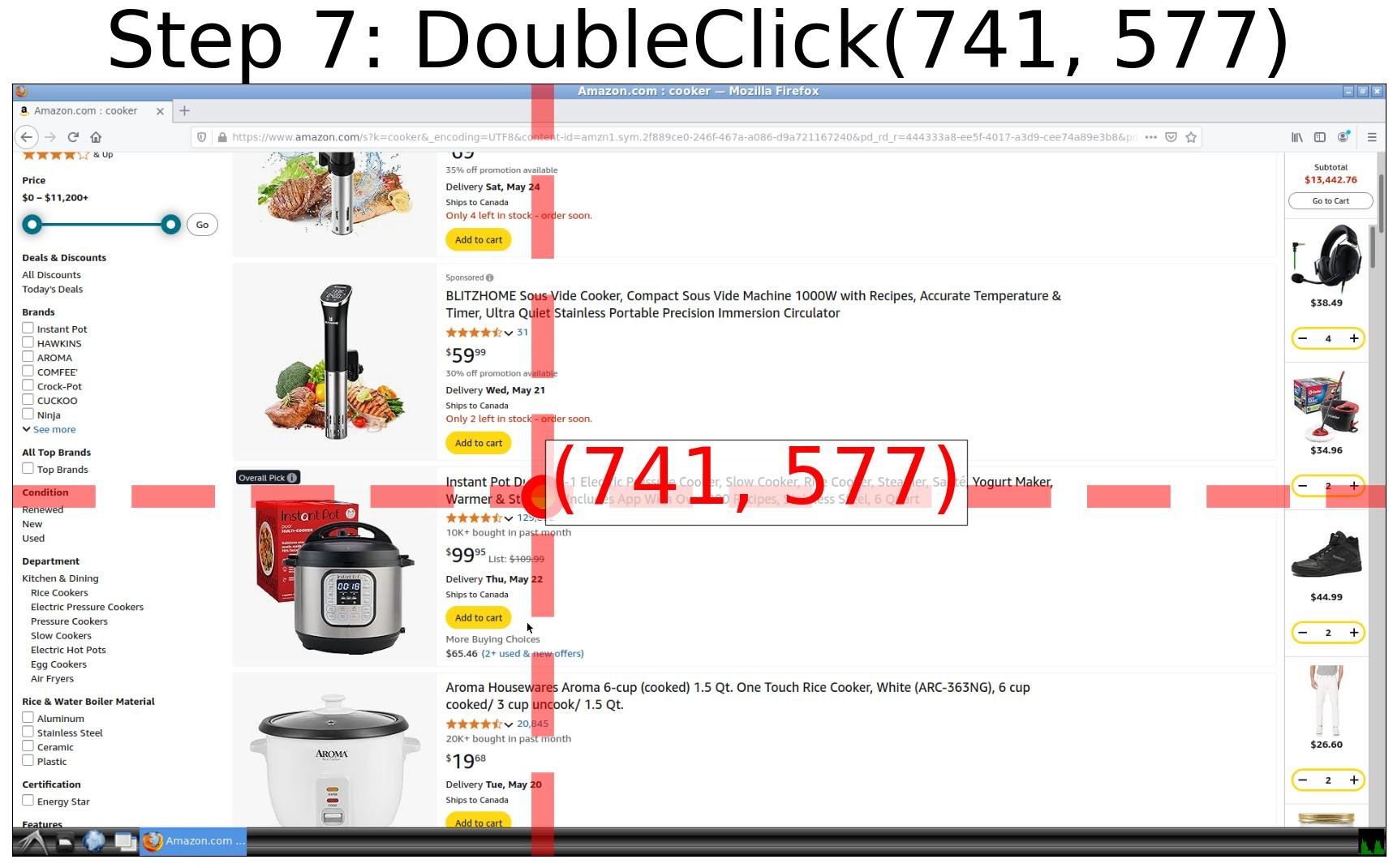} &
    \includegraphics[width=\linewidth]{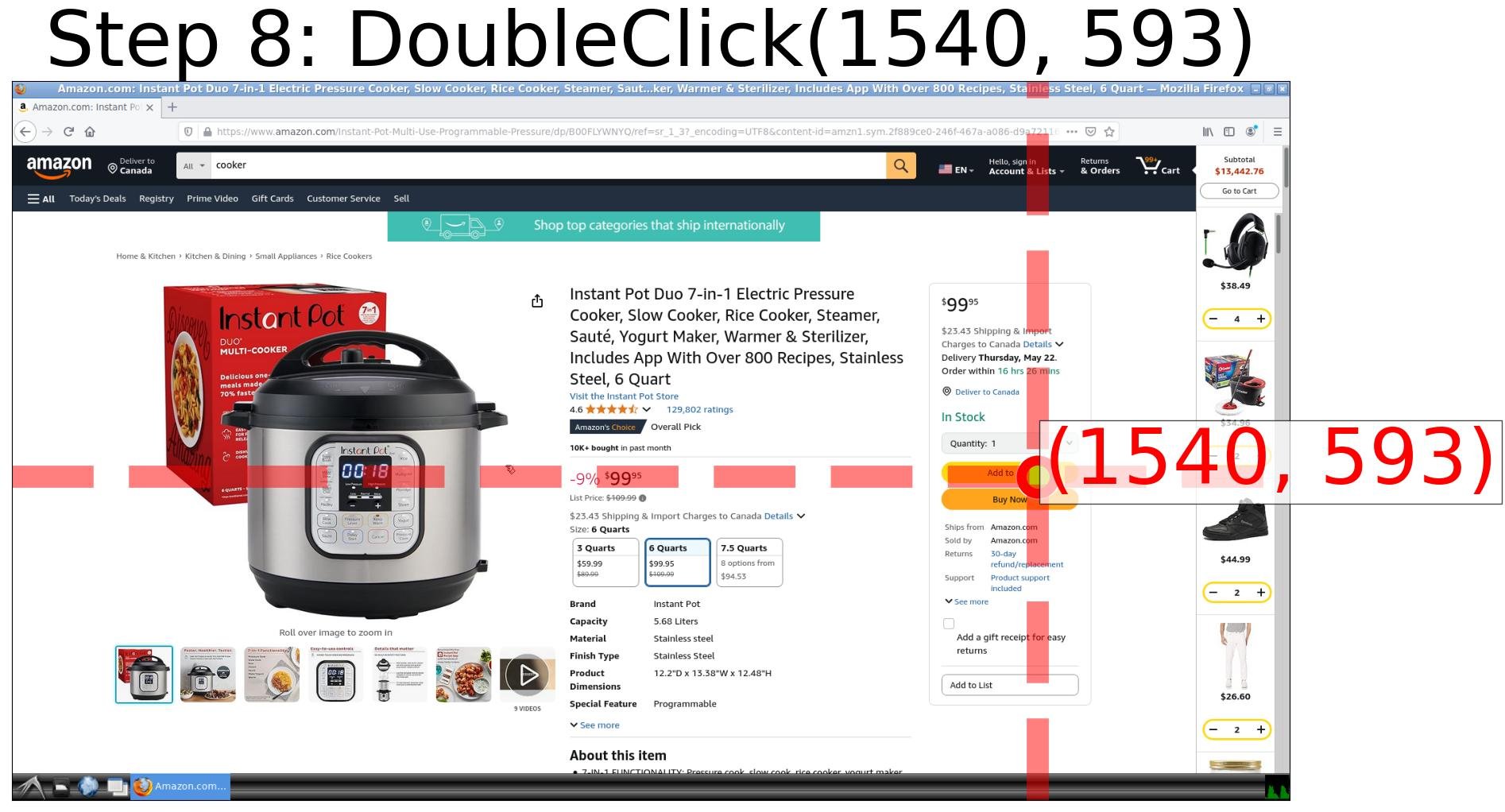} &
    \includegraphics[width=\linewidth]{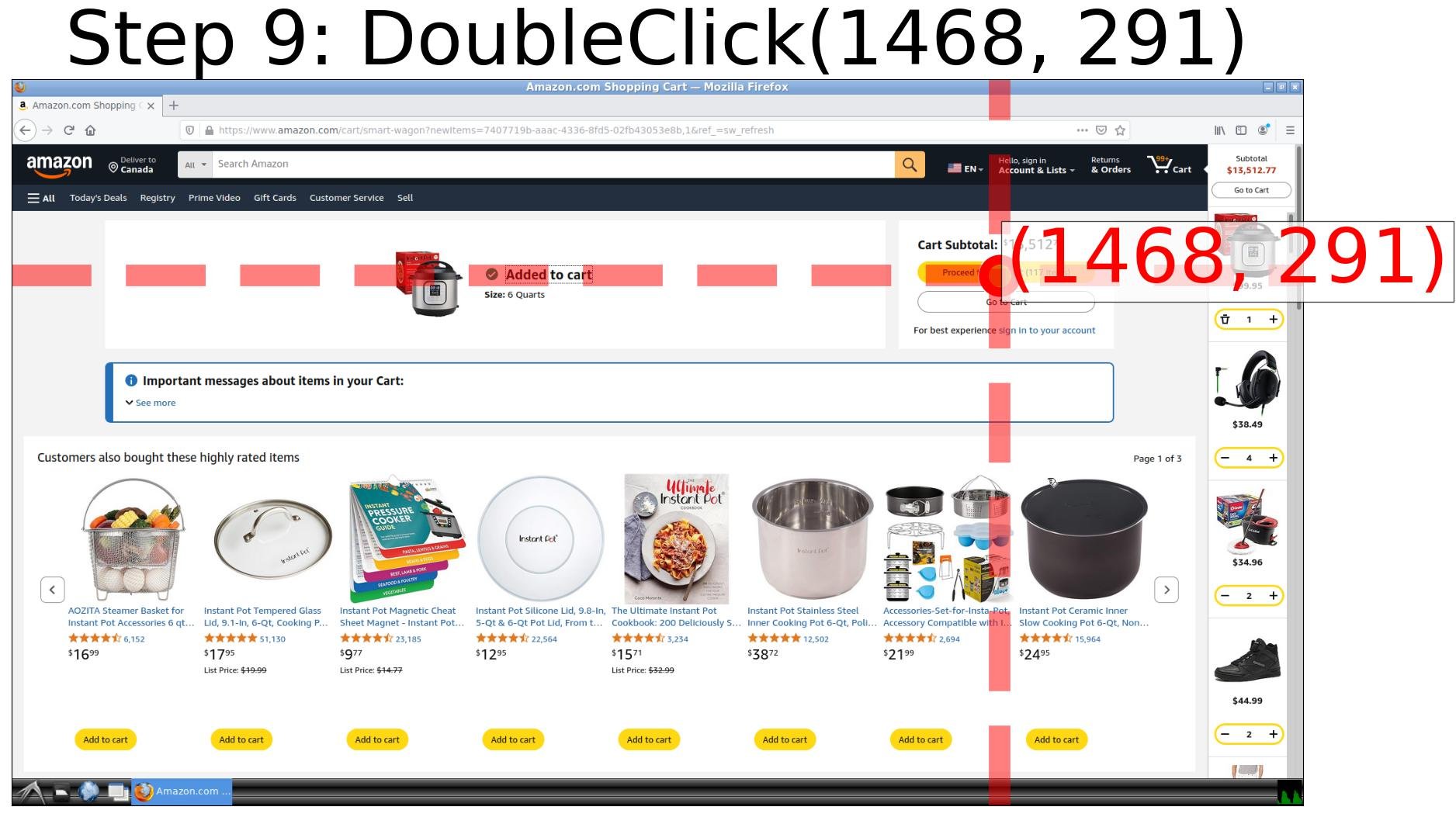}
  \end{tabular}
  \caption{Episode-100 of \textit{Qwen2.5-VL-7B}: The 7B model successfully completed a product purchase flow, from product selection to adding items to the shopping cart.}\label{fig:7B-case-study-4}
\end{figure}

\begin{figure}[h]
  \centering
  \begin{tabular}{@{} m{0.5\textwidth}  m{0.45\textwidth} @{}}
    \includegraphics[width=\linewidth]{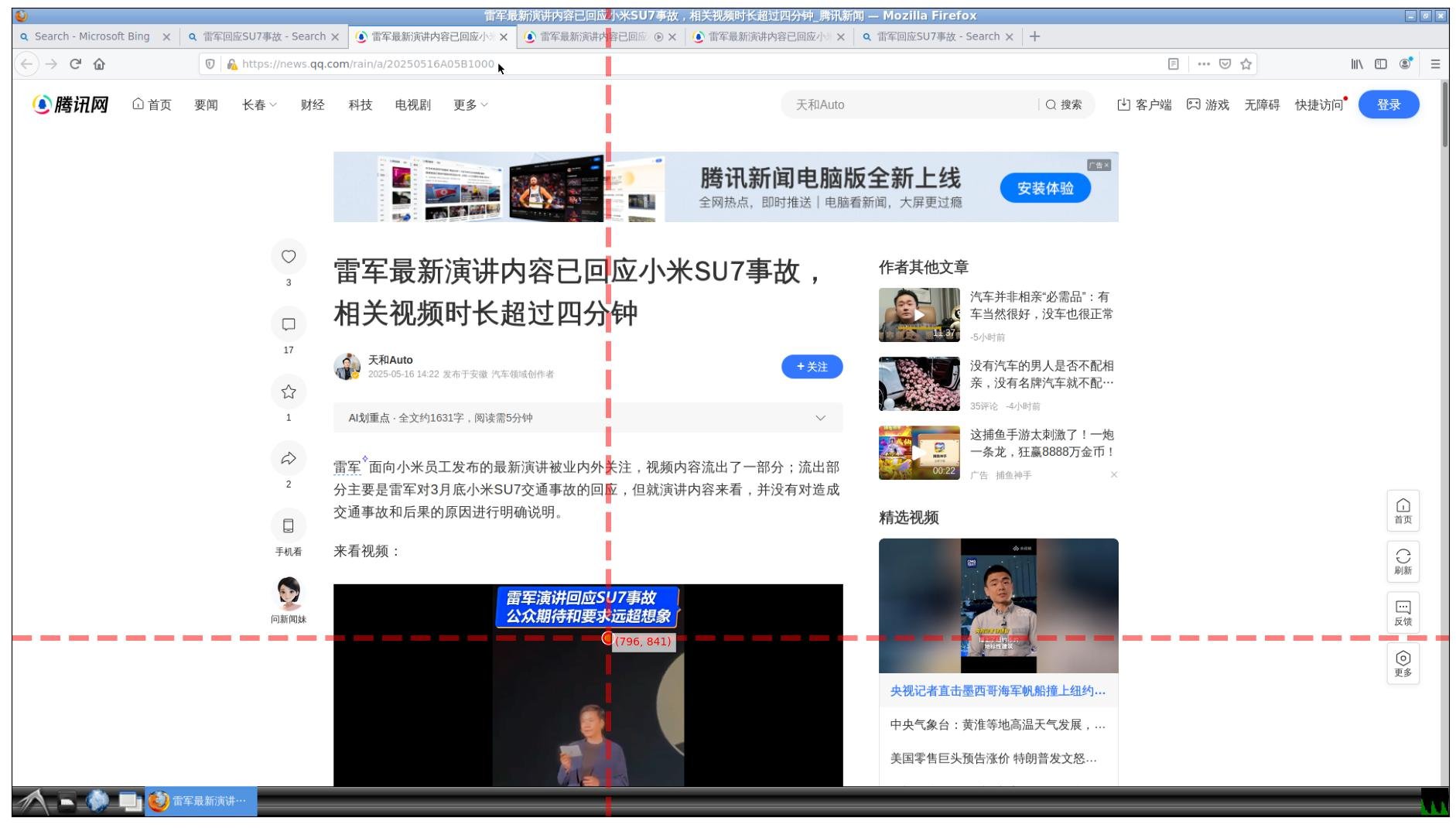} &
    \shortstack[l]{%
      \textbf{Intent:} “To continue exploring the webpage \\ 
      and find relevant information about the incident \\ 
      and the corresponding video. Clicking the video \\ 
      link will allow us to watch the related content.”\\[0.5ex]
      \textbf{Action:} “DoubleClick(796, 841)”
    }%
  \end{tabular}
  \caption{At episode 160, the 7B model demonstrated superior image reasoning capabilities and more precise action execution. In its intent description, it first comprehended that the news content pertained to an incident, then identified the potential presence of a "corresponding video" below, subsequently articulated a clear action intent of "clicking the video," and accurately targeted the video's location in its action output.}\label{fig:7B-case-study-5}
\end{figure}

Figures \ref{fig:7B-case-study-1} through \ref{fig:7B-case-study-5} present several case studies of the 7B model, demonstrating its superior GUI knowledge, environmental interaction and scene comprehension capabilities compared to the 3B model. Through our training process, we were able to enhance its exploratory behavior, enabling it to discover more diverse and deeper environmental states.

\end{document}